\documentclass{article}
\usepackage[top=1in,bottom=1in,left=1in,right=1in]{geometry}
\usepackage{amssymb}
\usepackage{amsmath}
\usepackage{mathtools}
\usepackage{xcolor}
\usepackage{subfigure}
\usepackage{hyperref}

\DeclareMathOperator*{\argmin}{arg\,min}

\providecommand{\keywords}[1]
{
  \small	
  \textbf{\textit{Keywords---}} #1
}

\title{Learning and discovering multiple solutions using physics-informed neural networks with random initialization and deep ensemble}
\author{Zongren Zou$^{1,*}$, Zhicheng Wang$^{1,*}$, George Em Karniadakis$^{1,\dagger}$}
\date{}

\begin{document}

\maketitle

\begin{center}
    \begin{minipage}{0.8\linewidth}
        \begin{center}
            $^{1}$ Division of Applied Mathematics, Brown University, Providence, 02906 RI, USA \\
            $^{*}$ These two authors contributed equally to this work. \\
            $^{\dagger}$ Corresponding author: george\_karniadakis@brown.edu.
        \end{center}
    \end{minipage}
\end{center}

\begin{abstract}

We explore the capability of physics-informed neural networks (PINNs) to discover multiple solutions. Many real-world phenomena governed by nonlinear differential equations (DEs), such as fluid flow, exhibit multiple solutions under the same conditions, yet capturing this solution multiplicity remains a significant challenge. A key difficulty is giving appropriate initial conditions or initial guesses, to which the widely used time-marching schemes and Newton's iteration method are very sensitive in finding solutions for complex computational problems.

While machine learning models, particularly PINNs, have shown promise in solving DEs, their ability to capture multiple solutions remains underexplored. In this work, we propose a simple and practical approach using PINNs to learn and discover multiple solutions. We first reveal that PINNs, when combined with random initialization and deep ensemble method -- originally developed for uncertainty quantification -- can effectively uncover multiple solutions to nonlinear ordinary and partial differential equations (ODEs/PDEs). Our approach highlights the critical role of initialization in shaping solution diversity, addressing an often-overlooked aspect of machine learning for scientific computing.

Furthermore, we propose utilizing PINN-generated solutions as initial conditions or initial guesses for conventional numerical solvers to enhance accuracy and efficiency in capturing multiple solutions. Extensive numerical experiments, including the Allen-Cahn equation and cavity flow, where our approach successfully identifies both stable and unstable solutions, validate the effectiveness of our method. These findings establish a general and efficient framework for addressing solution multiplicity in nonlinear differential equations.

\end{abstract}


\section{Introduction}\label{sec:1}

Many real-world phenomena, including fluid dynamics, quantum mechanics, and chemical reactions, are governed by nonlinear differential equations (DEs) that can exhibit multiple solutions under identical conditions \cite{hao2013completeness, lin2008enclosing, lemee2015multiple, kolvsek2007numerical, perumal2011multiplicity, del2010multiple}. For example, \cite{kuhlmann1997flow} demonstrated experimentally and numerically that steady flow in a 2D rectangular cavity can develop into two distinct patterns when the opposing walls move in opposite directions. Moreover, obtaining these multiple solutions numerically often requires prior knowledge, such as prescribing flow patterns like spanwise vortices \cite{osman2004multiple}. In general, capturing solution multiplicity in complex computational problems remains challenging, as numerical solvers typically converge to a single solution based on initialization or discretization choices. A key difficulty lies in selecting appropriate initial conditions or initial guesses in time-marching schemes and Newton’s method, which, while effective, are highly sensitive to these choices \cite{allgower2003introduction, cai1998parallel, cai2002nonlinearly, lo2012robust}.

In recent years, machine learning and neural network (NN) models have emerged as powerful tools for solving nonlinear ordinary and partial differential equations (ODEs/PDEs). These models offer key advantages, such as mesh-free formulations and the ability to approximate complex functions \cite{raissi2019physics, sirignano2018dgm, yu2018deep}. Among them, physics-informed neural networks (PINNs) \cite{raissi2019physics} have gained significant attention for tackling computational science and engineering problems. PINNs have been successfully applied across various domains, including fluid mechanics \cite{raissi2020hidden, almajid2022prediction, wessels2020neural, jin2021nsfnets}, solid mechanics \cite{goswami2020transfer, shukla2021physics, li2021physics, bastek2023physics}, epidemiology \cite{linka2022bayesian, kharazmi2021identifiability, qian2025physics}, and uncertainty quantification \cite{yang2021b, zou2025uncertainty, zou2024leveraging}. Beyond applications, recent theoretical advances have enhanced our understanding of PINNs, including convergence analysis and error estimation for specific PDEs \cite{shin2020convergence, wu2022convergence, qian2023physics}, multi-stage training techniques to achieve machine precision \cite{wang2024multi}, and the use of neural tangent kernels to study training dynamics \cite{wang2022and}. For a comprehensive overview of PINNs, their developments, and applications, readers may refer to \cite{karniadakis2021physics, toscano2024pinns, cuomo2022scientific, huang2022applications, cai2021physics}.

Despite these advancements, the ability of PINNs to capture multiple solutions of nonlinear DEs remains underexplored. Prior studies suggest that PINNs often exhibit bias toward a single solution due to the nature of their optimization process, making it difficult to recover solution multiplicity \cite{huang2022hompinns, di2020finding}. Moreover, standard PINN formulations typically employ a single NN to approximate the solution, limiting their capacity to uncover multiple valid solutions. 

In this work, we reveal that PINNs can effectively address solution multiplicity by leveraging random initialization of NNs and deep ensemble -- a simple and scalable approach originally developed for uncertainty quantification \cite{lakshminarayanan2017simple, psaros2023uncertainty, zou2024neuraluq}. Surprisingly, by introducing randomness at the initialization stage and aggregating solutions from an ensemble of networks, PINNs can effectively learn or discover multiple solutions to nonlinear ODEs/PDEs.
We demonstrate that the initialization method plays a crucial role in obtaining multiple solutions by shaping the diversity and bifurcation of approximated solutions, and PINNs are able to obtain both stable and unstable solutions.
% -- not only in influencing the training dynamics, as discussed in the literature \cite{glorot2010understanding, he2015delving, wang2022and, wang2021understanding}, but also in shaping the diversity and bifurcation of approximated solutions. 
To the best of our knowledge, this aspect has been largely overlooked in the literature of machine learning for scientific computing, and we are the first to systematically investigate it.

Furthermore, we propose using PINN solutions as initial conditions or initial guesses for conventional numerical solvers \cite{kierzenka2001bvp, shampine2000solving, bangerth2007deal, MATLAB, cook2007concepts, karniadakis2005spectral, xu2023conservation, africa2024deal, xu2024local}, such as finite difference methods (FDMs), finite element methods (FEMs), and spectral element methods (SEMs), to improve the accuracy and efficiency of computing multiple solutions. Due to the differentiability and mesh-free nature of NNs -- allowing evaluation of values and derivatives at arbitrary points -- PINNs can be seamlessly integrated with traditional numerical solvers \cite{shukla2025neurosem}. A key advantage of this approach is that, while PINNs may not always achieve the accuracy of high-fidelity solvers, their primary role is to identify distinct solution patterns or modes, rather than providing highly precise approximations. The final accuracy and convergence guarantees are ensured by the conventional solvers. As a result, exhaustive training of PINNs is not required, significantly enhancing computational efficiency.

As far as we know, existing machine learning approaches for handling solution multiplicity in nonlinear DEs impose restrictions on the type of equation being solved. For instance, \cite{di2020finding} introduces an interaction loss term in the PINN loss function to prevent solutions from coinciding, but this method is limited to ordinary differential equations (ODEs). In \cite{huang2022hompinns}, the homotopy continuation method was integrated into the PINN framework, termed HomPINNs, to obtain multiple solutions of nonlinear elliptic differential equations. This approach first approximates a starting function with a NN and then defines a homotopy function to transition from the starting function to the target solutions, enhancing the ability to find diverse solutions effectively while leveraging PINN training techniques. The HomPINNs method was further extended to solve inverse problems involving nonlinear differential equations with solution multiplicity in \cite{zheng2024hompinns}, where the authors addressed the challenge of unlabeled observations due to non-unique solutions. Additionally, \cite{hao2024newton} incorporates a pretrained deep operator network \cite{lu2021learning} into Newton’s method to efficiently and accurately handle solution multiplicity. However, these methods impose certain restrictions, such as assuming a linear differential operator.
In contrast, our approach imposes no such restrictions on the form of the DE, allowing for broader applicability.

The main contributions of this work are as follows:
\begin{enumerate}
    \item \textbf{Ensemble PINNs for multiple solutions}: We introduce a novel perspective of using PINNs with random initialization and deep ensemble to learn and discover solution patterns or modes of nonlinear DEs. This approach is simple yet remarkably effective, as evidenced by extensive numerical experiments.
    \item \textbf{Generality beyond prior methods}: Unlike previous works on PINNs for solution multiplicity \cite{huang2022hompinns, zheng2024hompinns, di2020finding, hao2024newton}, our framework accommodates a broader class of DEs with fewer restrictions.
    \item \textbf{Combination with conventional solvers}: We propose using ensemble PINNs to generate initial conditions or initial guesses for high-fidelity (FDM-, FEM-, and SEM-based) numerical solvers. This combination (1) improves PINN training efficiency by requiring only reasonable approximations, (2) enhances solution accuracy with guaranteed convergence through conventional numerical solvers, and (3) validates the computed solutions.
\end{enumerate}
The remainder of this paper is structured as follows. In Section \ref{sec:2}, we describe our proposed approach in detail, covering PINNs, random initialization and deep ensembles, and the procedure for integrating PINNs with conventional numerical solvers. Section \ref{sec:3} presents extensive numerical experiments, including 1D and 2D ODEs/PDEs, to validate the effectiveness of our approach in learning and discovering multiple solutions. Finally, we summarize our findings in Section \ref{sec:4}.

\section{Methodology}\label{sec:2}

\begin{figure}[ht]
    \centering
    \includegraphics[scale=0.6]{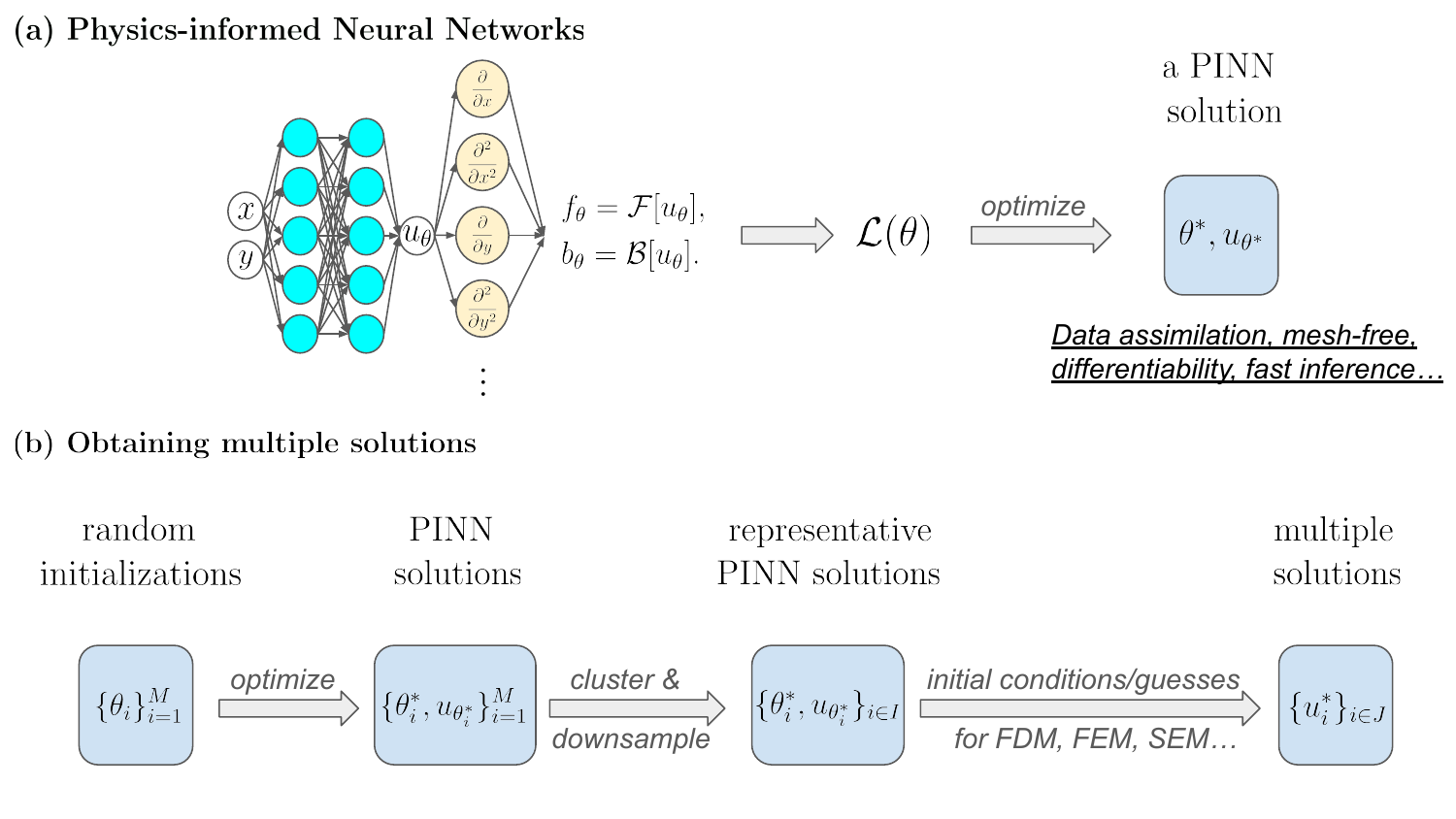}
    \caption{A schematic illustration of the proposed approach for solving ODEs/PDEs with solution multiplicity. \textbf{(a)} Traditional applications of PINNs typically focus on obtaining a single solution by optimizing the PINN loss function $\mathcal{L}(\theta)$ with respect to the NN parameters $\theta$. \textbf{(b)} When applied to ODEs/PDEs with multiple solutions, the approach consists of three key steps: (1) PINNs are randomly initialized to generate an ensemble of $M$ PINN solutions $\{u_{\theta^*_i}\}_{i=1}^M$; (2) these solutions are clustered based on their distinct patterns or modes and downsampled to a set of representative PINN solutions, $\{u_{\theta_i^*}\}_{i\in I}$ where $I$ denotes their index set; and (3) these representative solutions serve as initial conditions or guesses for conventional numerical solvers (FDM-, FEM-, or SEM-based), refining them into highly accurate solutions, $\{u_i^*\}_{i\in J}$, with high precision and guaranteed convergence.}
    \label{fig:1}
\end{figure}

In this work, we consider the nonlinear DE with solution multiplicity described as follows:
\begin{equation}\label{eq:problem}
\begin{dcases}
\mathcal{F}[u](x) = f(x),  x\in \Omega, \\ 
\mathcal{B}[u](x) = b(x), x \in \partial\Omega,
\end{dcases}
\end{equation}
where $\Omega$ is a bounded domain, $x$ represents the spatial-temporal coordinate, $\mathcal{F}$ and $\mathcal{B}$ represent differential and boundary operators, respectively, and $f$ and $b$ are source and boundary terms.

\subsection{Physics-informed neural networks (PINNs)}

The PINNs method, originally proposed in \cite{raissi2019physics}, utilizes neural networks (NNs) and modern machine learning techniques to solve ODE/PDE problems. In this approach, the sought solution is approximated by a NN, denoted as $u_\theta$ where $\theta$ represents the NN parameter. The underlying physics defined in \eqref{eq:problem} is encoded into the loss function via automatic differentiation, leading to the following optimization problem.
\begin{equation}\label{eq:loss}
    \theta^* = \argmin_{\theta}\mathcal{L}(\theta), \text{ where } \mathcal{L}(\theta) := \frac{w_f}{N_f}\sum_{i=1}^{N_f} |\mathcal{F}[u_\theta](x_i^f) - f_i|^2 + \frac{w_b}{N_f}\sum_{i=1}^{N_b} |\mathcal{B}[u_\theta](x_i^b) - b_i|^2.
\end{equation}
Here, $\{x_i^f, f_i\}_{i=1}^{N_f}, \{x_i^b, b_i\}_{i=1}^{N_b}$ are data for the equation and the boundary, $w_f$ and $w_b$ are belief weights, $|\cdot|$ denotes the $l^2$ norm, and $\theta^*$ denotes the minimizer. In this work, we refer to $u_{\theta^*}(x), x\in\Omega$ as a \textbf{PINN solution}. A schematic overview of the PINN framework is illustrated in Figure \ref{fig:1}(a).

While PINNs have proven effective for solving nonlinear ODEs and PDEs, they may lack high accuracy and convergence guarantees in certain scenarios.  Unlike prior studies focused on improving, accelerating, or stabilizing PINN training \cite{mcclenny2023self, lu2021deepxde, yu2022gradient, wang2022and, wang2021understanding, wang2024respecting}, our work introduces a novel perspective: leveraging random NN initializations to systematically learn and discover multiple solutions or patterns of DEs.

\subsection{Random initialization of NNs and deep ensemble}\label{sec:2_2}

\begin{figure}[ht]
    \centering
    \includegraphics[scale=0.42]{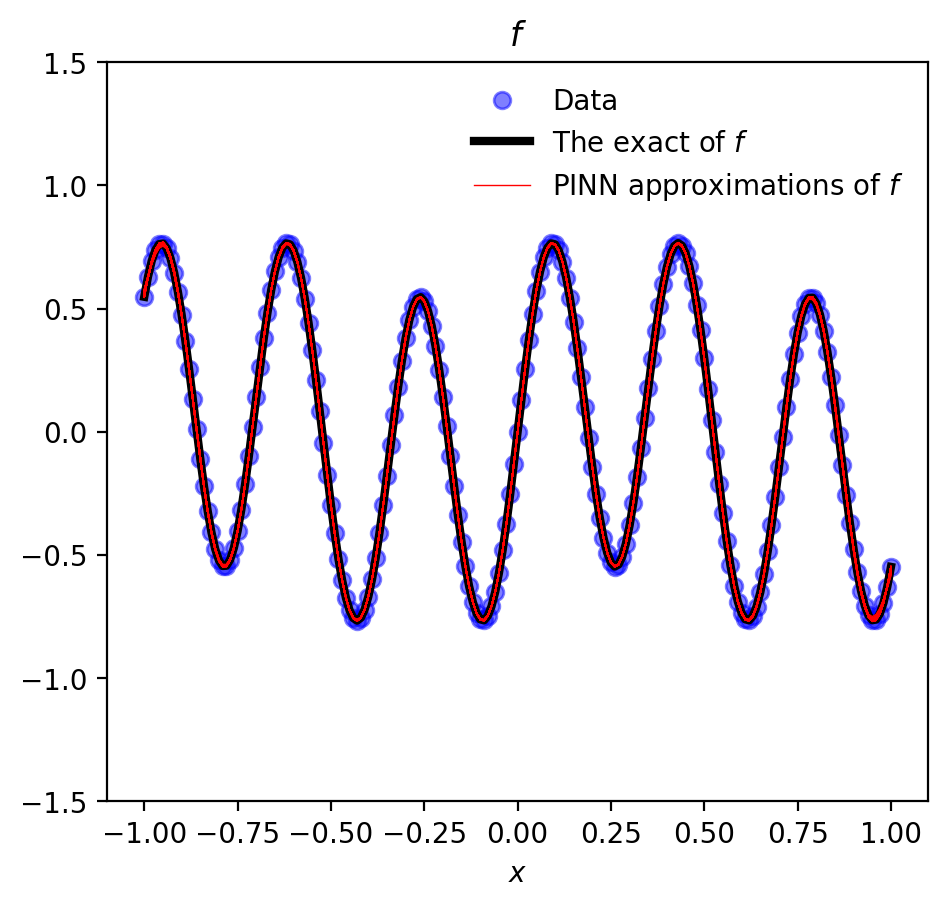}
    \includegraphics[scale=0.42]{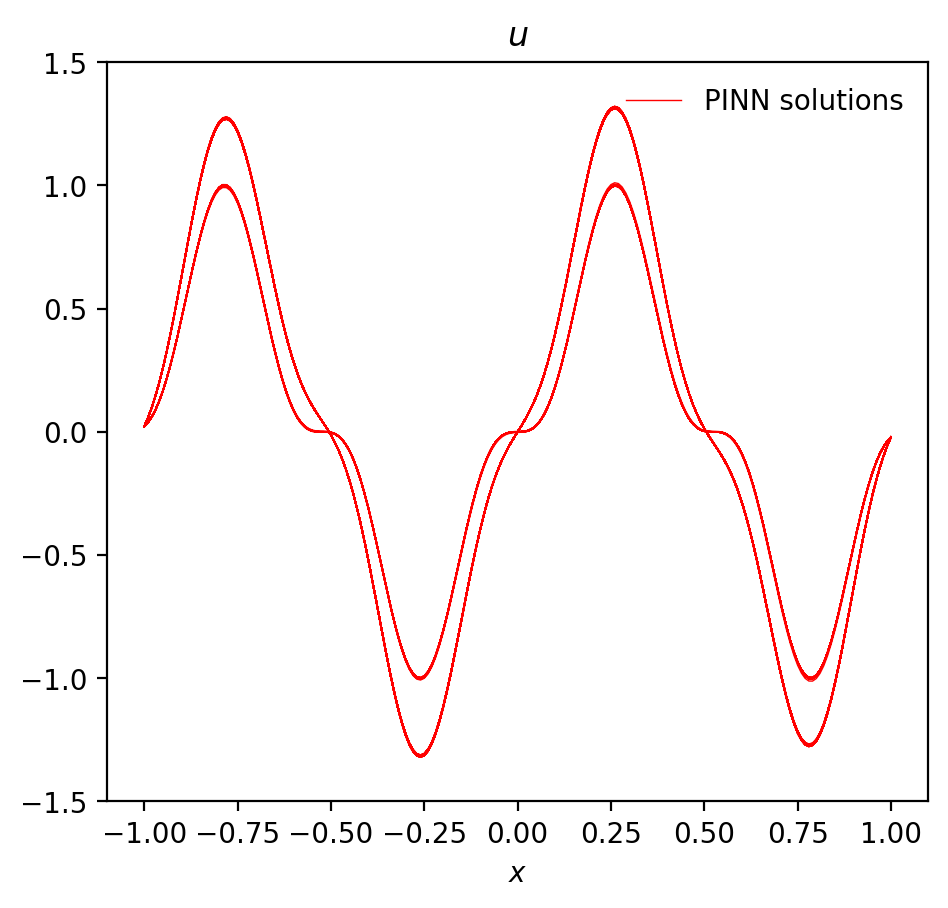}
    \includegraphics[scale=0.42]{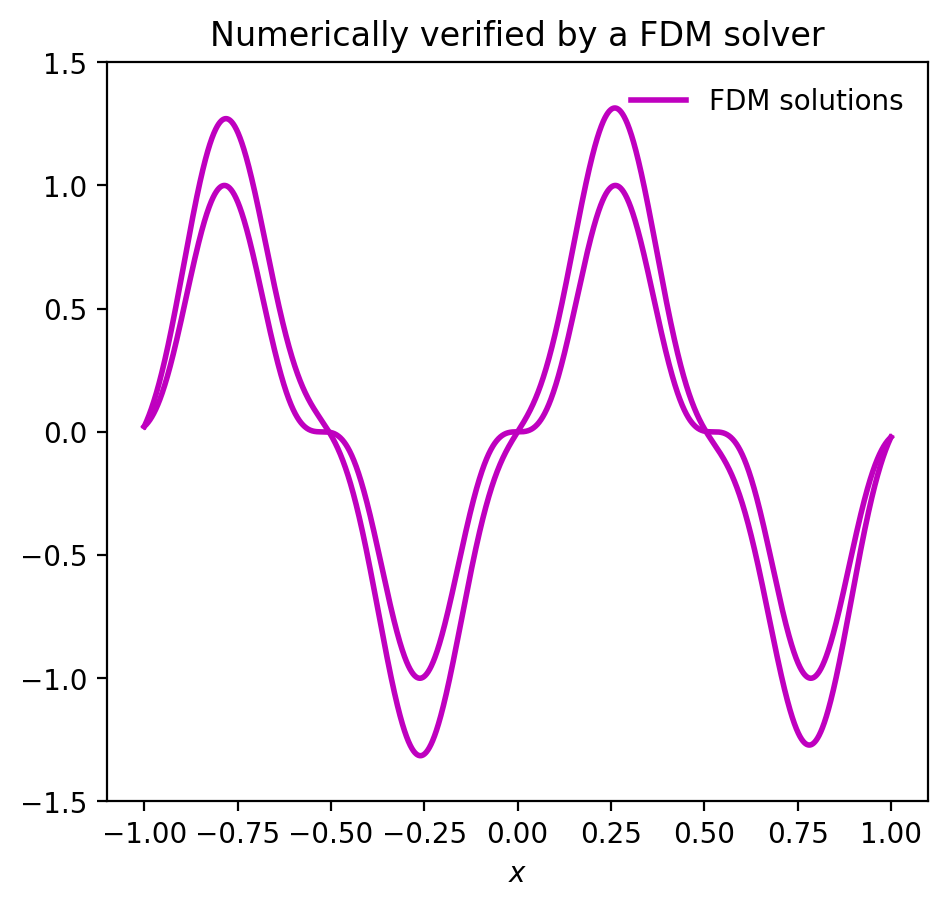}
    \caption{Solving a 1D steady-state PDE using PINNs with $1,000$ randomly initialized NNs, which results in the same approximation of the source term $f$ (on the \textbf{left}) but two distinct approximations of the solution $u$ (in the \textbf{middle}). These two distinct PINN solutions are then numerically verified by using them as the initial guesses for a finite-difference-method-based solver on the uniform mesh with mesh size $h=1/1600$ (on the \textbf{right}). Details and more results can be found in Section 3.3. }
    \label{fig:first_example}
\end{figure}

The initialization method for PINNs remains an open question and is rarely explored in the literature. The scientific machine learning community often adopts initialization techniques developed for conventional machine learning applications \cite{lu2021deepxde, psaros2023uncertainty, wang2023expert}, such as the method proposed in \cite{glorot2010understanding}, which helps mitigate vanishing gradients and accelerate training for image classification tasks. However, when NNs are used to solve nonlinear differential equations, the analysis in \cite{glorot2010understanding} may not be directly applicable, as the loss function involves derivatives of the network and introduces greater nonlinearity. While in some cases the choice of initialization may have little impact on obtaining accurate approximations for ODEs/PDEs, it becomes crucial when identifying solutions with different patterns in problems exhibiting solution multiplicity. We demonstrate this by employing $1,000$ PINNs with random initialization to solve a prototype 1D steady-state nonlinear PDE, with results presented in Figure \ref{fig:first_example}. As shown, the PINN approximations of the source term $f$ closely match both the data and the exact, while the PINN solutions ($u$) themselves naturally cluster into two distinct patterns. These patterns are then numerically verified as two separate numerical solutions using a FDM solver. This simple example highlights the ability of PINNs to capture solution multiplicity through random initialization and deep ensemble. Details and more results are discussed in Section \ref{sec:3_3}.

Random initialization of NNs forms the foundation of a widely used uncertainty quantification (UQ) method known as deep ensemble. Originally proposed by \cite{lakshminarayanan2017simple} as a simple and scalable approach to quantifying uncertainty in NN models, this method was later introduced to scientific machine learning (SciML) due to its practicality and flexibility \cite{psaros2023uncertainty, zou2024neuraluq}. It has since been applied to various SciML problems involving UQ (e.g., \cite{zou2024correcting, haitsiukevich2022improved, de2024quantification, jiang2023practical, soibam2024inverse}).
While the randomness of NN initialization serves as the primary source of uncertainty in this approach, the impact of different initialization methods on the quantified uncertainty remains underexplored. Many existing works rely on default initialization settings provided by modern machine learning platforms. Given that random initialization enables the discovery of multiple solutions, we will empirically investigate and demonstrate in Section \ref{sec:3} that the choice of initialization method plays a crucial role in the diversity of these solutions as well.

Training randomly initialized neural networks to obtain multiple solutions can be efficiently vectorized on a single processing unit or parallelized across multiple units, leveraging modern machine learning software and hardware, as each network is trained independently. In this section, we propose two alternative approaches to further enhance the efficiency of our method for solving PDEs with solution multiplicity:
\begin{enumerate}
    \item Parameter-sharing NN architectures such as multi-head ones \cite{zou2023hydra} -- reducing the total number of parameters in the optimization problem significantly.
    \item Hybrid approach with conventional numerical solvers -- using PINN solutions as initial conditions and initial guesses for time-marching and Newton's methods, respectively, thereby relaxing the requirement for high accuracy in PINN solutions.
\end{enumerate}

\subsection{Initialization from a PINN solution for solution multiplicity}\label{sec:2_3}

In this section, we propose a general procedure for solving ODEs/PDEs with solution multiplicity by integrating PINNs with conventional numerical methods:
\begin{enumerate}
    \item \textbf{Step 1}: Construct multiple randomly initialized NNs and train them to obtain multiple PINN solutions.
    \item \textbf{Step 2}: Analyze the obtained PINN solutions and downsample them to a set of representative PINN solutions.
    \item \textbf{Step 3}: Use the representative PINN solutions as initial condition or initial guesses for conventional numerical solvers.
\end{enumerate}
A schematic illustration of the proposed approach is shown in Figure \ref{fig:1}(b). Notably, due to the differentiability and mesh-free nature of NNs, PINN solutions can be efficiently evaluated on any given mesh and at any order of derivatives, making them easy and flexible to integrate with various existing numerical solvers. For example, certain boundary-value-problem (BVP) solvers require the initial guess to include both the function value and its higher-order derivatives \cite{kierzenka2001bvp, shampine2000solving}. We also note that conventional numerical solvers in this work serve two key purposes: (1) reducing the training cost of PINNs, and (2) verifying whether the solutions discovered by PINNs with random initialization and deep ensembles are true numerical solutions by assessing their convergence.

\section{Computational results}\label{sec:3}

In this section, we numerically investigate random initializations of NNs in solving ODEs/PDEs with solution multiplicity, and demonstrate the effectiveness of the presented approach in obtaining multiple solutions. We consider five 1D and 2D examples, including: (1) the 1D Bratu problem, (2) a boundary layer problem, (3) a 1D steady-state PDE, (4) a 2D steady-state Allen-Cahn equation, and (5) a 2D steady-state lid-driven cavity flow problem. Details of the computations (e.g., training steps, NN architectures, etc.) can be found in Appendix \ref{sec:appendix_1}. Another example of a 2D PDE with multiple solutions and additional results are presented in Appendix \ref{sec:appendix_2} for verification of the proposed approach and Appendix \ref{sec:appendix_3}, respectively.
The code to reproduce the results from this section will be made publicly available at \url{ https://github.com/ZongrenZou/PINNs_multiple_solutions} upon acceptance of this paper.

\subsection{1D Bratu problem}\label{sec:3_1}

\begin{figure}[ht]
    \centering
    \subfigure[]{
        \includegraphics[scale=0.25]{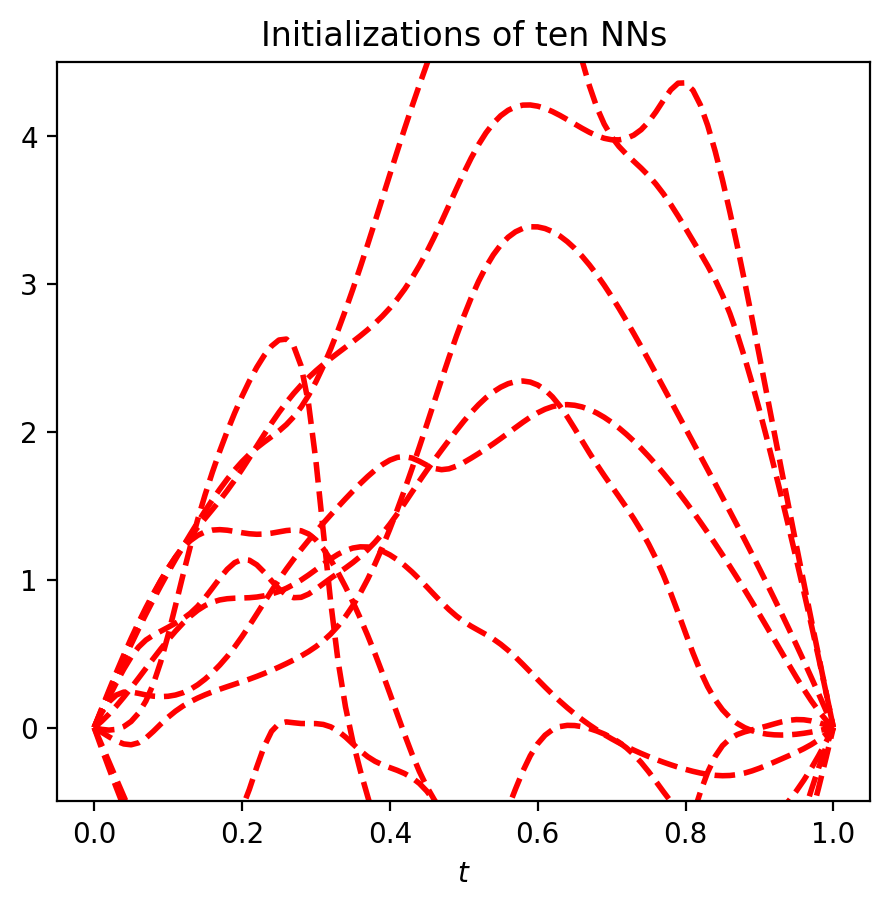}
    }
    \subfigure[]{
        \includegraphics[scale=0.25]{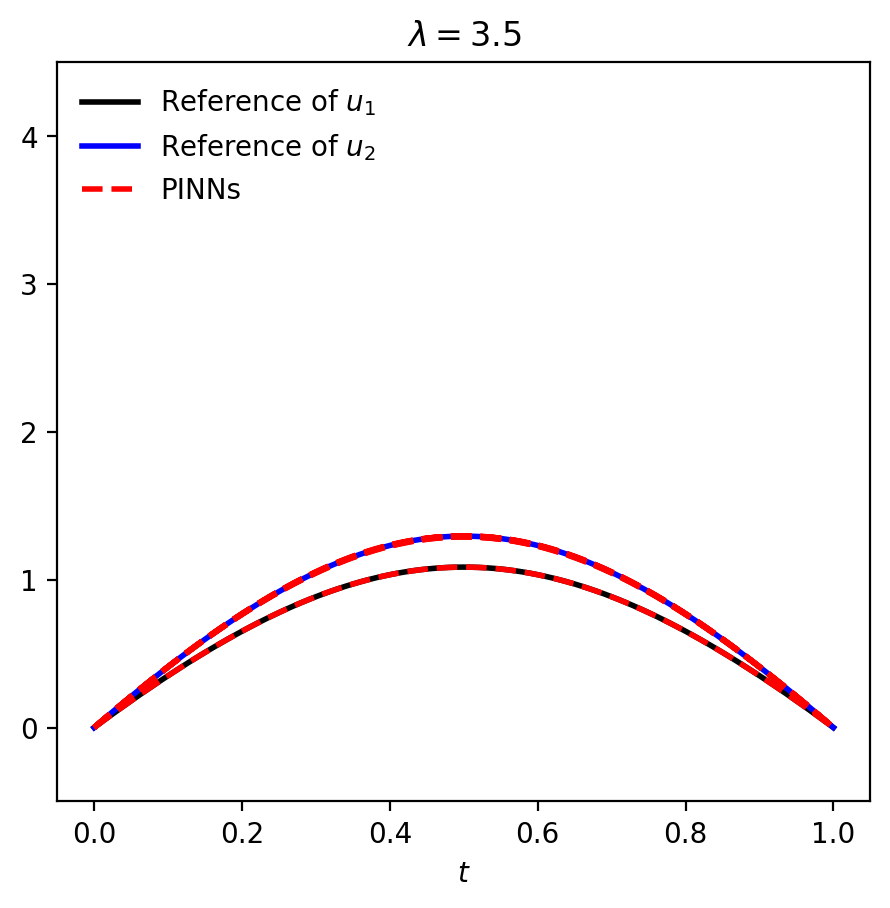}
        \includegraphics[scale=0.25]{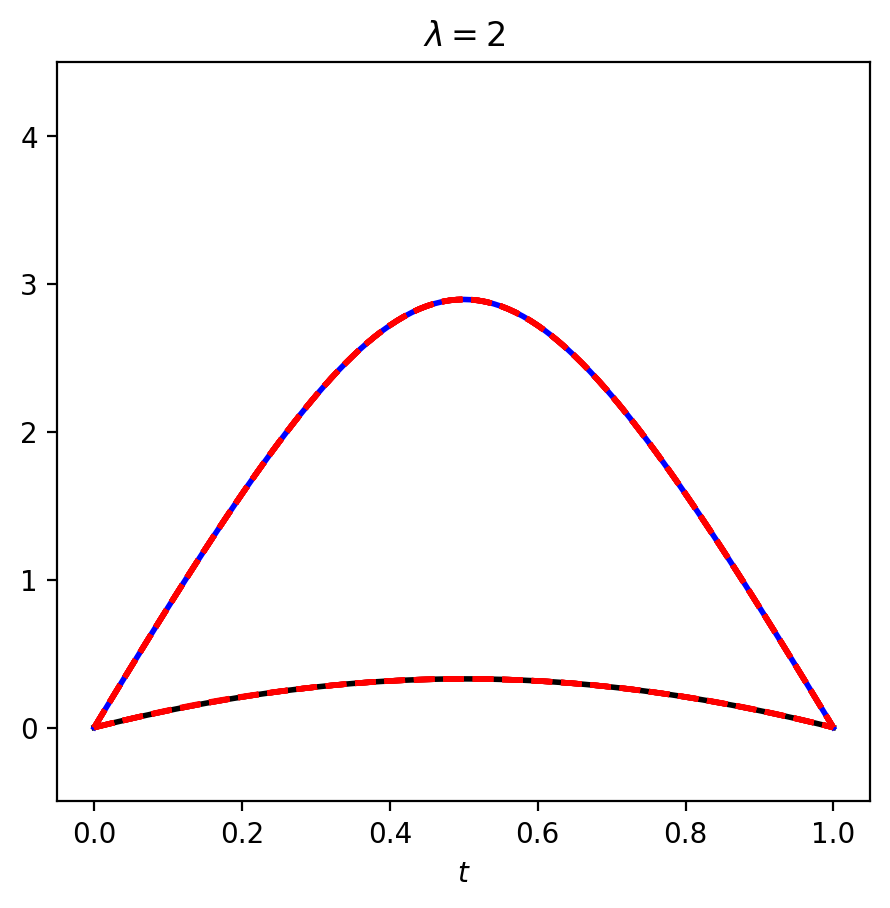}
        \includegraphics[scale=0.25]{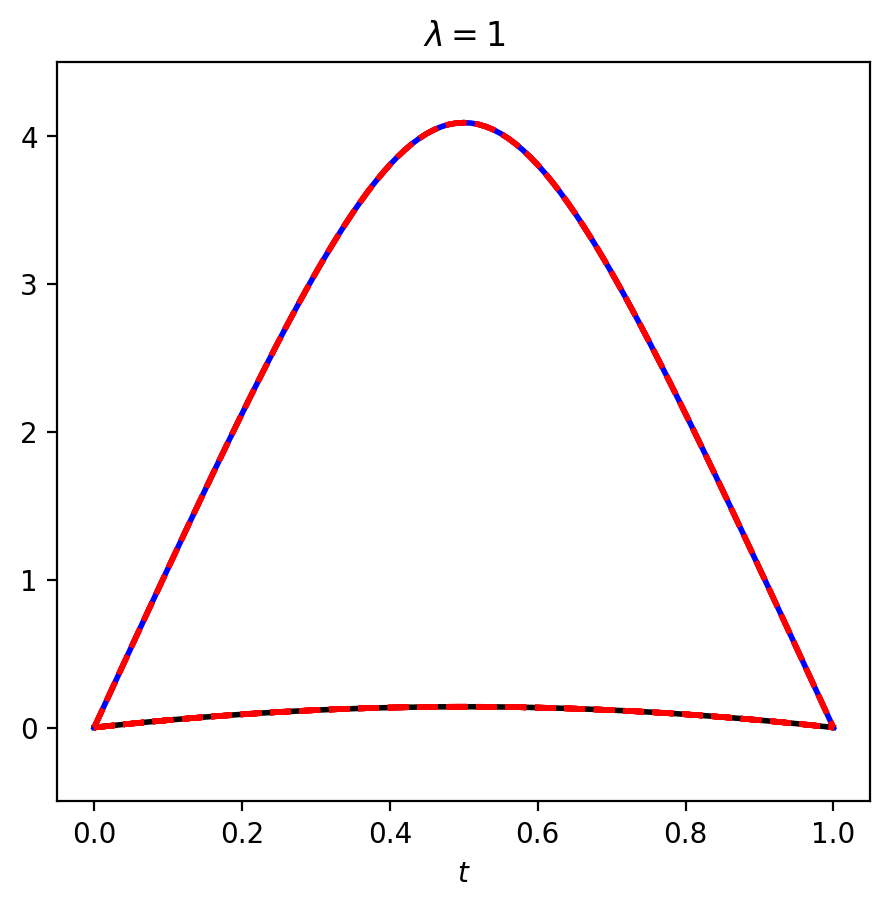}
        \includegraphics[scale=0.25]{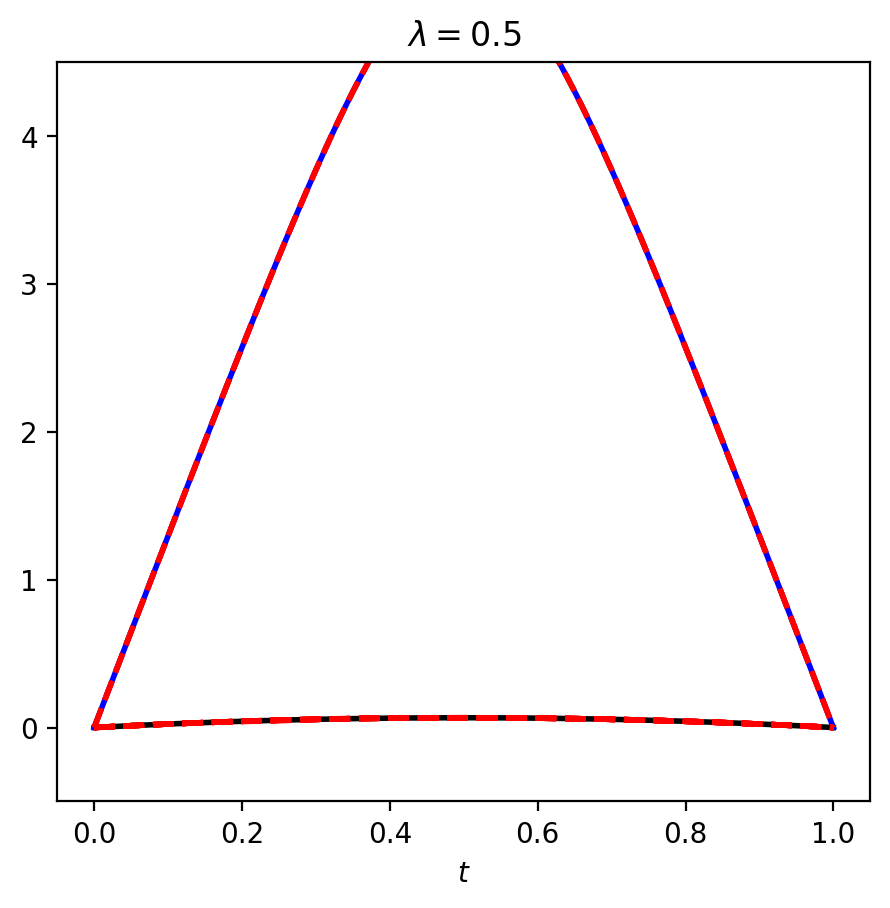}
    }
    \caption{Solving the 1D Bratu problem using PINNs with (a) ten randomly initialized NNs and (b) different values of $\lambda$ (from right to left are $\lambda=3.5, 2, 1, 0.5$). For each value of $\lambda$, ten PINNs are trained starting from these ten randomly initialized NNs displayed in (a).}
    \label{fig:example_1_1}
\end{figure}

We start with the classic 1D Bratu problem described by the following ODE:
\begin{equation}\label{eq:example_1_1}
    \frac{d^2 u}{d t^2} + \lambda \exp(u) = 0, t\in(0,1),
\end{equation}
with boundary conditions $u(0) = u(1) = 0$ and parameter $\lambda>0$. The existence and uniqueness of the solution to this BVP have been extensively discussed in the literature (e.g., \cite{mohsen2014simple, abbasbandy2011lie, khuri2004new, raja2014numerical}). The solutions to the 1D Bratu problem can be expressed as:
\begin{equation}\label{eq:example_1_2}
    u(x; \alpha) = 2 \log \left(\frac{\cosh(\alpha)}{\cosh(\alpha(1-2x))}\right),
\end{equation}
where $\alpha$ satisfies the following equation:
\begin{equation}\label{eq:example_1_3}
    \cosh(\alpha) - \frac{4}{\sqrt{2\lambda}}\alpha = 0.
\end{equation}
We consider $\lambda<3.513830719$, such that \eqref{eq:example_1_3} has two distinct solutions, denoted by $\alpha_1, \alpha_2$ where $\alpha_1 < \alpha_2$. As a result, \eqref{eq:example_1_1} has two solutions, denoted by $u_1(\cdot) := u(\cdot; \alpha_1)$ and $u_2(\cdot) := u(\cdot; \alpha_2)$. Throughout this example, the boundary conditions are explicitly enforced in the NN model using $u_\theta(t) = t(1-t)v_\theta(t)$, where $v_\theta$ represents the NN. 

The capability of PINNs with randomly initialized NNs in identifying multiple solutions to \eqref{eq:example_1_1} is demonstrated in Figure \ref{fig:example_1_1}, where values of $\lambda\in\{3.5, 2, 1, 0.5\}$ are chosen such that there are two distinct solutions. We note that, for each value of $\lambda$, the training of PINNs begins with the same set of ten NNs, which are initialized randomly in advance (displayed in Figure \ref{fig:example_1_1}(a)).

\begin{figure}[ht]
    \centering
    \subfigure[]{
        \includegraphics[scale=0.33]{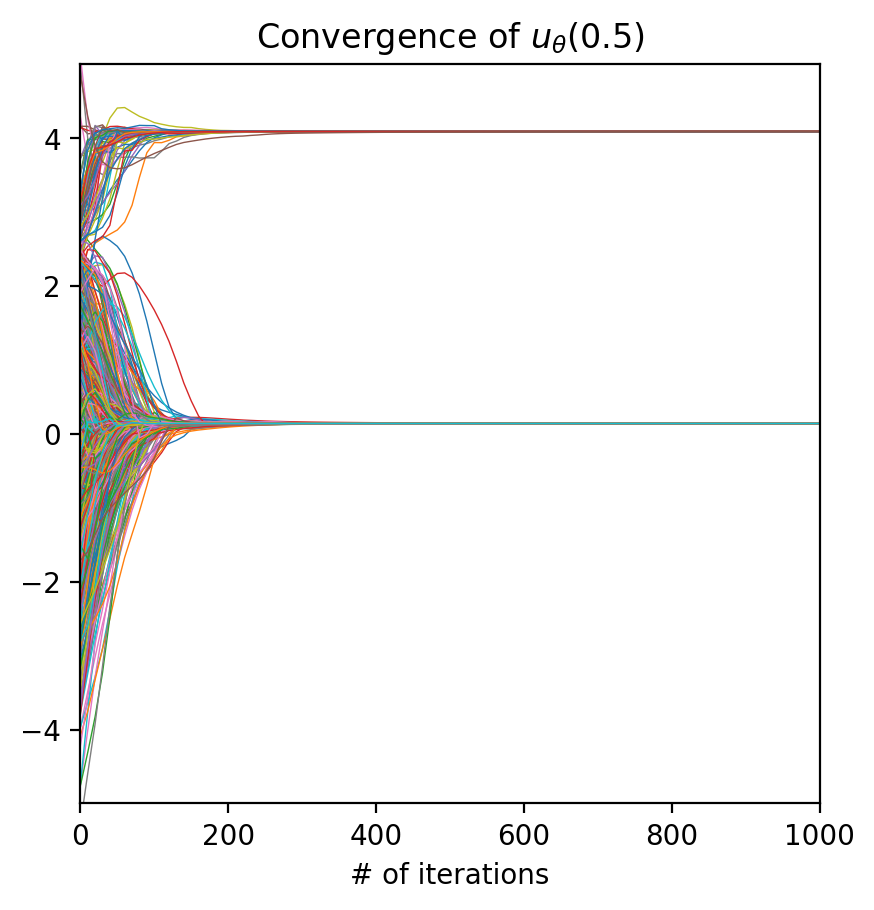}
    }
    \subfigure[]{
        \includegraphics[scale=0.33]{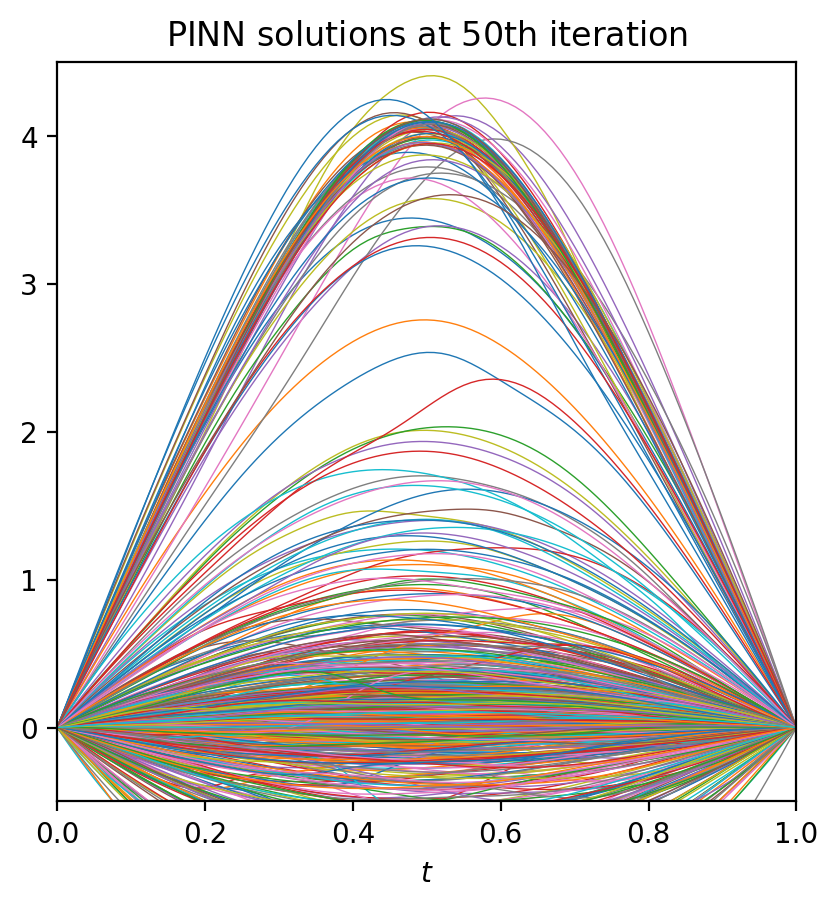}
        \includegraphics[scale=0.33]{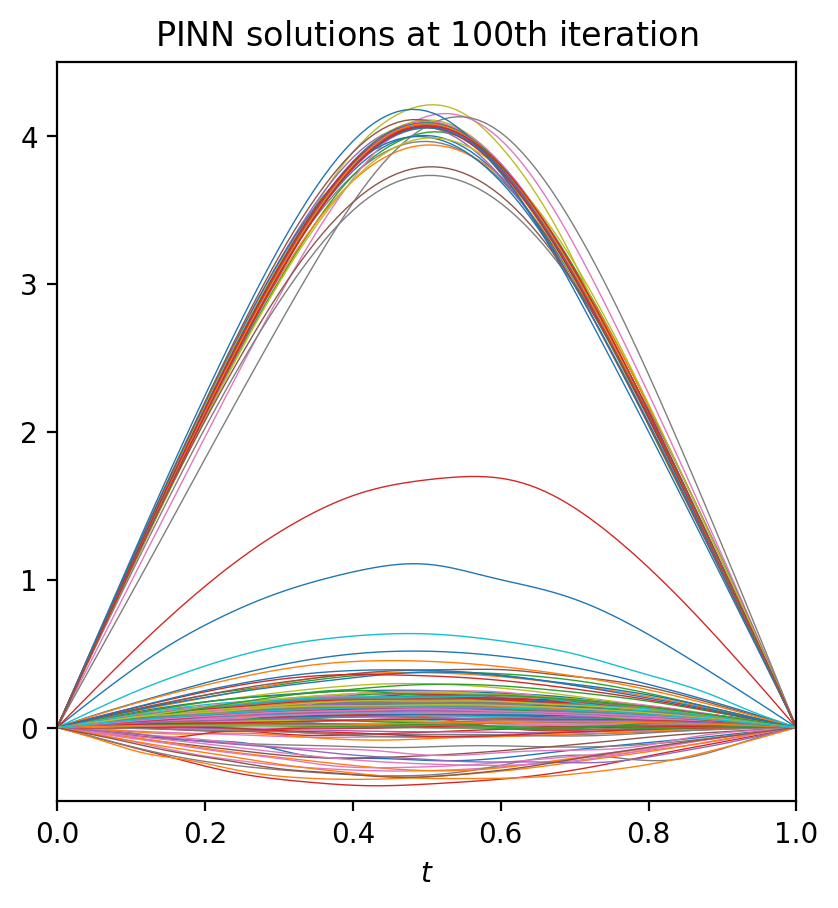}
        \includegraphics[scale=0.33]{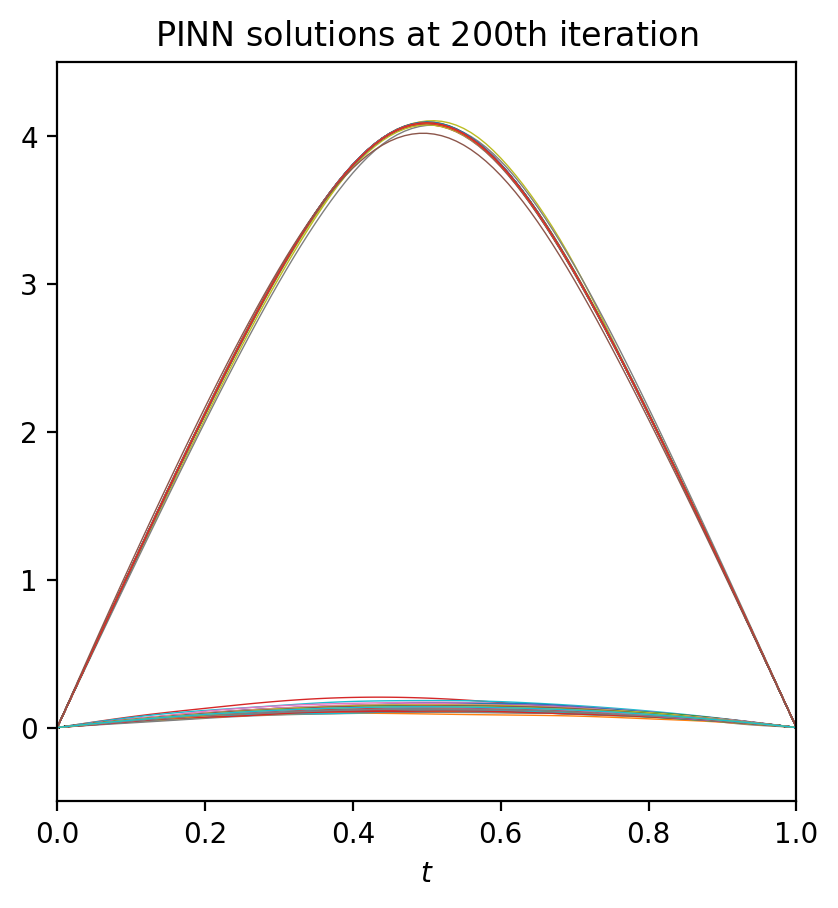}
    }
    \subfigure[]{
        \includegraphics[scale=0.33]{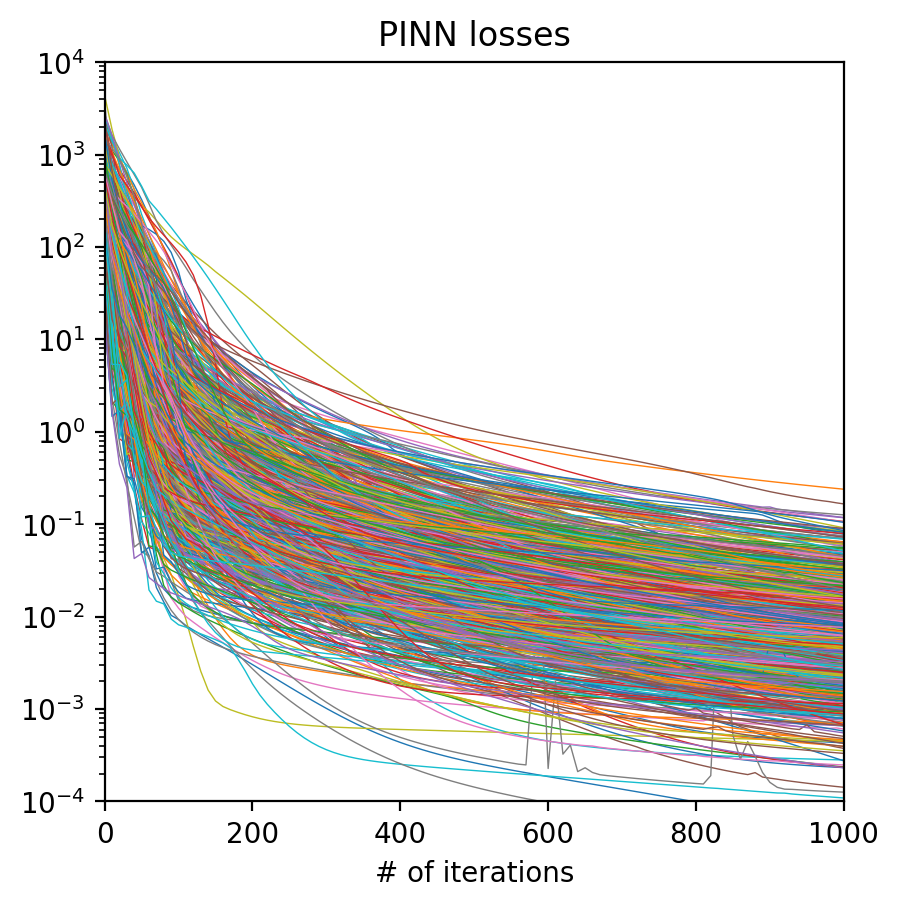}
    }
    \subfigure[]{
        \includegraphics[scale=0.33]{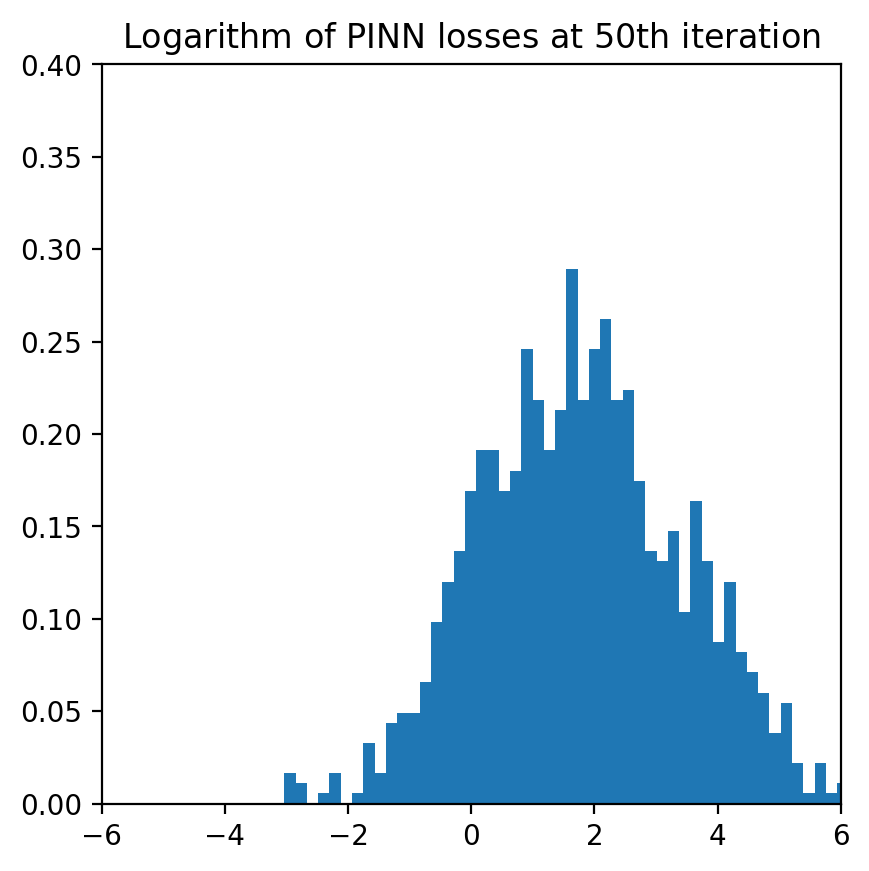}
        \includegraphics[scale=0.33]{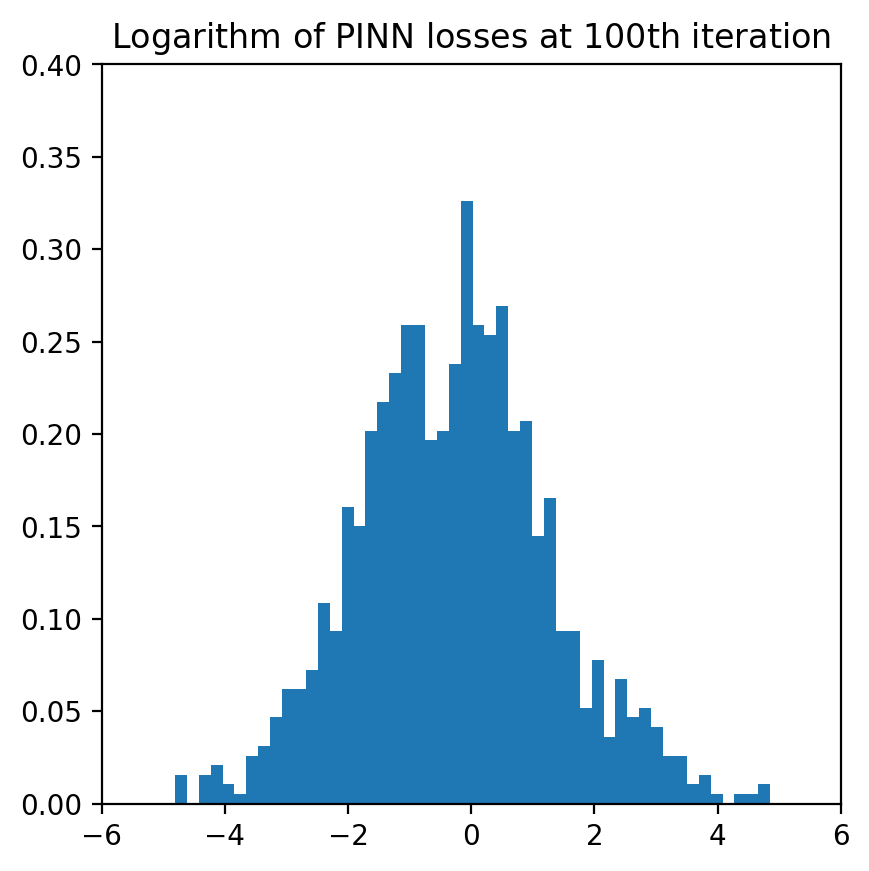}
        \includegraphics[scale=0.33]{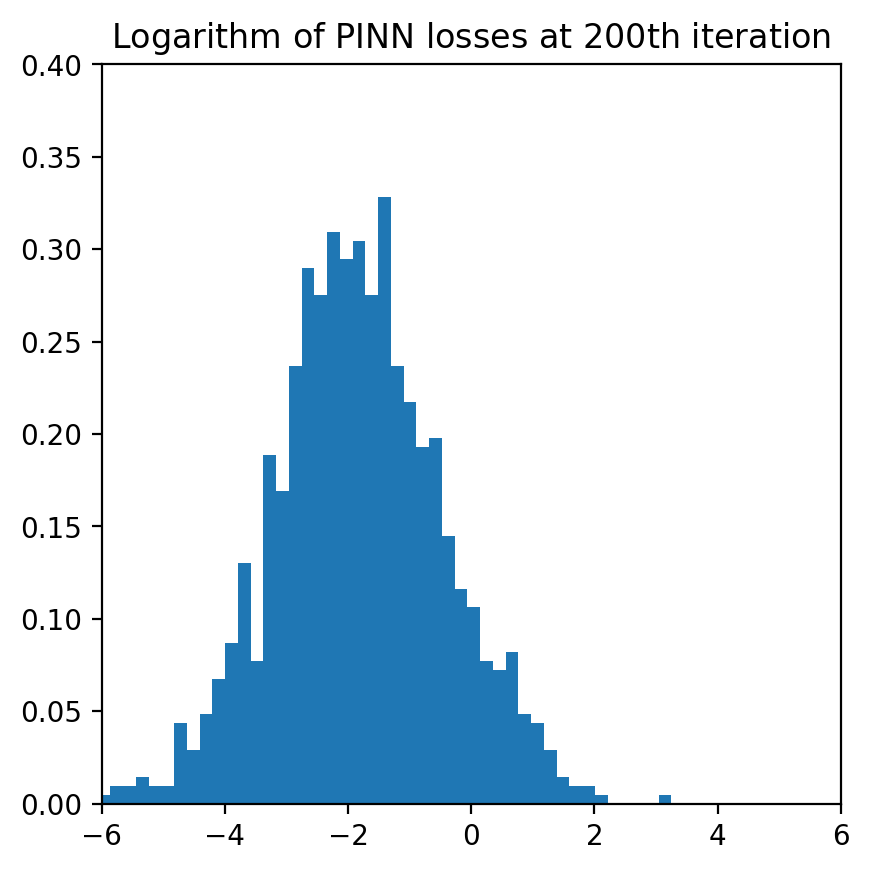}
    }
    \caption{Training of PINNs in solving the 1D Bratu problem with $\lambda=1$ and $1,000$ randomly initialized NNs. In (a), we show the bifurcation of the predicted values of $u(0.5)$ across the $1,000$ NNs. In (b), we present PINN solutions at different iterations. In (c) and (d), we display the PINN loss for these $1,000$ NNs.}
    \label{fig:example_1_2}
\end{figure}

\begin{figure}[ht]
    \centering
    \subfigure[]{
        \includegraphics[scale=0.5]{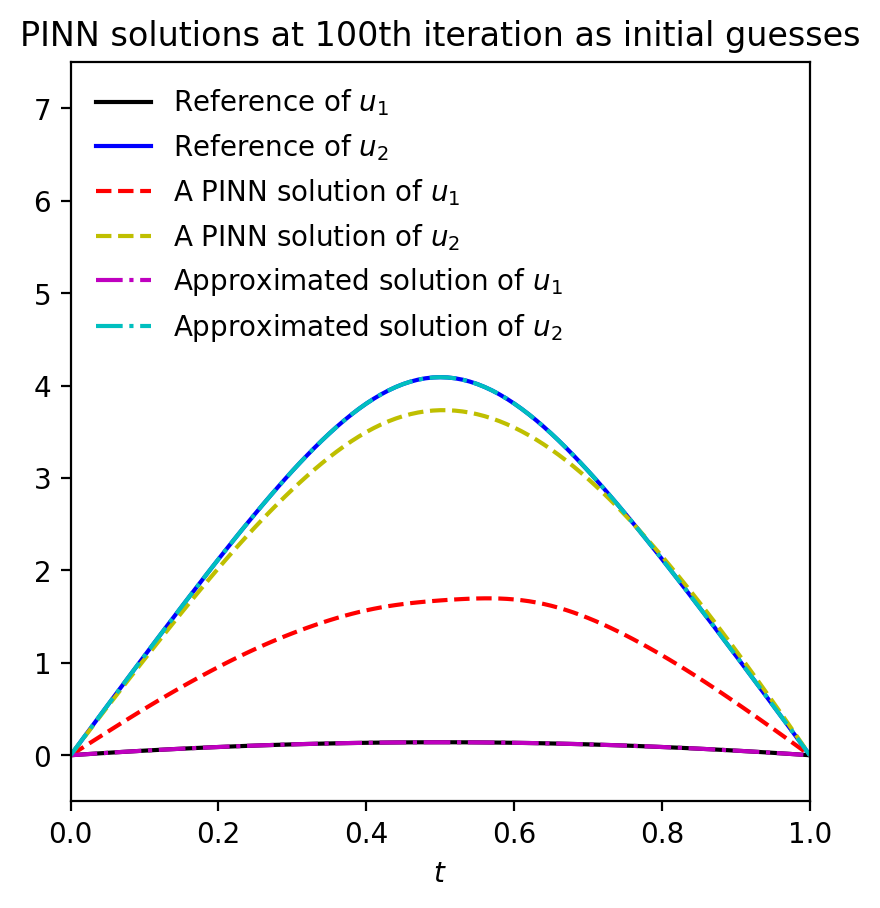}
    }
    \subfigure[]{
        \includegraphics[scale=0.5]{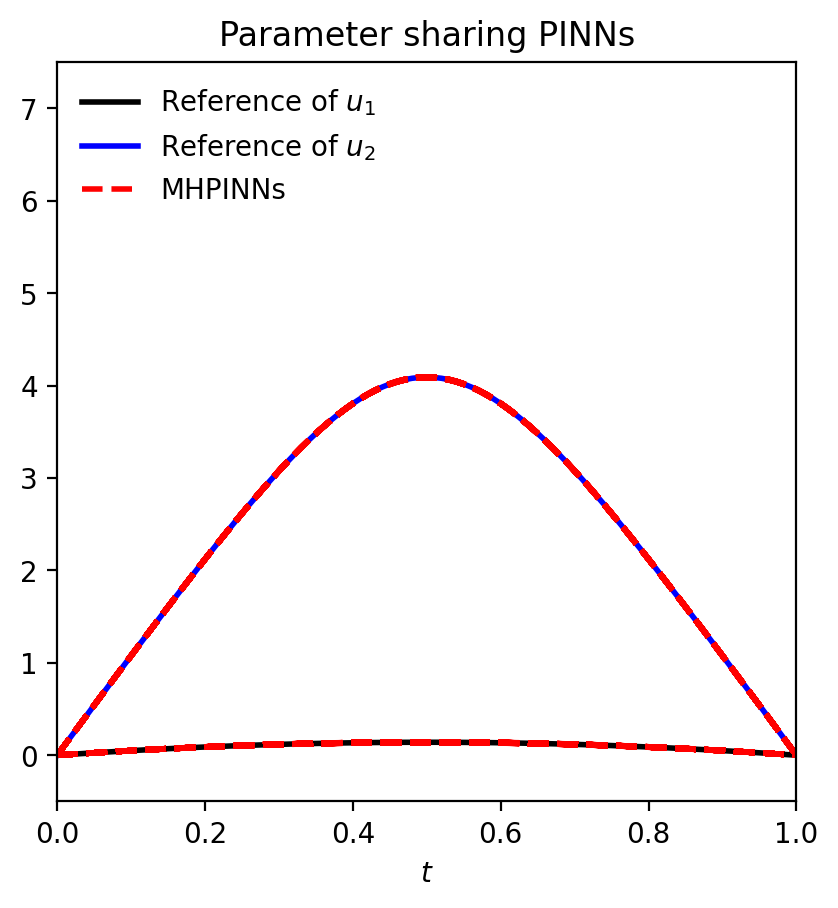}
    }
    \caption{Results for solving the 1D Bratu problem with $\lambda=1$ of (a) using representative PINN solutions at 100th iteration as the initial guesses for a BVP numerical solver and (b) parameter sharing PINNs.}
    \label{fig:example_1_3}
\end{figure}

We then focus on a specific value of $\lambda=1$ and examine the training of PINNs in solving \eqref{eq:example_1_1} with randomly initialized NNs. In particular, we initialize $1,000$ NNs randomly, each of which is then trained following the PINN method, and we investigate the predictions of these NNs for every iteration. We specifically keep track of the predictions at $x=0.5$ where the difference between $u_1$ and $u_2$ reaches maxima. Results are presented in Figure \ref{fig:example_1_2}, in which we present the training dynamics of $u_\theta(0.5)$ and the predictions of these $1,000$ NNs at $50^{th}, 100^{th}$ and $200^{th}$ iterations. As shown, the PINNs method is able to identify two solution patterns quickly and these $1,000$ NNs clearly can be categorized into two classes after a short time of training, e.g., $200$ iterations. While PINNs typically require a large number of iterations to achieve high-precision solution approximations - an issue widely recognized as a significant computational bottleneck in the literature (e.g., \cite{lu2021deepxde, mcclenny2023self, wang2023expert}) -- we observe that they do not require as many iterations to successfully identify solution patterns when addressing solution multiplicity. Specifically for this example, $200$ iterations of training are not enough for solving \eqref{eq:example_1_1} accurately but are sufficient to give us useful estimate for the two distinct solutions, which can be used as initial guesses for conventional numerical solvers, in the format of either discrete values at any mesh or functions, as PINN solutions are given as trained NNs. As demonstration, we use NNs trained for $100$ iterations and follow the procedure presented in Section \ref{sec:2_3}:
\begin{enumerate}
    \item categorize these $1,000$ NNs into two classes by $u_\theta(0.5)<3$,
    \item randomly pick one NN from each class,
    \item use these two NNs as two initial guesses to solve \eqref{eq:example_1_1} using a conventional boundary value problem solver.
\end{enumerate}
In this example, we use MATLAB \textit{bvp4c} \cite{kierzenka2001bvp, shampine2000solving}, which is based on a finite difference method (FDM), and to demonstrate the effectiveness of our approach, we deliberately pick the worst PINN solutions at $100^{th}$ iteration as the initial guesses for the MATLAB solver. The results of the BVP solver with PINN solutions as the initial guesses are displayed in Figure \ref{fig:example_1_3}(a), from which we can see that PINNs are able to provide useful initial guesses, which lead to accurate approximation of both solutions to \eqref{eq:example_1_1}.

\begin{table}[h]
    \footnotesize
    \centering
    \begin{tabular}{c|c|c|c|c|c|c|c}
    \hline
    \hline
    NN architecture & $\mathcal{N}(0, 0.5^2)$ & $\mathcal{N}(0, 1^2)$ & 
    $\mathcal{N}(0, 1.5^2)$ & $\mathcal{U}[-1, 1)$ &
    $\mathcal{U}[-2, 2)$ & $\mathcal{U}[-2.5, 2.5)$ & $\mathcal{U}[-3, 3)$\\
    \hline 
    $[1, 50, 50, 1]$ & 0.000 & 0.062 & 0.186 & 0.001 & 0.109 & 0.179 & 0.240\\
    \hline
    $[1, 50, 50, 50, 1]$ & 0.000 & 0.074 & 0.224 & 0.003 & 0.112 & 0.205 & 0.335\\
    \hline
    $[1, 100, 100, 1]$ & 0.004 & 0.154 & 0.302 & 0.018 & 0.222 & 0.290 & 0.347\\
    \hline
    \hline
    \end{tabular}
    \caption{Ratio of PINN solutions (out of $10,000$) whose predictions at $x=0.5$ are larger than $3$ after $20,000$ iterations of training. Qualitative results are presented in Figure \ref{fig:example_1_4}} 
    \label{tab:example_1}
\end{table}

We further conduct an ablation study of PINNs for solving \eqref{eq:example_1_1} with different NN architectures and initialization methods, with results presented in Table \ref{tab:example_1} and Figure \ref{fig:example_1_4}. Specifically, we consider:
\begin{enumerate}
    \item two initialization methods, the random normal initialization with mean zero and different standard deviations (denoted as $\mathcal{N}(0, \sigma^2)$ where $\sigma$ represents the standard deviation) and the random uniform initialization with mean zero and different upper/lower bounds (denoted as $\mathcal{U}[-B, B)$ where $B>0$ represents the bound),
    \item three NN architectures, two hidden layers with $50$ neurons for each layer (denoted as $[1, 50, 50, 1]$), three hidden layers with $50$ neurons ($[1, 50, 50, 50, 1]$), and two hidden layers with $100$ neurons ($[1, 100, 100, 1]$), 
\end{enumerate}
and train $10,000$ PINNs independently for a fixed number of iterations ($20,000$).
We observe that while a larger variance in the initialization method or a wider and deeper architecture consistently leads to a higher ratio of identified $u_2$, NNs initialized this way are empirically more challenging to train, exhibiting issues such as slow convergence and a higher likelihood of getting stuck in local minima. Conversely, NNs initialized with relatively small weight and bias values are easier to train but tend to converge to $u_1$. 

Apart from randomly initialized NNs being trained independently, we also consider the multi-head architecture \cite{zou2023hydra} in solving \eqref{eq:example_1_1}. In this setup, NNs share the same body, which is constructed as the first hidden layer, and are associated independent heads, which referred to the last two layers, i.e. the second hidden layer and the output linear layer. We randomly initialize the body and $1,000$ heads and train them simultaneously following the multi-head PINNs method \cite{zou2023hydra}. We note that this setup prohibits parallel computation due to parameter sharing but can still be vectorized. As presented in Figure \ref{fig:example_1_3}(b), we observe that multiple solutions can still be recovered from this multi-head structure. 

\subsection{1D boundary layer problem}\label{sec:3_2}

\begin{figure}[ht]
    \centering
    \subfigure[]{
        \includegraphics[scale=0.33]{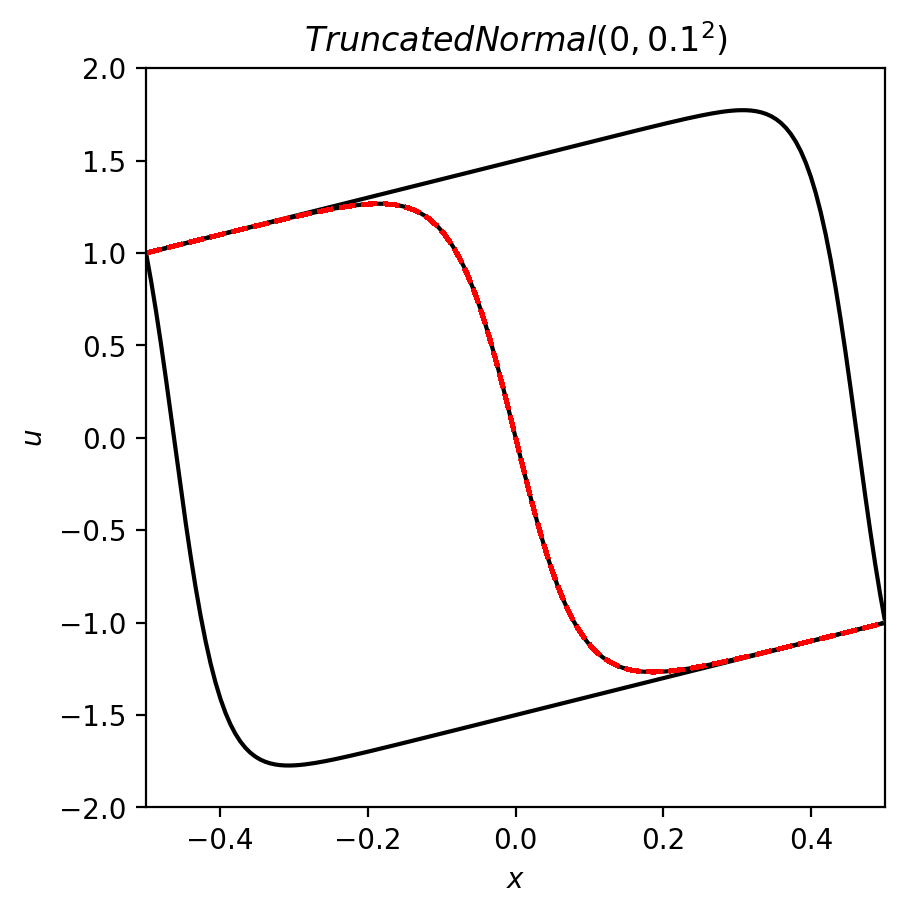}
        \includegraphics[scale=0.33]{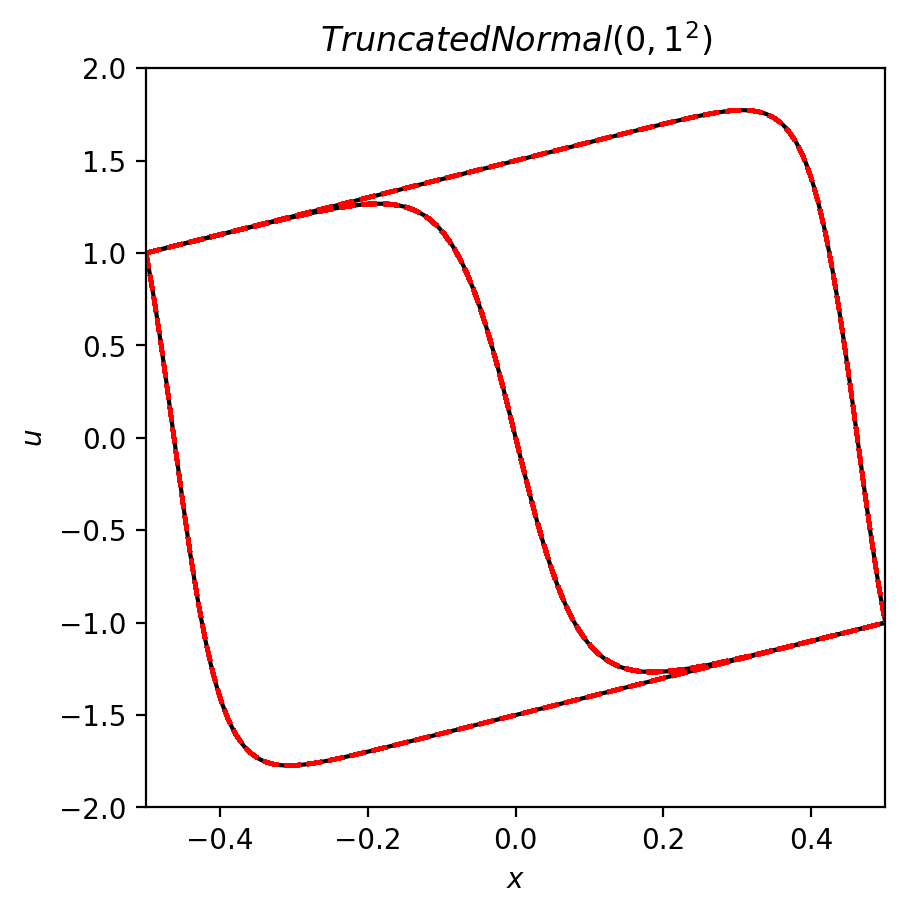}
    }
    \subfigure[]{
        \includegraphics[scale=0.33]{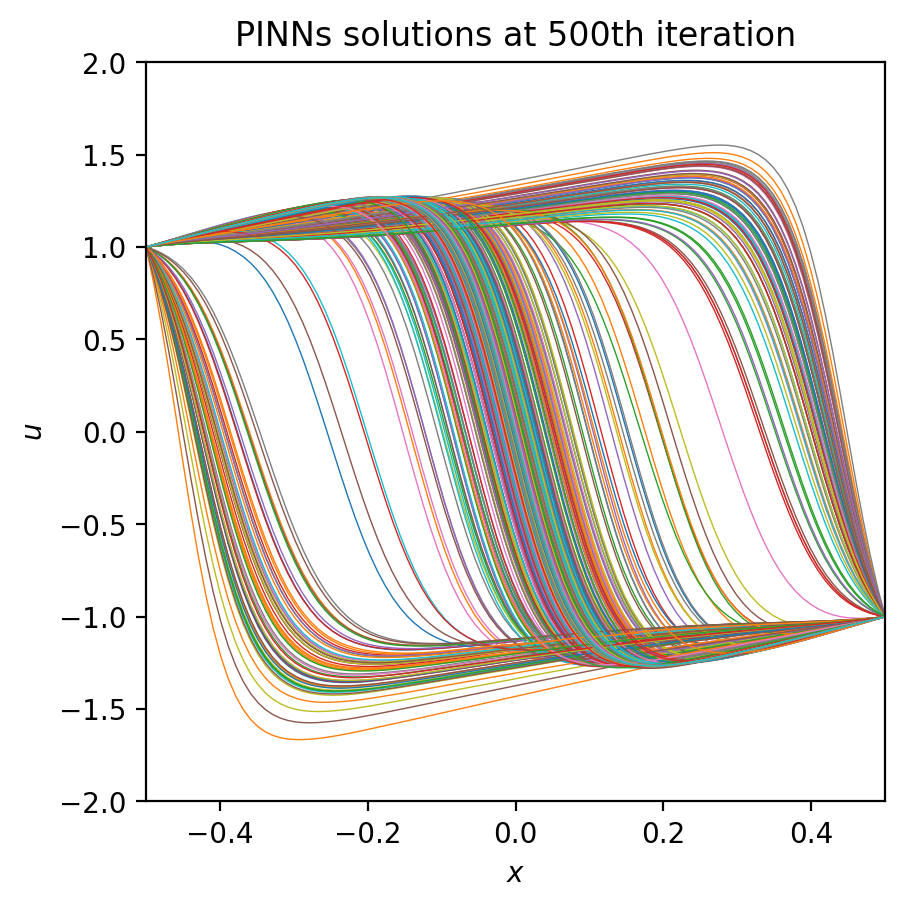}
        \includegraphics[scale=0.33]{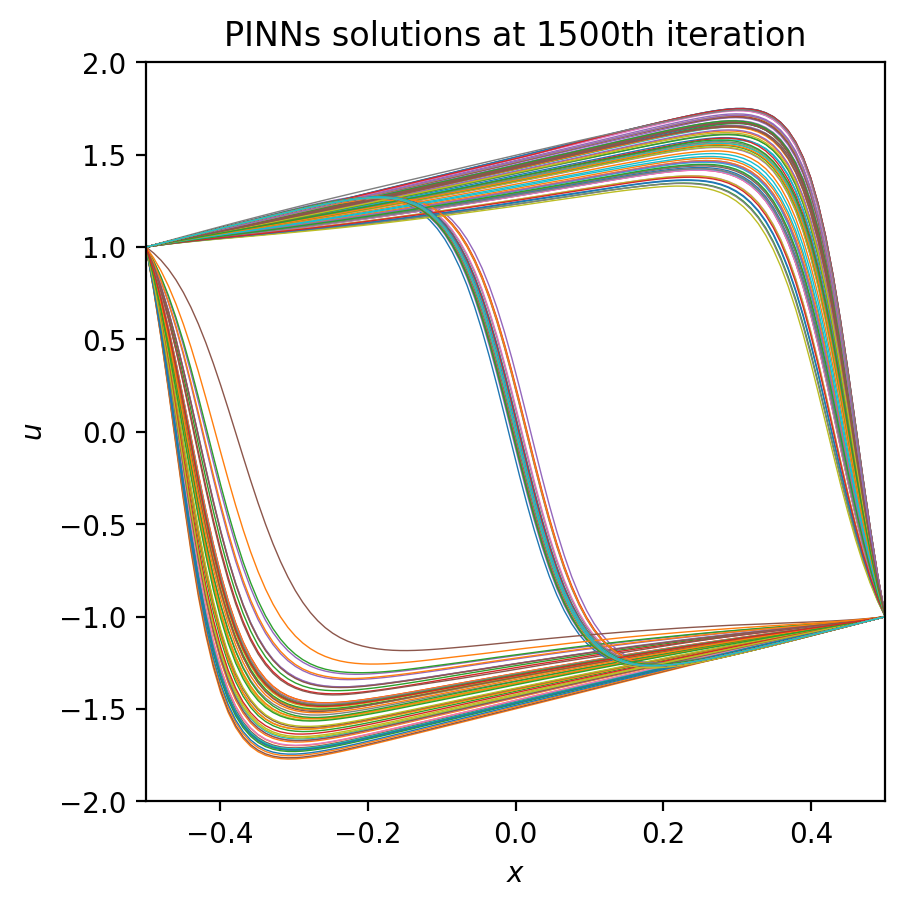}
    }
    \caption{Solving the boundary layer problem defined in \eqref{eq:boundary_layer} using PINNs with $1,000$ randomly initialized NNs. In (a), {\color{red}\textbf{red}} dashed lines are $1,000$ PINN solutions and \textbf{black} solid lines are the reference solutions. In (b), PINN solutions at different iterations are presented.}
    \label{fig:example_2_1}
\end{figure}

\begin{figure}[ht]
    \centering
    \subfigure[]{
        \includegraphics[scale=0.5]{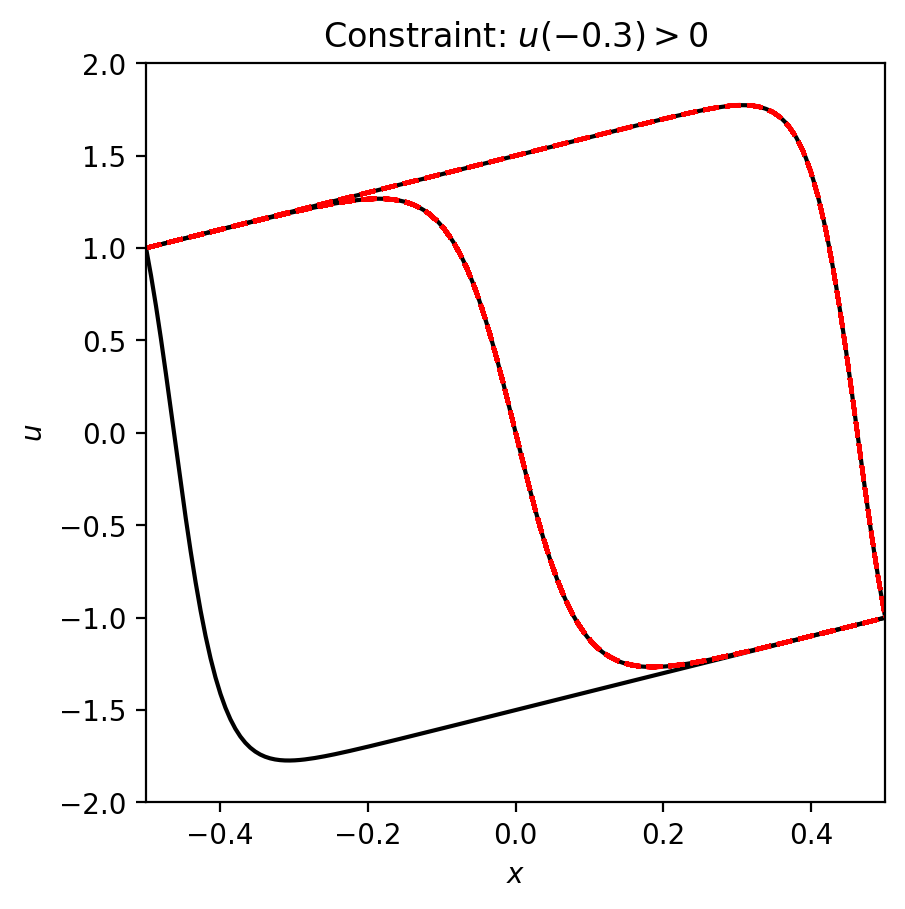}
    }
    \subfigure[]{
        \includegraphics[scale=0.5]{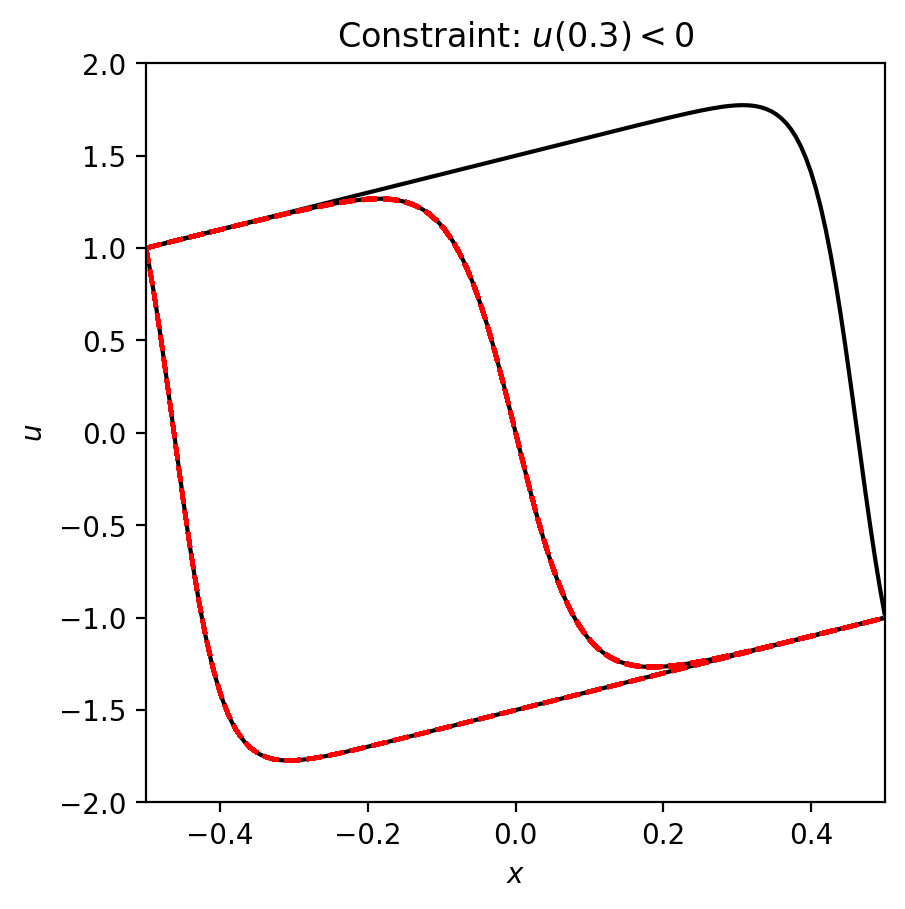}
    }
    \caption{Learning and discovering solutions to \eqref{eq:boundary_layer} that satisfy certain constraints, i.e. $u(-0.3)>0$ in (a) and $u(0.3)<0$ in (b), using PINNs with ($1,000$) randomly initialized NNs and deep ensemble. {\color{red}\textbf{Red}} dashed lines are in $1,000$ PINN solutions. The reference three solutions to \eqref{eq:boundary_layer} without any constraint are in \textbf{black} solid lines for reference.}
    \label{fig:example_2_2}
\end{figure}

In this example, we examine a boundary layer problem governed by the following DE:
\begin{subequations}\label{eq:boundary_layer}
    \begin{align}
        & \epsilon\frac{d^2 u}{dx^2} - u \frac{du}{dx} + u  = 0, x \in (-0.5, 0.5),\\
        & u(-0.5) = 1, u(0.5) = -1,
    \end{align}
\end{subequations}
where $\epsilon$ is a positive constant. While it was previously argued in \cite{holmes2012introduction} that \eqref{eq:boundary_layer} has a unique solution, a recent study \cite{clark2023surprises} revealed that for certain values of $\epsilon$, the equation actually admits three distinct solutions: one featuring a boundary layer at $x=-0.5$, another with an interior layer near the center, and a third exhibiting a boundary layer at $x=0.5$. The three reference solutions are obtained by using the code associated with \cite{clark2023surprises}.

Here, we apply the proposed approach to numerically identify these solutions for $\epsilon = 0.06$. Specifically, we explicitly enforce the boundary conditions into the NN model using $u_\theta(x) = (0.5 + x) (0.5 - x) v_\theta(x) - 2x$, where $v_\theta(x)$ represents the NN. We employ the random truncated normal initialization method with mean zero and two different standard deviations, $0.1$ and $1$, and train $1,000$ NNs independently for $20,000$ iterations. As shown in Figure \ref{fig:example_2_1}(a), the choice of initialization significantly impacts solution multiplicity, enabling us to learn all three solutions to \eqref{eq:boundary_layer}. Furthermore, Figure \ref{fig:example_2_1}(b) illustrates that the bifurcation emerges early in the training, requiring relatively fewer iterations.

Additionally, we test the ensemble PINNs method in handling additional constraints when identifying multiple solutions. Specifically, we assume a priori that we seek solutions either $u(-0.3)>0$ or $u(0.3)<0$. These constraints are incorporated into the PINN loss function as soft penalties. For instance, for the constraint $u(-0.3)>0$, the new PINN loss function (denoted as $\mathcal{L}_{new}(\theta)$) is defined as:
\begin{equation}\label{eq:loss_new}
     \mathcal{L}_{new}(\theta) := \mathcal{L}(\theta) + \frac{\text{stop\_gradient}\{\mathcal{L}(\theta)\}}{M}\sum_{k=1}^M\exp(-u_{\theta_k}(-0.3)),
\end{equation}
where $\text{stop\_gradient}\{\mathcal{L}(\theta)\}$ means the backpropagation through this term is prohibited, allowing the current loss function value to serve as a weighting coefficient for the penalty term. A similar approach was employed in \cite{wang2023asymptotic} to mitigate non-smoothness in finding the self-similar solution of the Boussinesq equations with PINNs. Results of ensemble PINNs with additional constraints are presented in Figure \ref{fig:example_2_2}, from which we can see that this method effectively discovers multiple solutions while adhering to the imposed constraints.

\subsection{1D steady-state reaction-diffusion equation}\label{sec:3_3}

\begin{figure}[ht!]
    \centering
    \subfigure[Case (A): $w=6$.]{
        \includegraphics[scale=0.33]{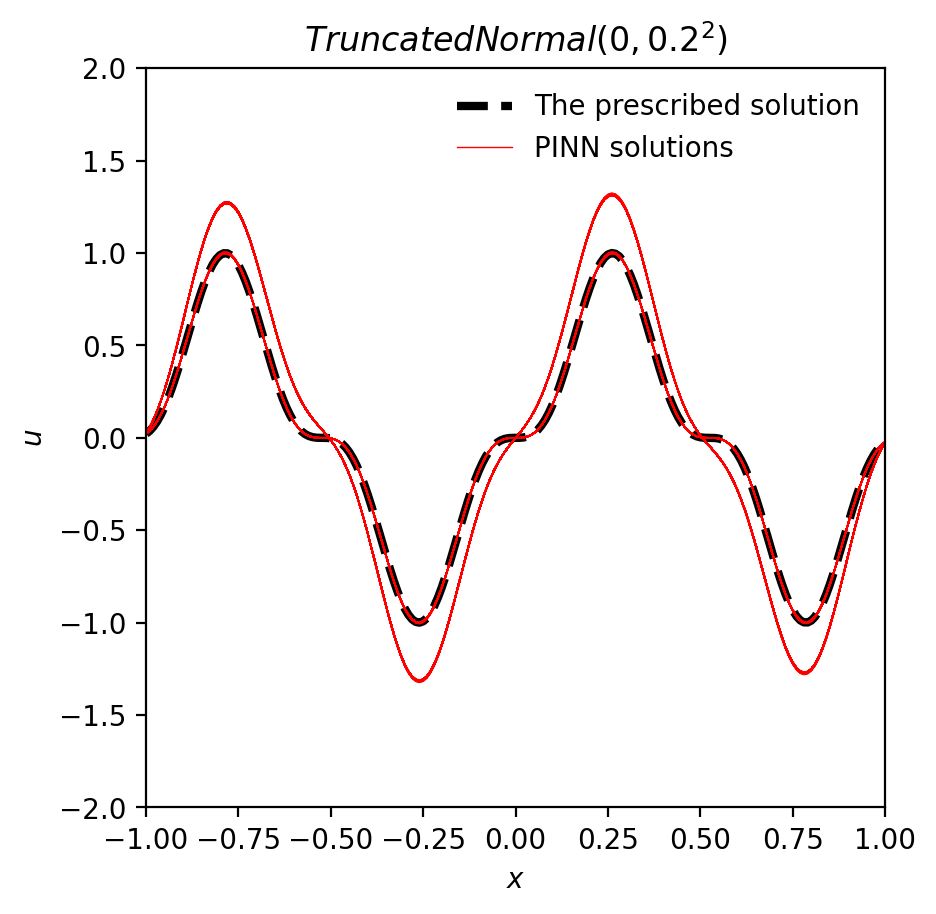}
        \includegraphics[scale=0.33]{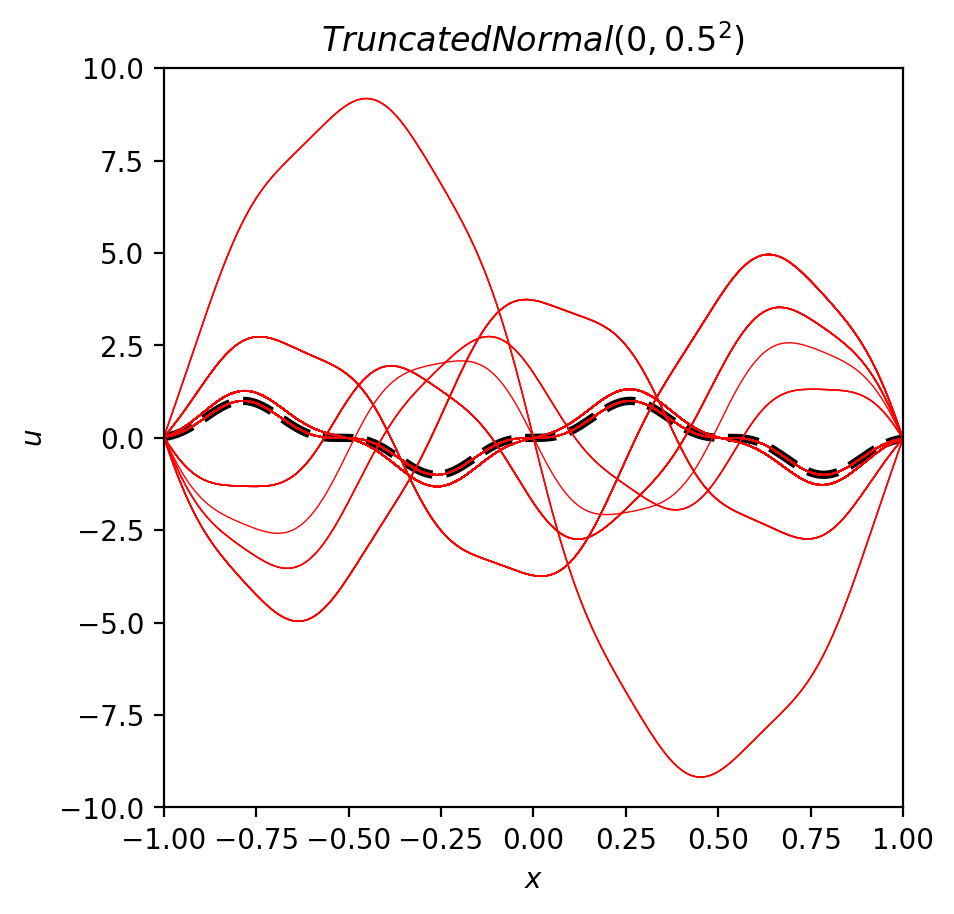}
        \includegraphics[scale=0.33]{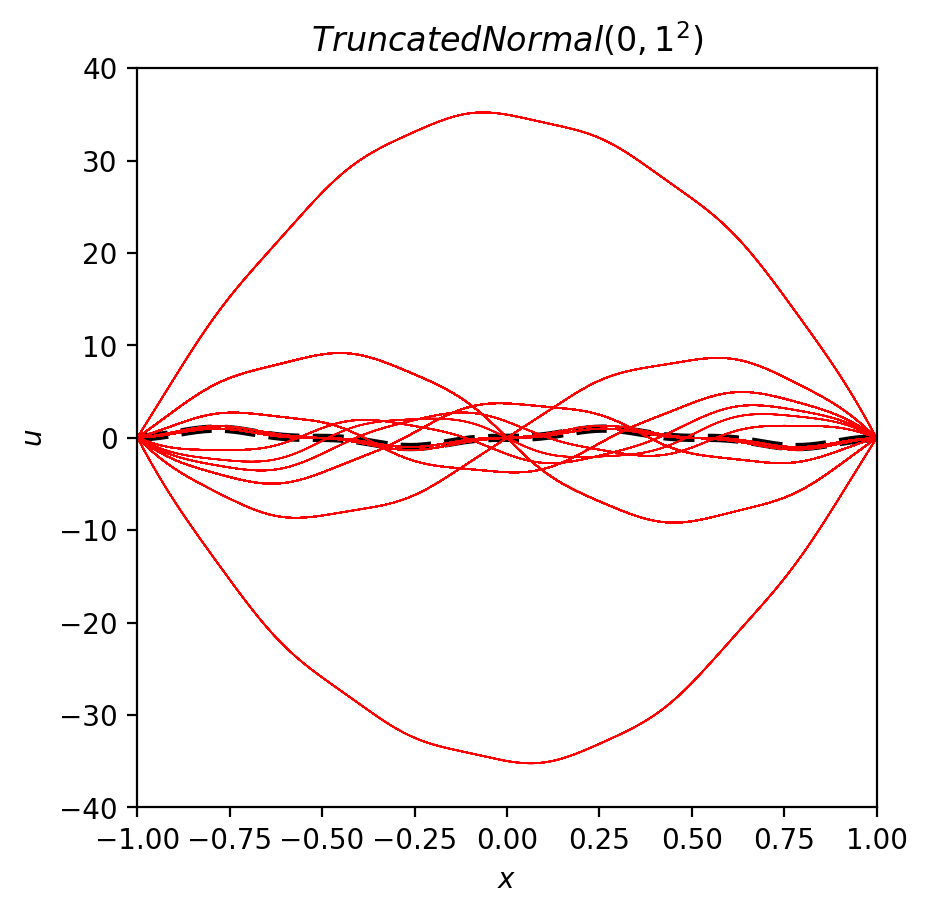}
        \includegraphics[scale=0.33]{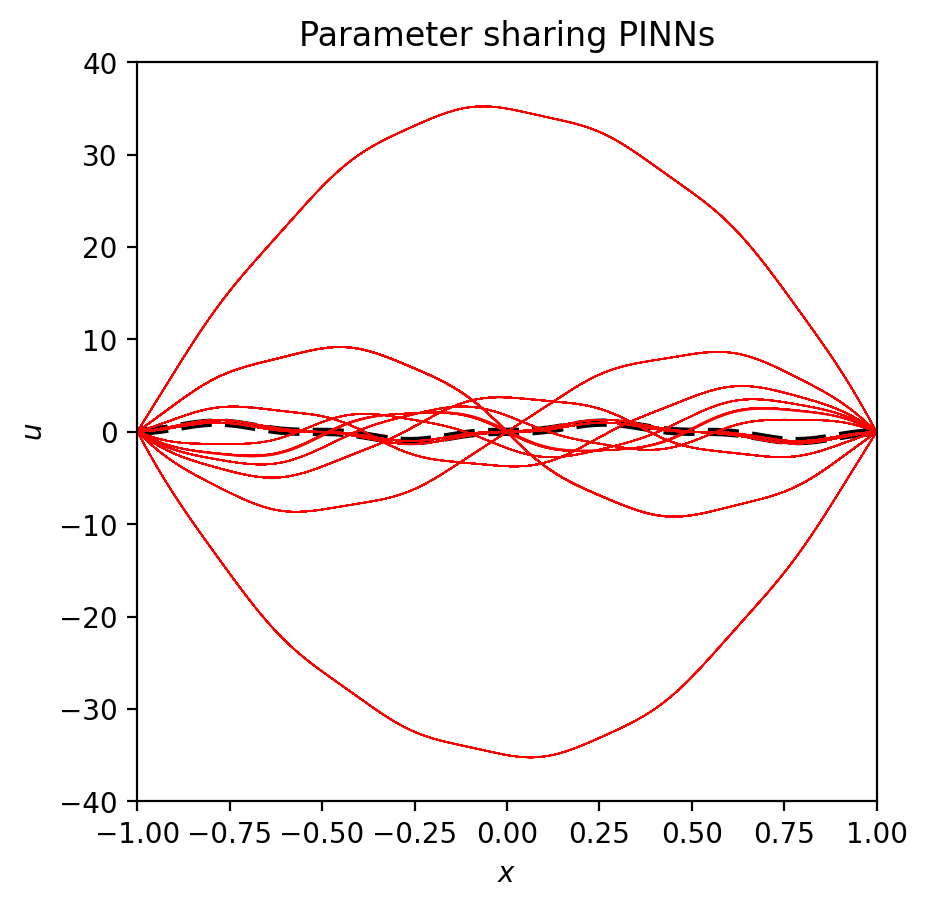}
    }
    \subfigure[Case (B): $w=10$.]{
        \includegraphics[scale=0.33]{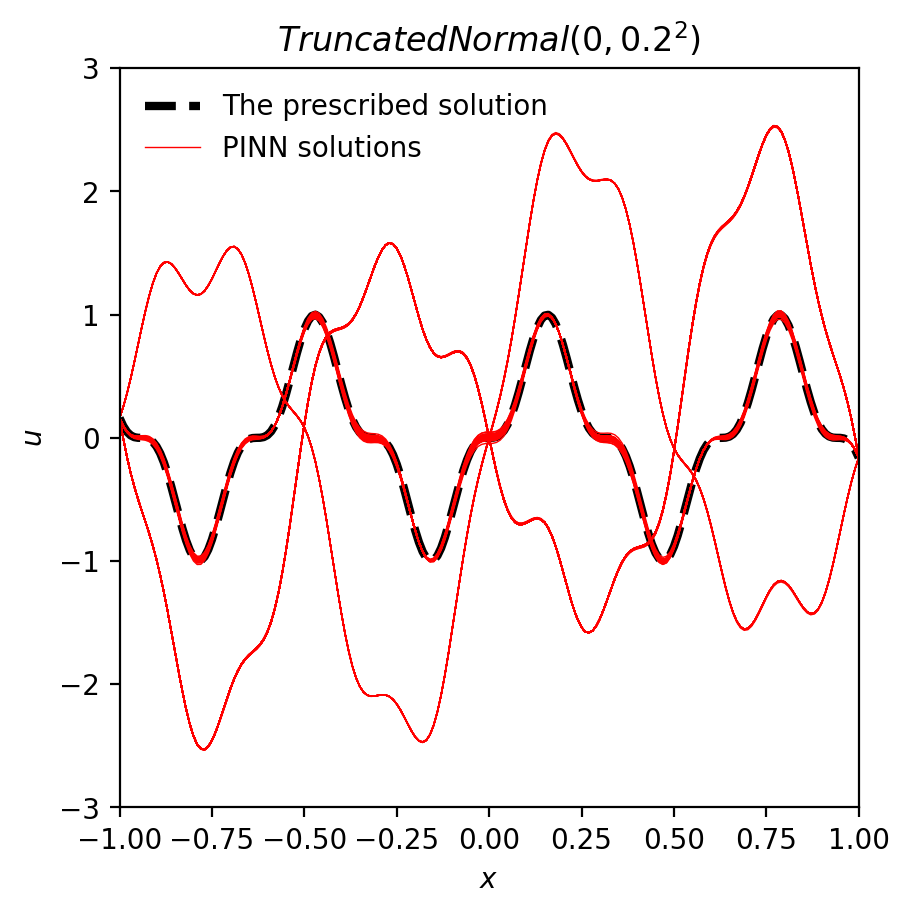}
        \includegraphics[scale=0.33]{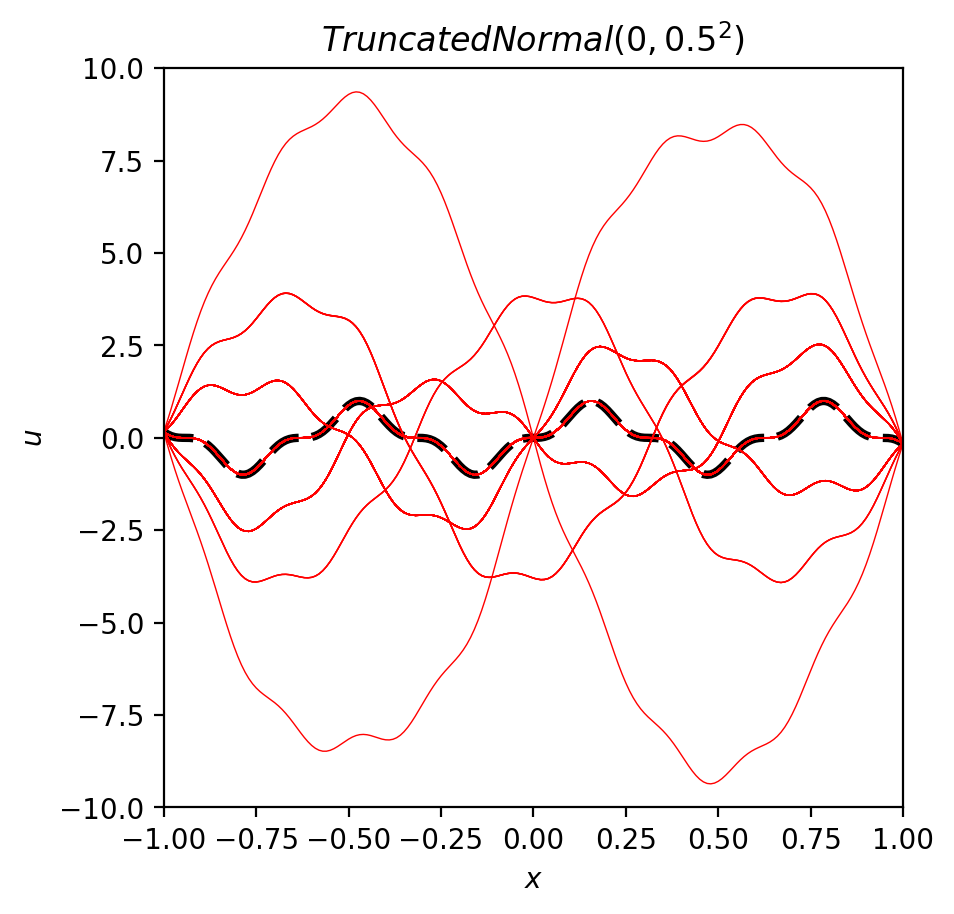}
        \includegraphics[scale=0.33]{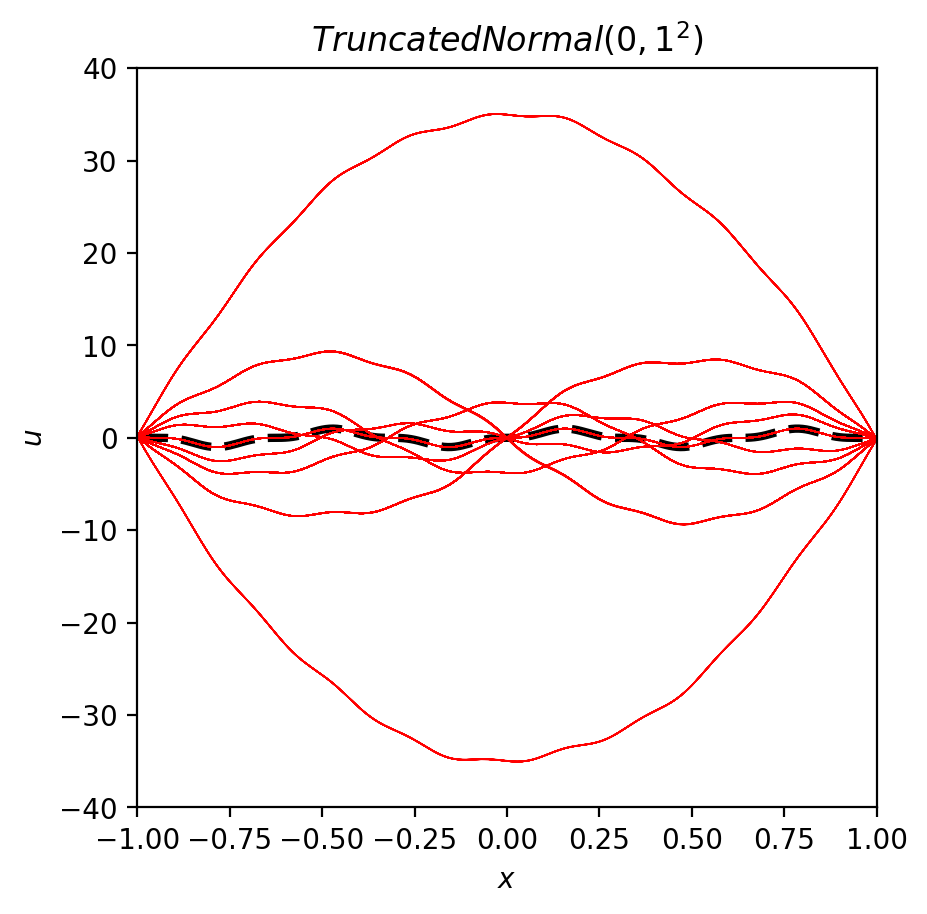}
        \includegraphics[scale=0.33]{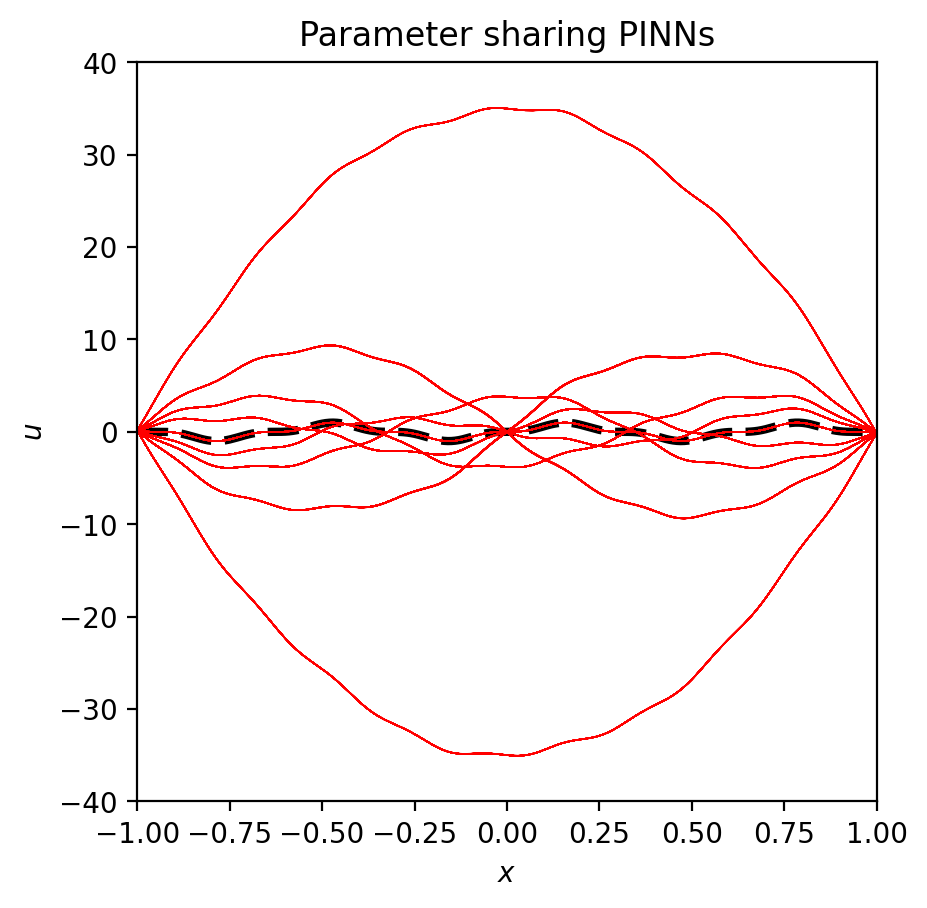}
    }
    \caption{PINN solutions to \eqref{eq:reaction_diffusion} with $w=6, 10$. The first three results (on the \textbf{left}) are obtained by training independently $1,000$ randomly initialized NNs (ensemble PINNs) using different initialization methods, specifically the truncated normal initialization, denoted as $TruncatedNormal(\mu, \sigma^2)$, where $\mu$ is the mean and $\sigma$ is the standard deviation. The results on the \textbf{right} are obtained by training parameter sharing PINNs (a multi-head architecture with $1,000$ heads). }
    \label{fig:example_3_1}
\end{figure}

\begin{figure}[ht!]
    \centering
    \subfigure[Case (A): $w=6$.]{
        \includegraphics[scale=0.45]{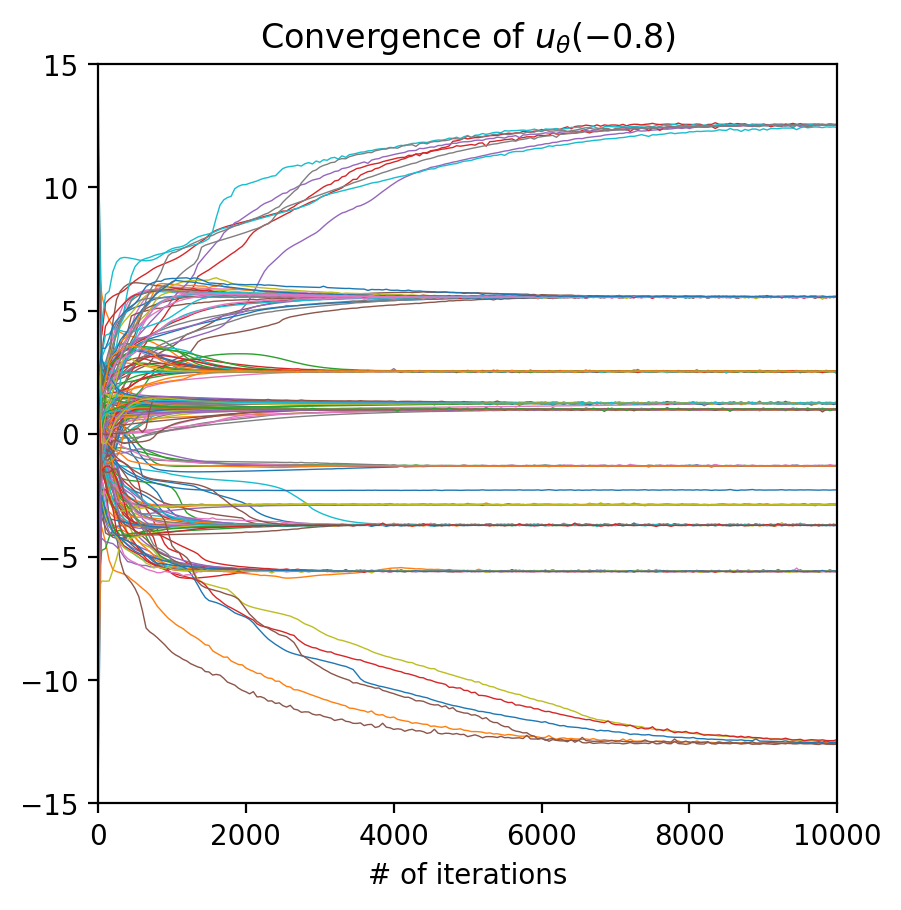}
        \includegraphics[scale=0.45]{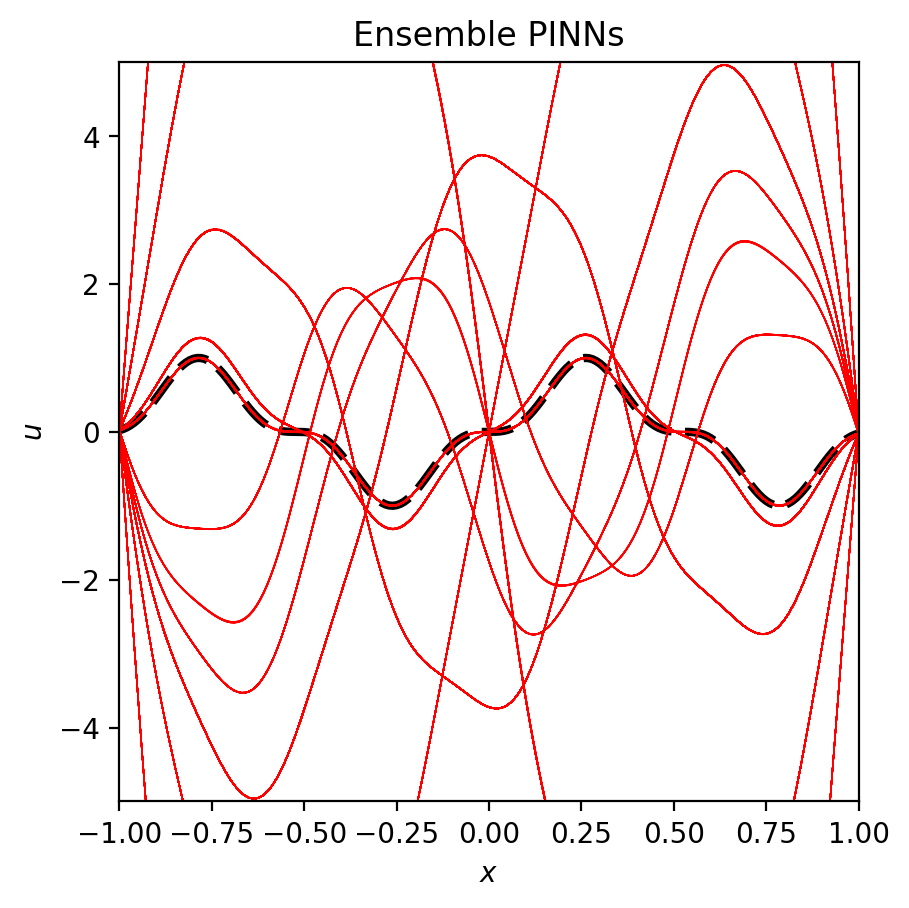}
        \includegraphics[scale=0.45]{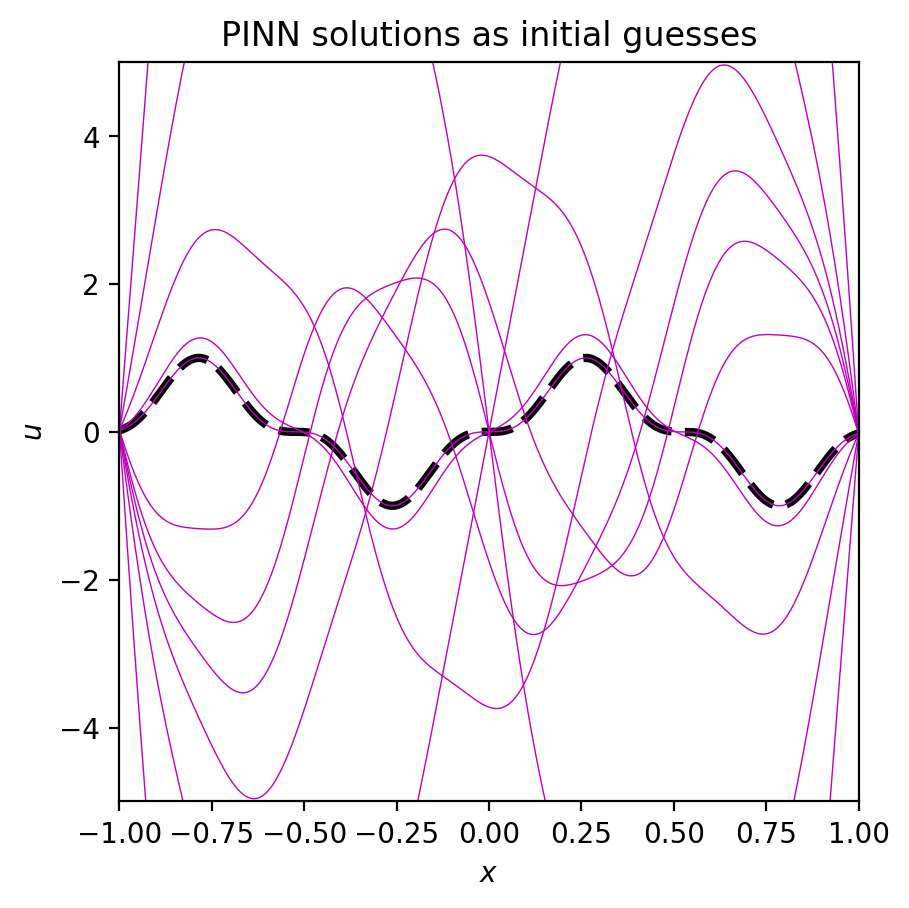}
    }
    \subfigure[Case (B): $w=10$.]{
        \includegraphics[scale=0.45]{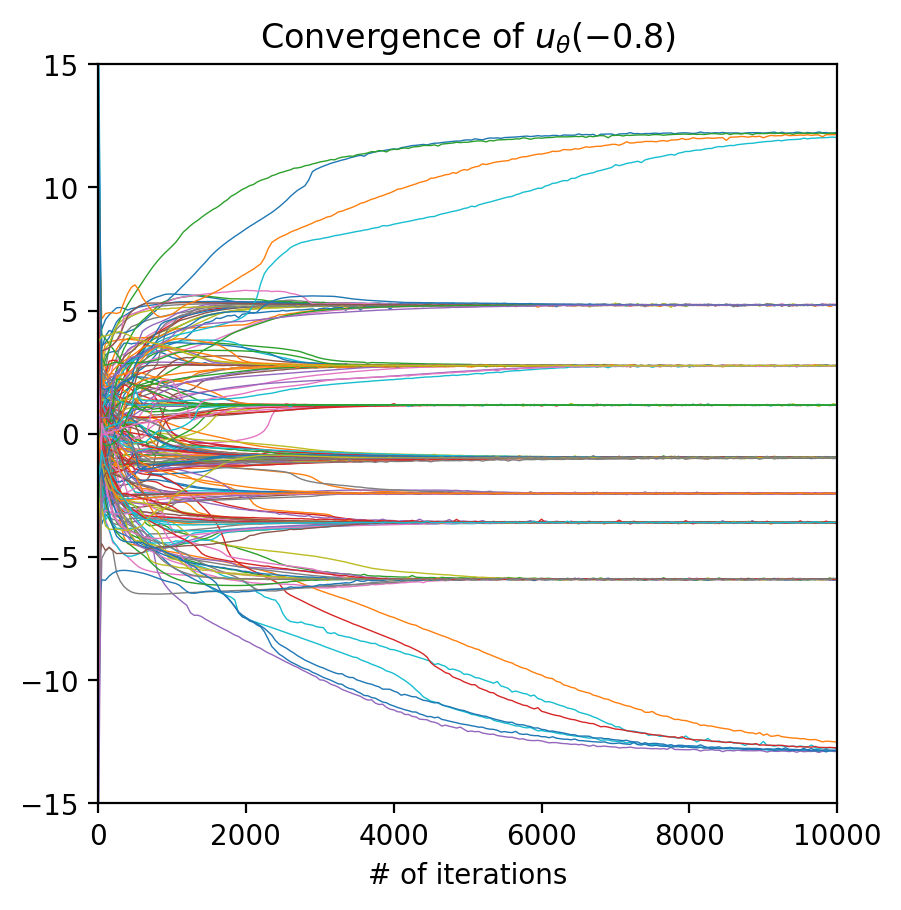}
        \includegraphics[scale=0.45]{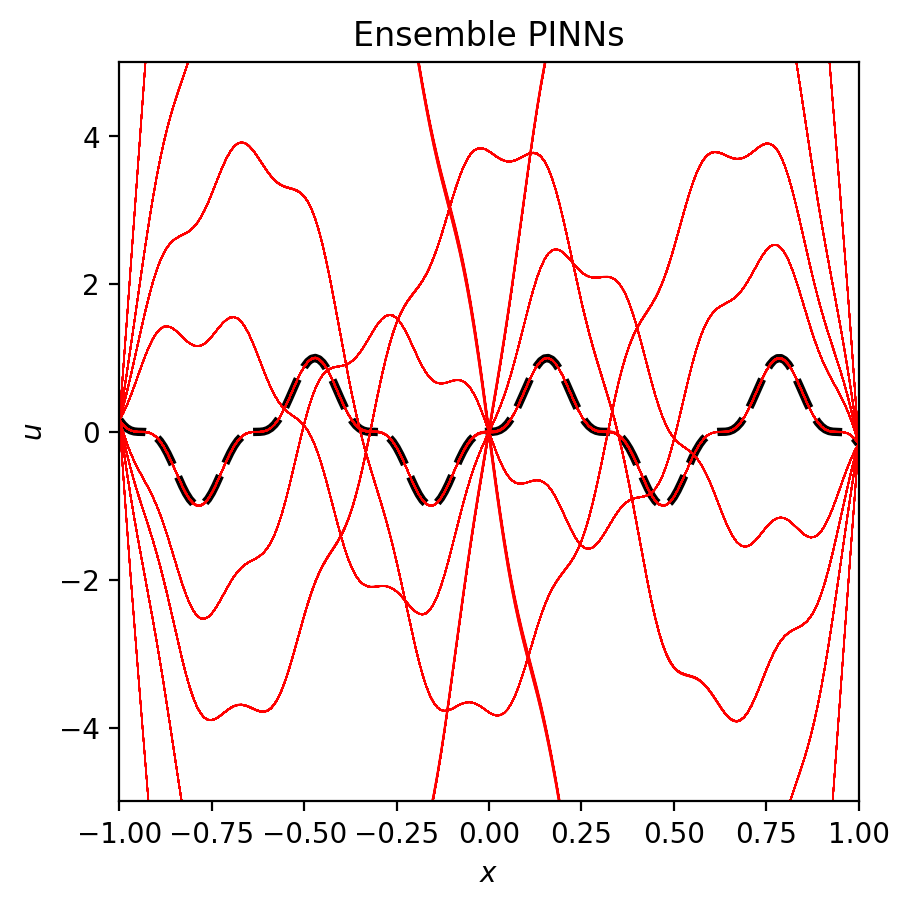}
        \includegraphics[scale=0.45]{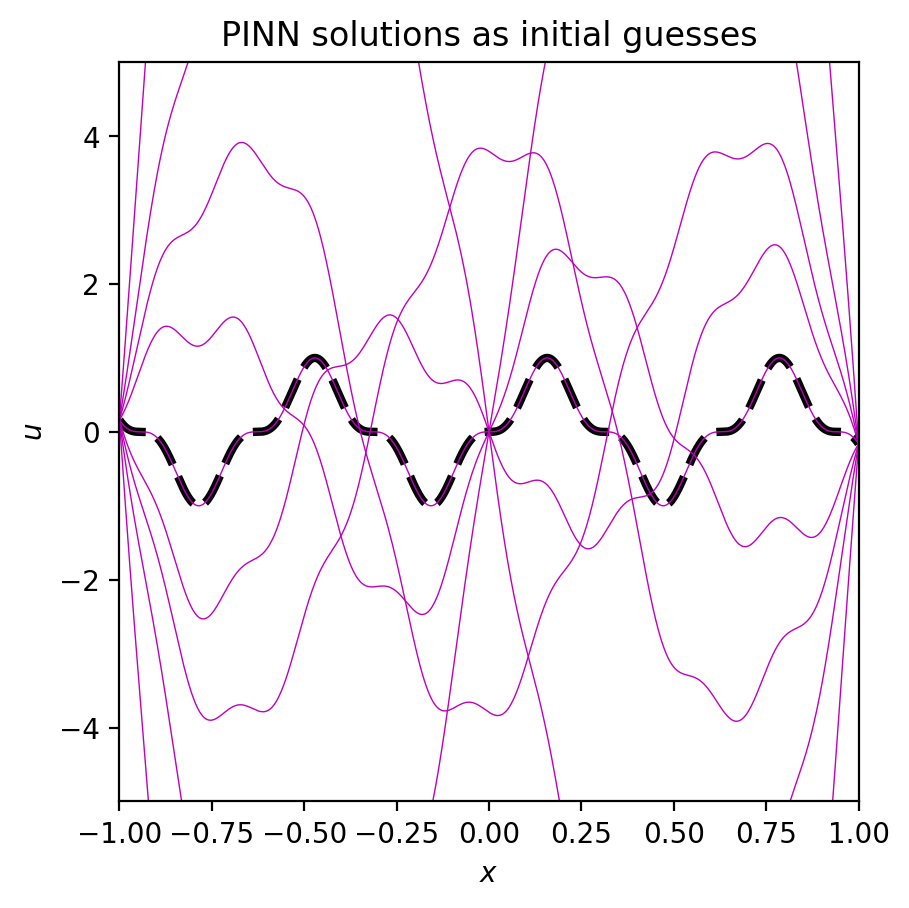}
    }
    \caption{Training dynamics of the predicted values of $u(-0.8)$ across these $1,000$ NNs (shown on the \textbf{left}) and multiple solutions to \eqref{eq:reaction_diffusion} with $w=6, 10$, which are obtained by using representative PINN solutions after $20,000$ iterations of training (shown in the \textbf{middle}) as the initial guesses for a BVP solver (shown on the \textbf{right}). The prescribed solutions are in \textbf{black} dashed lines while the PINN solutions and BVP solutions are in {\color{red}\textbf{red}} and {\color{cyan}\textbf{cyan}} solid lines, respectively. Quantitative results can be found in Tables \ref{tab:example_3_1} and \ref{tab:example_3_2}.}
    \label{fig:example_3_2}
\end{figure}

\noindent\textbf{Numerical Discovery of Multiple Solutions}

In this section, we apply the proposed method to discover multiple solutions to the following 1D steady-state PDE:
\begin{subequations}\label{eq:reaction_diffusion}
    \begin{align}
        & D\frac{d^2u}{dx^2}(x) + \kappa \tanh(u(x)) = f(x), x\in(-1, 1).\\
        & u(-1) = u_l, u(1) = u_r,
    \end{align}
\end{subequations}
where $D=0.01, \kappa=0.7$ are known constants. This equation can be interpreted as a steady-state reaction-diffusion equation with a nonlinear reaction term $\kappa\tanh(u)$ \cite{schafer2021bayesian, zhang2024discovering}. Here, we analytically derive the source term $f(x)$ and boundary conditions $u_l, u_r$ are from the prescribed solution:
\begin{equation}
    u^*(x) = \sin^3(wx), x\in[-1, 1].
\end{equation}
Thus, the source term is defined as $f(x) = D\frac{d^2u^*}{dx^2}(x) + \kappa \tanh(u^*(x))$ and the boundary conditions are set as $u_r = u^*(-1), u_r = u^*(1)$, ensuring $u^*(x)$ is a solution to \eqref{eq:reaction_diffusion}. As presented in Figure \ref{fig:example_3_1}, solving \eqref{eq:reaction_diffusion} with $w=6$ using PINNs with randomly initialized NNs not only yields an accurate approximation of $u^*$, as expected, but also reveals an additional distinct solution. A BVP solver \cite{kierzenka2001bvp, shampine2000solving} confirms this as another valid solution to \eqref{eq:reaction_diffusion}.

Notably, unlike the previous examples, where the existences of multiple solutions were well-documented in the literature, the solution multiplicity of this equation is, to the best of our knowledge, not previously established. We consider two cases:
\begin{enumerate}
    \item \textbf{Case (A)}: $w=6$, in which 11 distinct patterns are discovered.
    \item \textbf{Case (B)}: $w=10$, in which nine distinct patterns are discovered.
\end{enumerate}
Following the methodology of the previous example, we initialize 1,000 NNs using a truncated normal distribution with a mean of zero and different standard deviations ($\sigma\in\{0.2, 0.5, 1\}$) and train them for $20,000$ iterations to clearly resolve solution patterns.
The results, presented in Figure \ref{fig:example_3_1}, demonstrate that different random initializations lead to varying solution diversity:
\begin{enumerate}
    \item $\sigma=0.2$ yields two and three solution patterns in Case (A) and Case (B), respectively.
    \item $\sigma=0.5$ yields eight and seven patterns.
    \item $\sigma=1$ yields 11 and nine patterns.
\end{enumerate}
These results highlight the critical role of random initialization in discovering diverse multiple solutions. Furthermore, parameter-sharing PINNs (implemented using a multi-head architecture \cite{zou2023hydra}) successfully recover all solutions found by standard ensemble PINNs, as shown in Figure \ref{fig:example_3_1}. 

To analyze training dynamics, we track the predicted values of $u(-0.8)$ across these $1,000$ NNs. As depicted in Figure \ref{fig:example_3_2}, the ensemble PINNs identify the 11 and nine distinct solutions relatively early in training (around 8,000 iterations), though 20,000 iterations are empirically required for all solutions to reach high precision.

\noindent\textbf{Numerical Verification of discovered PINN solutions}

\begin{table}[h]
    \footnotesize
    \centering
    \begin{tabular}{c|c|c|c|c|c|c|c|c|c|c|c}
    \hline
    \hline
    & $u_1$ & $u_2$ & $u_3$ & $u_4$ & $u_5$ & $u_6$ & $u_7$ & $u_8$ & $u_9$ & $u_{10}$ & $u_{11}$\\
    \hline 
    ($\times 10^{-16}$) & $40.9$ & $6.14$ & $2.76$ & $1.79$ & $1.26$ & $1.79$ & $0.33$ & $0.31$ & $2.76$ & $6.93$ & $40.9$\\
    \hline
    \hline
    \end{tabular}
    \caption{Case (A): Maximum of the residuals of \eqref{eq:reaction_diffusion} on the uniform mesh of mesh size $h=1/3200$. Here, $u_i, i=1,...,11$ are in ascending order of their values at $x=-0.8$: $u_i(-0.8)<u_j(-0.8), \forall 1\leq i<j\leq 11$.} 
    \label{tab:example_3_1}
\end{table}

\begin{table}[h]
    \footnotesize
    \centering
    \begin{tabular}{c|c|c|c|c|c|c|c|c|c}
    \hline
    \hline
    & $u_1$ & $u_2$ & $u_3$ & $u_4$ & $u_5$ & $u_6$ & $u_7$ & $u_8$ & $u_9$\\
    \hline 
    ($\times 10^{-16}$) & $37.7$ & $5.42$ & $5.62$ & $5.25$ & $4.21$ & $4.38$ & $5.62$ & $9.47$ & $37.7$ \\
    \hline
    \hline
    \end{tabular}
    \caption{Case (B): Maximum of the residuals of \eqref{eq:reaction_diffusion} on the uniform mesh of mesh size $h=1/3200$. Here, $u_i, i=1,...,9$ are in ascending order of their values at $x=-0.8$: $u_i(-0.8)<u_j(-0.8), \forall 1\leq i<j\leq 9$.} 
    \label{tab:example_3_2}
\end{table}

Following the framework in Section \ref{sec:3_3}, we:
\begin{enumerate}
    \item categorize these $1,000$ trained PINNs into 11 or nine classes by the value of $u_\theta(-0.8)$,
    \item randomly select one NN from each class,
    \item use these selected NNs as initial guesses to solve \eqref{eq:reaction_diffusion} using a BVP solver.
\end{enumerate}
The same numerical solver as in the previous example is employed. To verify that the distinct PINN solutions correspond to valid numerical solutions of \eqref{eq:reaction_diffusion}, we solve the equation using the BVP solver with these PINN solutions as initial conditions. The computations are performed on a uniform mesh of $[-1, 1]$ with a mesh size of $1/3200$.

Qualitative results in Figure \ref{fig:example_3_2} confirm that using PINN solutions as initial guesses leads to ten and eight additional numerical solutions in Cases (A) and (B), respectively. To quantitatively assess these solutions, we evaluate the maximum residual of \eqref{eq:reaction_diffusion} across the uniform mesh. The results, summarized in Tables \ref{tab:example_3_1} and \ref{tab:example_3_2}, show that the maximum residual reaches machine precision, verifying that the discovered PINN solutions are indeed valid numerical solutions to \eqref{eq:reaction_diffusion}.

\subsection{2D Allen-Cahn equation}\label{sec:3_5}

\begin{figure}[ht]
    \centering
    \subfigure[FEM solutions obtained from initializing  distinct PINN solutions (shown in Figure \ref{fig:example_5_3}) for the Gauss-Newton method.]{
        \includegraphics[scale=0.33]{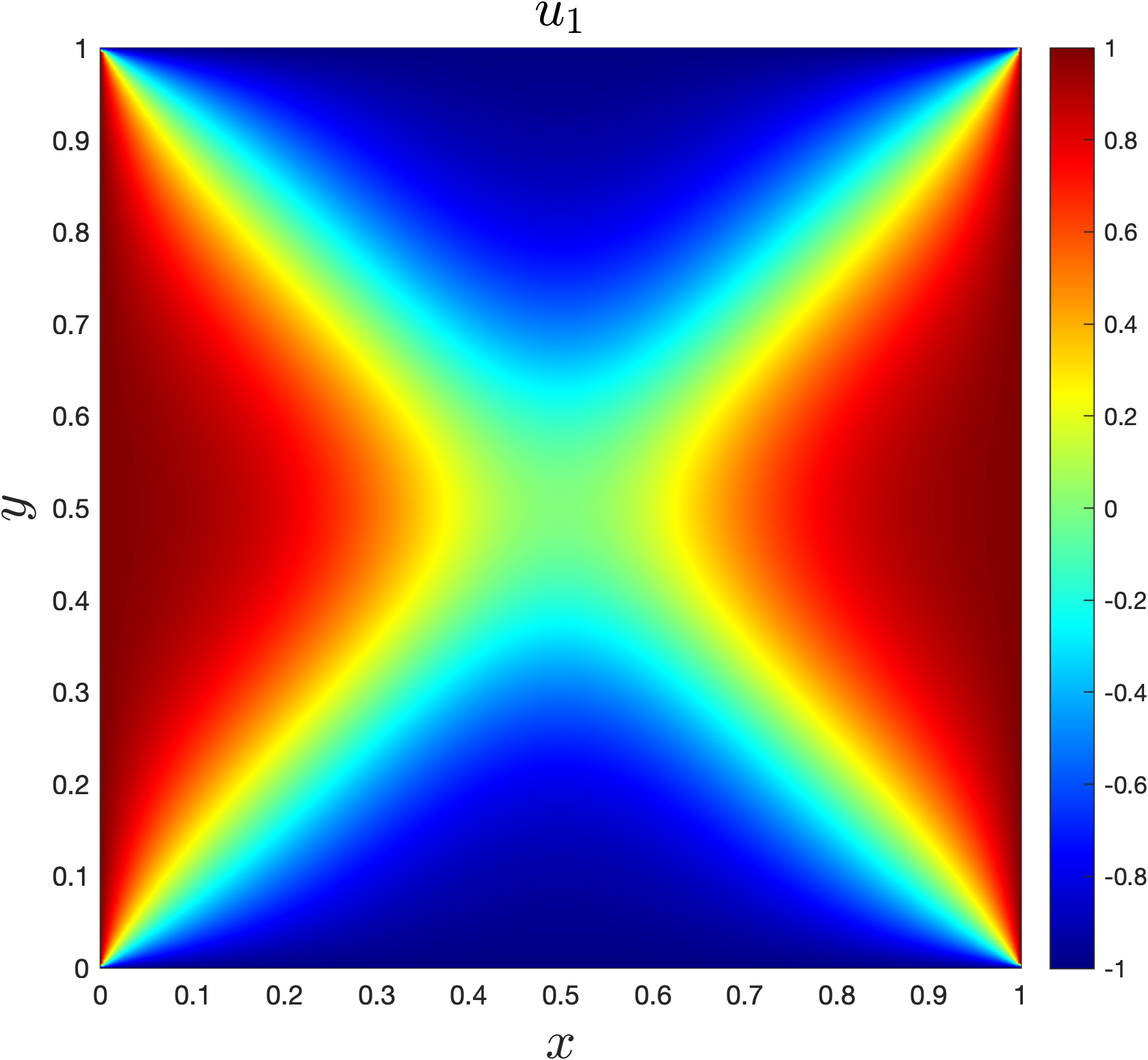}
        \includegraphics[scale=0.33]{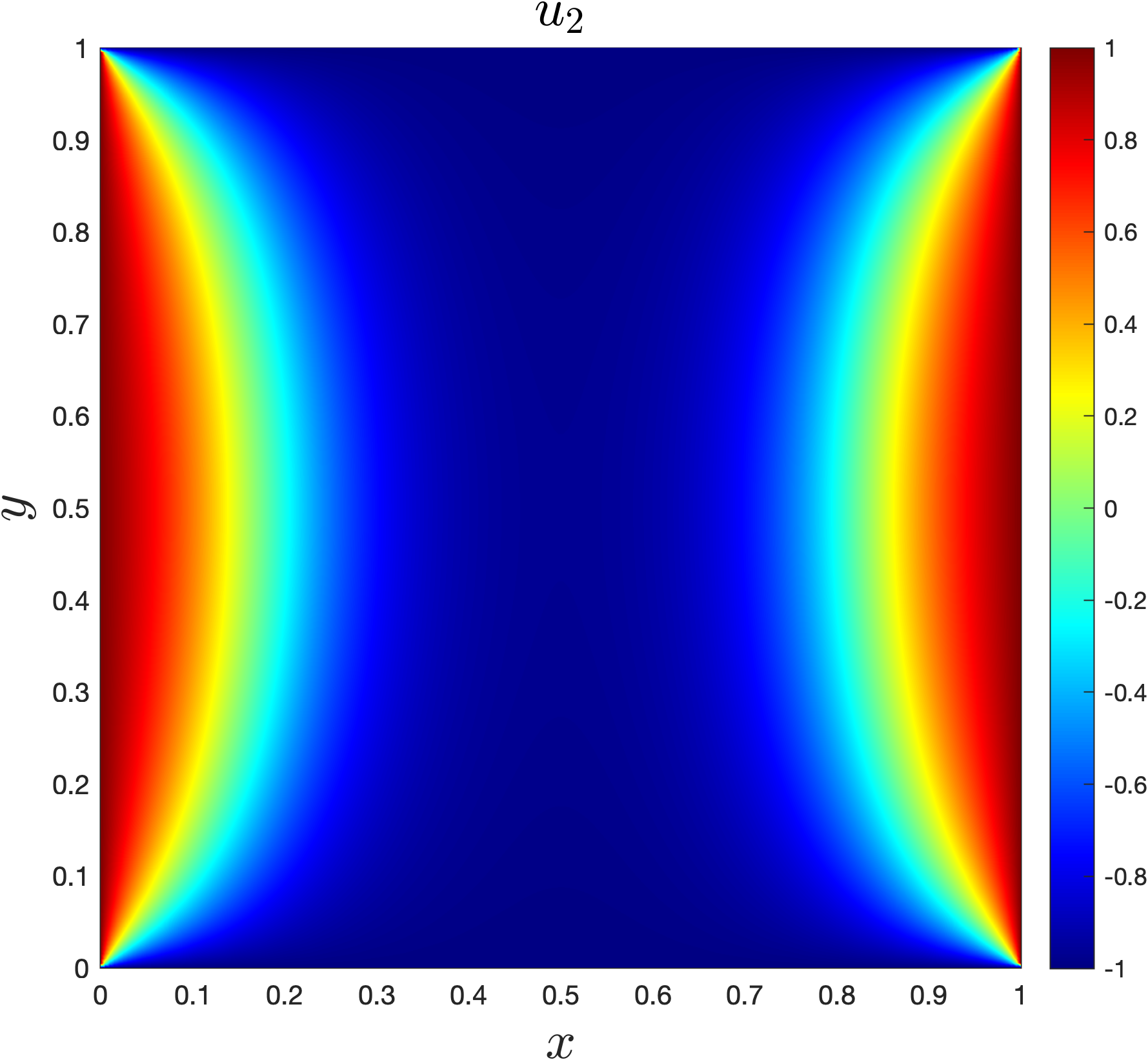}
        \includegraphics[scale=0.33]{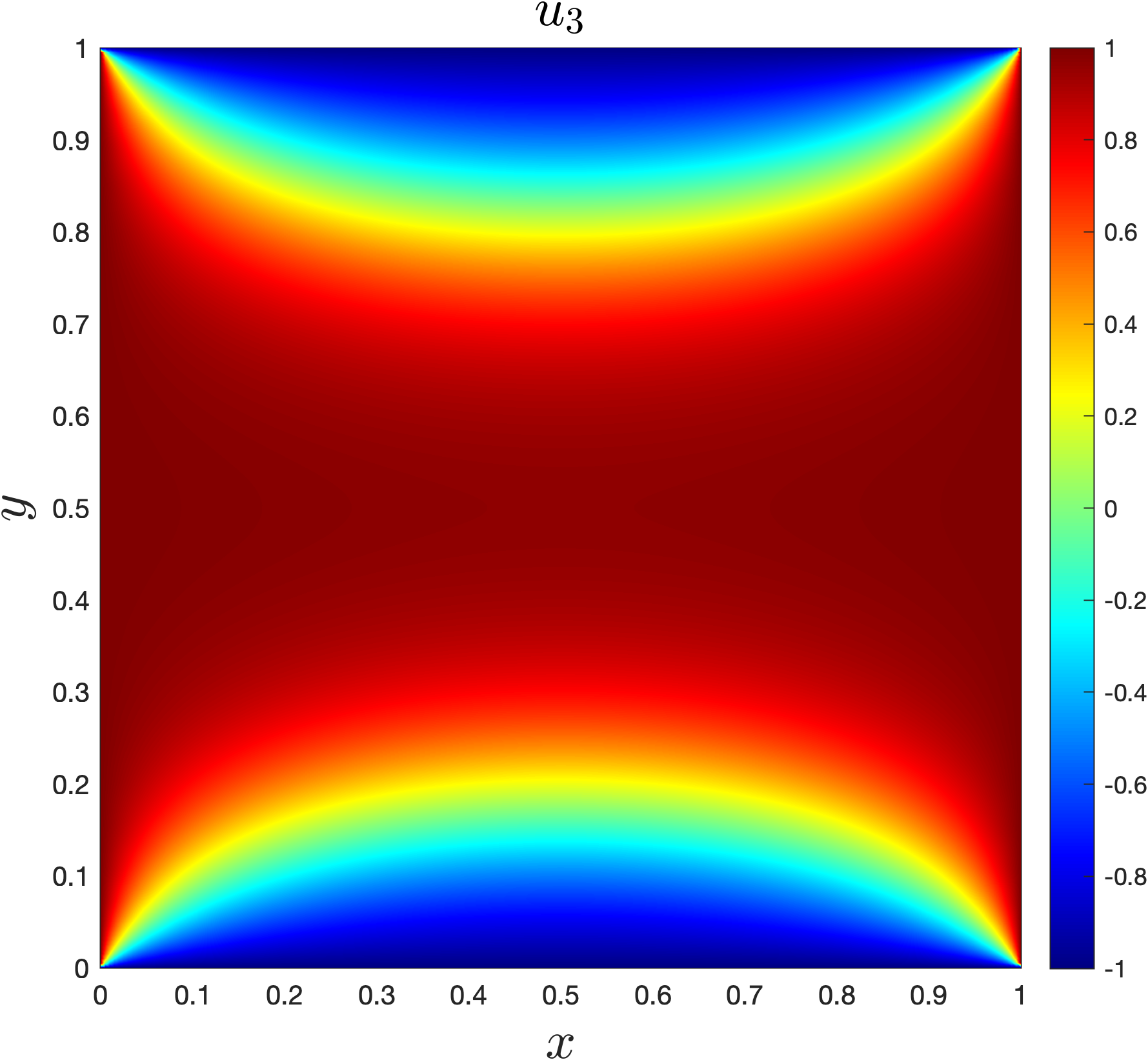}
    }
    \subfigure[The mesh.]{
        \includegraphics[scale=0.30]{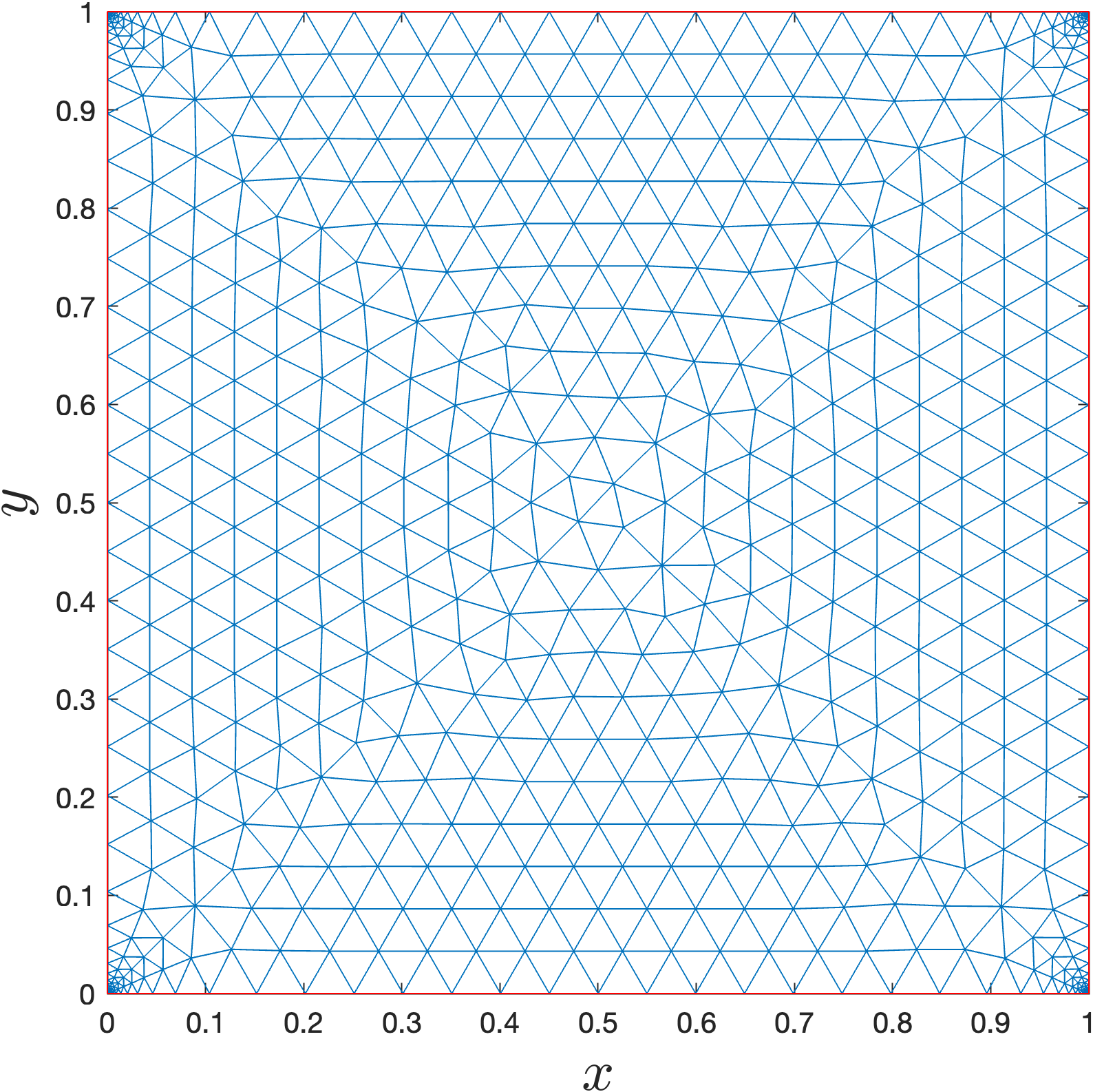}
    }
    \subfigure[Convergence of residual.]{
        \includegraphics[scale=0.30]{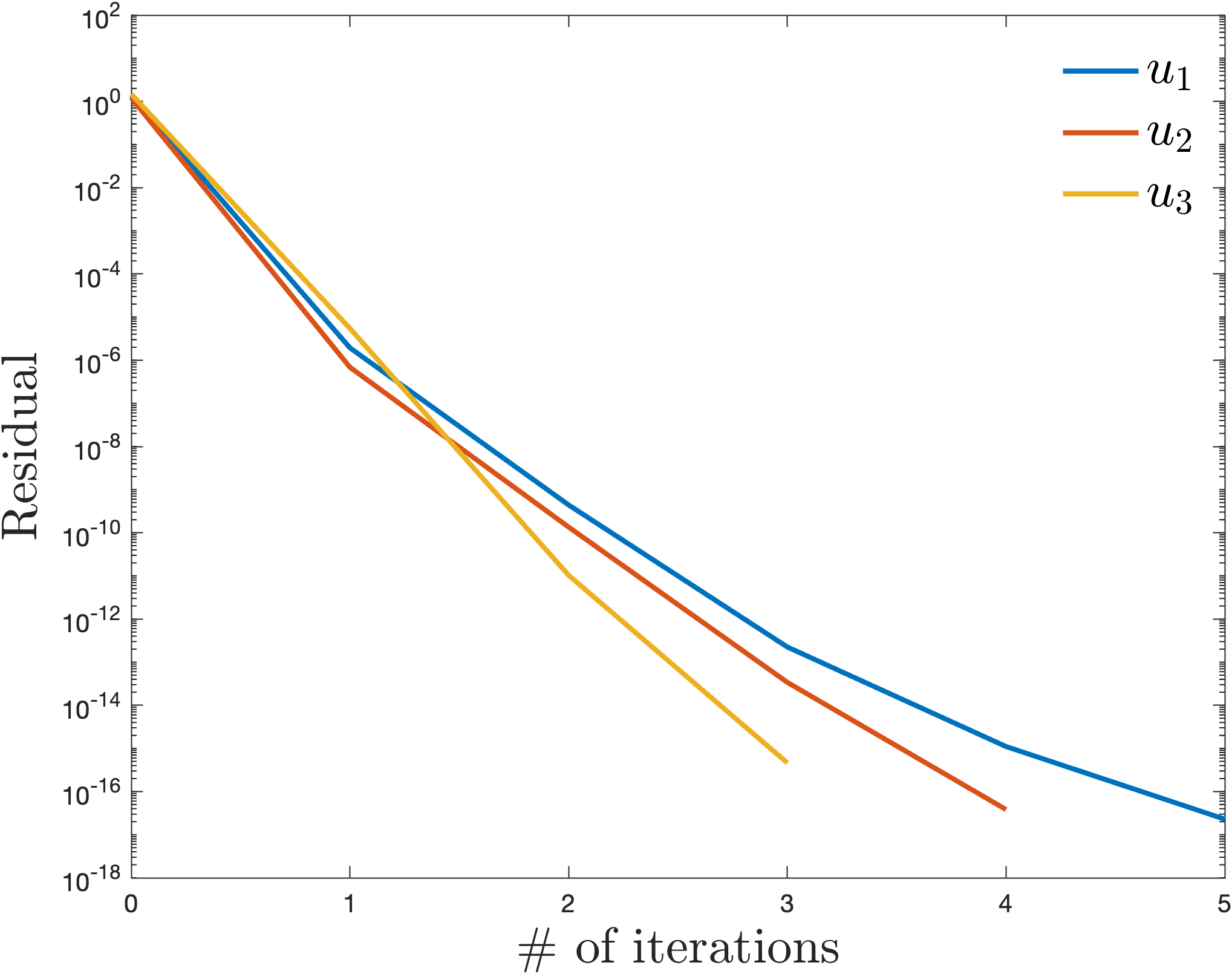}
    }
    \caption{Results of solving \eqref{eq:2d_allen} with $\epsilon=0.01$ using the presented approach. In (a), we show three distinct found solutions to \eqref{eq:2d_allen} ($u_1, u_2, u_3$ from left to right). The mesh is presented in (b) and the convergence of residuals in (c). The convergence criterion is set to residual decreasing below $1\times10^{-15}$. We note that $u_1$ in (a) is an unstable solution while $u_2$ and $u_3$ are stable ones.}
    \label{fig:example_5}
\end{figure}

\begin{figure}[ht]
    \centering
    \includegraphics[scale=0.25]{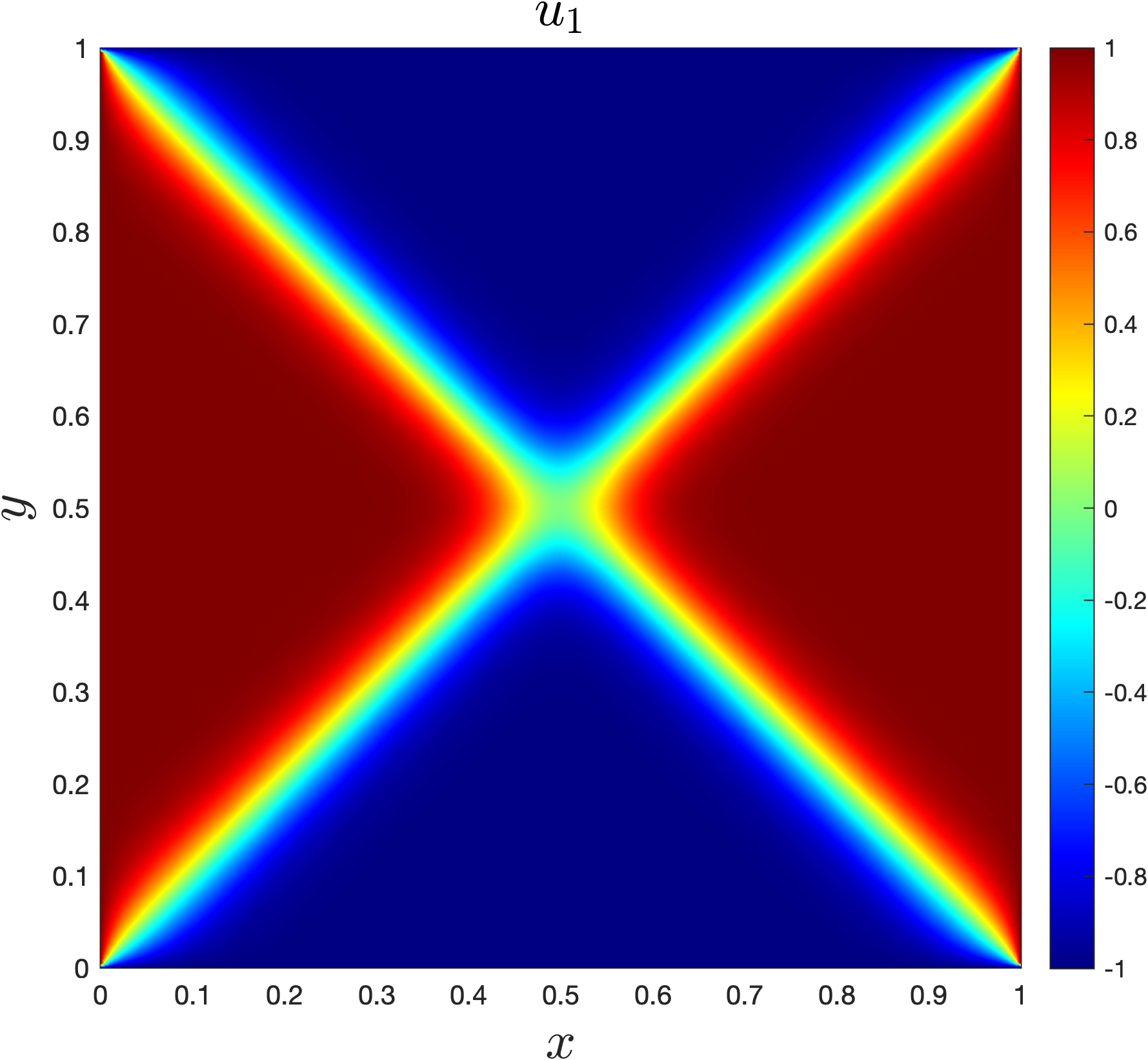}
    \includegraphics[scale=0.25]{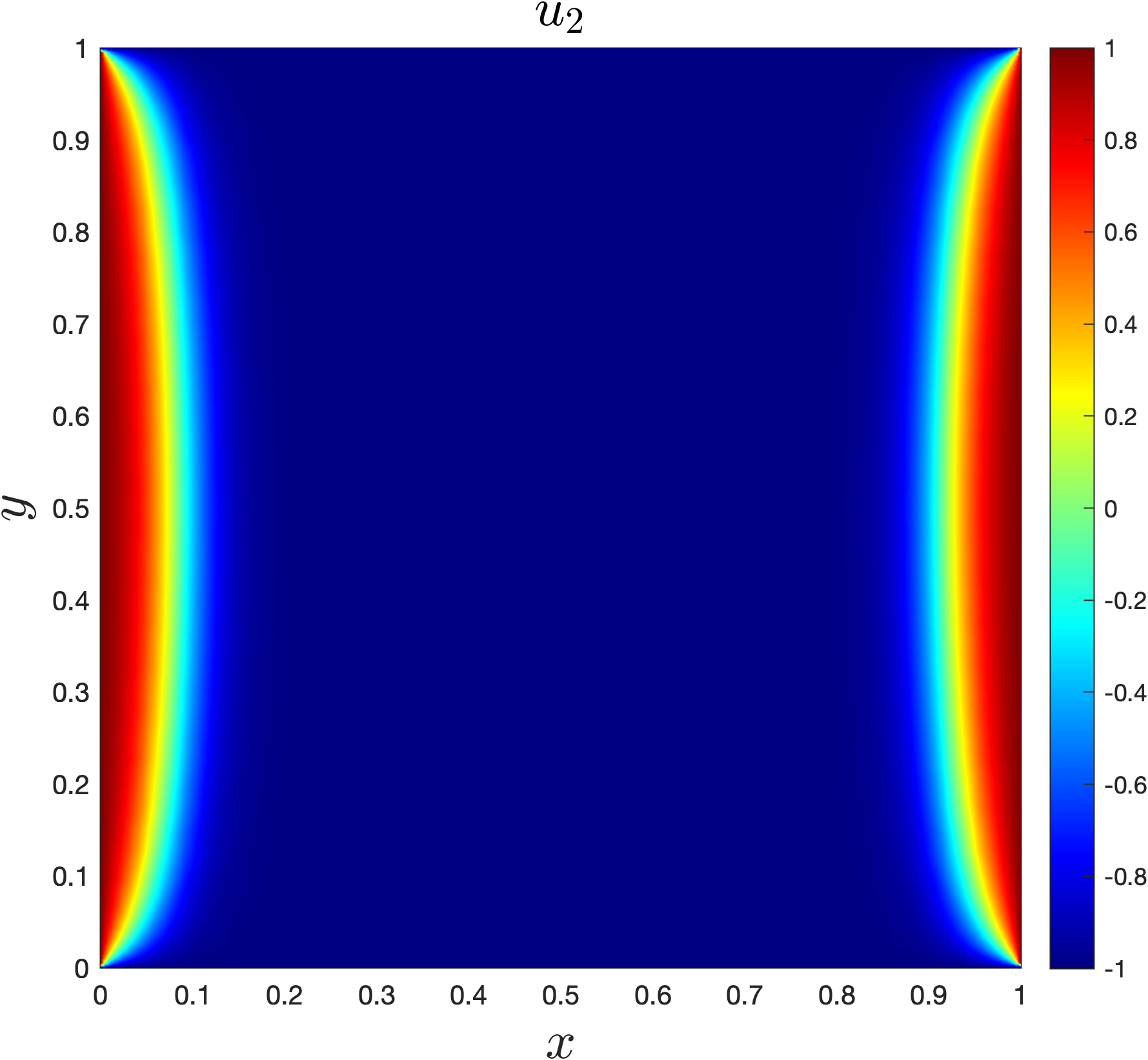}
    \includegraphics[scale=0.25]{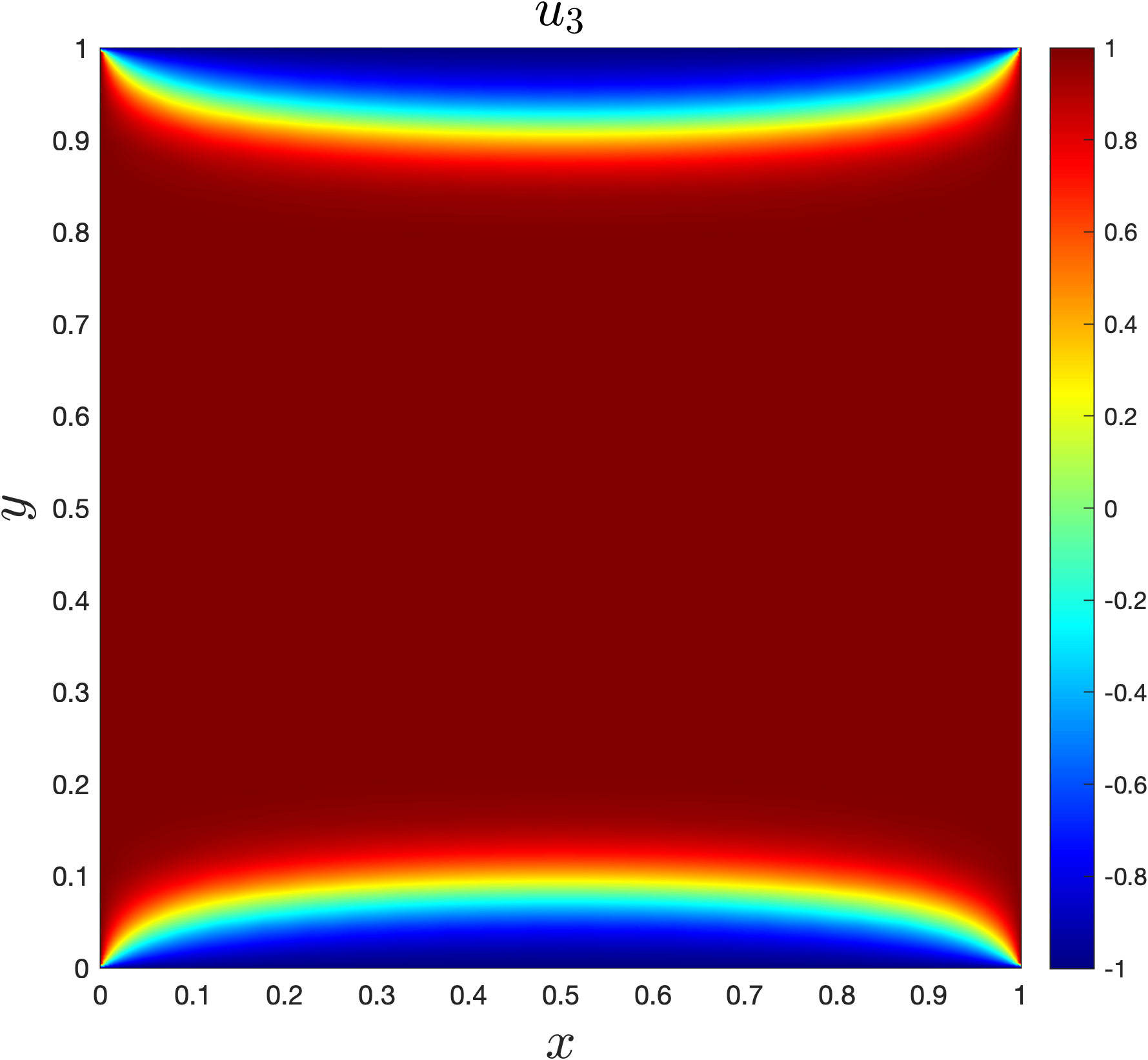}
    \includegraphics[scale=0.25]{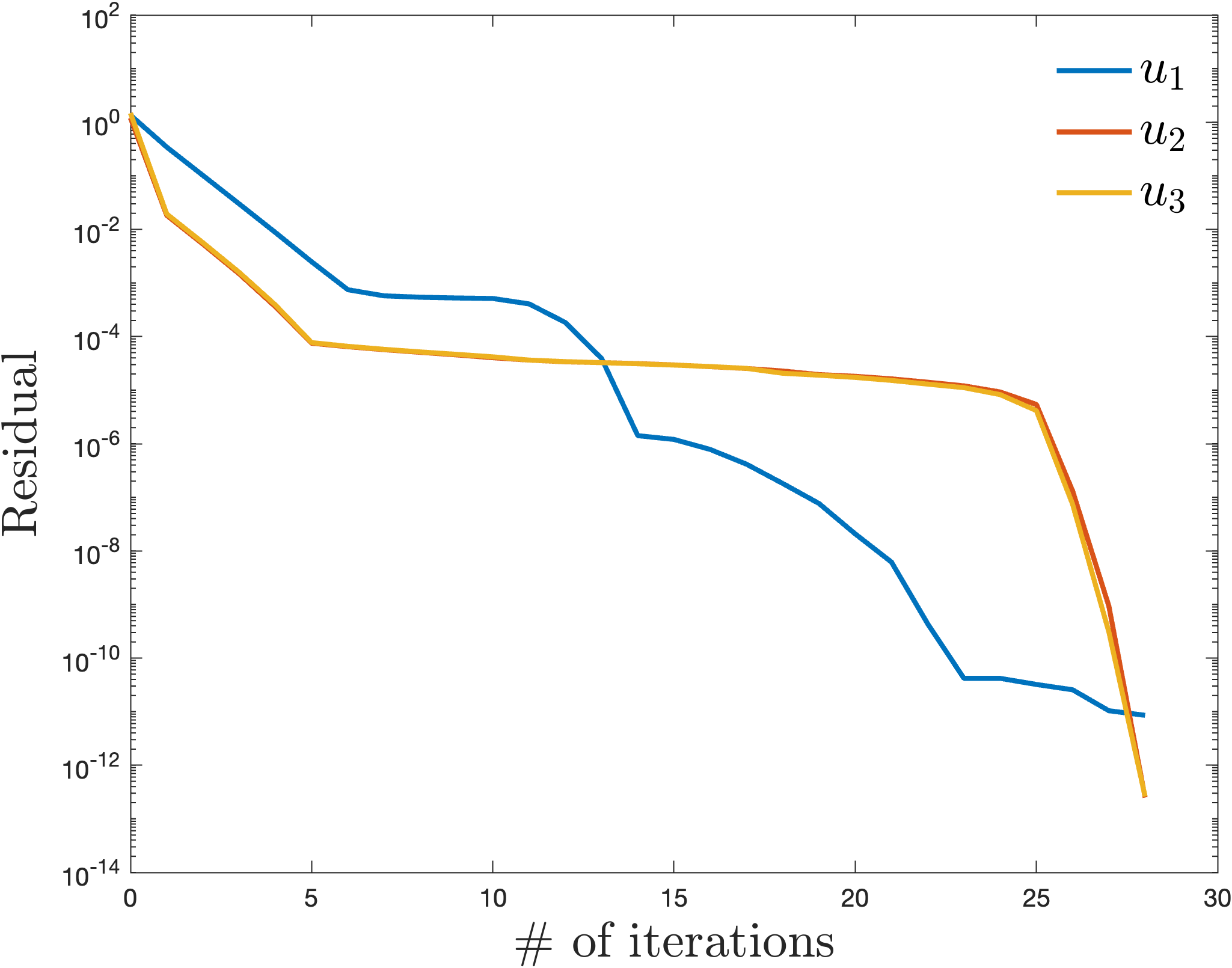}
    \caption{FEM solutions to \eqref{eq:2d_allen} with $\epsilon=0.001$ obtained from initializing at distinct PINN solutions to \eqref{eq:2d_allen} with $\epsilon=0.01$ (shown in Figure \ref{fig:example_5_3}) for the Gauss-Newton method. The mesh is displayed in Figure \ref{fig:example_4}(b). The convergence criterion is set to residual decreasing below $1\times10^{-11}$.}
    \label{fig:example_5_2}
\end{figure}

\begin{table}[h]
    \footnotesize
    \centering
    \begin{tabular}{c|c|c|c}
    \hline
    \hline
    & $u_1$ & $u_2$ & $u_3$ \\
    \hline 
    $\epsilon=0.01$ & $2.29\times 10^{-17}$ & $3.82\times 10^{-17}$ & $4.61\times 10^{-16}$\\
    \hline 
    $\epsilon=0.001$ & $8.55\times 10^{-12}$ & $ 2.50\times 10^{-13}$ & $2.61\times 10^{-13}$\\
    \hline
    \hline
    \end{tabular}
    \caption{Maximum of the residuals of \eqref{eq:2d_allen} with different values of $\epsilon$ of three found solutions on the mesh displayed in Figure \ref{fig:example_5}(b).} 
    \label{tab:example_5}
\end{table}

The Allen-Cahn equation is a fundamental PDE that describes phase separation in multi-component systems, such as phase formation in alloys and interface evolution in reaction-diffusion systems \cite{allen1979microscopic, allen1975coherent}. It is also known to exhibit solution multiplicity due to the presence of metastable states and bifurcations \cite{weinan2004minimum, farrell2015deflation, li2024adaptive}. The 2D formulation considered in this example is given by:
\begin{subequations}\label{eq:2d_allen}
    \begin{align}
        &-\epsilon\Delta u + u^3 - u = 0, (x, y) \in (0, 1)^2.\\
        & u(x, 0) = u(x, 1) = -1, x \in (0, 1),\\
        & u(0, y) = u(1, y) = 1, y\in (0, 1),
    \end{align}
\end{subequations}
where $\epsilon>0$ is a constant. This equation was investigated in \cite{weinan2004minimum} for determining minimal action pathways and transition times between stable states. Without noise, the system settles into one of its stable states and remains there indefinitely. However, in the presence of stochastic noise, the system can transition between metastable states when the noise pushes it out of the basin of attraction of its steady-state solution. 

We first apply the proposed approach to discover multiple solutions to \eqref{eq:2d_allen} with $\epsilon=0.01$. Specifically, we initialize $50$ NNs using a truncated normal distribution with a mean of zero and a standard deviation of $0.1$, then train them for $20,000$ iterations to choose representative PINN solutions. Ensemble PINNs are able to identify three distinct solutions, as displayed in Figure \ref{fig:example_5_3}. 
Following the approach outlined in Section \ref{sec:2_3}, we use a conventional numerical solver to solve \eqref{eq:2d_allen}, employing representative PINN solutions as initial guesses. In this example, we utilize MATLAB’s \textit{Partial Differential Equation Toolbox} \cite{MATLAB}, which implements a finite element method (FEM) combined with the Gauss-Newton algorithm for solving nonlinear steady-state PDEs.
Figure \ref{fig:example_5} presents the results alongside the computational mesh (quadratic element geometric order) and the convergence of the Gauss-Newton algorithm. The maximum of residuals of these solutions can be found in Table \ref{tab:example_5}. 
Notably, the solution shown on the left in Figure \ref{fig:example_5} ($u_1$) is in fact an unstable solution to \eqref{eq:2d_allen} while the other two ($u_2$ and $u_3$) are stable. 

To verify this, we apply the time-marching approach on the same mesh by introducing an artificial time derivative term to \eqref{eq:2d_allen} and solving the resulting time-dependent PDE until it reaches a steady state. Further details and results can be found in \ref{sec:appendix_3}. PINN solutions representing $u_1$ eventually transition into either $u_2$ or $u_3$.

Next, we extend our investigation by decreasing the value of $\epsilon$ in \eqref{eq:2d_allen} from $\epsilon=0.01$ to $\epsilon=0.001$. We use the previously obtained PINN solutions ($\epsilon=0.01$) as initial guesses for the Gauss-Newton algorithm, following a continuation method approach \cite{allgower2012numerical}. Despite the fact that PINNs are trained on a different $\epsilon$, multiple solutions are still accurately obtained for $\epsilon=0.001$, as shown in Figure \ref{fig:example_5_2} and Table \ref{tab:example_5}, using the same computational mesh and FEM solver. However, we note that this strategy does not facilitate the discovery of additional solution patterns.

\subsection{2D lid-driven cavity flow}\label{sec:3_6}

\begin{figure}[ht!]
    \centering
    \subfigure[PINN solutions of the cavity flow with random initialization. ]{
        \includegraphics[scale=0.1]{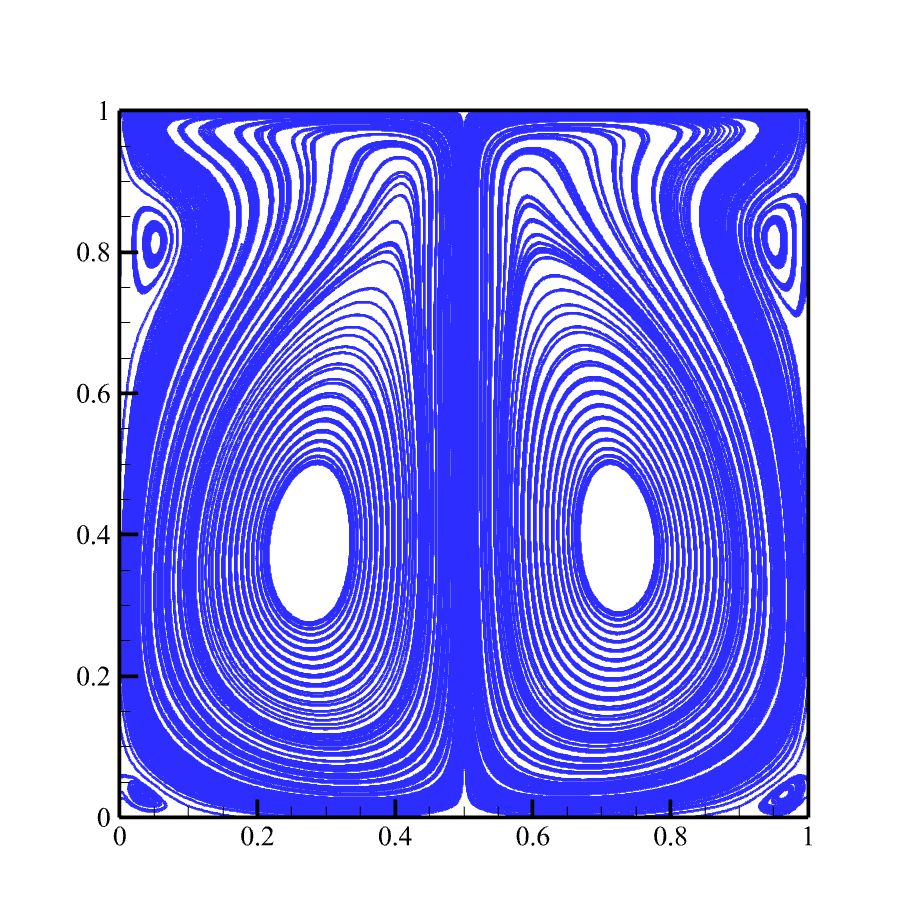}
        \includegraphics[scale=0.1]{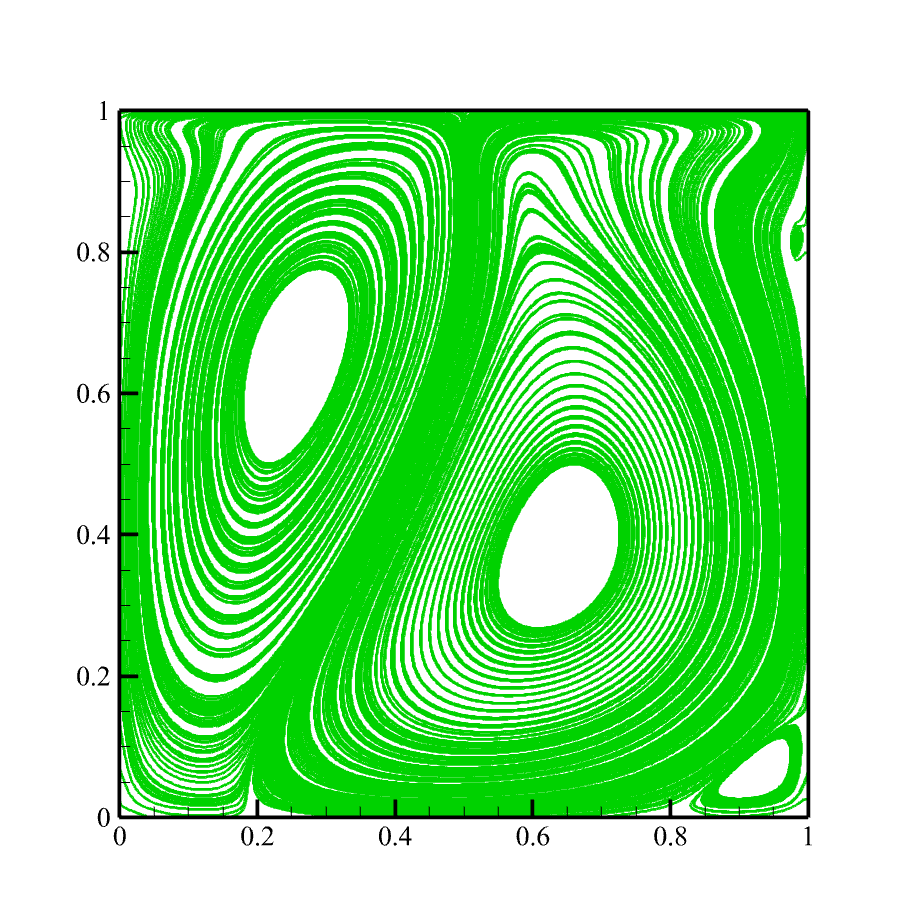}
        \includegraphics[scale=0.1]{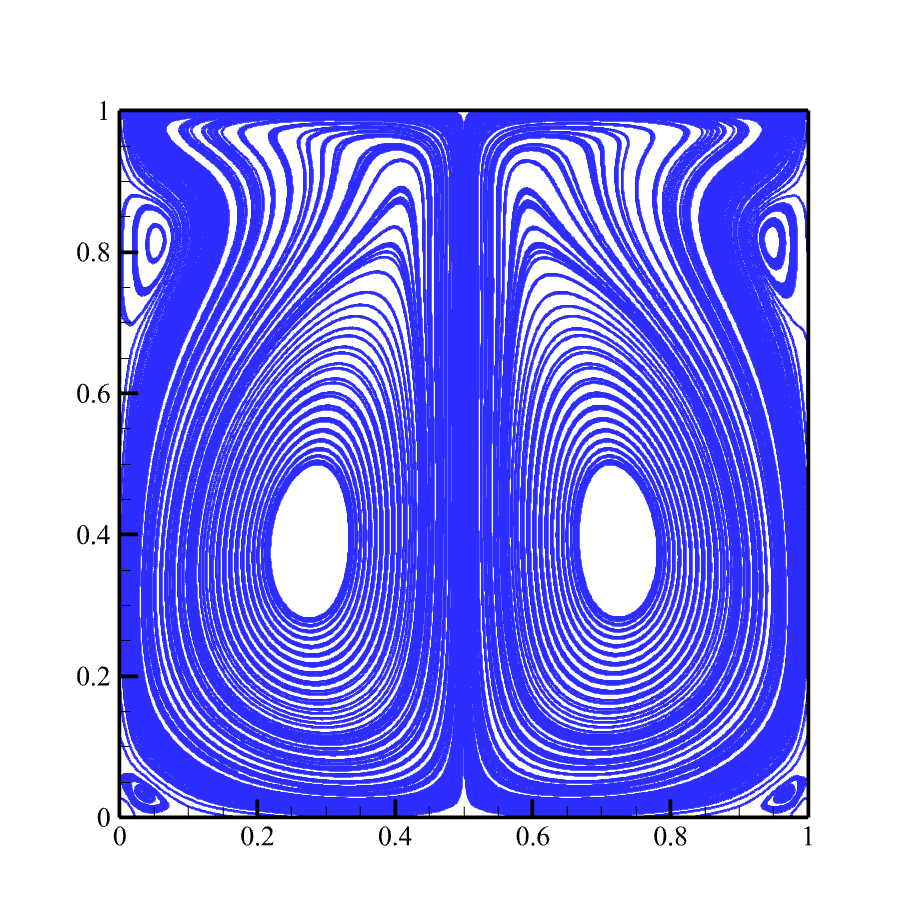}
        \includegraphics[scale=0.1]{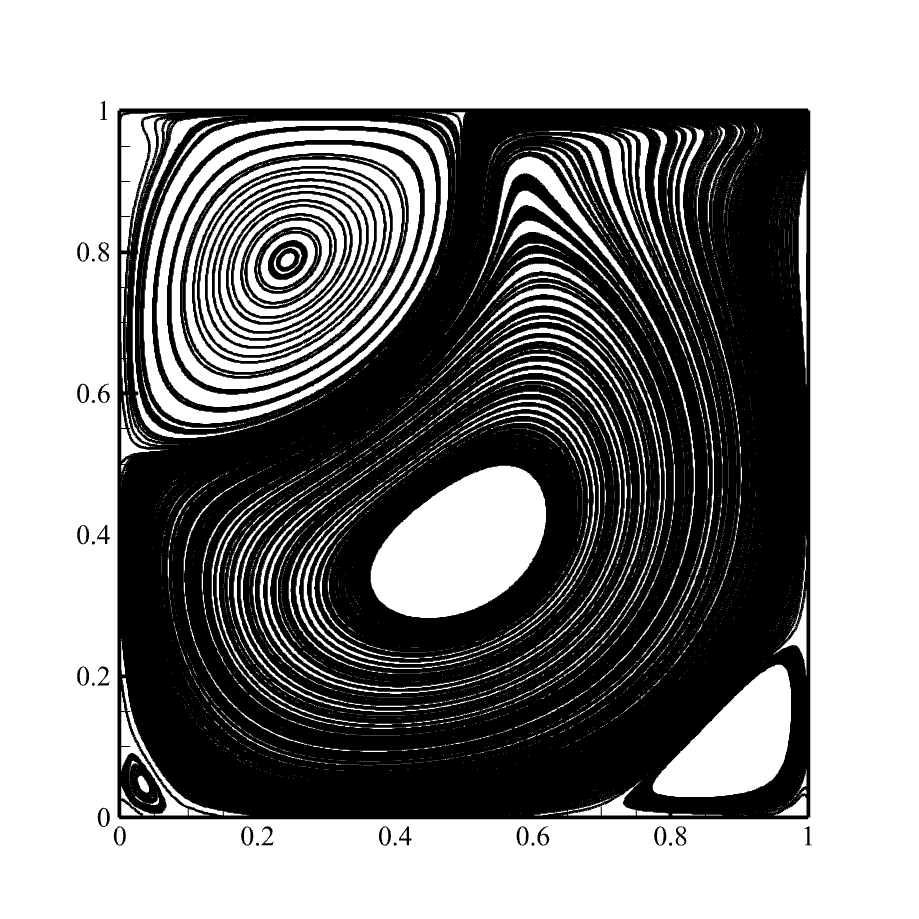}
        \includegraphics[scale=0.1]{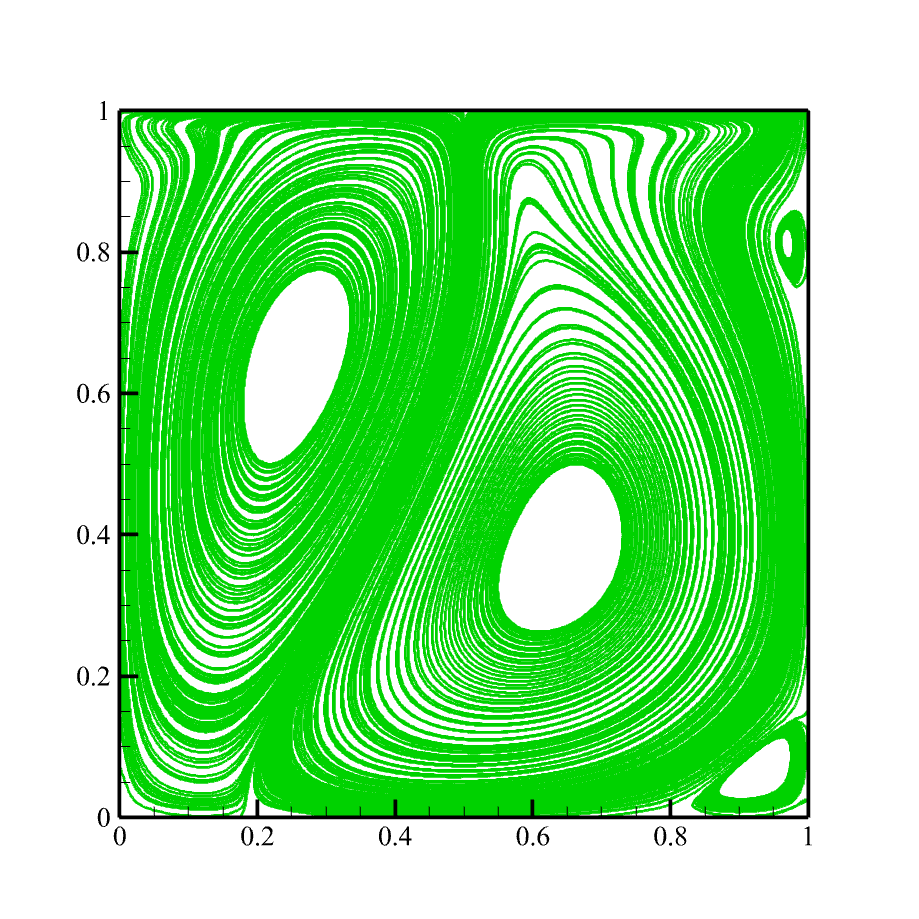}
    }
    \subfigure[Solutions from finite element method solver of the \emph{steady} Navier-Stokes equations, implemented on Deal.II. The initial condition of each case is the the corresponding PINN solution.]{
        \includegraphics[scale=0.1]{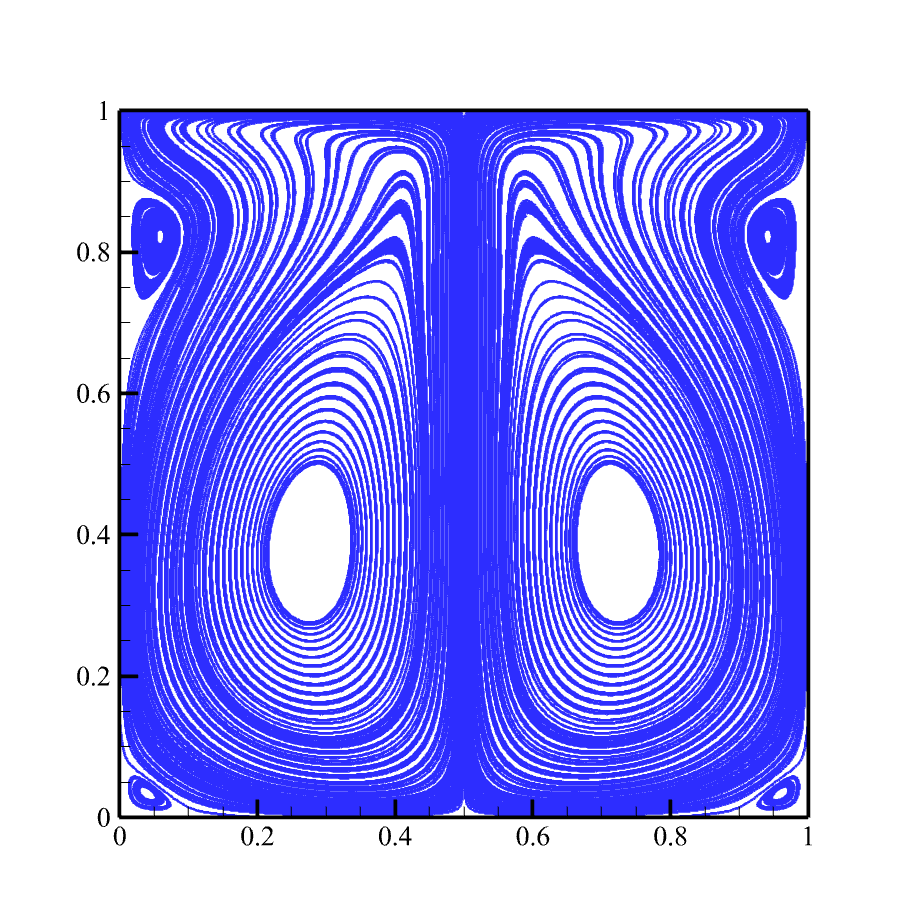}
        \includegraphics[scale=0.1]{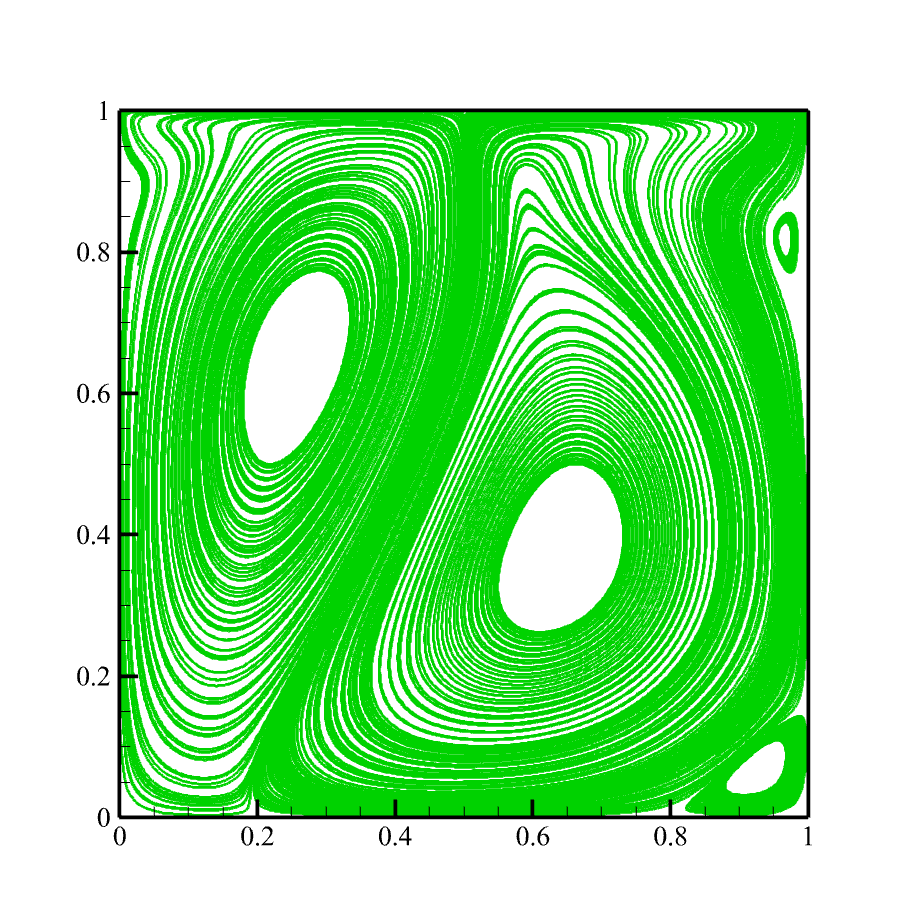}
        \includegraphics[scale=0.1]{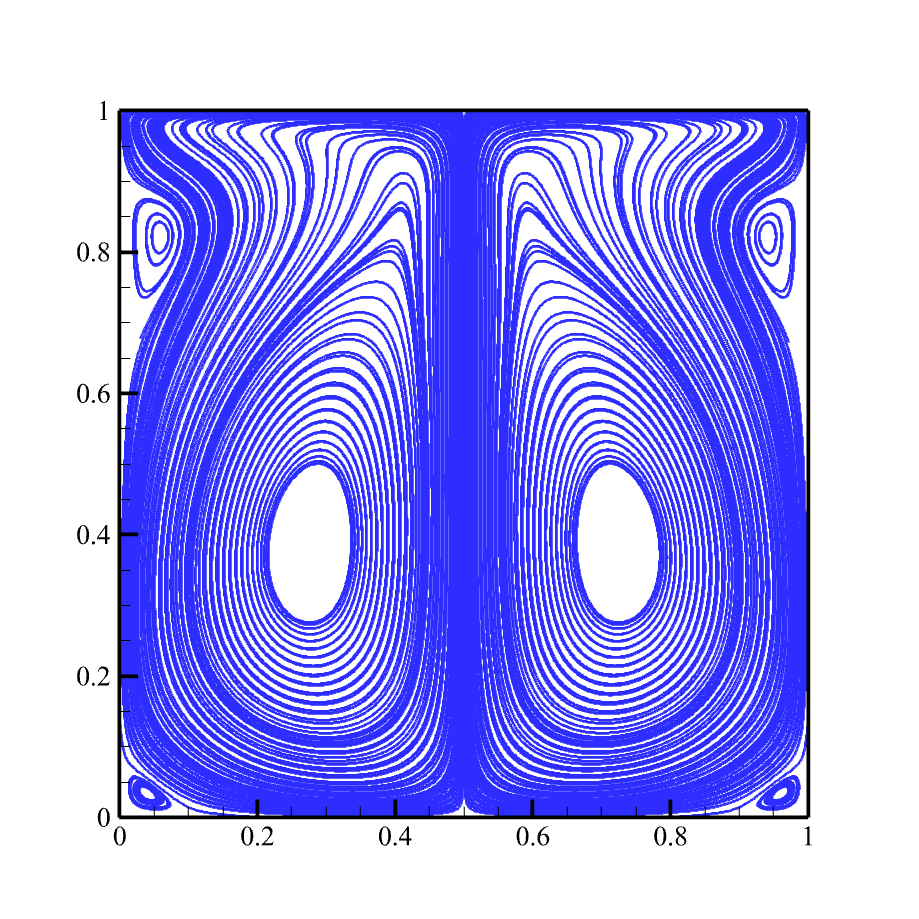}
        \includegraphics[scale=0.1]{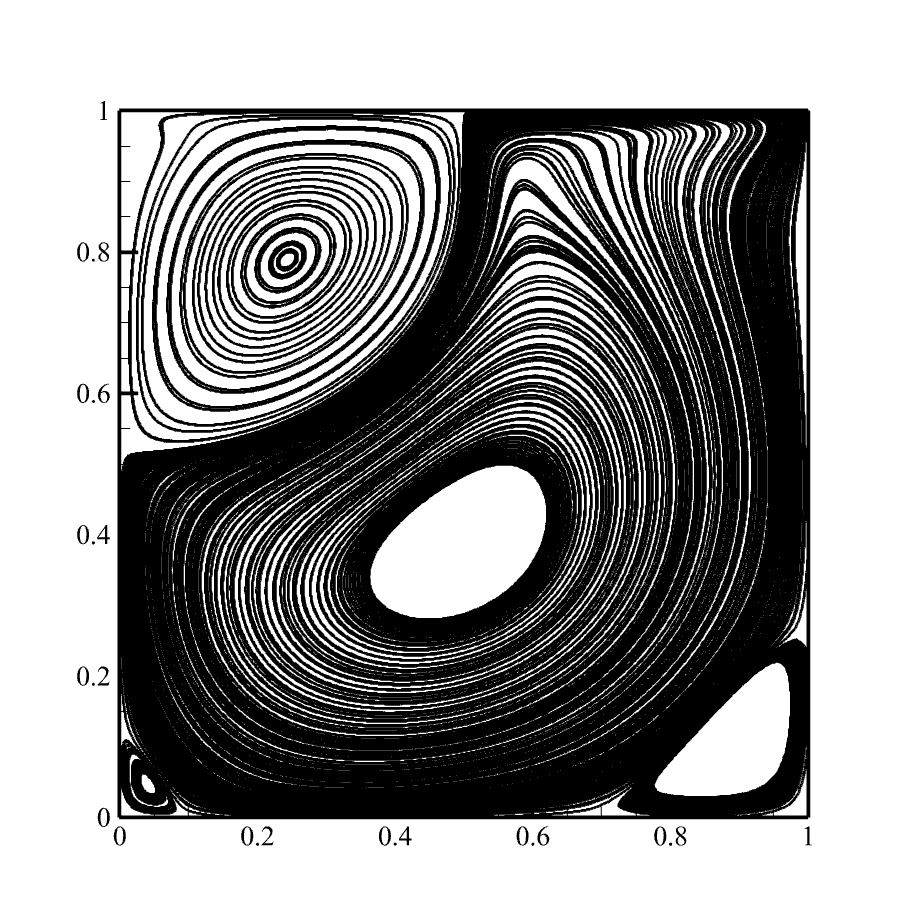}
        \includegraphics[scale=0.1]{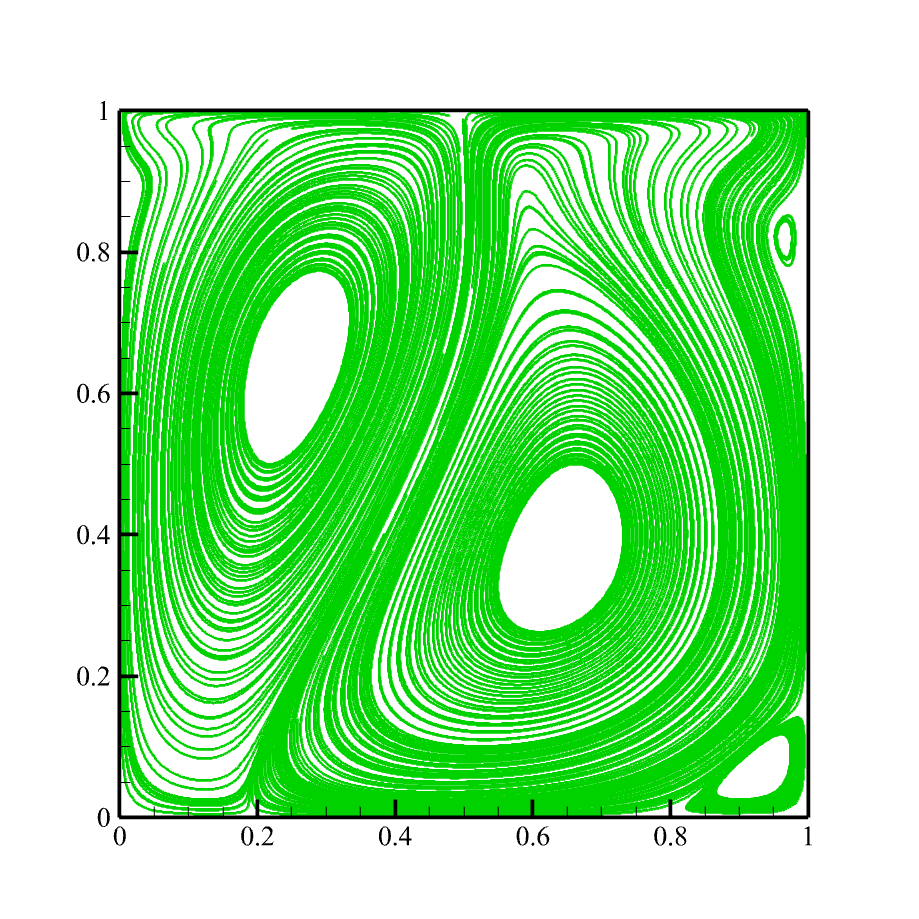}
    }
    \subfigure[Solutions from the spectral element method solver of the \emph{unsteady} Navier-Stokes equations, implemented on Nektar. The solutions presented here are the results after $2\times 10^4$ time steps, with the initial condition from the the corresponding PINN solution.]{
        \includegraphics[scale=0.1]{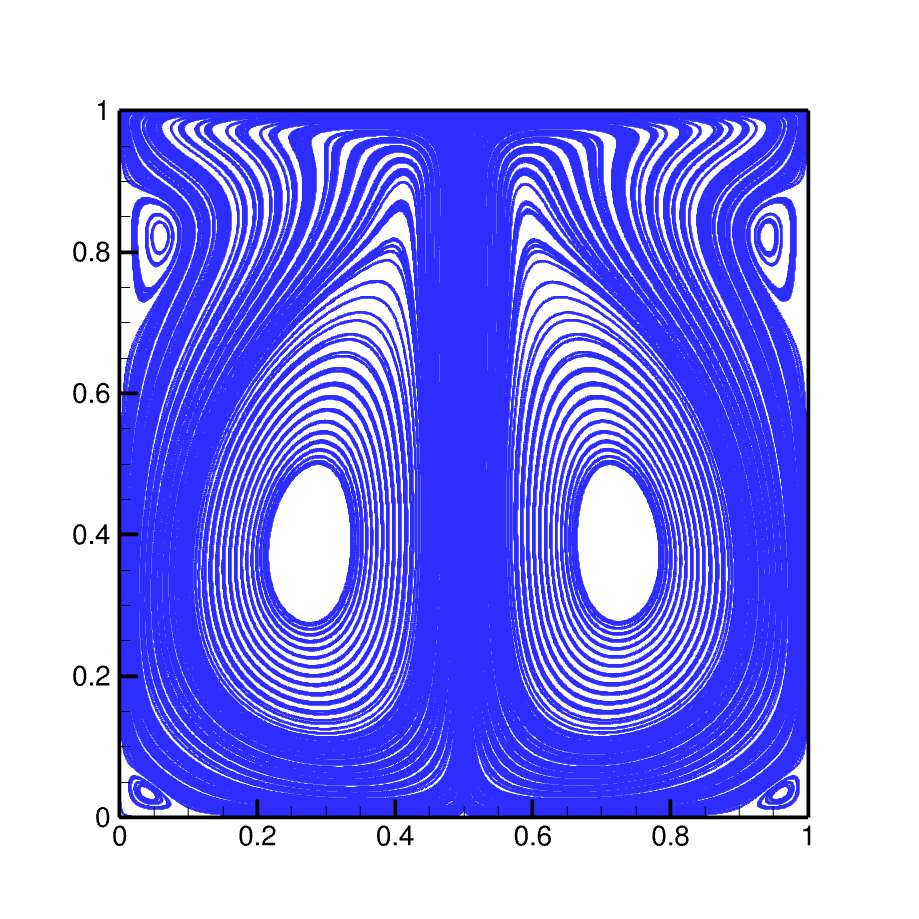}
        \includegraphics[scale=0.1]{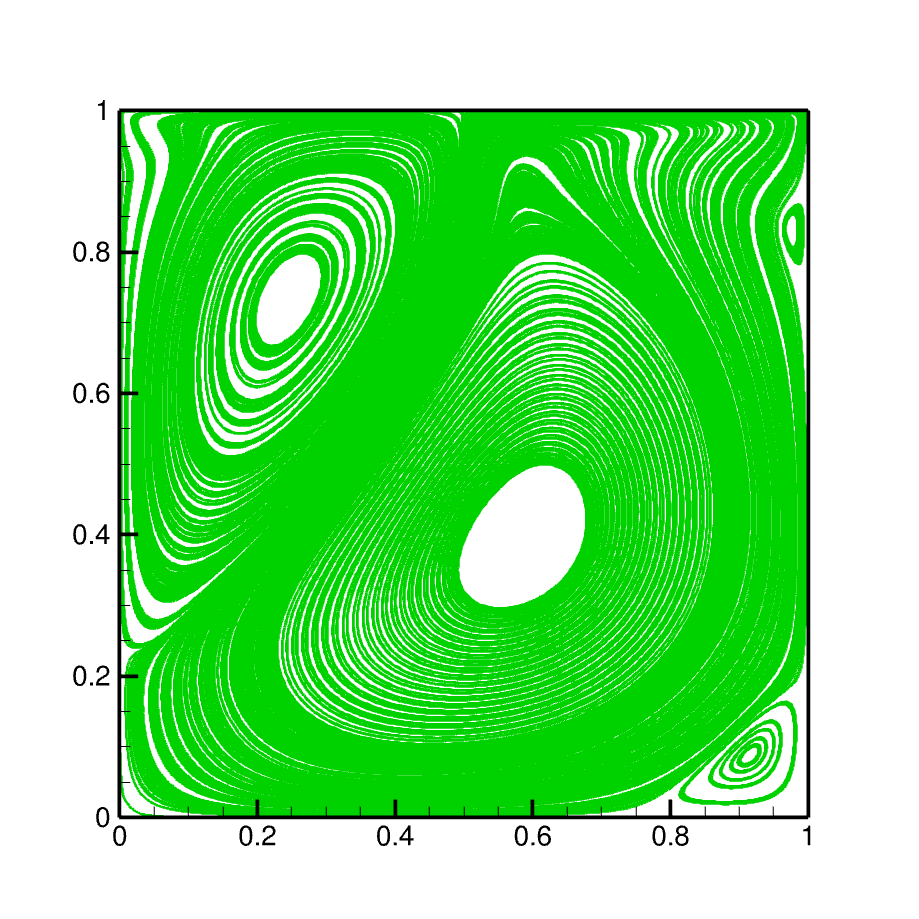}
        \includegraphics[scale=0.1]{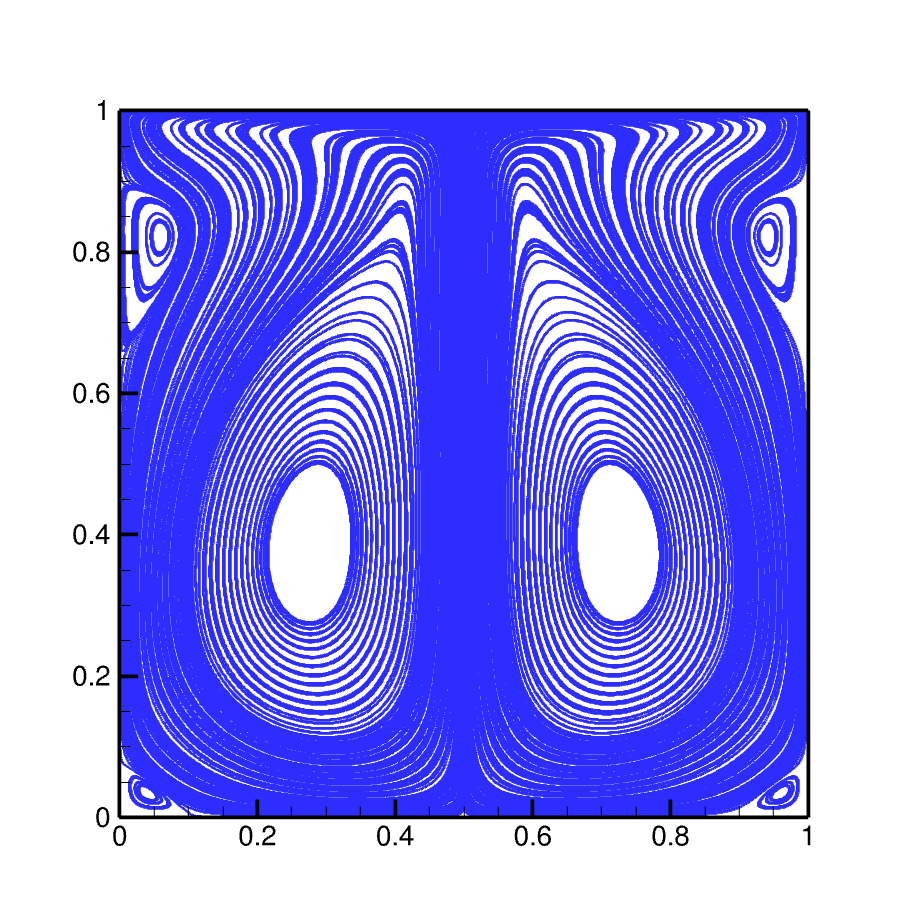}
        \includegraphics[scale=0.1]{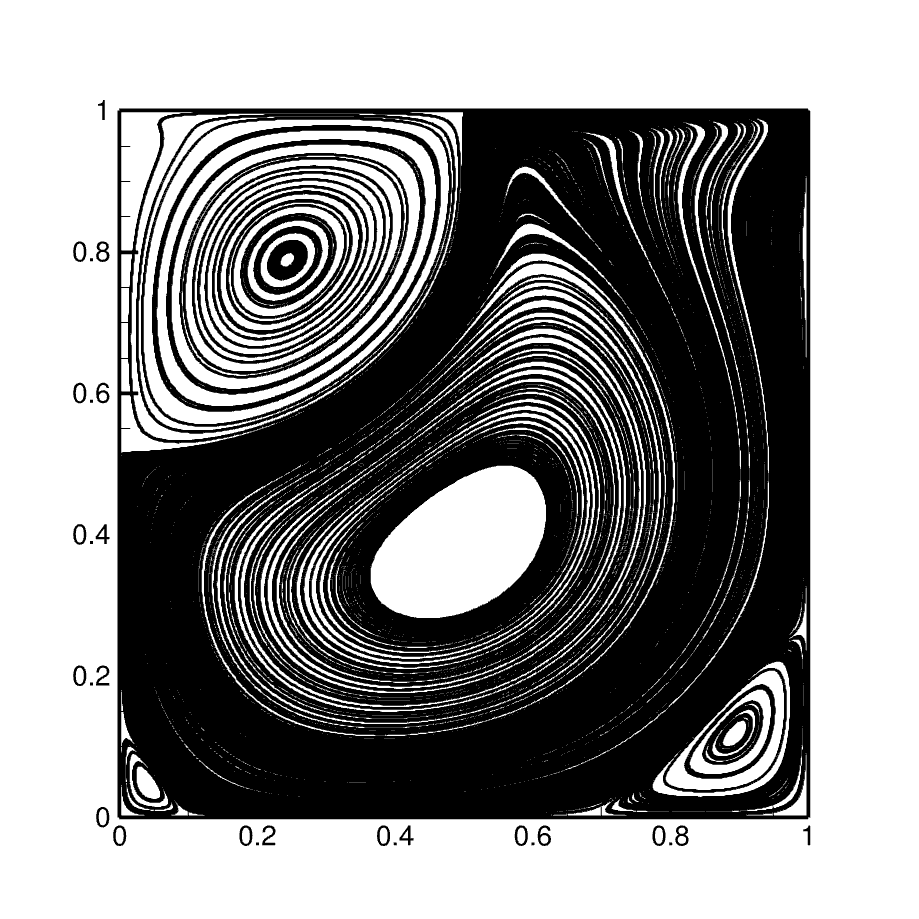}
        \includegraphics[scale=0.1]{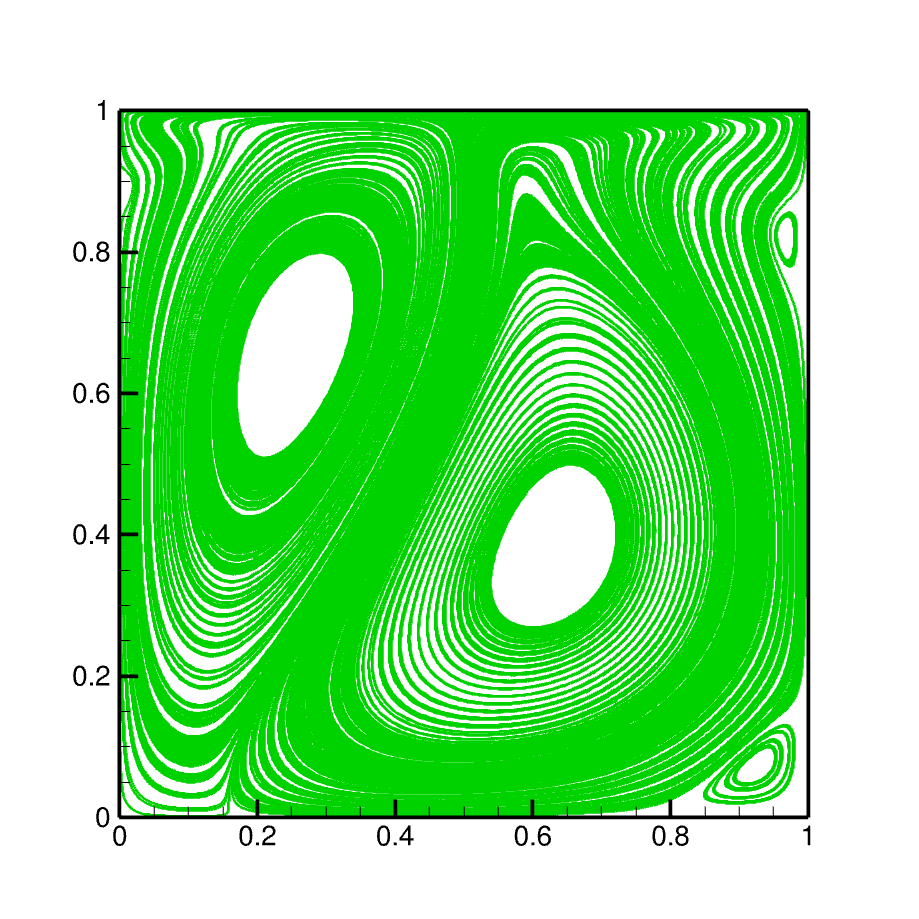}
    }
    \subfigure[Solutions from the unsteady spectral element method solver. The solutions presented here are the results after $2\times 10^5$ time steps, and it is a continuation of above simulation.]{
        \includegraphics[scale=0.1]{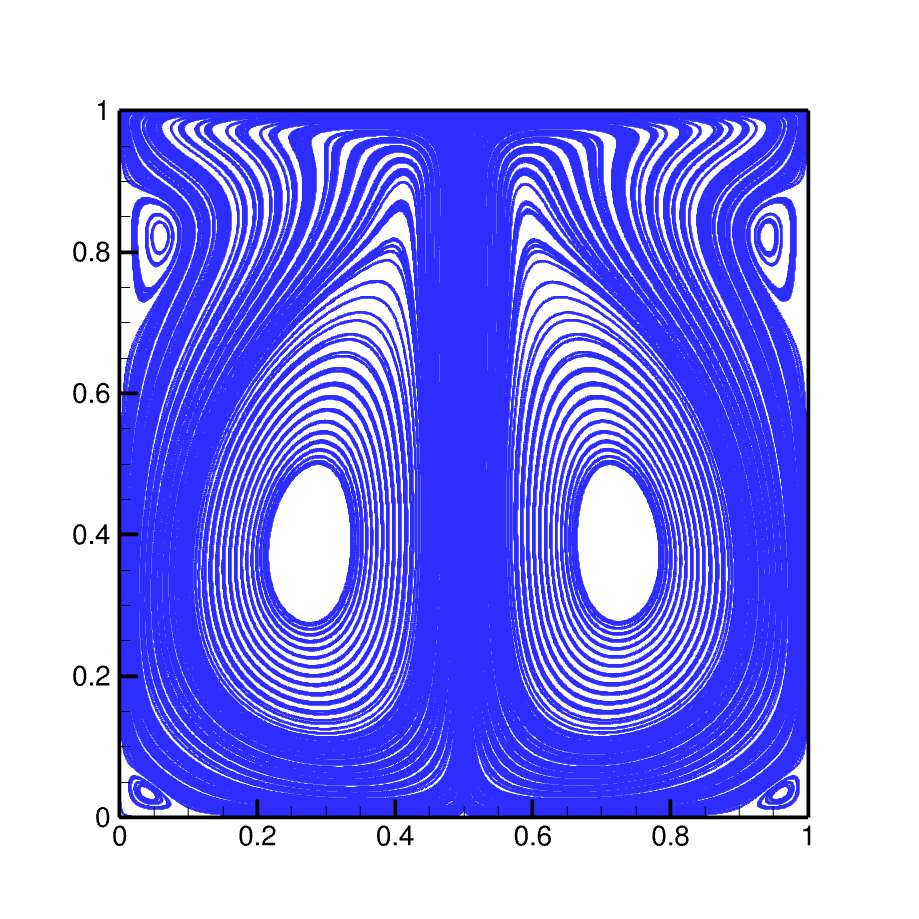}
        \includegraphics[scale=0.1]{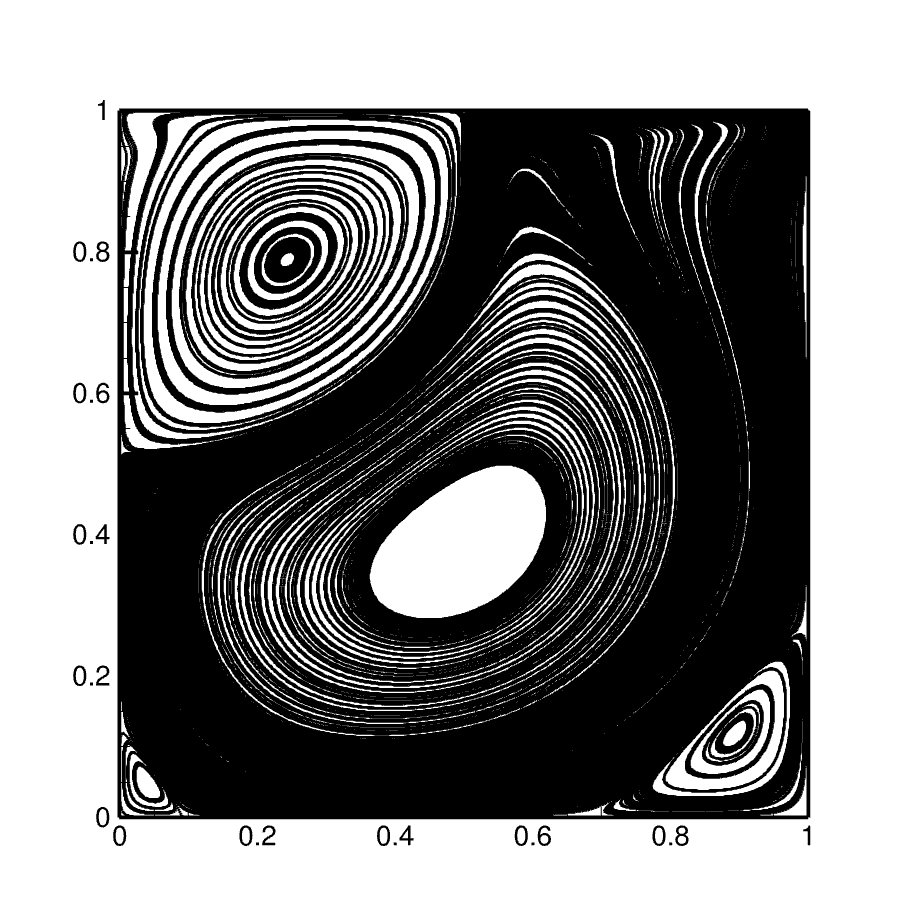}
        \includegraphics[scale=0.1]{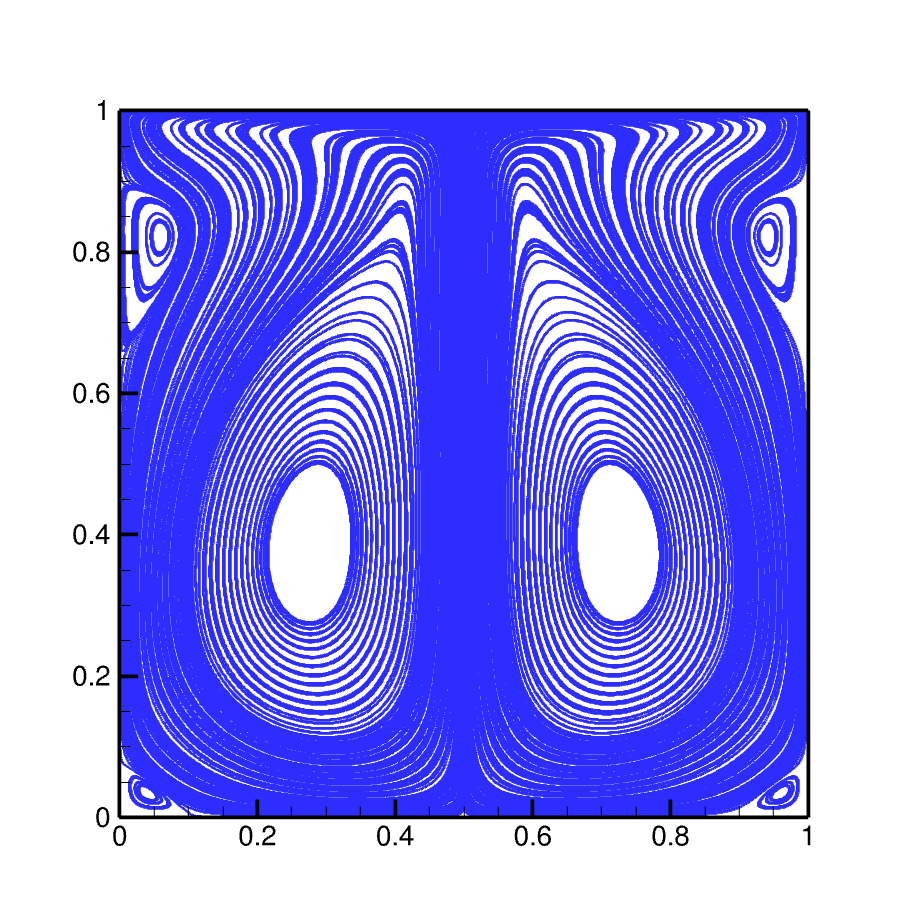}
        \includegraphics[scale=0.1]{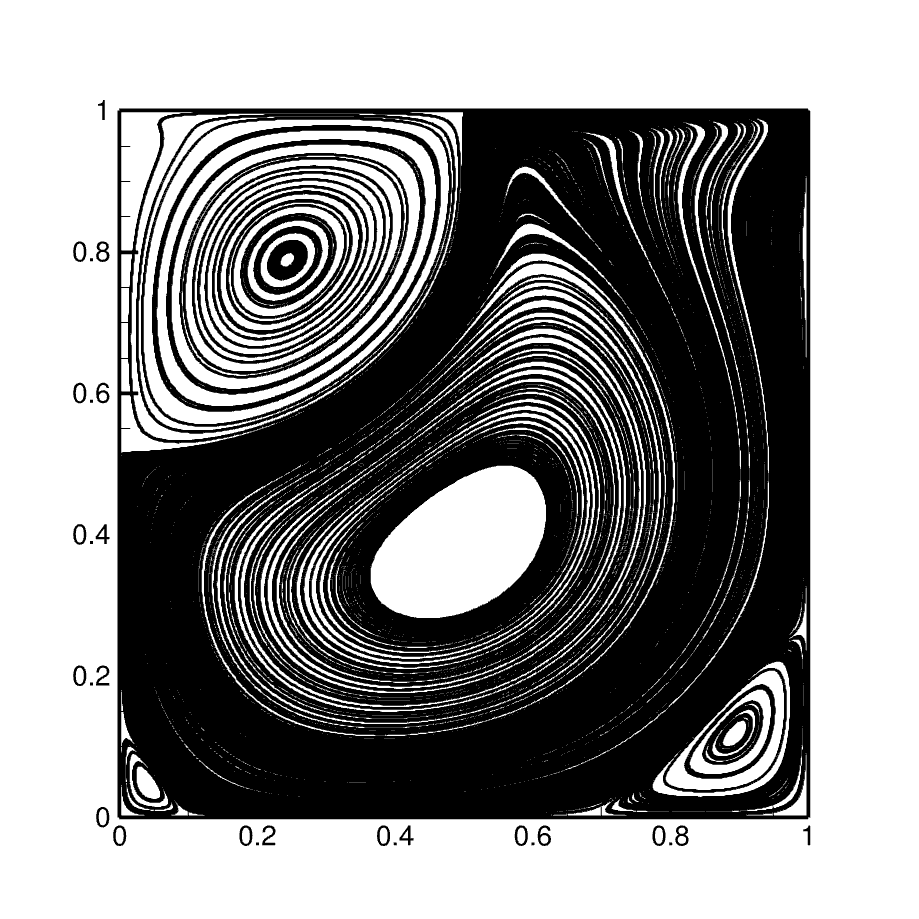}
        \includegraphics[scale=0.1]{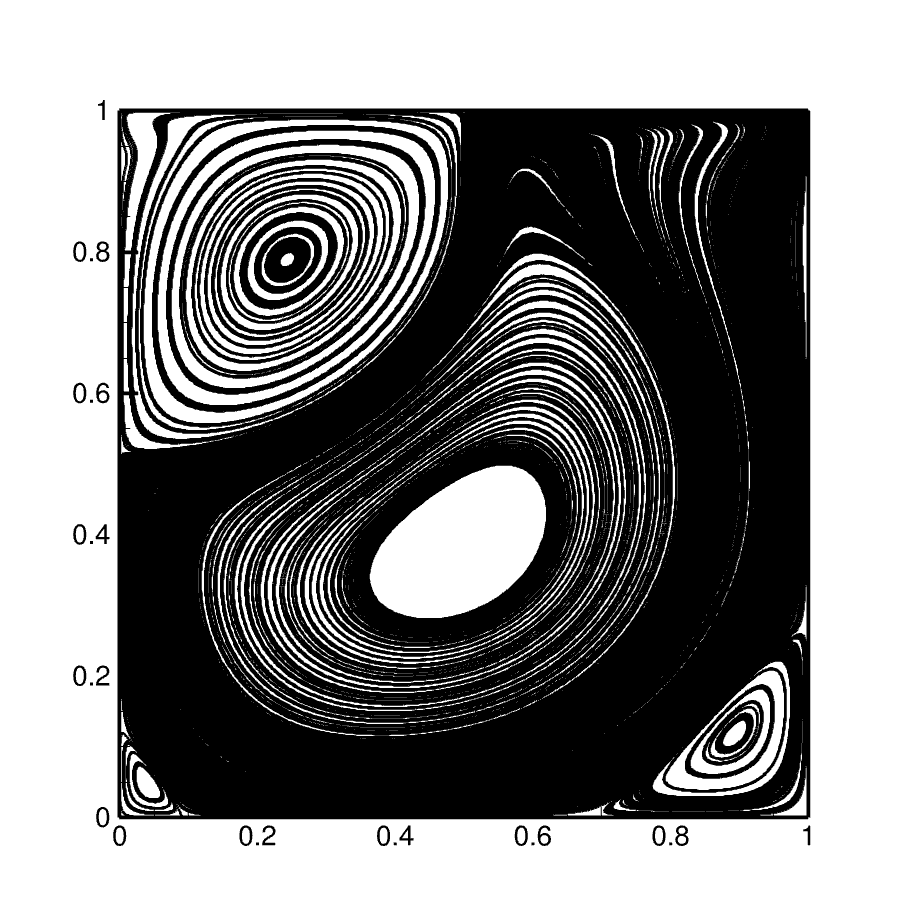}
    }
    \caption{Solution multiplicity for the cavity flow. The {\color{blue}\textbf{blue}} streamlines are from \emph{stable symmetric} solution, the {\color{green}\textbf{green}} streamlines are from the \emph{unstable asymmetric} solution, while the \textbf{black} streamlines are from the \emph{stable asymmetric} solution.}
    \label{fig:cavity_result}
\end{figure}

We consider a 2D steady-state lid-driven cavity flow described by the following Navier-Stokes equations:
\begin{subequations}\label{eq:2d_ns}
    \begin{align}
        \nabla\cdot\mathbf{u} &= 0, \text{in }\Omega,\\
        \mathbf{u}\cdot\nabla\mathbf{u} &= -\nabla p + \frac{1}{\text{Re}}\nabla^2 \mathbf{u}, \text{in }\Omega.
    \end{align}
\end{subequations}
The computational domain is a unit square $[0,1]\times [0,1]$, with the top wall ($x \in[0,1], y=1$) moving at a velocity $u=\sin{2\pi x},\,v=0$. The remaining boundaries are solid walls.

A fully-connected NN is employed to approximate the solution to \eqref{eq:2d_ns} and it takes the spatial coordinates $\mathbf{x}$ as input and outputs $\mathbf{u},\,p$ and an auxiliary variable $r$, defined as: 
\begin{equation}\label{eq:ev}
    r = (u-u_m)e_1+(v-v_m)e_2.
\end{equation}
In \eqref{eq:ev}, the constants $u_m=0.5, \text{and}\, \, v_m=0.5$, while $e_1$ and $e_2$ are residuals derived from \eqref{eq:2d_ns}:
\begin{subequations}\label{eq:losses}
    \begin{align}
        e_1&=u\frac{\partial u}{\partial x}+v\frac{\partial u}{\partial y}+\frac{\partial p}{\partial x}-(\frac{1}{Re}+\nu_E)(\frac{\partial^2 u}{\partial x^2}+\frac{\partial^2 u}{\partial y^2}),\label{eq:loss1} \\
        e_2&=u\frac{\partial v}{\partial x}+v\frac{\partial v}{\partial y}+\frac{\partial p}{\partial y}-(\frac{1}{Re}+\nu_E)(\frac{\partial^2 v}{\partial x^2}+\frac{\partial^2 v}{\partial y^2}).\label{eq:loss2} 
    \end{align}
\end{subequations}
The artificial viscosity $\nu_E$ is computed as:
\begin{equation}\label{eq:nu_e}
  \nu_E=\min(\beta \nu , \alpha |r|). 
\end{equation}
Note that $\nu_E$ is a scalar, constructed based on the entropy viscosity method \cite{guermond2011entropy, wang2019entropy} for numerical stabilization in the high-$Re$ flow simulations. In this study, we set $\alpha=0.03$ and $\beta=5$.
Additionally, the PINN optimization problem incorporates two more loss functions:
\begin{equation}
e_3 = \frac{\partial u}{\partial x}+\frac{\partial v}{\partial y},
\end{equation}
and we incorporate \eqref{eq:ev} into the neural network loss function as a residual term:
\begin{equation} \label{EV1}
e_4=(u-u_m)e_1+(v-v_m)e_2-r.
\end{equation} 
For further details and investigation on this parameterized entropy-viscosity method for solving the Navier-Stokes equations, refer to \cite{wang2023solution}.

In this example, we randomly initialize five NNs and train them independently. Their results are presented in Figure \ref{fig:cavity_result}(a). Three distinct types of solutions emerged from these five independently trained PINNs: a stable symmetric solution ({\color{blue}\textbf{blue}} streamlines), an unstable asymmetric solution ({\color{green}\textbf{green}} streamlines), and a stable asymmetric solution (\textbf{black} streamlines). Notably, these three solution types closely match numerical results from previous studies \cite{FariasPoF202, osman2004multiple, MA_Cavity_2008}. However, in traditional numerical simulations, obtaining multiple solutions requires prior knowledge to construct an appropriate initial guess—for instance, initializing the velocity field with multiple vortices \cite{osman2004multiple}. In contrast, PINNs, with different random initializations, can naturally discover all three solution types.

Furthermore, the five PINN-generated solutions were used as initial conditions in two numerical solvers: a 2D steady Navier-Stokes finite element solver implemented in Deal.II \cite{bangerth2007deal, africa2024deal} and a 2D unsteady Navier-Stokes spectral element solver implemented in Nektar \cite{karniadakis2005spectral}. In the steady solver, the computational mesh consists of $64\times 64$ uniform quadrilateral elements, using P2 Lagrange elements for velocity and P1 Lagrange elements for pressure. In the unsteady solver, the mesh contains the same number of quadrilateral elements, but third-order spectral elements are used for both velocity and pressure. The stopping criterion for the steady solver is an $L^2$-norm residual below $10^{-12}$, while for the unsteady solver, the simulation runs for $2\times 10^4$ steps with a time step of $2\times 10^{-3}$. 
The numerical results are shown in Figures \ref{fig:cavity_result}(b) and \ref{fig:cavity_result}(d). The steady Navier-Stokes solver preserves the same solution type as the corresponding PINN solution. However, in the unsteady Navier-Stokes solver, the unstable asymmetric solution gradually transitions into a stable asymmetric solution, as seen by comparing Figures \ref{fig:cavity_result}(c) and \ref{fig:cavity_result}(d).

\section{Summary}\label{sec:4}

We have explored the ability of physics-informed neural networks (PINNs) to capture multiple solutions of nonlinear differential equations (DEs). Many real-world problems governed by nonlinear DEs exhibit solution multiplicity, but traditional numerical methods often struggle to identify multiple valid solutions due to their sensitivity to initial conditions. Although one PINN can only represent one solution, an ensemble of PINNs, which are randomly initialized and independently trained, are capable of representing multiple distinct solutions. In this regard, we have proposed a general approach that leverages PINNs with random initialization and deep ensemble techniques -- originally developed for uncertainty quantification -- to systematically learn and discover multiple solutions of nonlinear ordinary and partial differential equations (ODEs/PDEs).

Our results demonstrate that randomness in initialization plays a critical role in capturing solution diversity, addressing a gap in the literature on machine learning-based solvers for DEs. Furthermore, we propose integrating PINN-generated solutions as initial conditions or guesses for conventional numerical solvers (finite difference, finite element, and spectral element methods) to enhance accuracy and efficiency. Through extensive numerical experiments, including the Allen-Cahn equation and cavity flow problems in which the proposed method successfully identifies both stable and unstable solutions, we validate our approach and establish a general framework for solving nonlinear DEs with solution multiplicity.

In summary, this work introduces a simple yet effective method for discovering multiple solutions using ensemble PINNs, extends the applicability of PINNs beyond existing solution-multiplicity approaches, and highlights their potential synergy with traditional numerical solvers. 
Future research could further enhance training efficiency by incorporating techniques such as mixed precision for scientific machine learning (SciML) \cite{hayford2024speeding} and developing strategies to improve the diversity of neural network outputs \cite{zhang2020diversified}, thereby reducing the required number of PINNs to achieve the same solution diversity.

\section*{Acknowledgments}

This work is supported by the MURI/AFOSR FA9550-20-1-0358 project and the DOE-MMICS SEA-CROGS DESC0023191 award.
ZZ thanks Dr. Ziyao Xu from University of Notre Dame and Professor Xuhui Meng from Huazhong University of Science and Technology for helpful discussions and suggestions.

\bibliography{main}

\appendix

\section{Details of computational results}\label{sec:appendix_1}

Here we include details of PINNs in Section \ref{sec:3}. In all examples, the hyperbolic tangent is chosen as the activation function and the Adam optimizer is employed, as suggested by the literature of PINNs (e.g., \cite{raissi2019physics, cai2021physics, jin2021nsfnets, wang2023expert}). We note that all weights and biases of NNs are initialized with the same initialization method in all examples excluding the cavity flows.
In Section \ref{sec:3_1}, we employ a fully-connected neural network (FNN) with two hidden layers, each of which is equipped with $50$ neurons. The number of training iterations is set to $20,000$ with $1\times 10^{-3}$ learning rate. For the results presented in Figure \ref{fig:example_1_1} in which different values of $\lambda$ are tested, these ten NNs are randomly initialized by the random normal distribution with mean zero and standard deviation $2$.
In Sections \ref{sec:3_2}, \ref{sec:3_3} and \ref{sec:3_5}, the FNN has three hidden layers and each layer has $50$ neurons, while in Section \ref{sec:3_6}, the FNN has four hidden layers and each layer has $120$ neurons. 
In Section \ref{sec:3_2}, NNs are first trained for $5,000$ iterations with $1\times 10^{-3}$ learning rate and then trained for $15,000$ iterations with $1\times 10^{-4}$ learning rate. In Sections \ref{sec:3_3} and \ref{sec:3_5}, NNs are first trained for $15,000$ iterations with $1\times 10^{-3}$ learning rate and then trained for $5,000$ iterations with $1\times 10^{-4}$ learning rate. In Section \ref{sec:3_6}, the training procedure is performed sequentially as follows:
\begin{enumerate}
    \item $20,000$ iterations with $1\times 10^{-3}$ learning rate,
    \item $20,000$ iterations with $2\times 10^{-4}$ learning rate,
    \item and $20,000$ iterations with $2\times 10^{-5}$ learning rate.
\end{enumerate}

Recall that we have tested the multi-head structure \cite{zou2023hydra} as one-parameter sharing architecture in Sections \ref{sec:3_1} and \ref{sec:3_3}. The multi-head NN consists of a shared body and multiple individual heads. Specifically, in Section \ref{sec:3_1}, the body is the first hidden layer while the head consists of the second hidden layer and the last output linear layer. In Section \ref{sec:3_3}, the body is the first two hidden layers and the head consists of the third hidden layer and the last output linear layer. We note that this setup is different from \cite{zou2023hydra}, in which the head is just the last output linear layer, making the body as a set of shared basis functions. 
In Section \ref{sec:3_1}, the multi-head PINNs are trained for $20,000$ iterations with $1\times 10^{-3}$ learning rate, while in Section \ref{sec:3_3}, the multi-head PINNs are first trained for $20,000$ iterations with $1\times 10^{-3}$ learning rate, then trained for $10,000$ iterations with $1\times 10^{-4}$ learning rate.

The data for training PINNs are generated as follows. In Section \ref{sec:3_1}, $101$ residual points $\{x_i^f, f_i\}_{i=1}^{N_f}$ are evenly sampled from the domain $[0, 1]$. In Section \ref{sec:3_2}, $100$ residual points are evenly sampled from the domain $[-0.5, 0.5]$. In Section \ref{sec:3_3}, $201$ residual points are evenly sampled from the domain $[-1, 1]$. In Section \ref{sec:3_5}, $3,000$ residual points are randomly sampled from the domain $(0, 1)^2$, following a uniform distribution, and $99$ boundary points are evenly sampled along each edge, excluding the four corner points. In Section \ref{sec:3_6}, $20,000$ residual points are randomly sampled from the domain $(0, 1)^2$.

\section{A 2D PDE}\label{sec:appendix_2}

\begin{figure}[ht]
    \centering
    \includegraphics[scale=0.10]{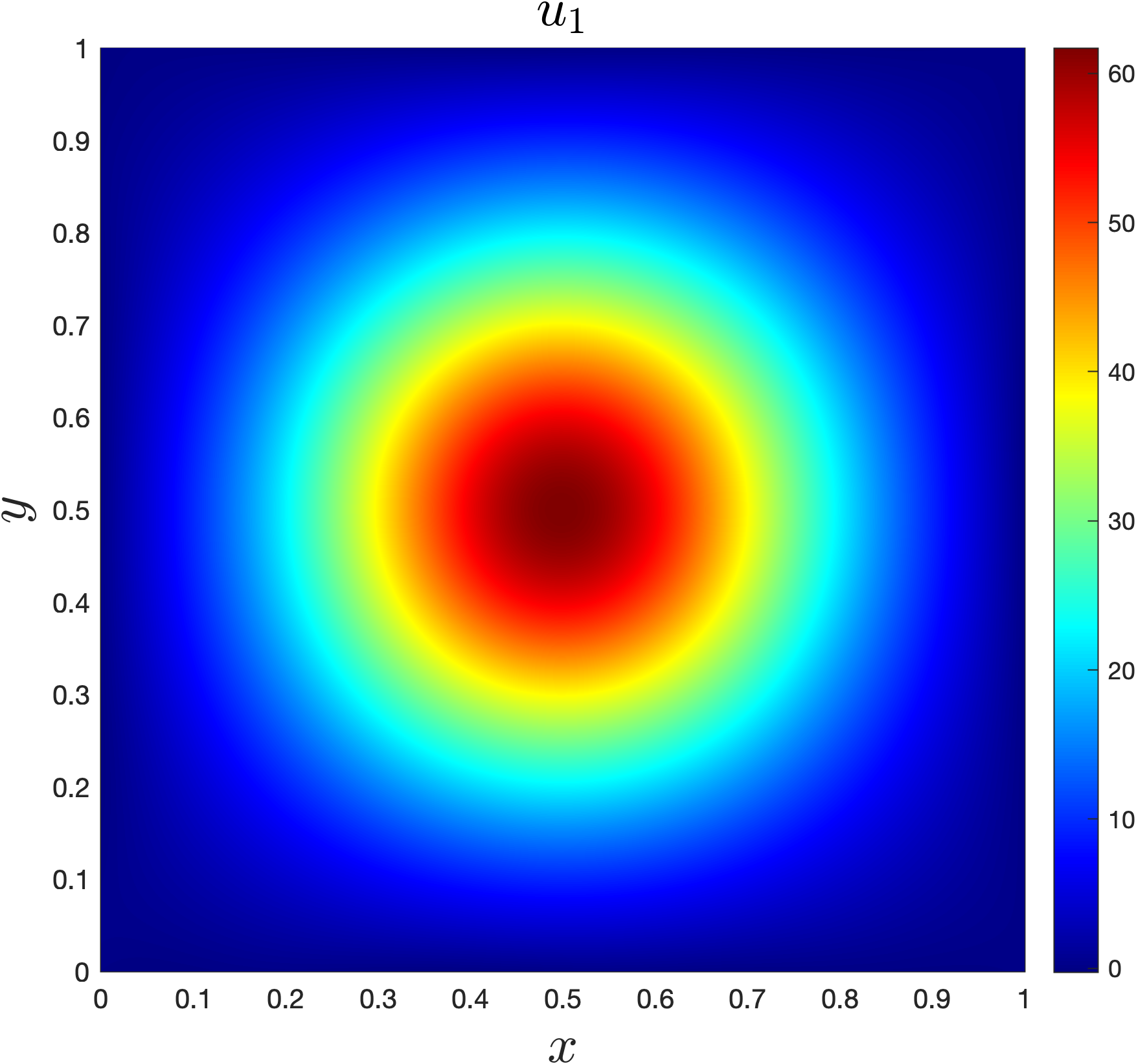}
    \includegraphics[scale=0.10]{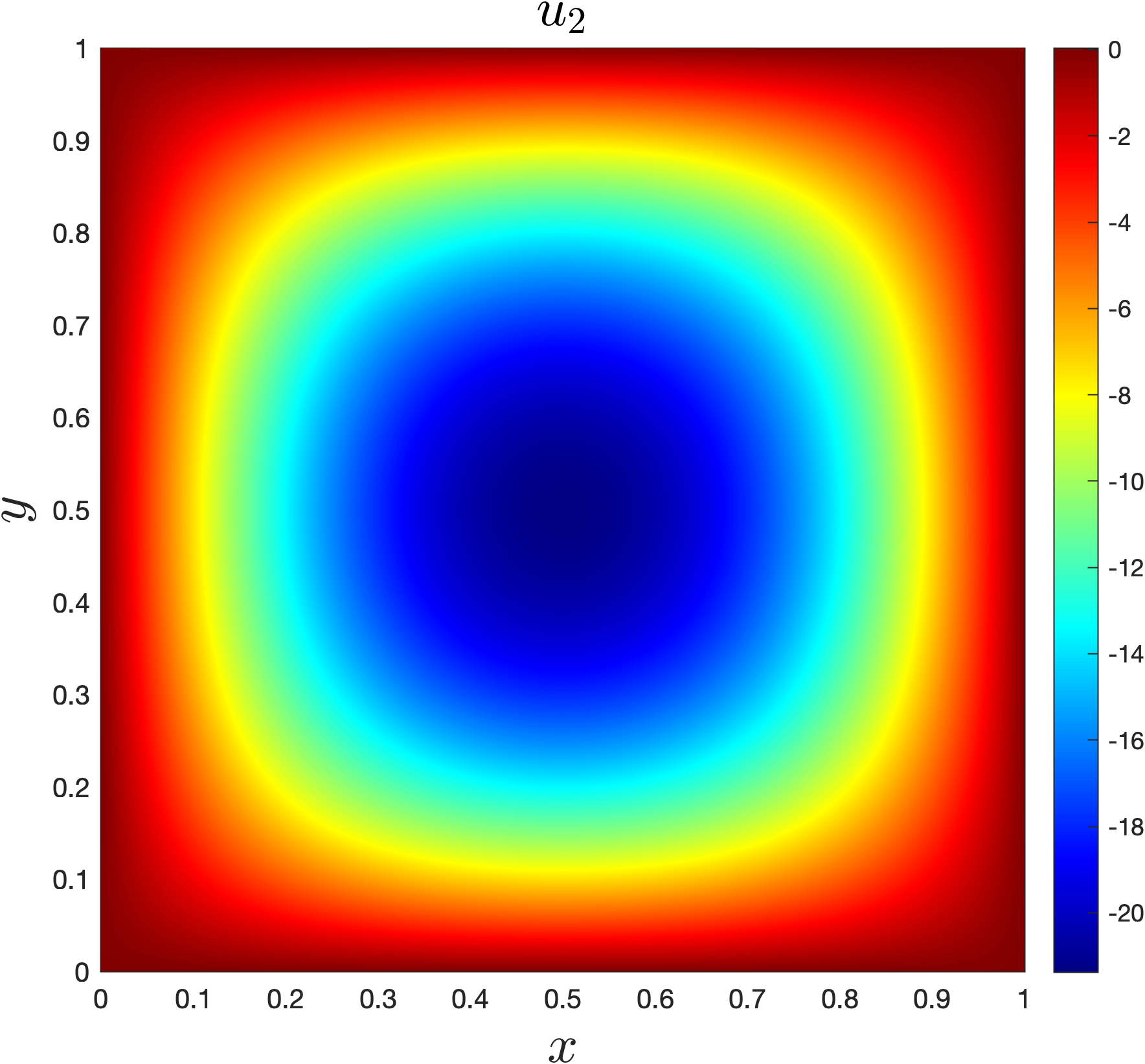}
    \includegraphics[scale=0.10]{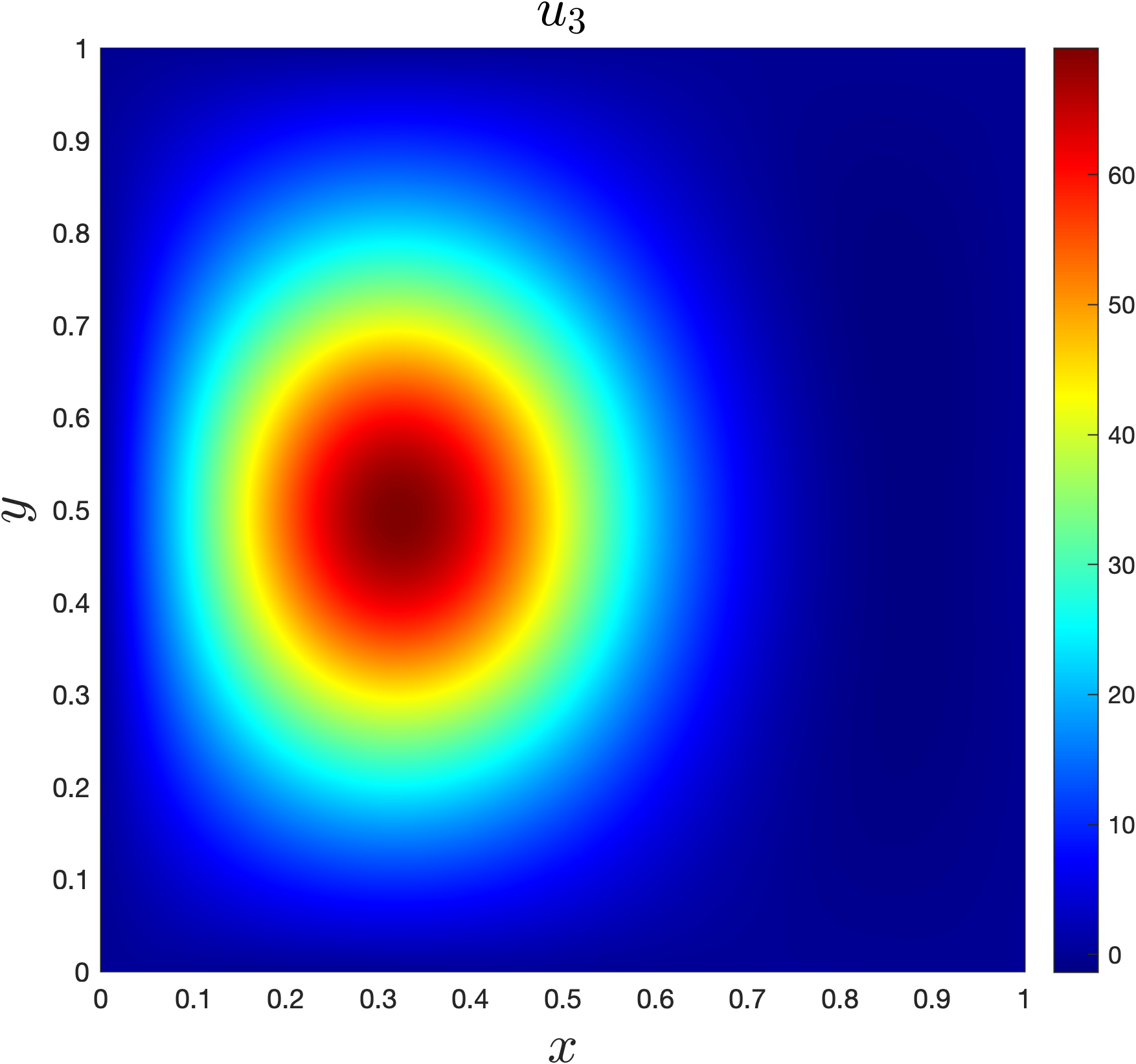}
    \includegraphics[scale=0.10]{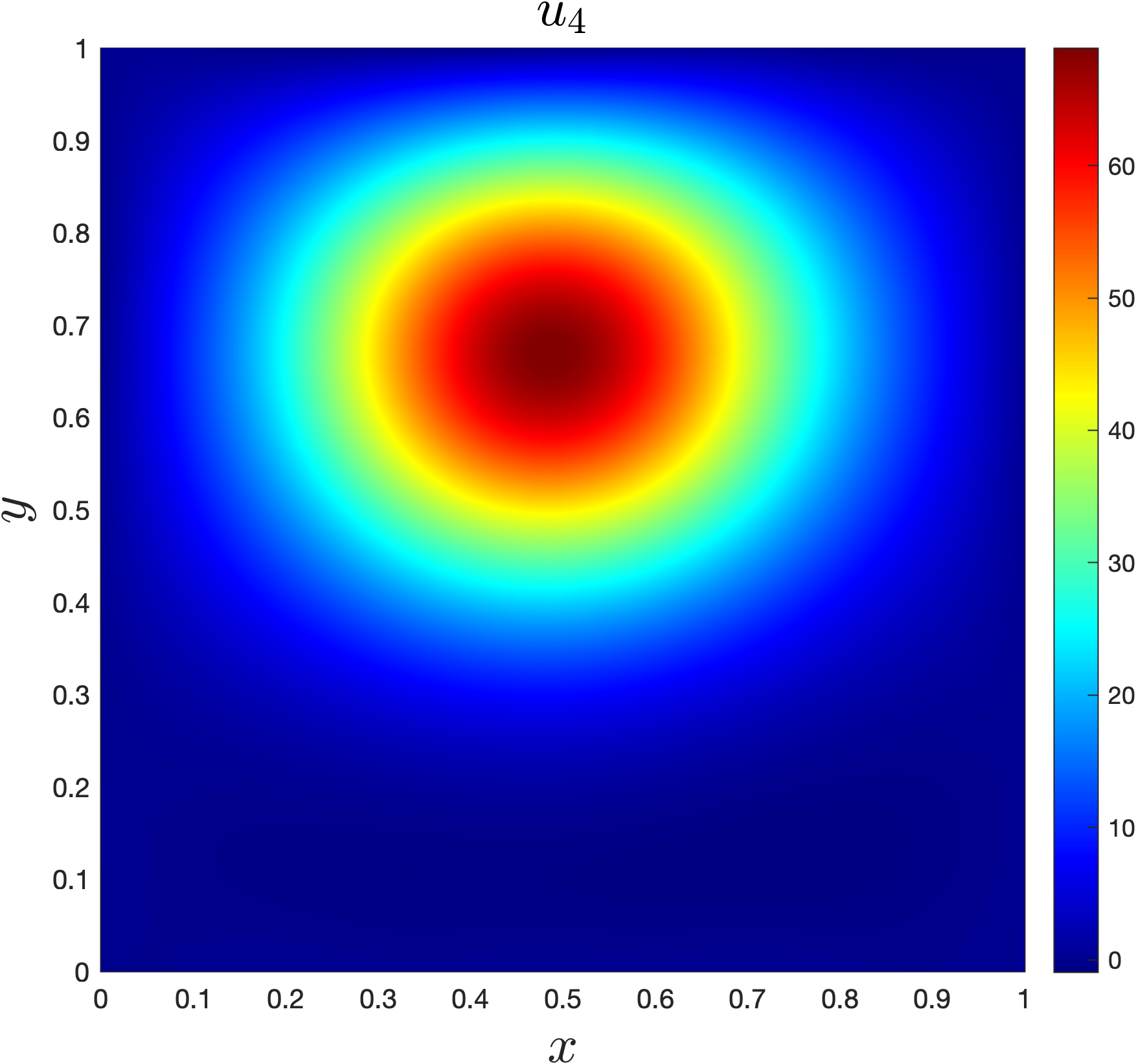}
    \includegraphics[scale=0.10]{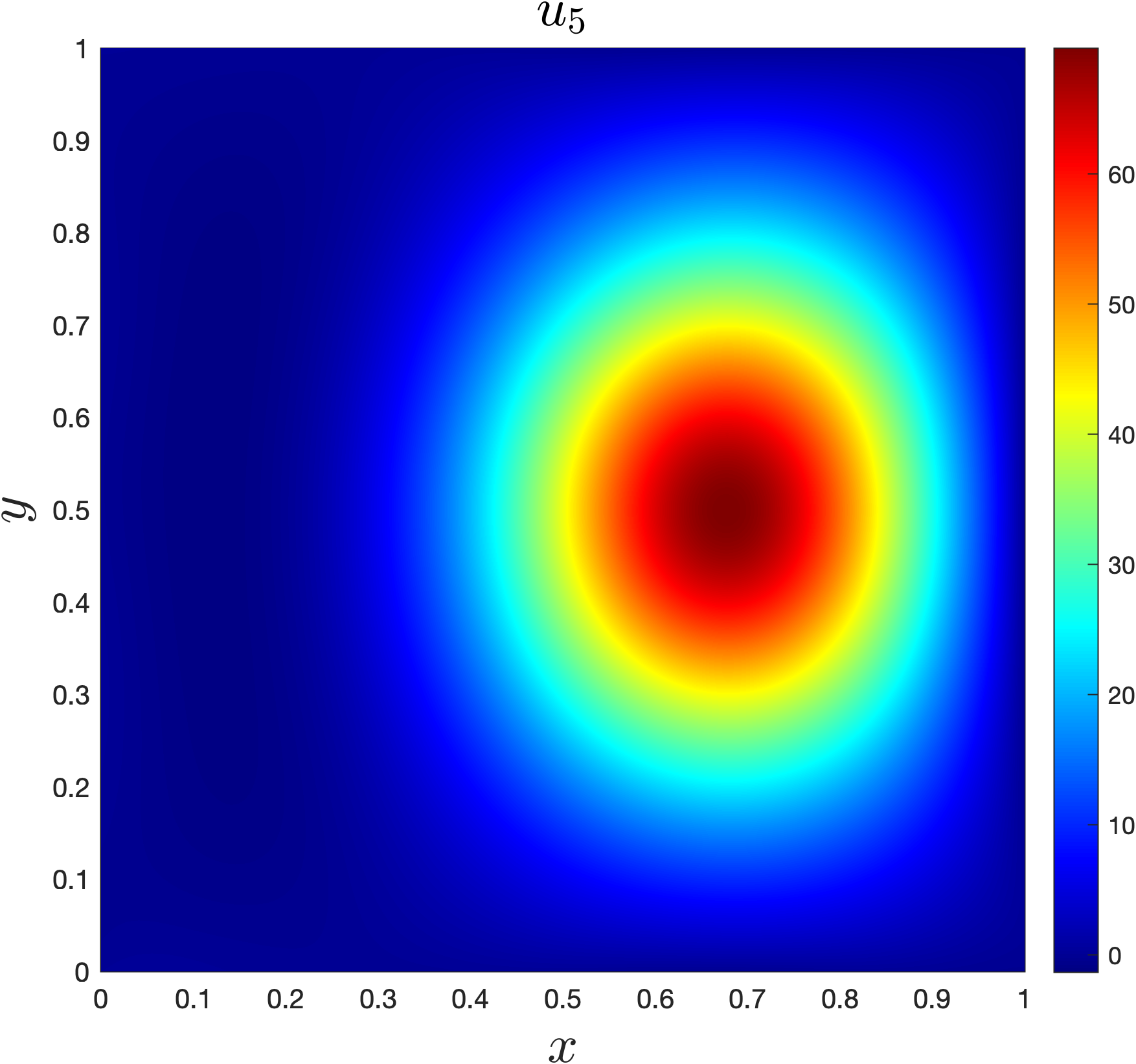}
    \includegraphics[scale=0.10]{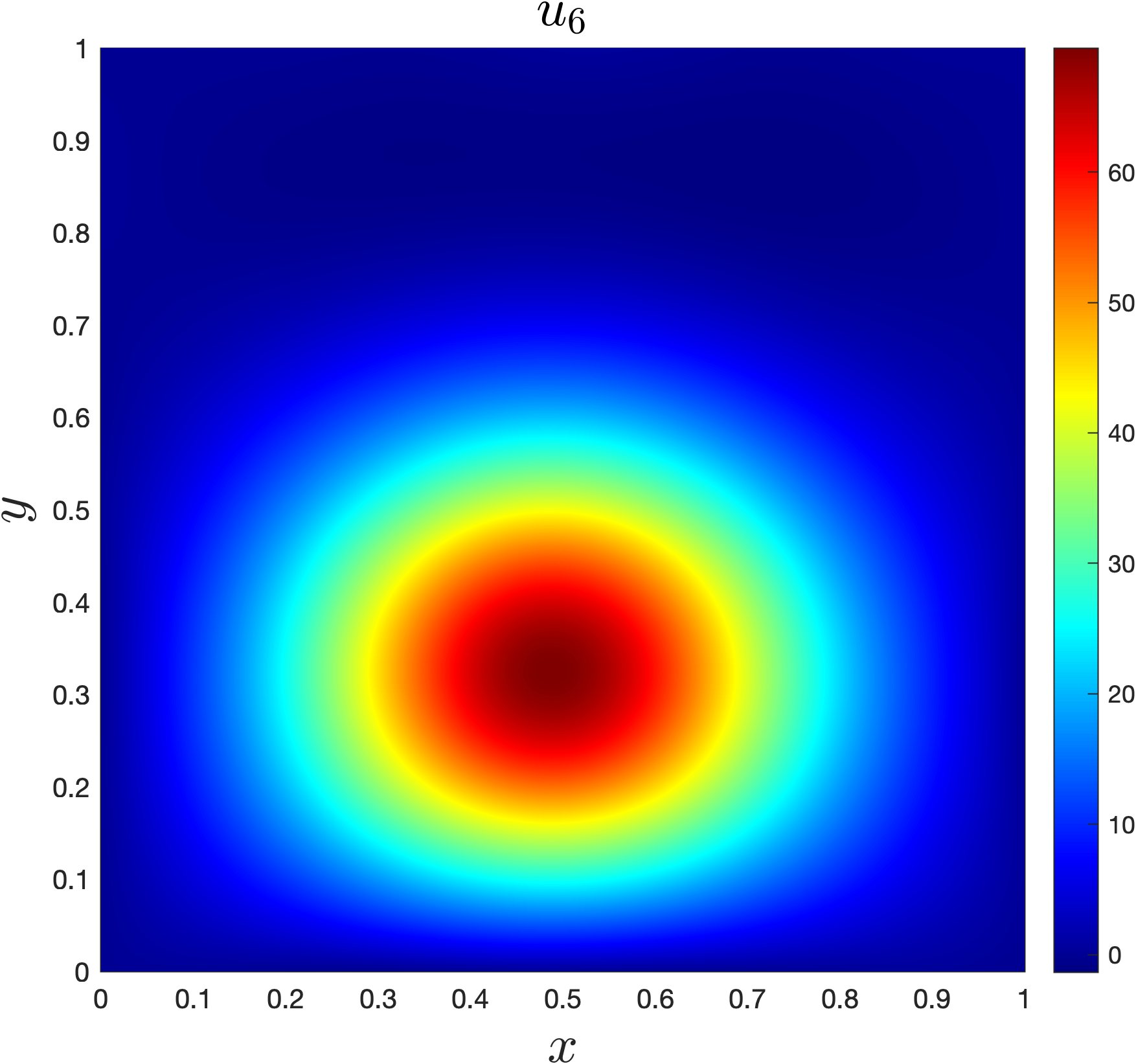}
    \includegraphics[scale=0.10]{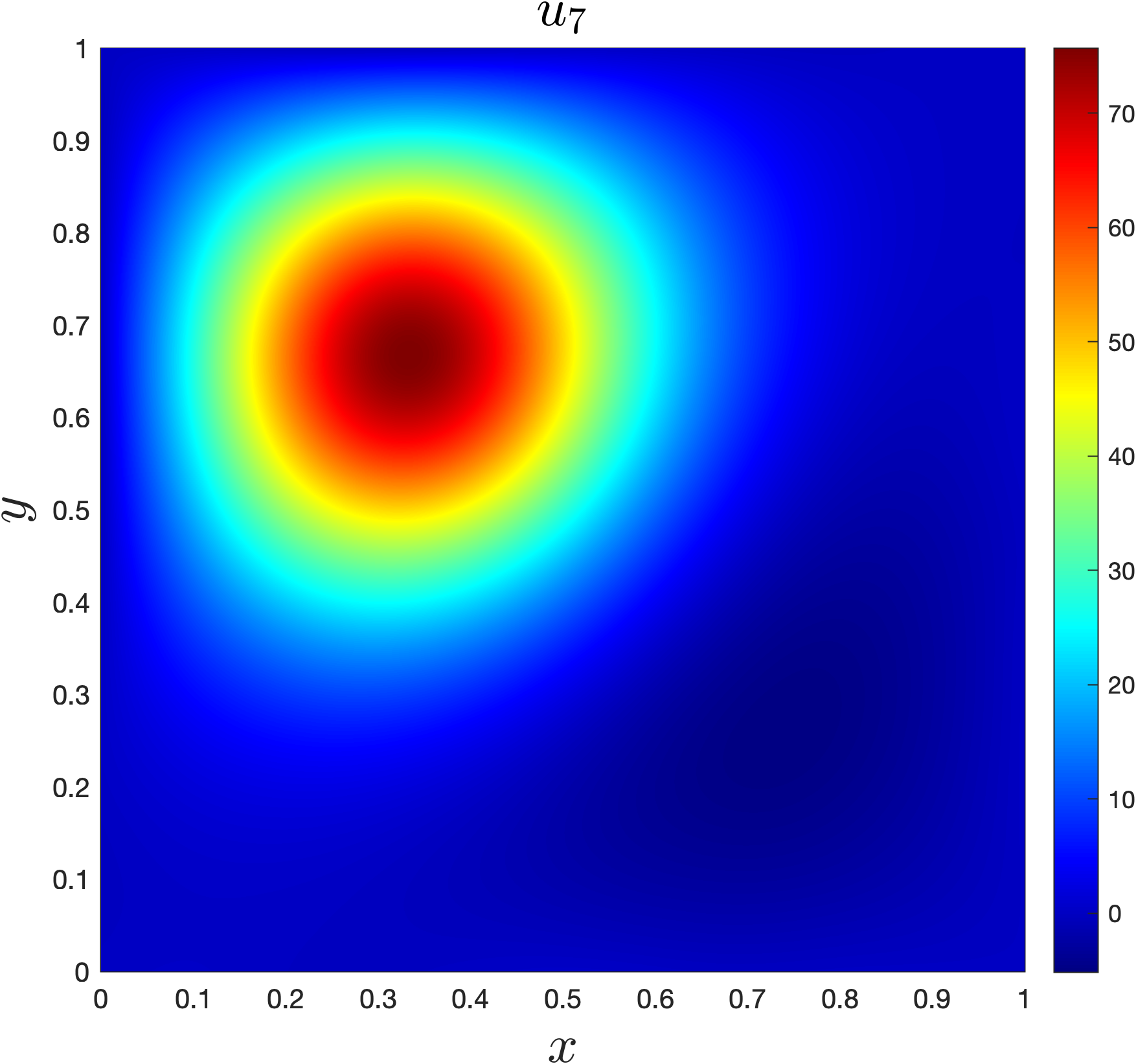}
    \includegraphics[scale=0.10]{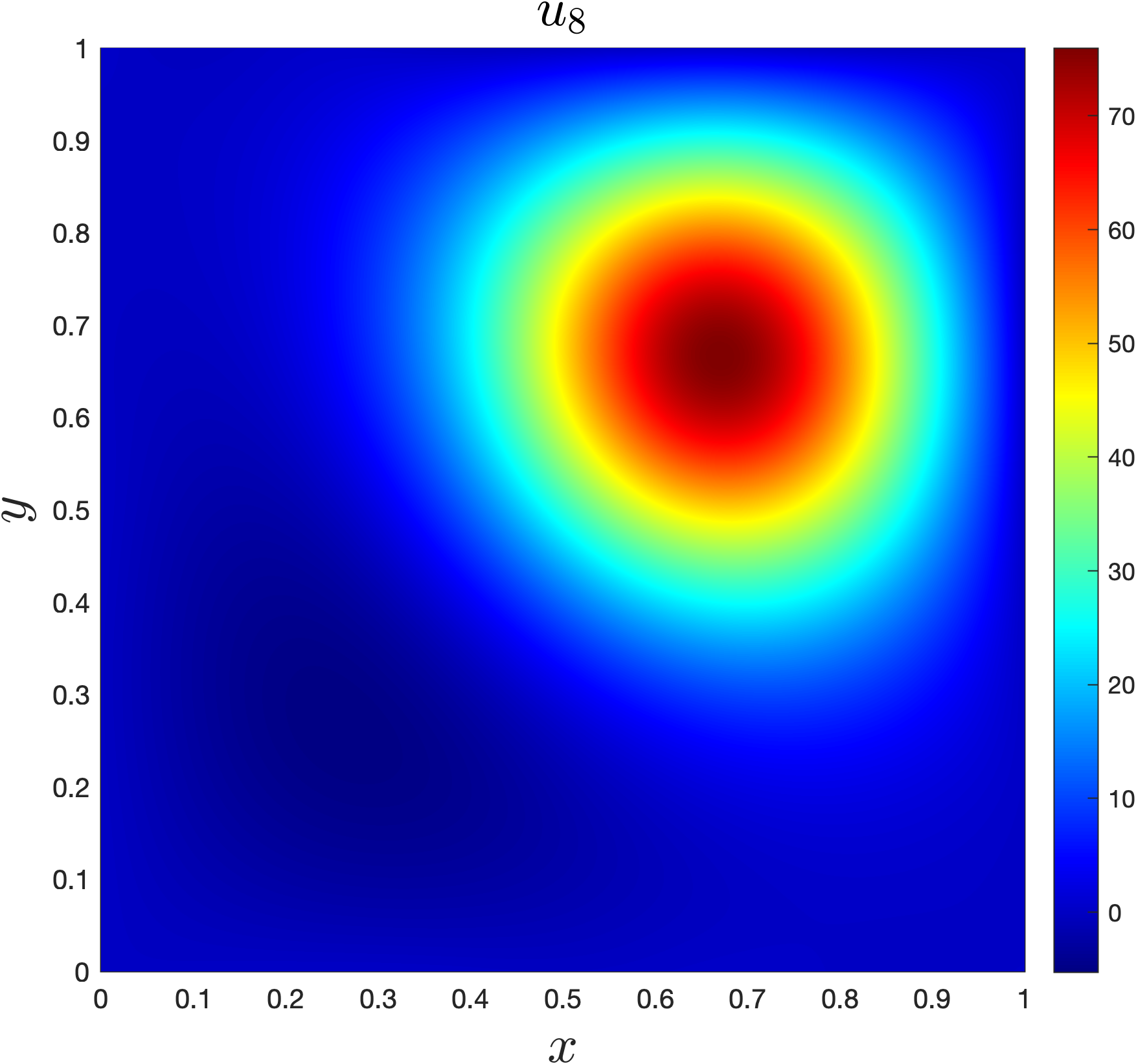}
    \includegraphics[scale=0.10]{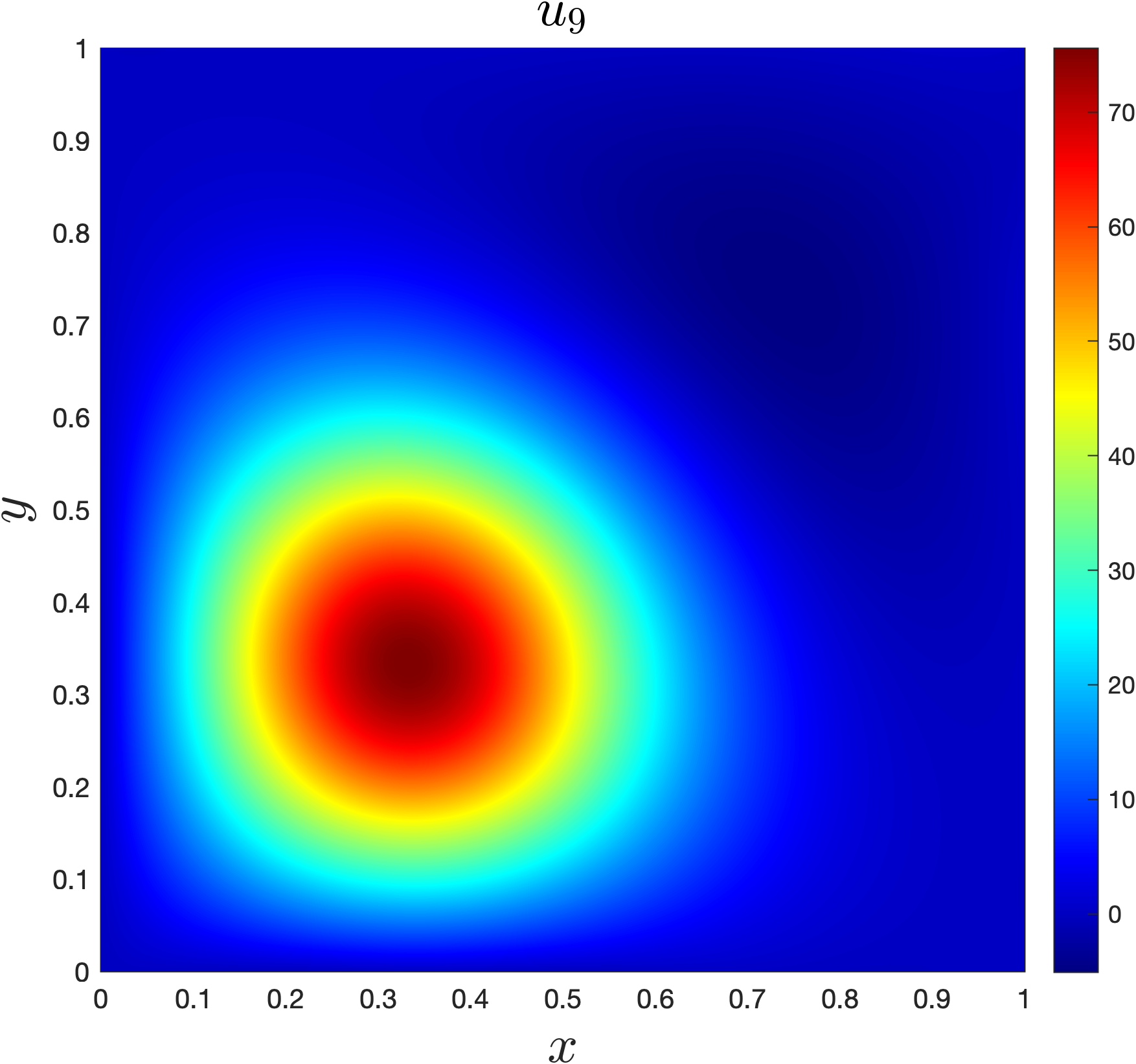}
    \includegraphics[scale=0.10]{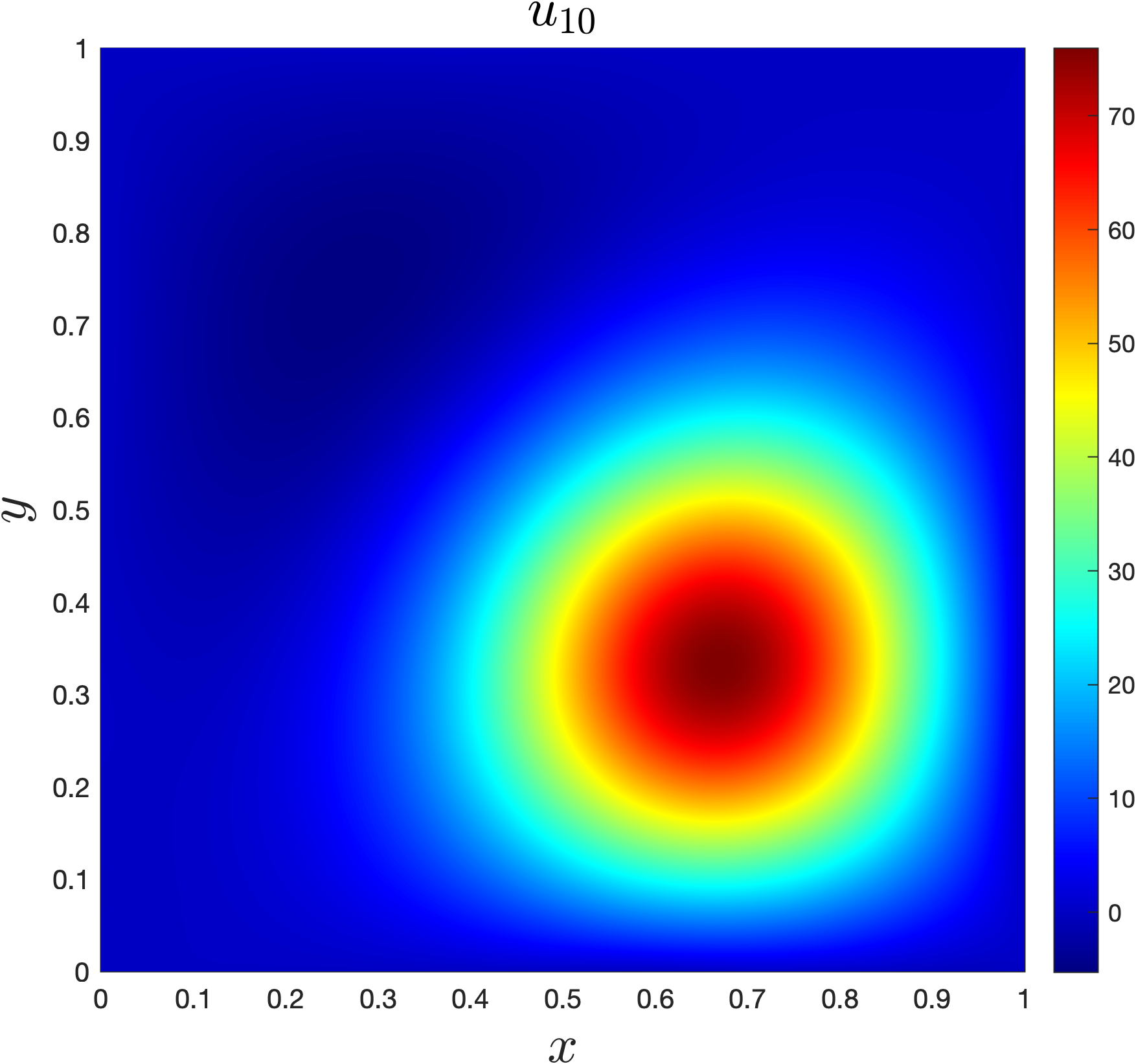}
    \caption{Ten distinct PINN solutions to \eqref{eq:2d_reaction}, which are chosen as the initial guesses for a FEM solver to solve \eqref{eq:2d_reaction}. Ten corresponding FEM solutions are presented in Fig. \ref{fig:example_4_2}.}
    \label{fig:example_4_2}
\end{figure}

\begin{figure}[ht]
    \centering
    \subfigure[FEM solutions obtained from initializing at distinct PINN solutions for the Gauss-Newton method.]{
        \includegraphics[scale=0.10]{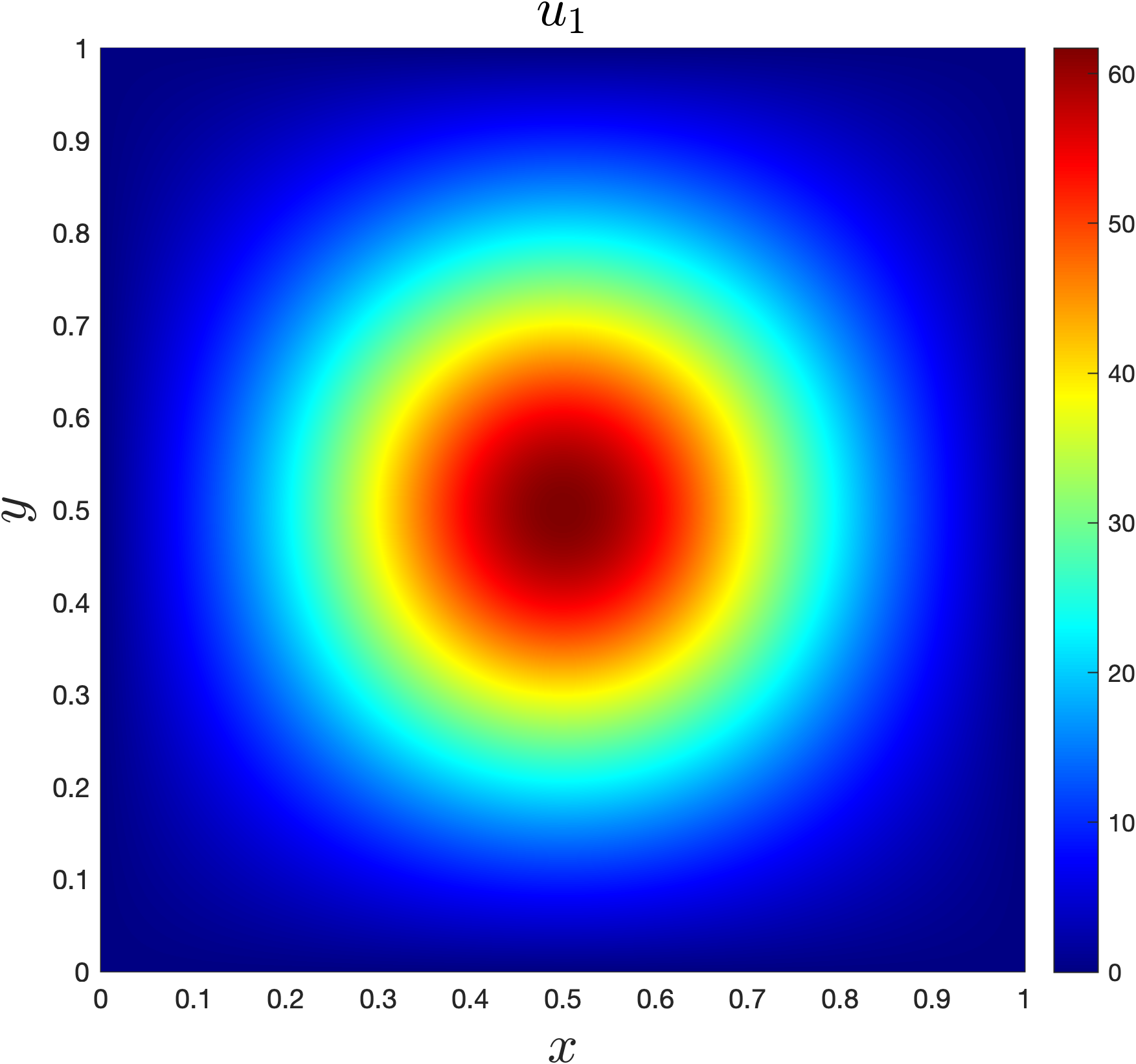}
        \includegraphics[scale=0.10]{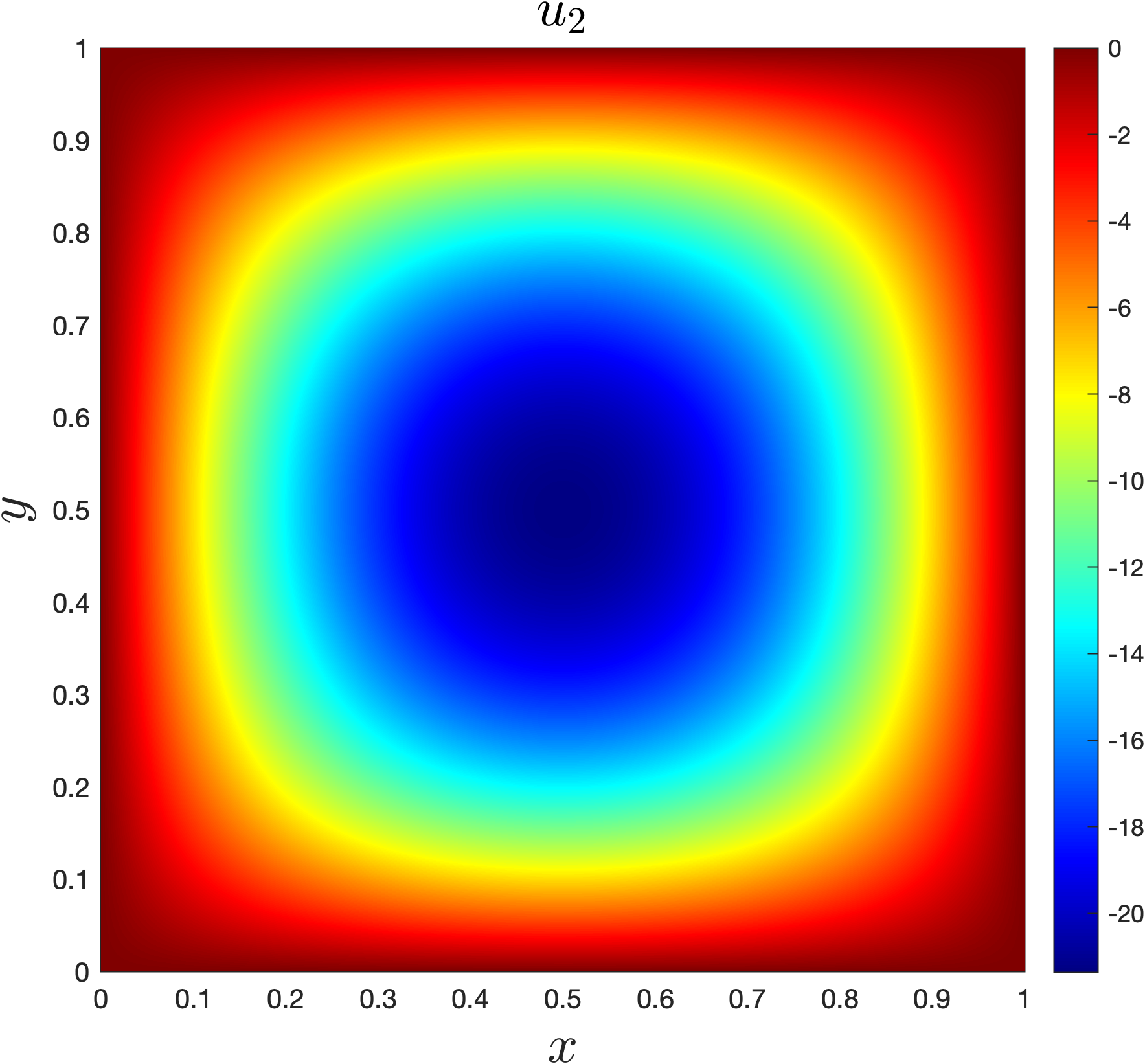}
        \includegraphics[scale=0.10]{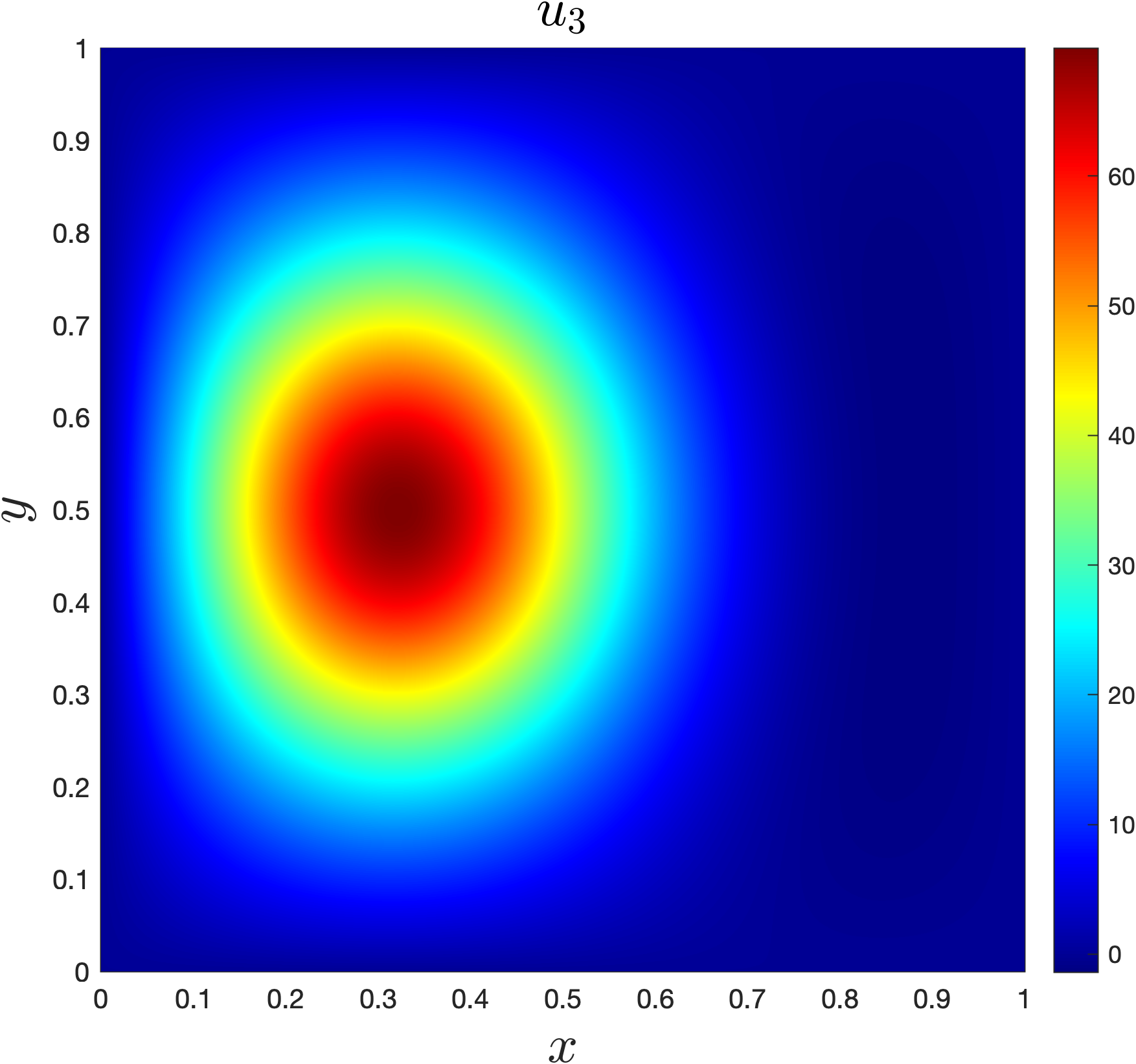}
        \includegraphics[scale=0.10]{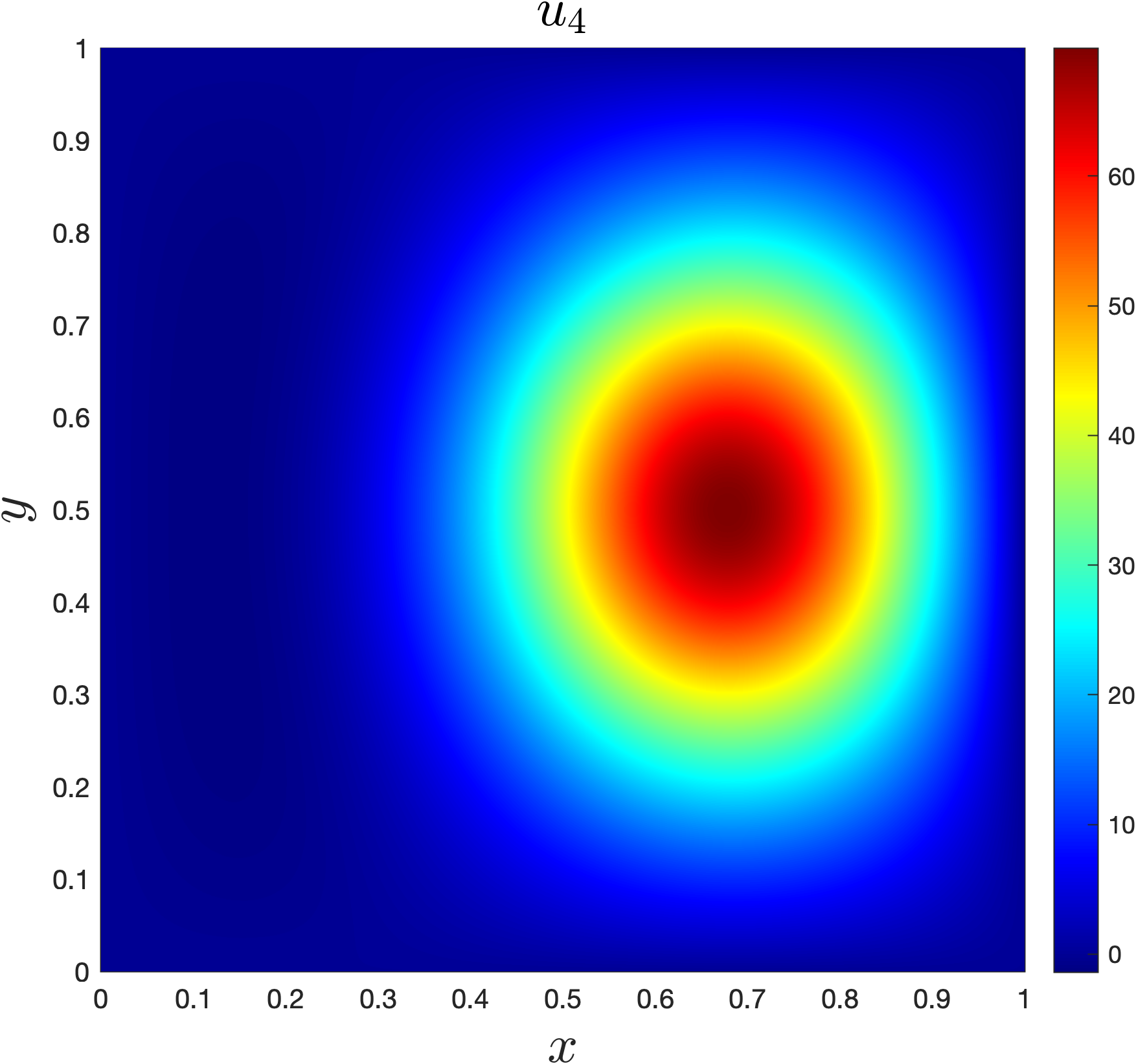}
        \includegraphics[scale=0.10]{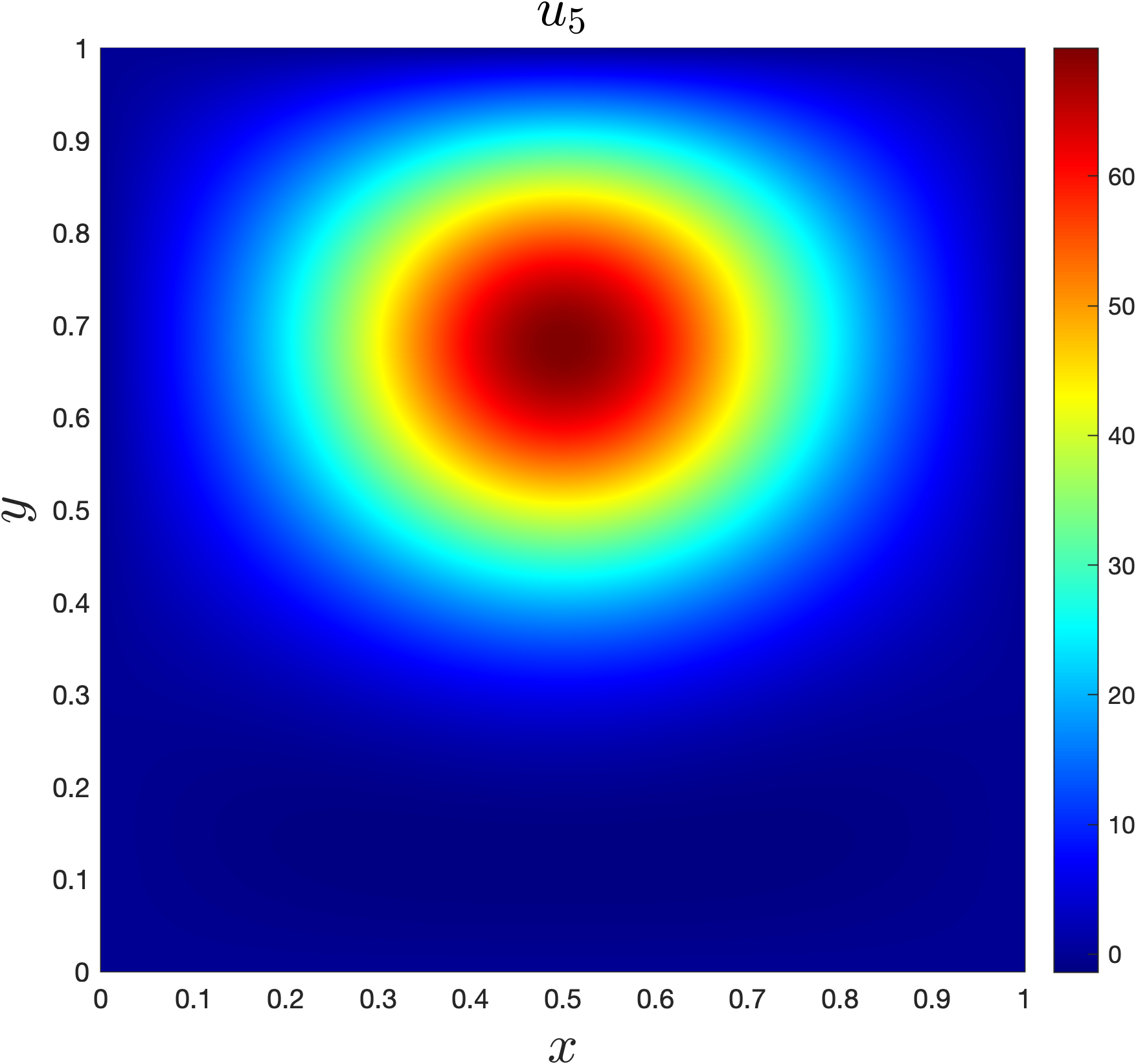}
        \includegraphics[scale=0.10]{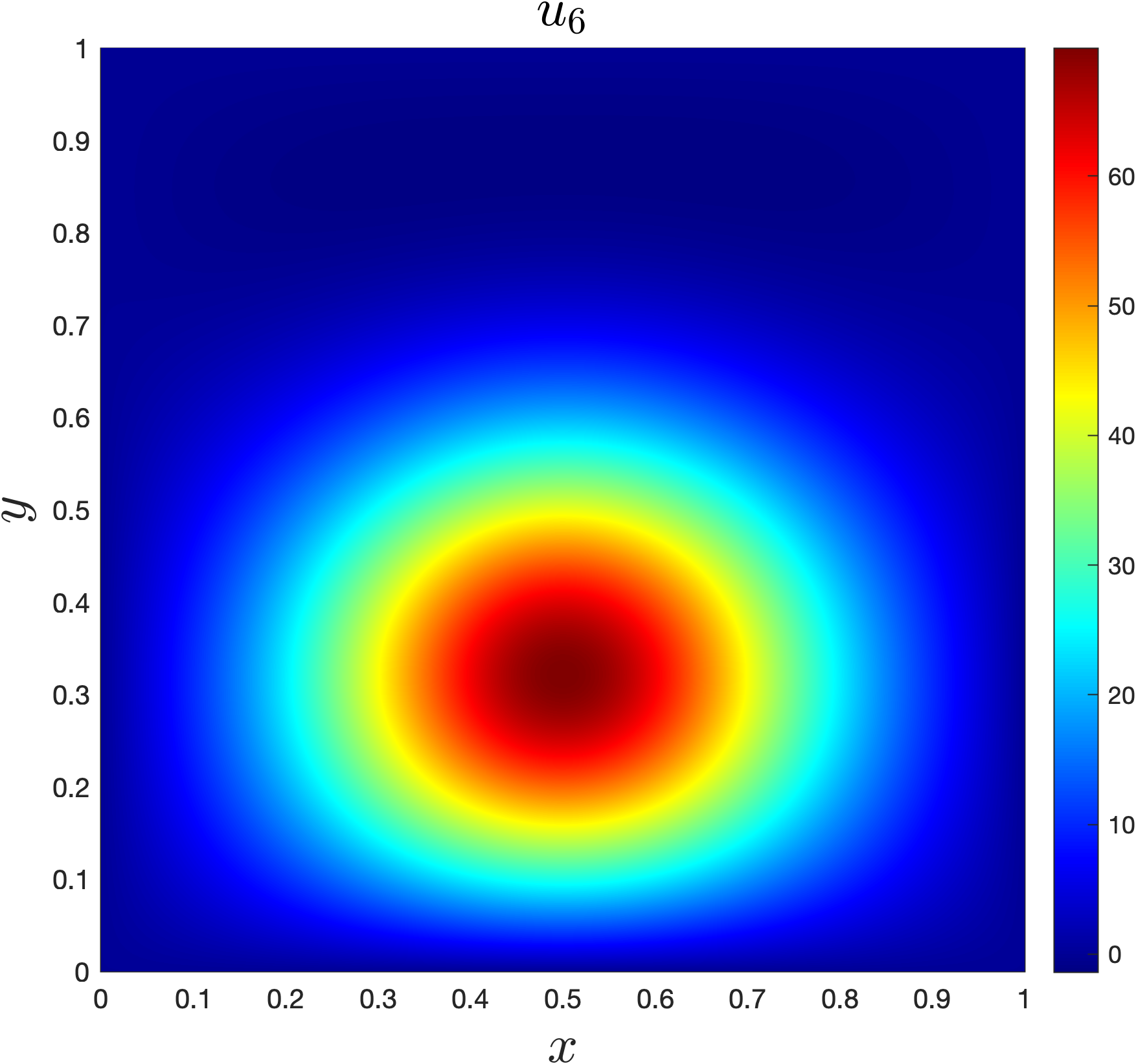}
        \includegraphics[scale=0.10]{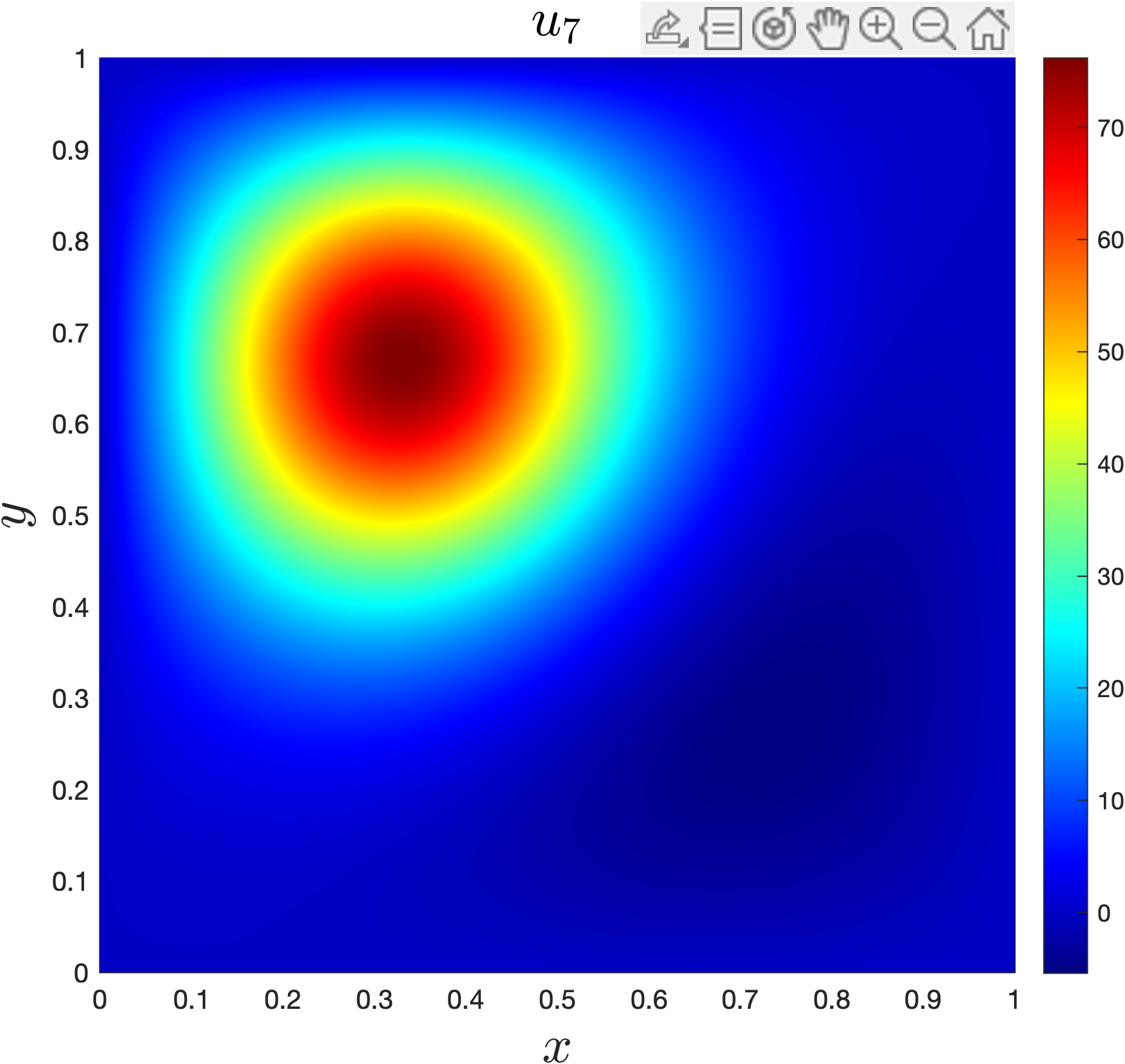}
        \includegraphics[scale=0.10]{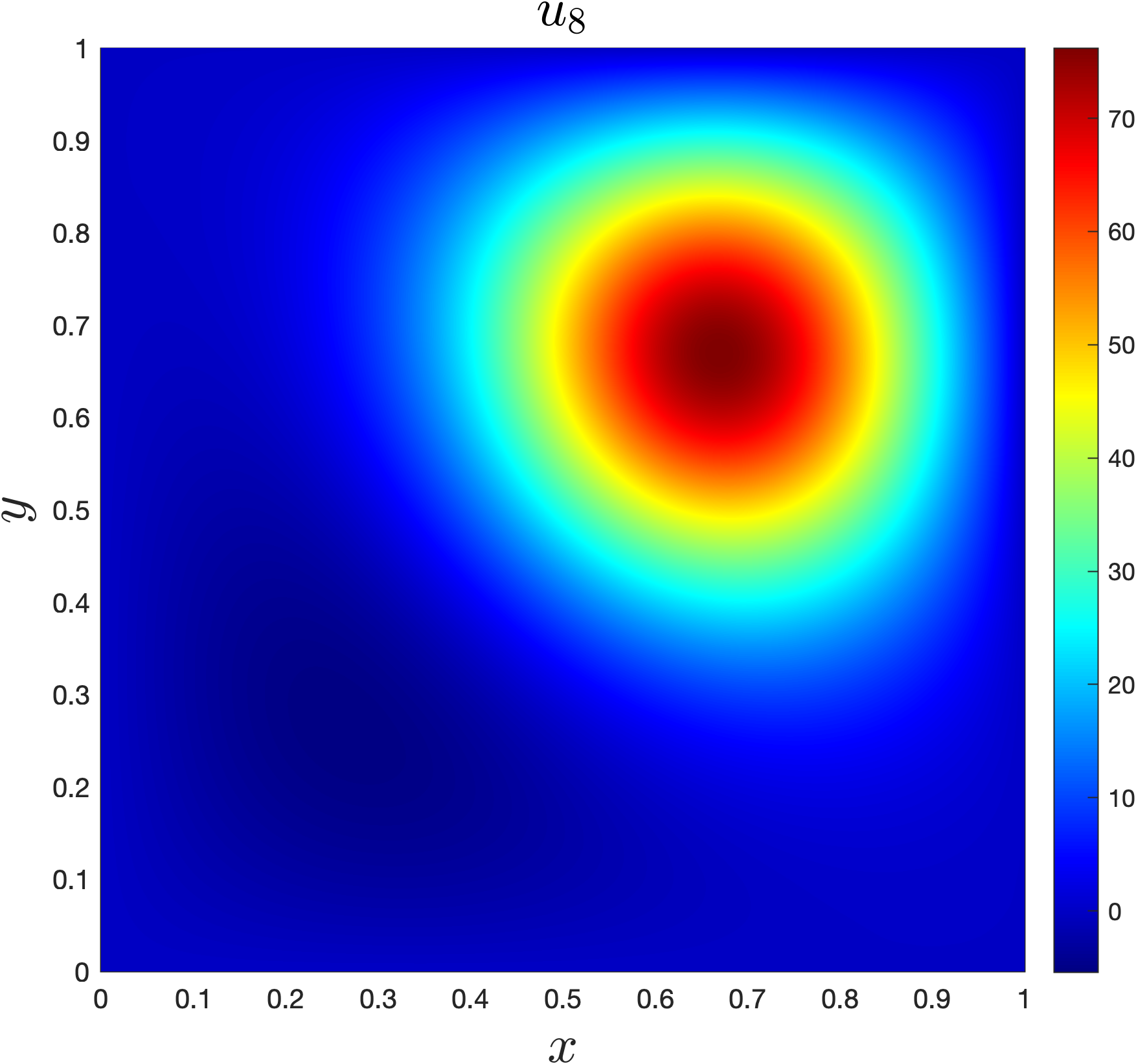}
        \includegraphics[scale=0.10]{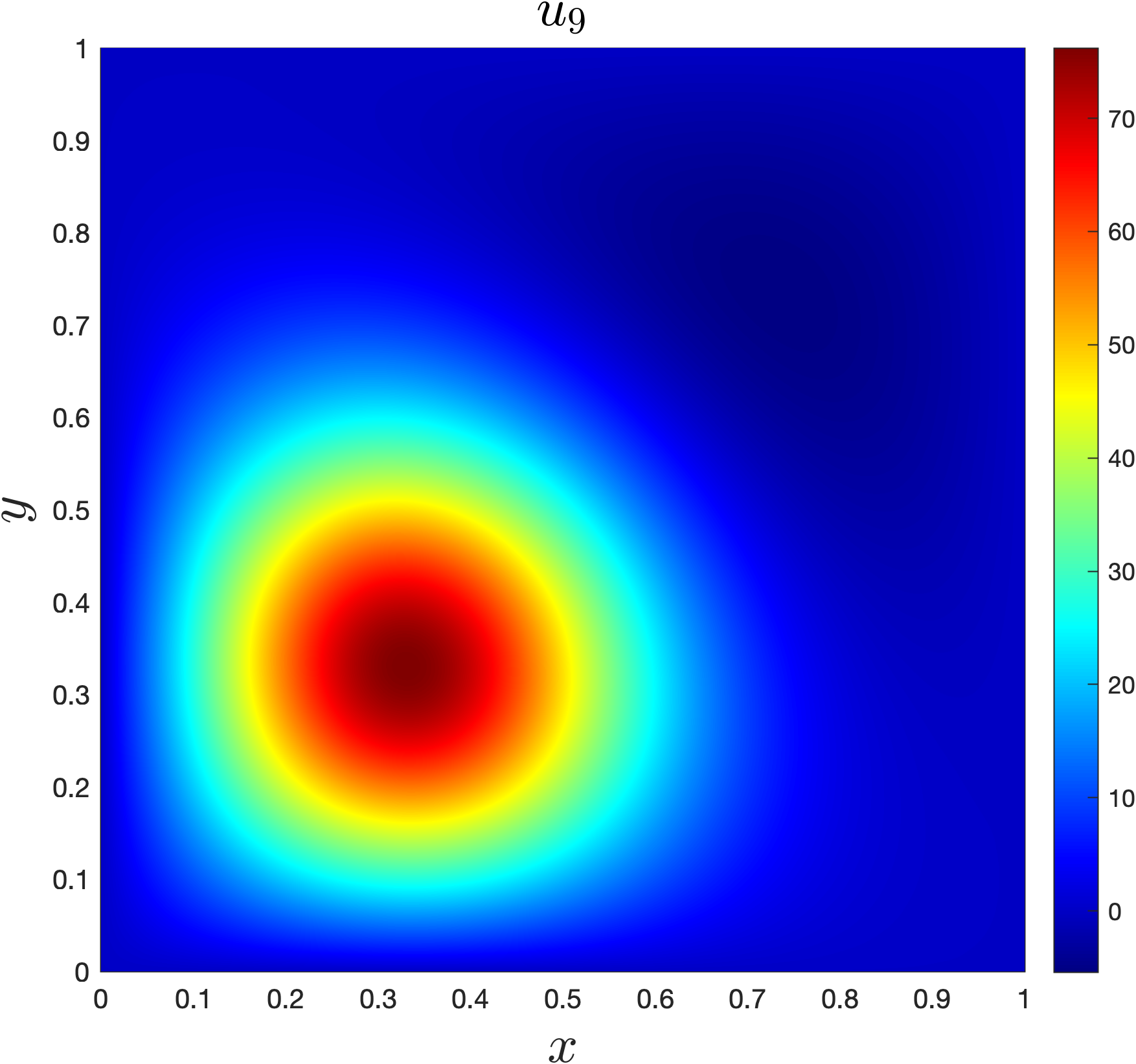}
        \includegraphics[scale=0.10]{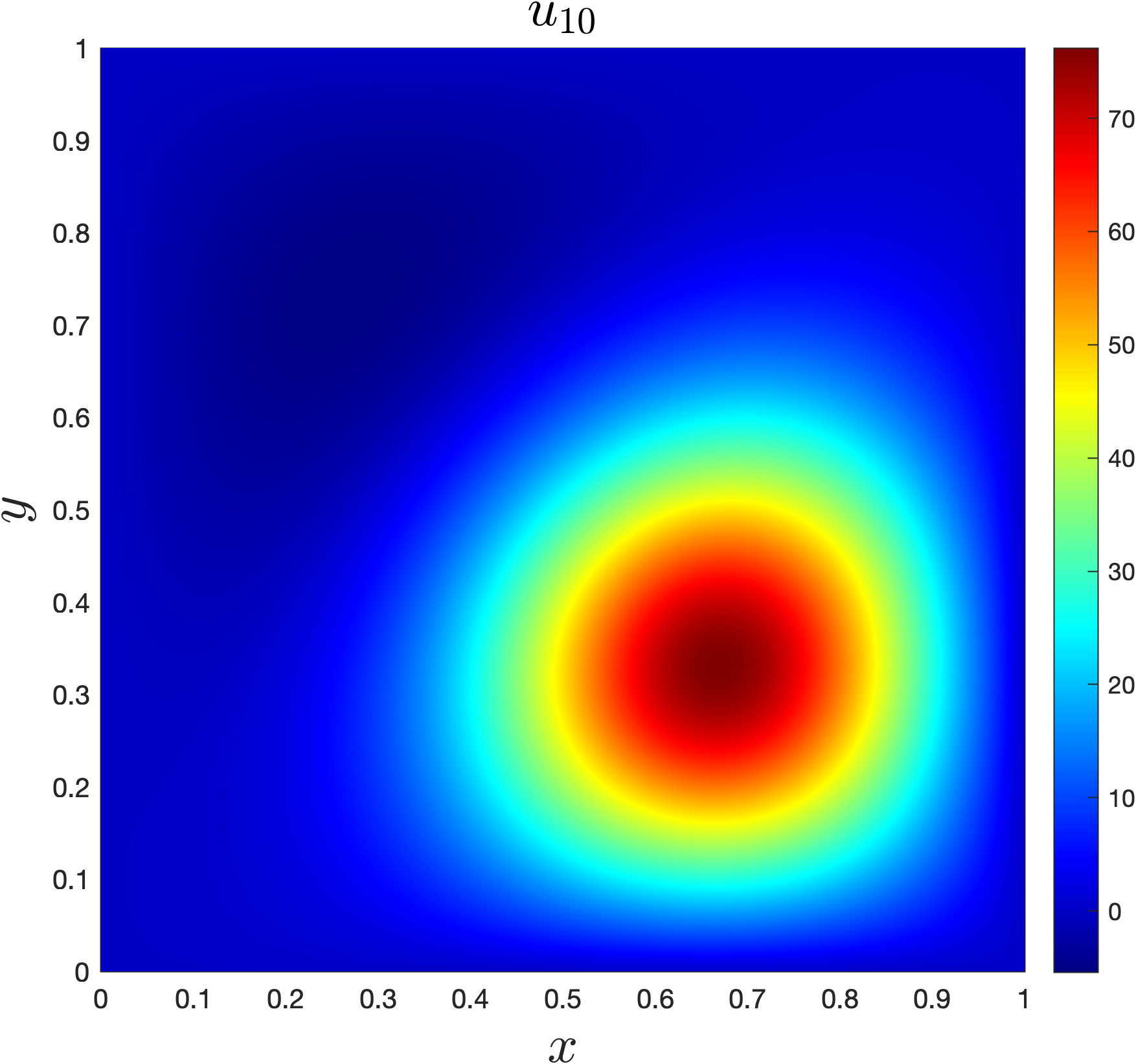}
    }
    \subfigure[The mesh.]{
        \includegraphics[scale=0.30]{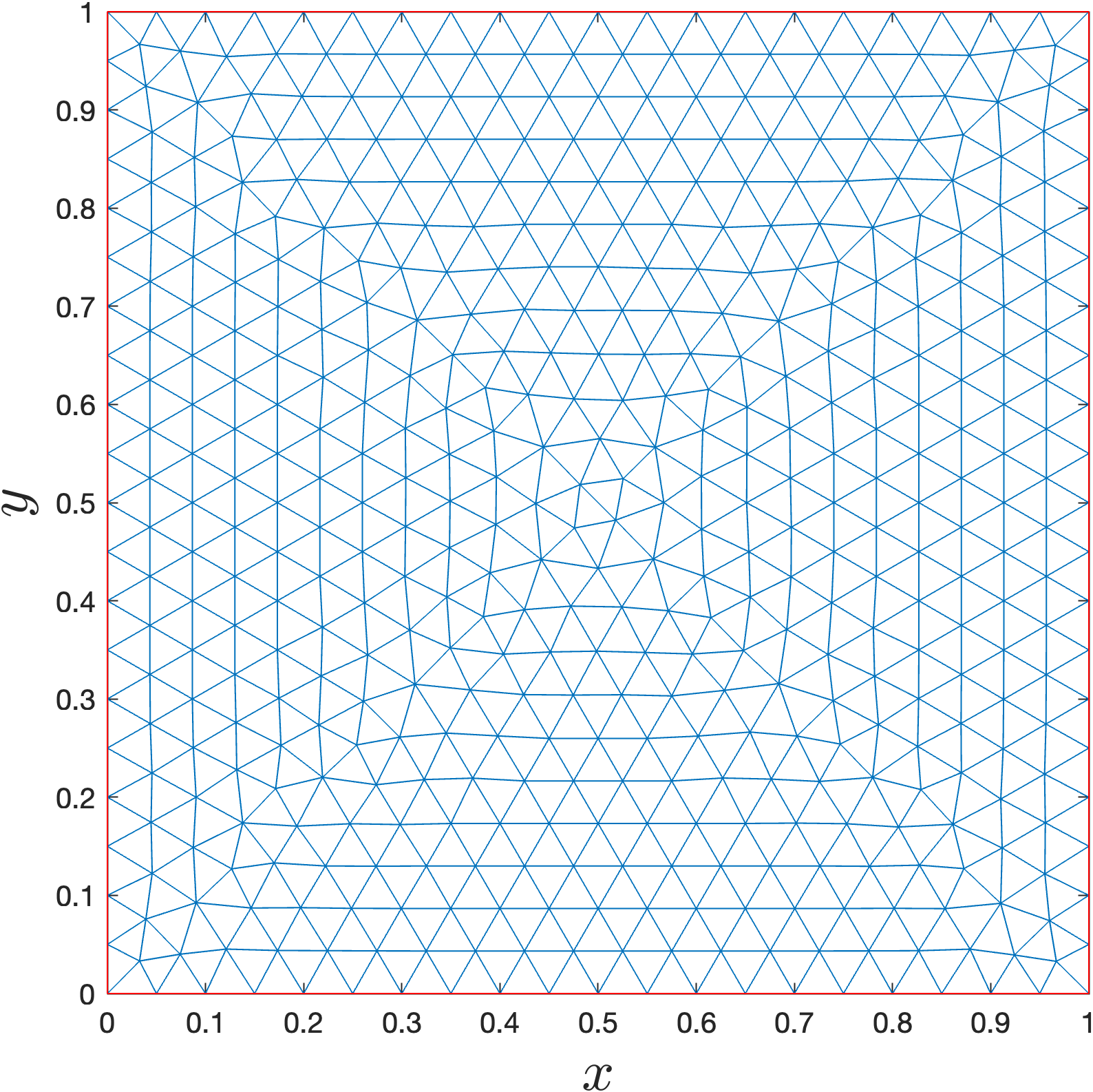}
    }
    \subfigure[Convergence of residual.]{
        \includegraphics[scale=0.30]{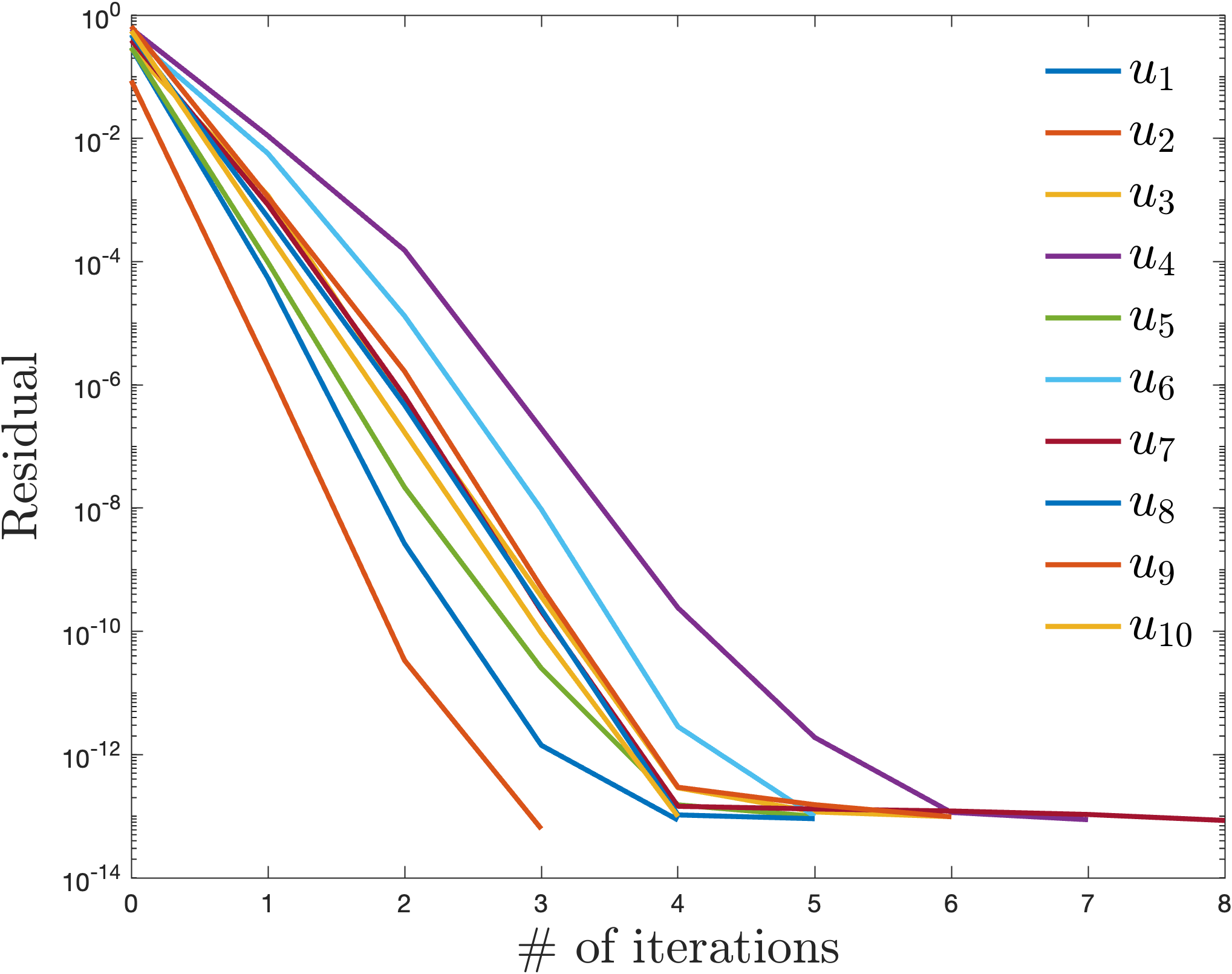}
    }
    \caption{Results of solving \eqref{eq:2d_reaction} using the presented approach. In (a), we show ten distinct found solutions to \eqref{eq:2d_reaction} with with $f(x, y) = 800\sin(\pi x)\sin(\pi y)$ ($u_1, u_2, ..., u_{10}$ from left to right). The mesh is presented in (b) and the convergence of residuals in (c). The convergence criterion is set to residual decreasing below $1\times10^{-13}$.}
    \label{fig:example_4}
\end{figure}

\begin{table}[h]
    \footnotesize
    \centering
    \begin{tabular}{c|c|c|c|c|c|c|c|c|c|c}
    \hline
    \hline
    & $u_1$ & $u_2$ & $u_3$ & $u_4$ & $u_5$ & $u_6$ & $u_7$ & $u_8$ & $u_9$ & $u_{10}$\\
    \hline 
     ($\times 10^{-14}$) & $8.66$ & $6.22$ & $9.82$ & $8.76$ & $9.94$ & $9.94$ & $8.53$ & $9.10$ & $9.77$ & $9.87$\\
    \hline
    \hline
    \end{tabular}
    \caption{Maximum of the residuals of \eqref{eq:2d_reaction} with $s=800$ of ten found solutions on the mesh displayed in Figure \ref{fig:example_4}(b).} 
    \label{tab:example_4}
\end{table}

We consider the following 2D PDE:
\begin{subequations}\label{eq:2d_reaction}
    \begin{align}
        &\Delta u + u^2 = f, (x, y) \in \Omega.\\
        & u|_{\partial\Omega} = 0.
    \end{align}
\end{subequations}
Here $\Omega = (0, 1)^2$ and the source term is defined as $f(x, y) = s\sin(\pi x)\sin(\pi y), (x, y)\in\Omega$  where $s$ is a scalar parameter. We apply our proposed approach to solve this equation with $s=800$. Specifically, we initialize $500$ NNs using a truncated normal distribution with a mean of zero and a standard deviation of 3, and then train them for $30,000$ iterations. As shown in Figure \ref{fig:example_4_2}, ensemble PINNs are able to capture ten solution patterns, consistent with those reported in \cite{breuer2003multiple, li2024adaptive, allgower2009application}. 

Following the approach outlined in Section \ref{sec:2_3}, we use a conventional numerical solver to solve \eqref{eq:2d_reaction}, employing representative PINN solutions as initial guesses. In this example, we utilize MATLAB’s \textit{Partial Differential Equation Toolbox} \cite{MATLAB}, which implements a finite element method (FEM) combined with the Gauss-Newton algorithm for solving nonlinear steady-state PDEs.
The results, including the computational mesh (quadratic element geometric order), are presented in Figures \ref{fig:example_4}(a) and \ref{fig:example_4}(b). Additionally, Figure \ref{fig:example_4}(c) and Table \ref{tab:example_4} depict the convergence of the Gauss-Newton algorithm and the maximum residuals, further validating the learned solutions to \eqref{eq:2d_reaction}. While our approach successfully identifies all solution patterns presented in \cite{breuer2003multiple, li2024adaptive, allgower2009application}, it does not recover the two additional patterns described in \cite{li2023efficient}.

\section{Additional results}\label{sec:appendix_3}

\subsection{Ablation study for the 1D Bratu problem}

\begin{figure}[ht]
    \centering
    \subfigure[\text{[1, 50, 50, 1]}.]{
        \includegraphics[scale=0.18]{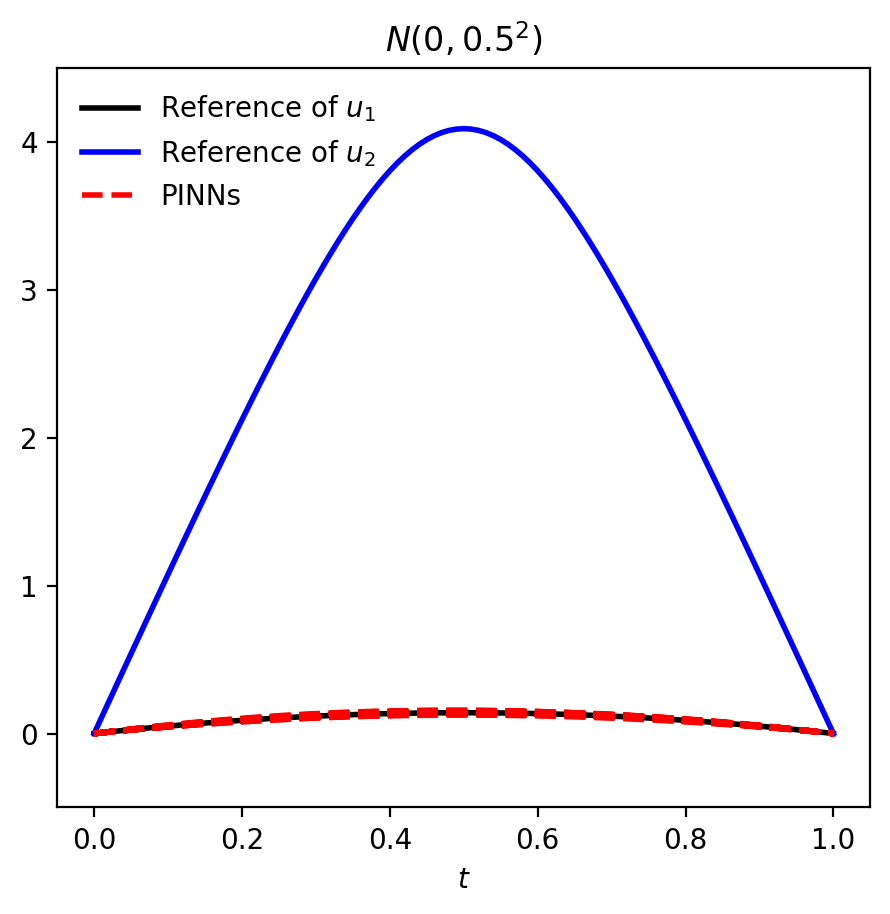}
        \includegraphics[scale=0.18]{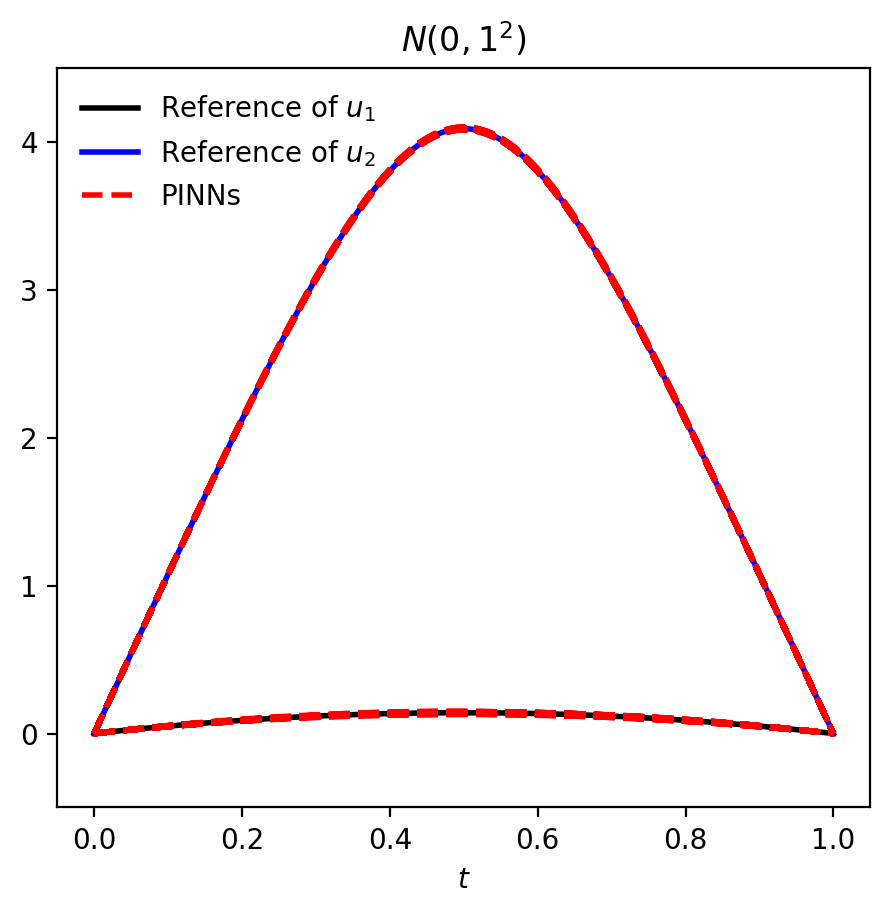}
        \includegraphics[scale=0.18]{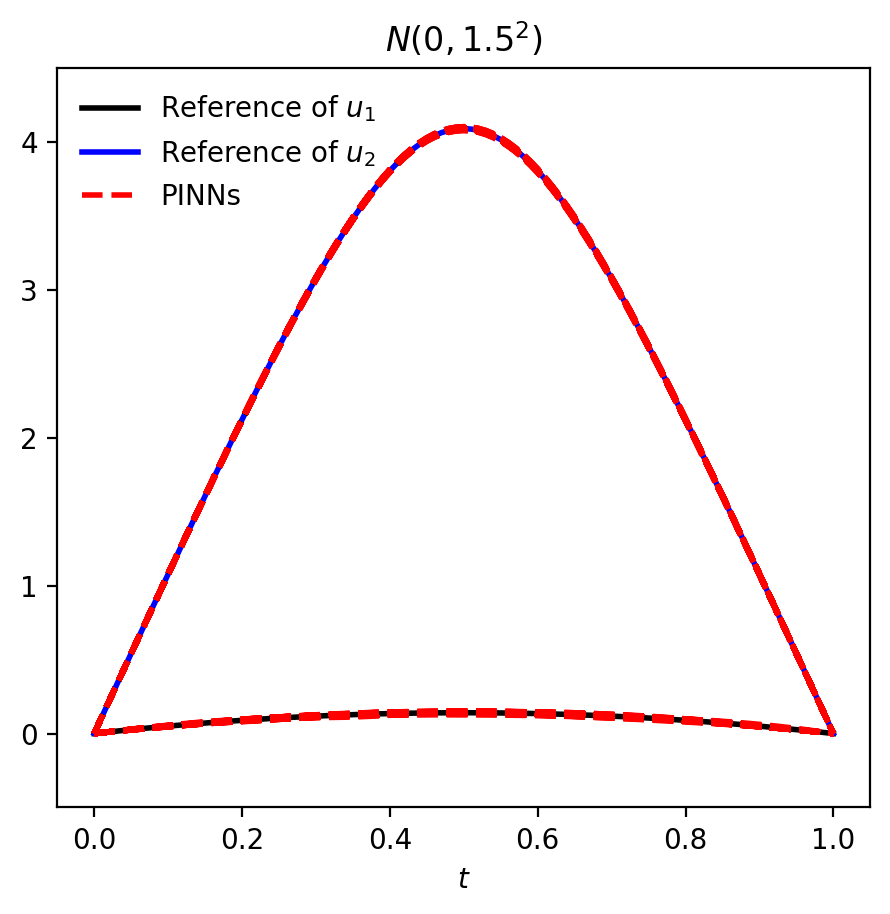}
        \includegraphics[scale=0.18]{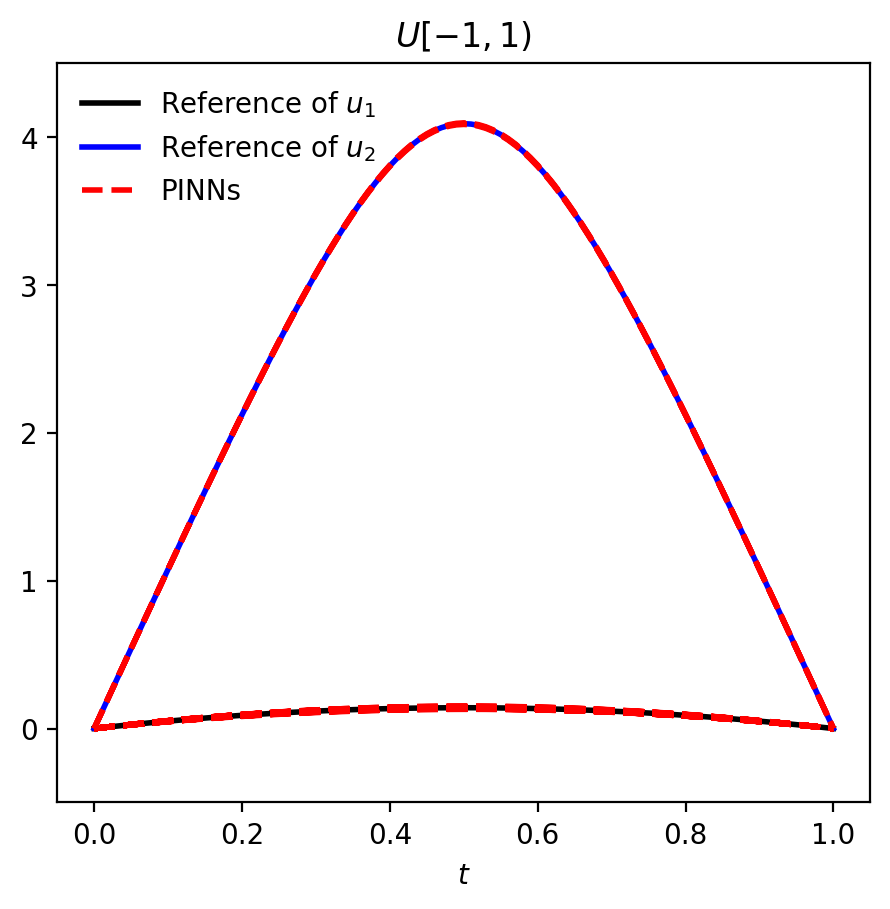}
        \includegraphics[scale=0.18]{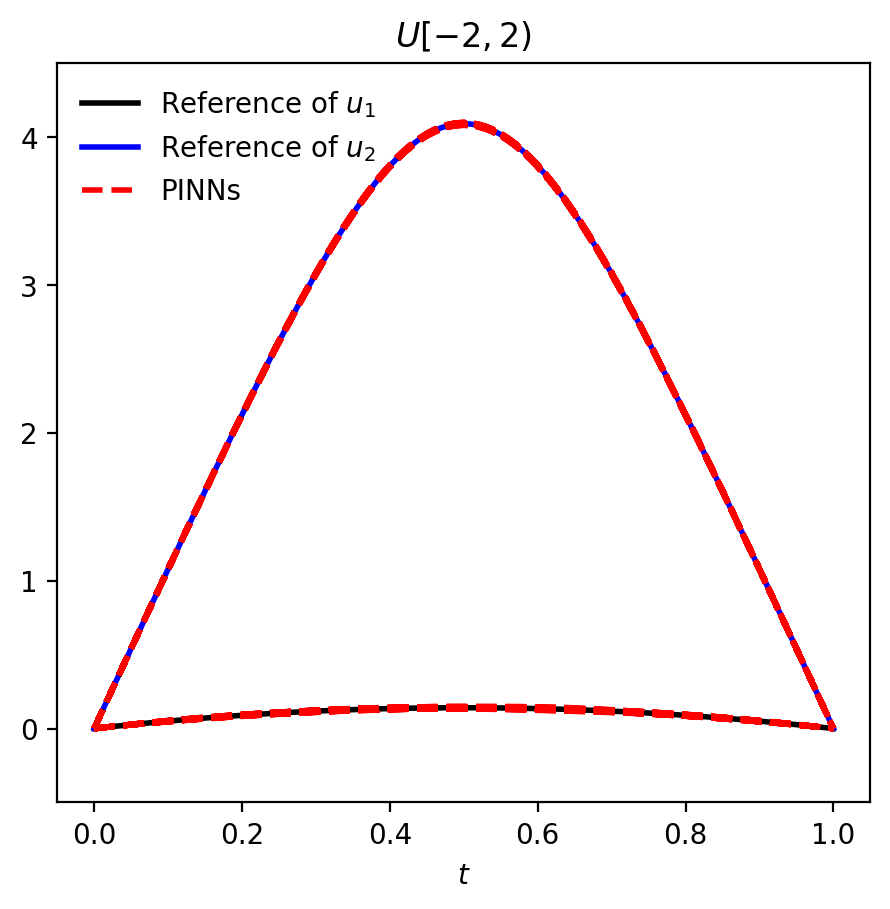}
        \includegraphics[scale=0.18]{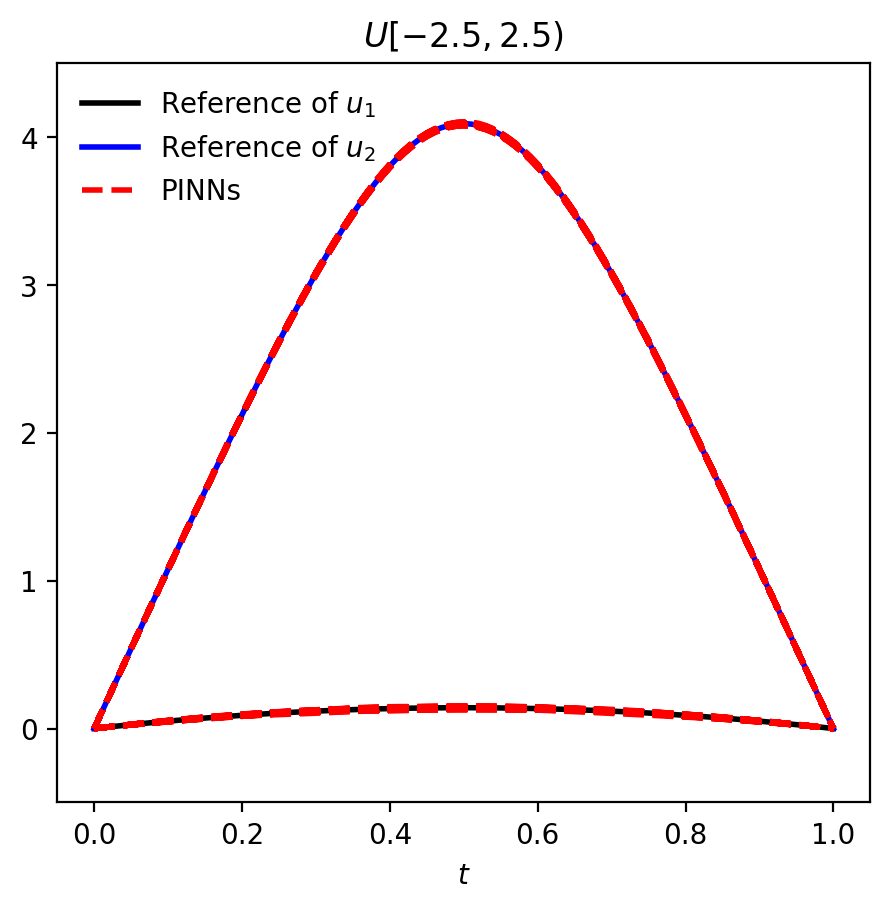}
        \includegraphics[scale=0.18]{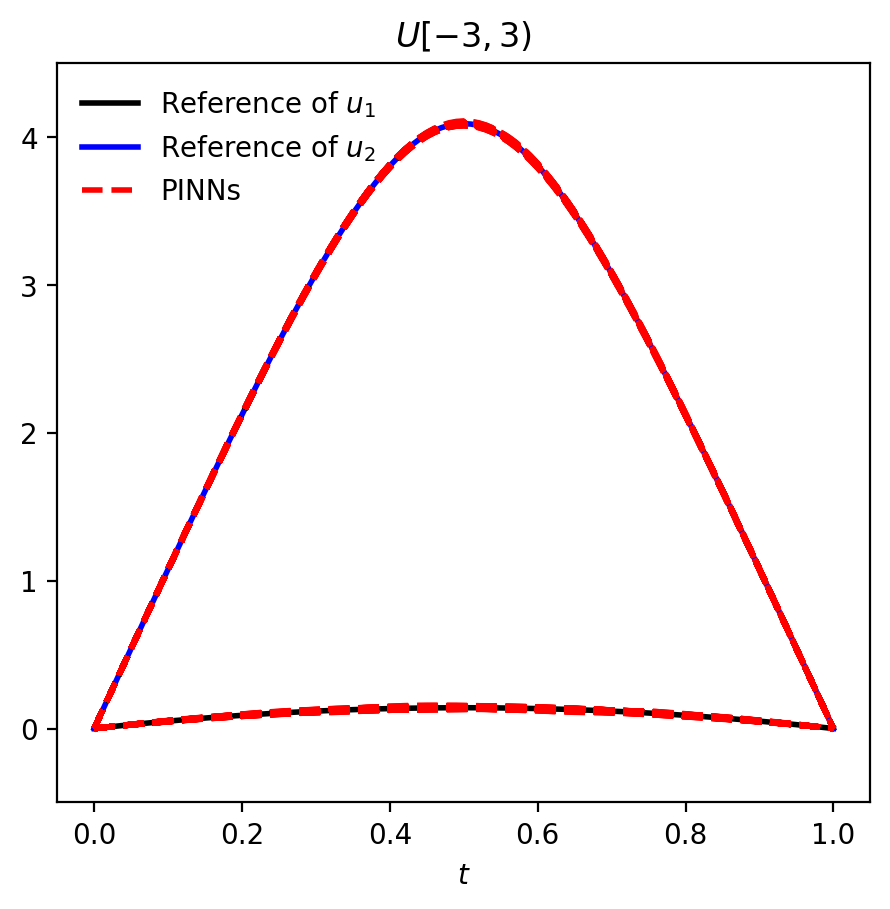}
    }
    \subfigure[\text{[1, 50, 50, 50, 1]}.]{
        \includegraphics[scale=0.18]{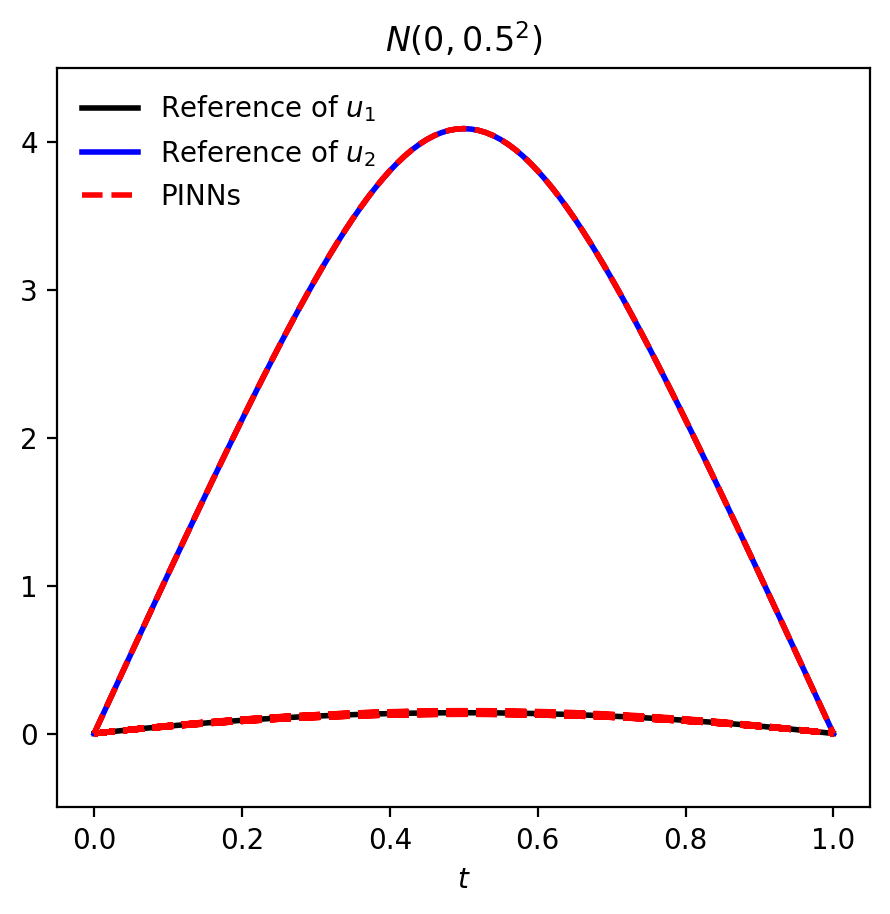}
        \includegraphics[scale=0.18]{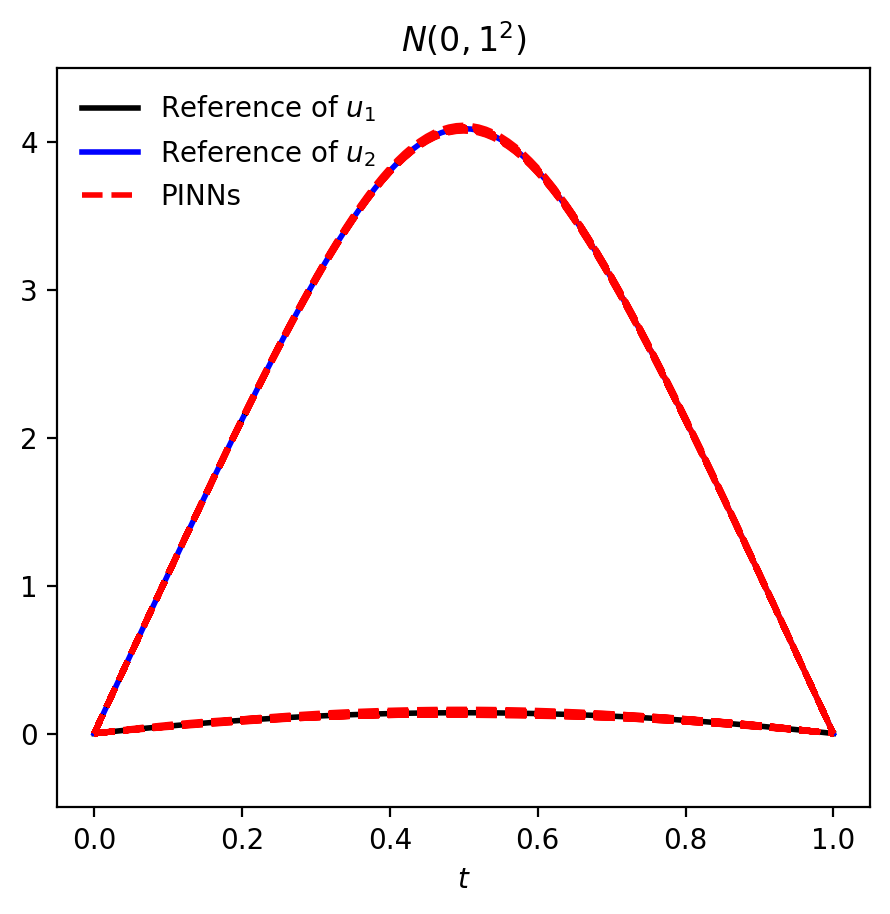}
        \includegraphics[scale=0.18]{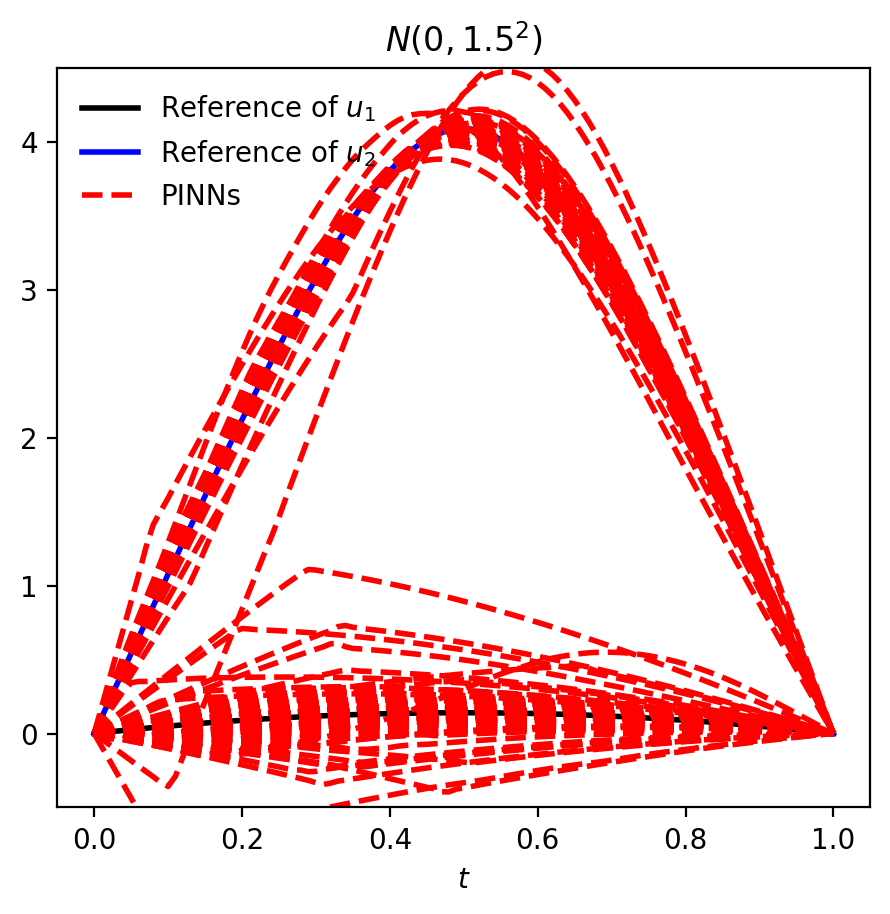}
        \includegraphics[scale=0.18]{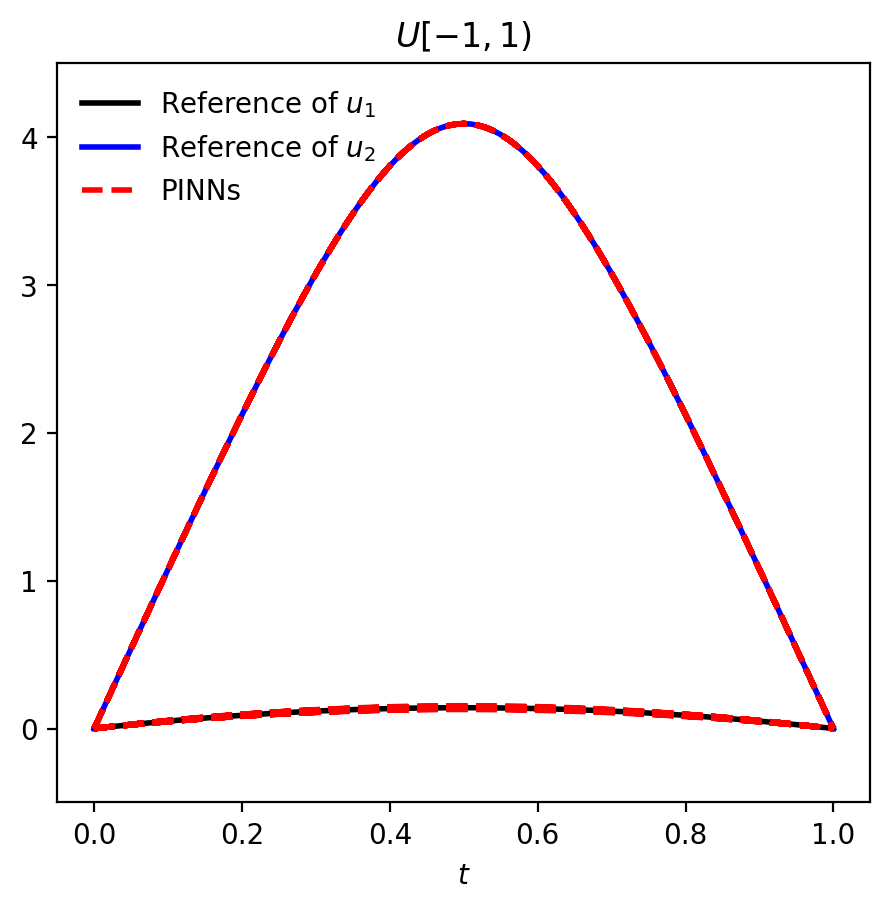}
        \includegraphics[scale=0.18]{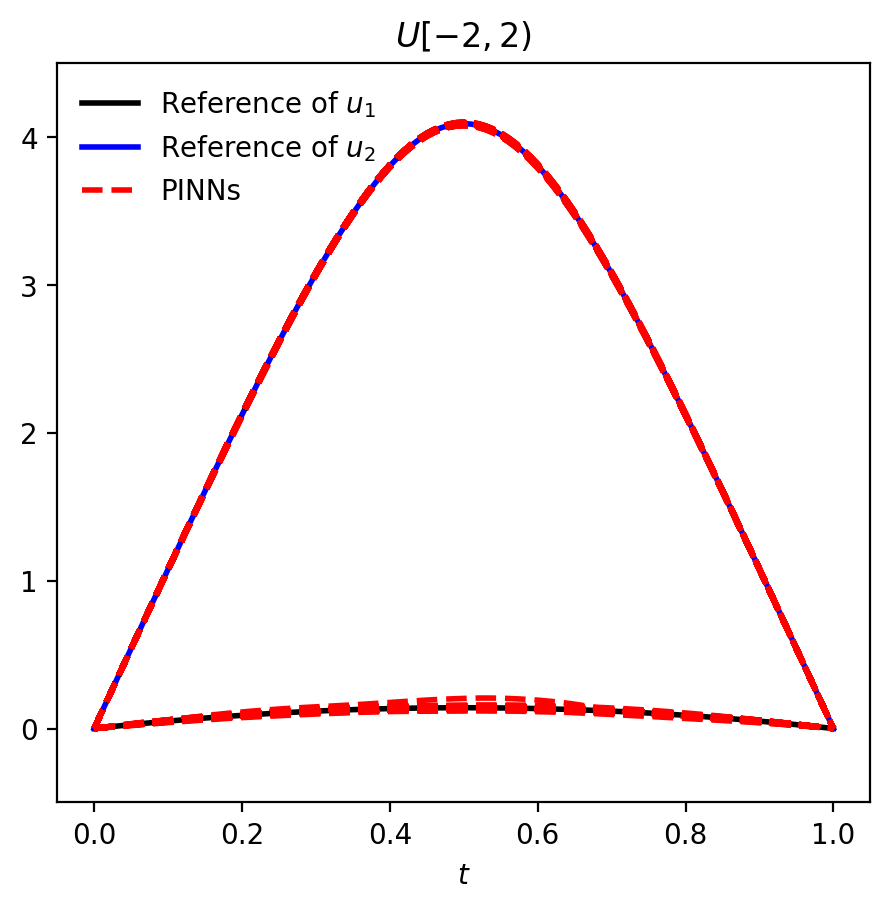}
        \includegraphics[scale=0.18]{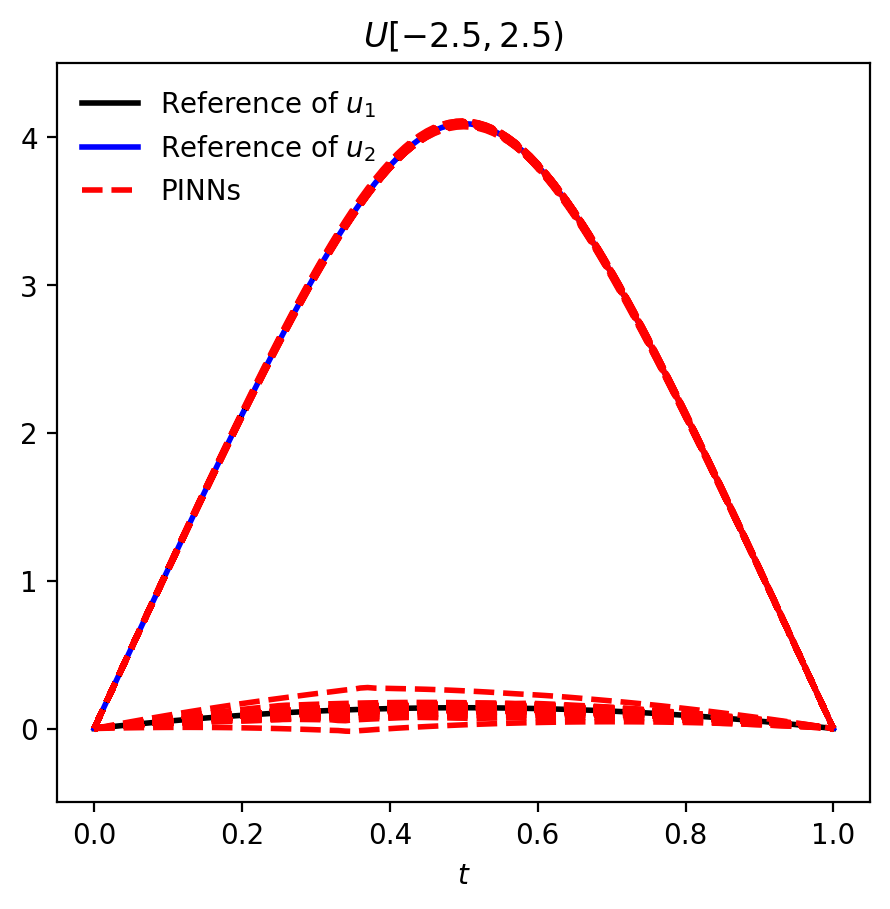}
        \includegraphics[scale=0.18]{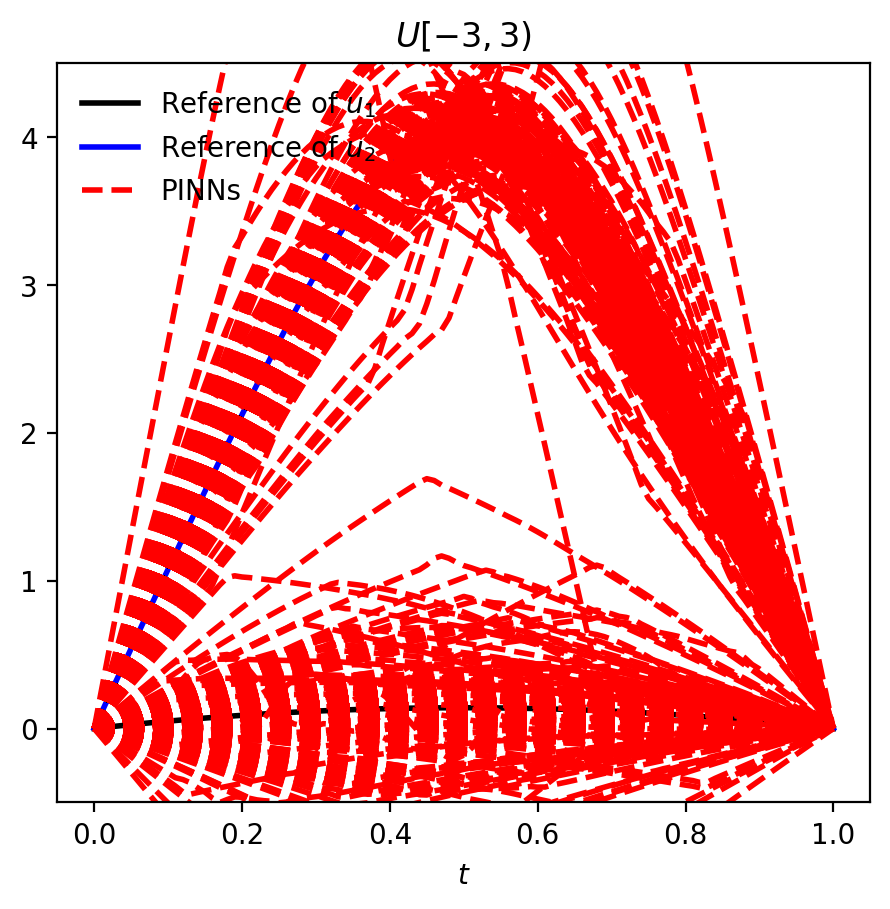}
    }
    \subfigure[\text{[1, 100, 100, 1]}.]{
        \includegraphics[scale=0.18]{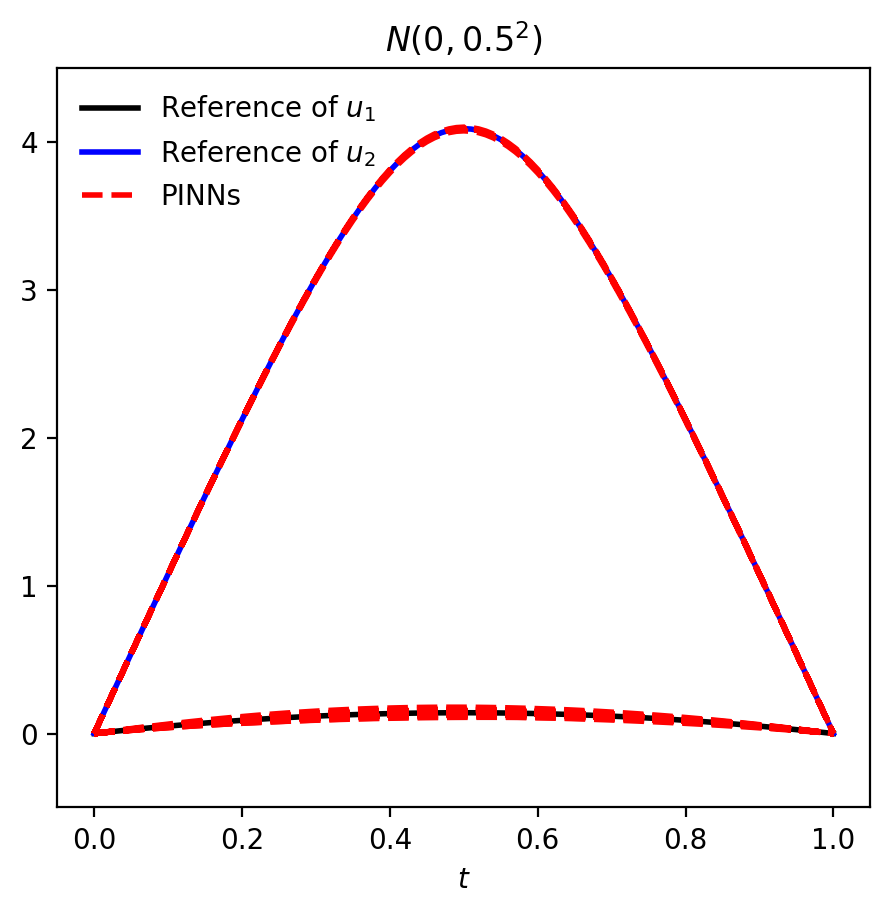}
        \includegraphics[scale=0.18]{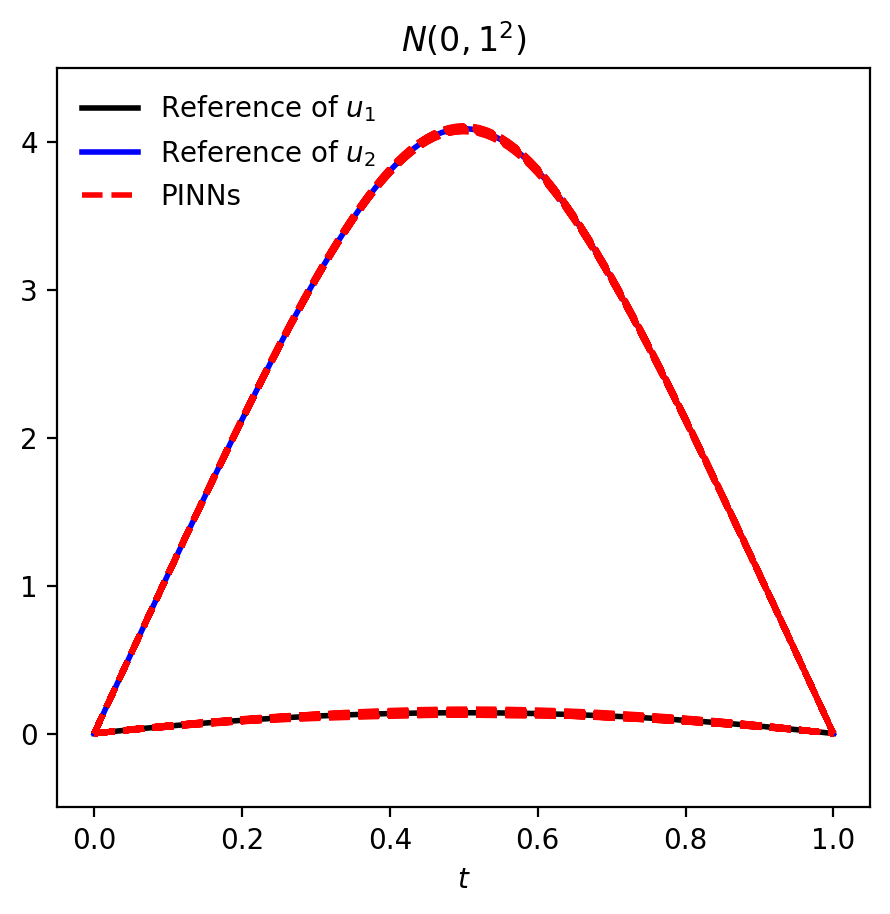}
        \includegraphics[scale=0.18]{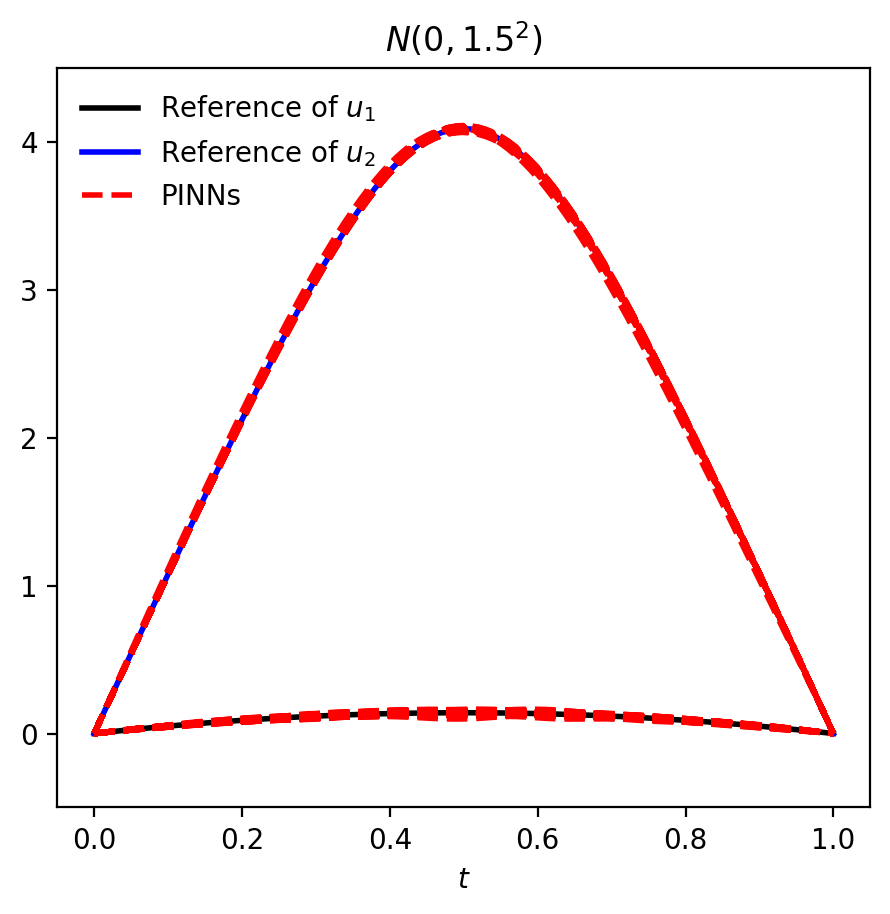}
        \includegraphics[scale=0.18]{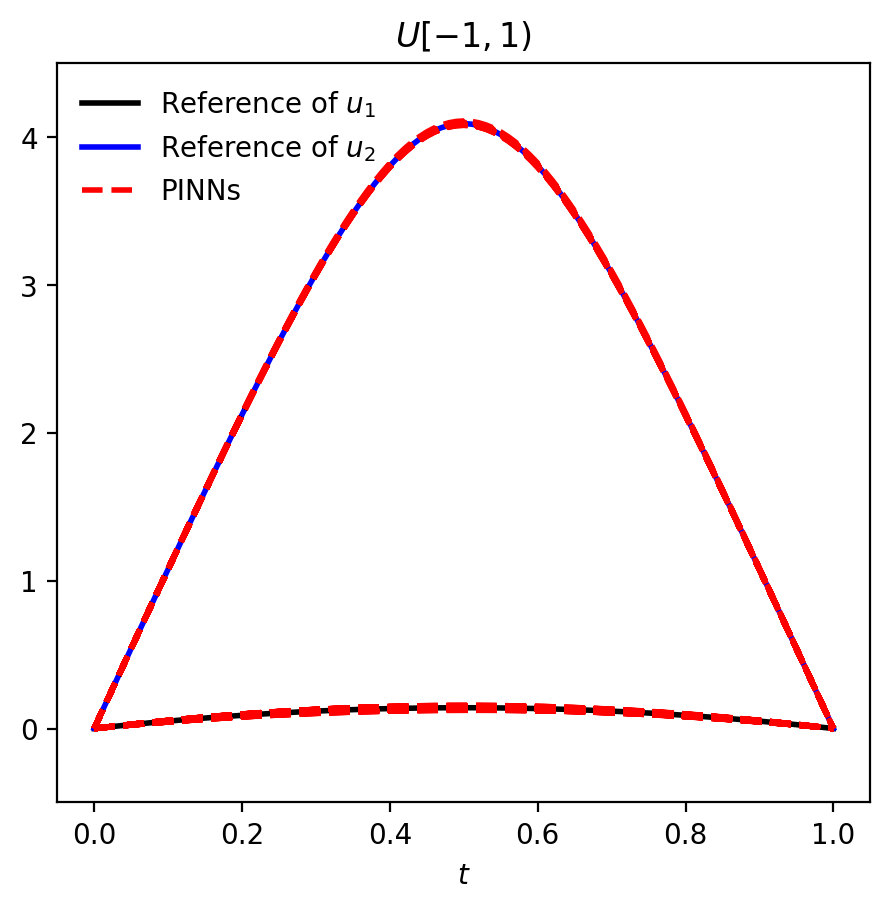}
        \includegraphics[scale=0.18]{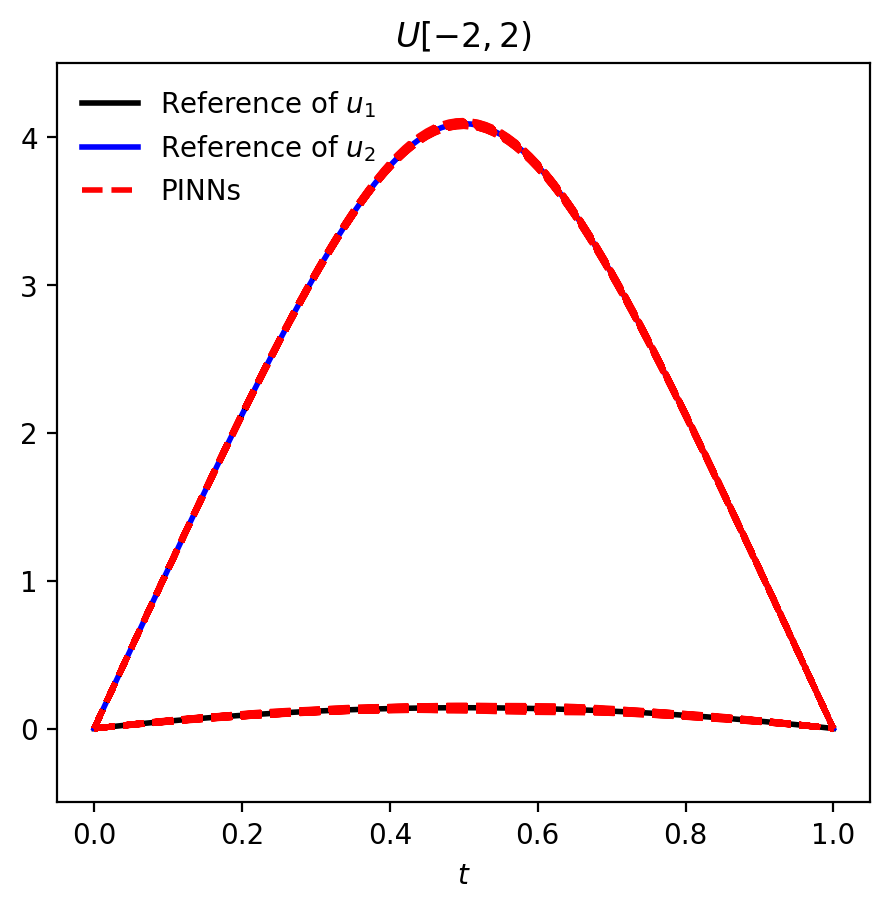}
        \includegraphics[scale=0.18]{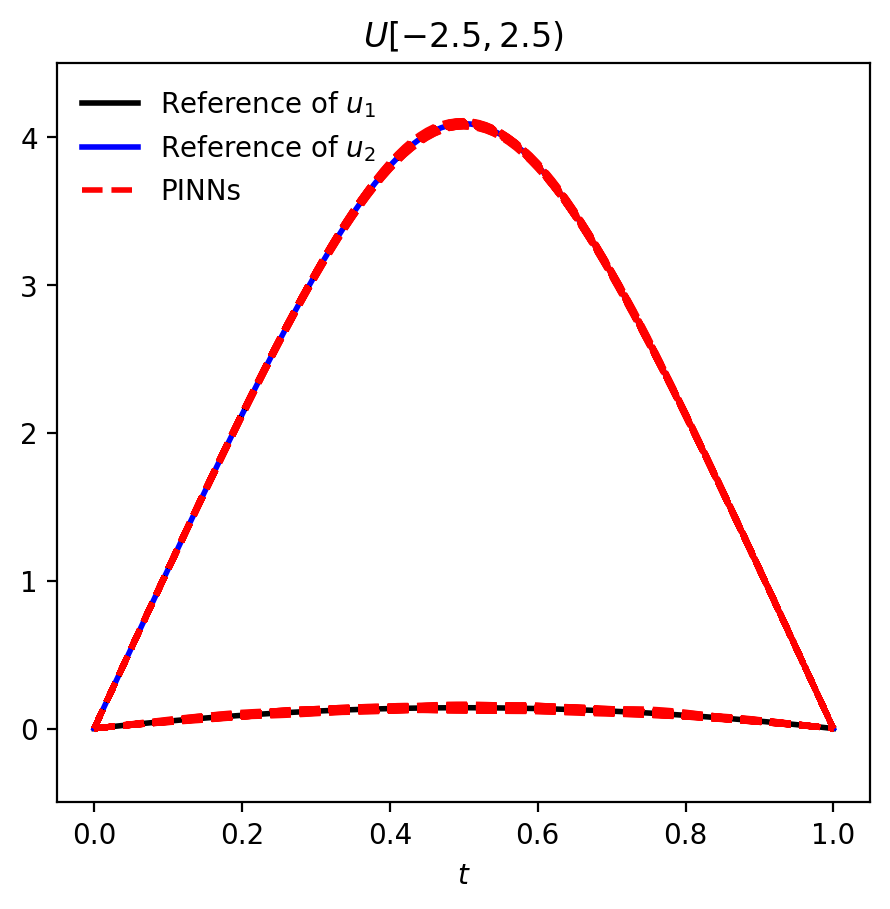}
        \includegraphics[scale=0.18]{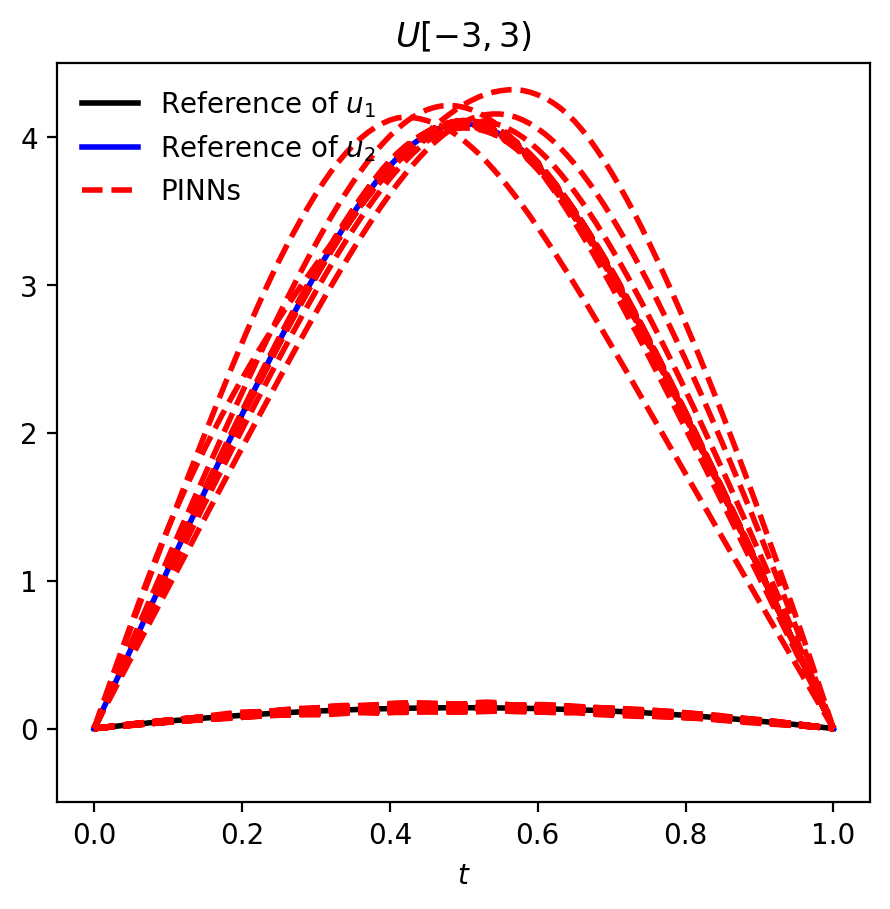}
    }
    \caption{Solving the 1D Bratu problem with $\lambda = 1$ using $10,000$ PINNs with randomly initialized NNs, differential architectures, and different initialization methods. The training is terminated at $20,000$ iterations. {\color{red}Red} dashed lines represent $10,000$ PINN solutions and \textbf{black} solid lines are the reference two solutions.}
    \label{fig:example_1_4}
\end{figure}

Here we present in Figure \ref{fig:example_1_4} the qualitative results of the ablation study conducted for the 1D Bratu problem using $10,000$ PINNs with randomly initialized NNs, varying architectures and initialization methods. The corresponding quantitative results are provided in Table \ref{tab:example_1}. Training was conducted for up to $20,000$ iterations. As shown in Figure \ref{fig:example_1_4}, while a larger variance in initialization or a wider and deeper NN architecture generally leads to more diverse PINN solutions after a fixed number of iterations, it also increases training difficulty. For instance, when initialized by $\mathcal{N}(0, 1.5^2)$ and $\mathcal{U}[-3, 3)$, networks with three hidden layers and 50 neurons per layer may fail to converge within 20,000 iterations.

\subsection{2D Allen-Cahn equation}

\begin{figure}[h!]
    \centering
    \includegraphics[scale=0.3]{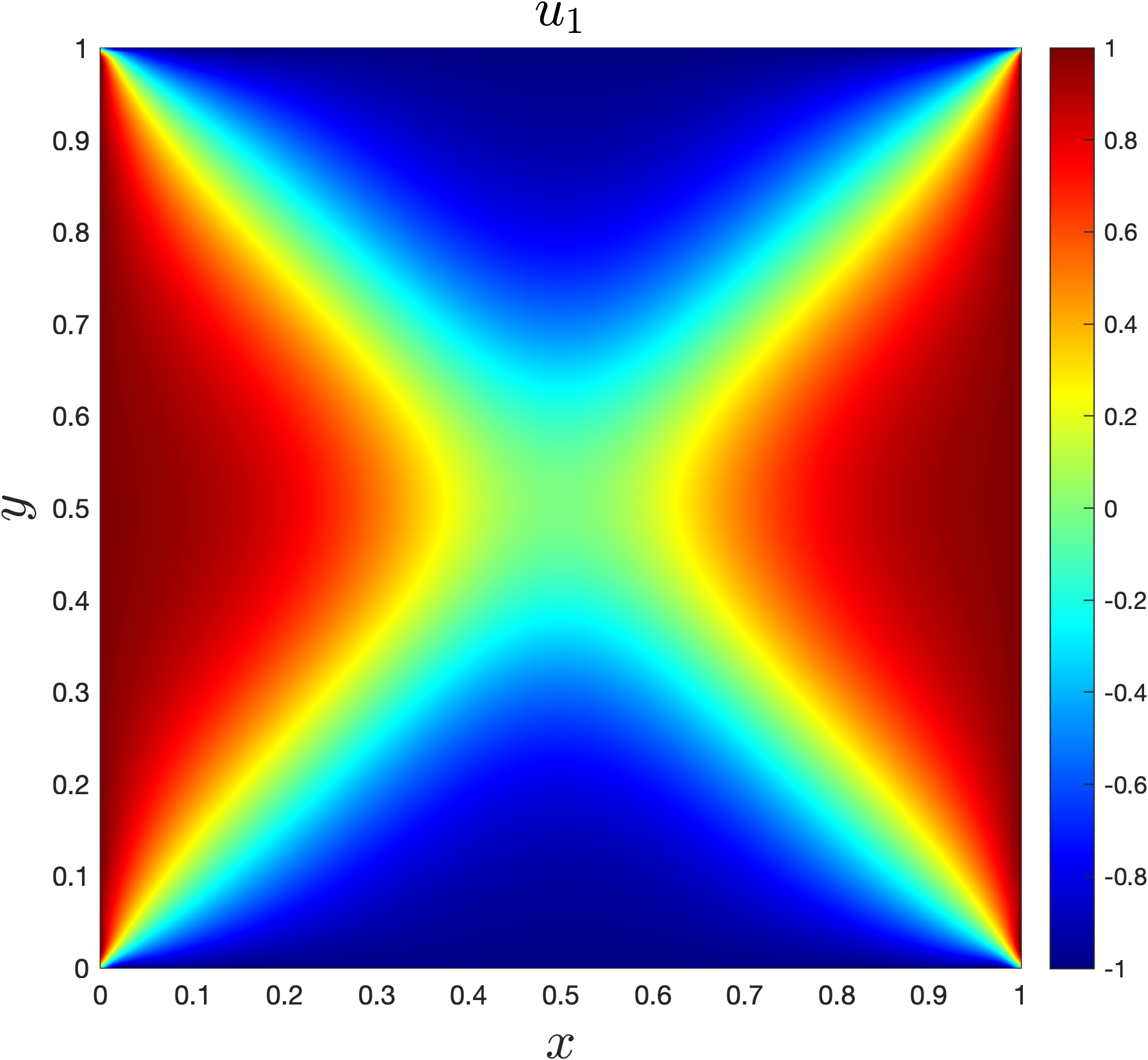}
   \includegraphics[scale=0.3]{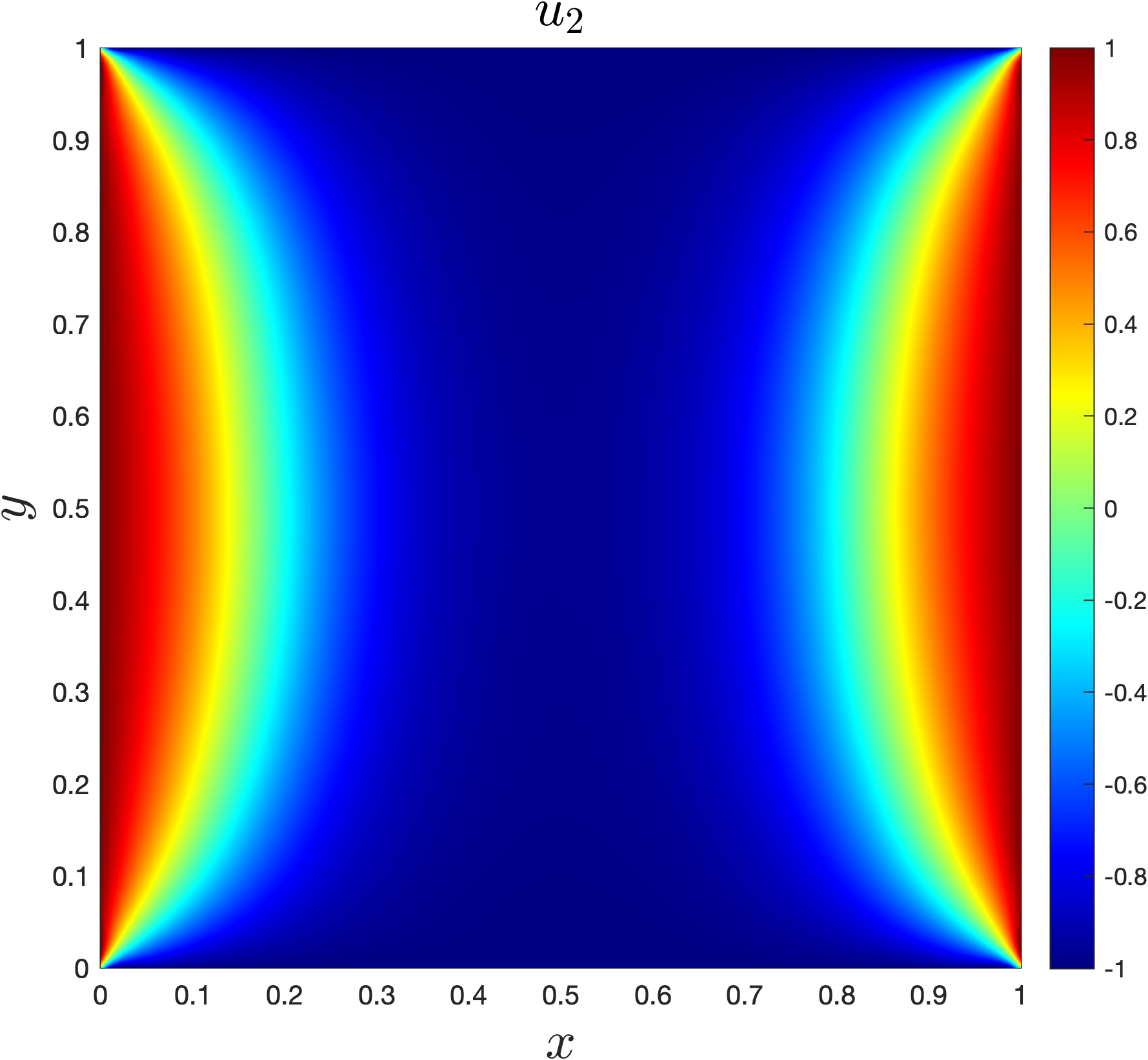}
   \includegraphics[scale=0.3]{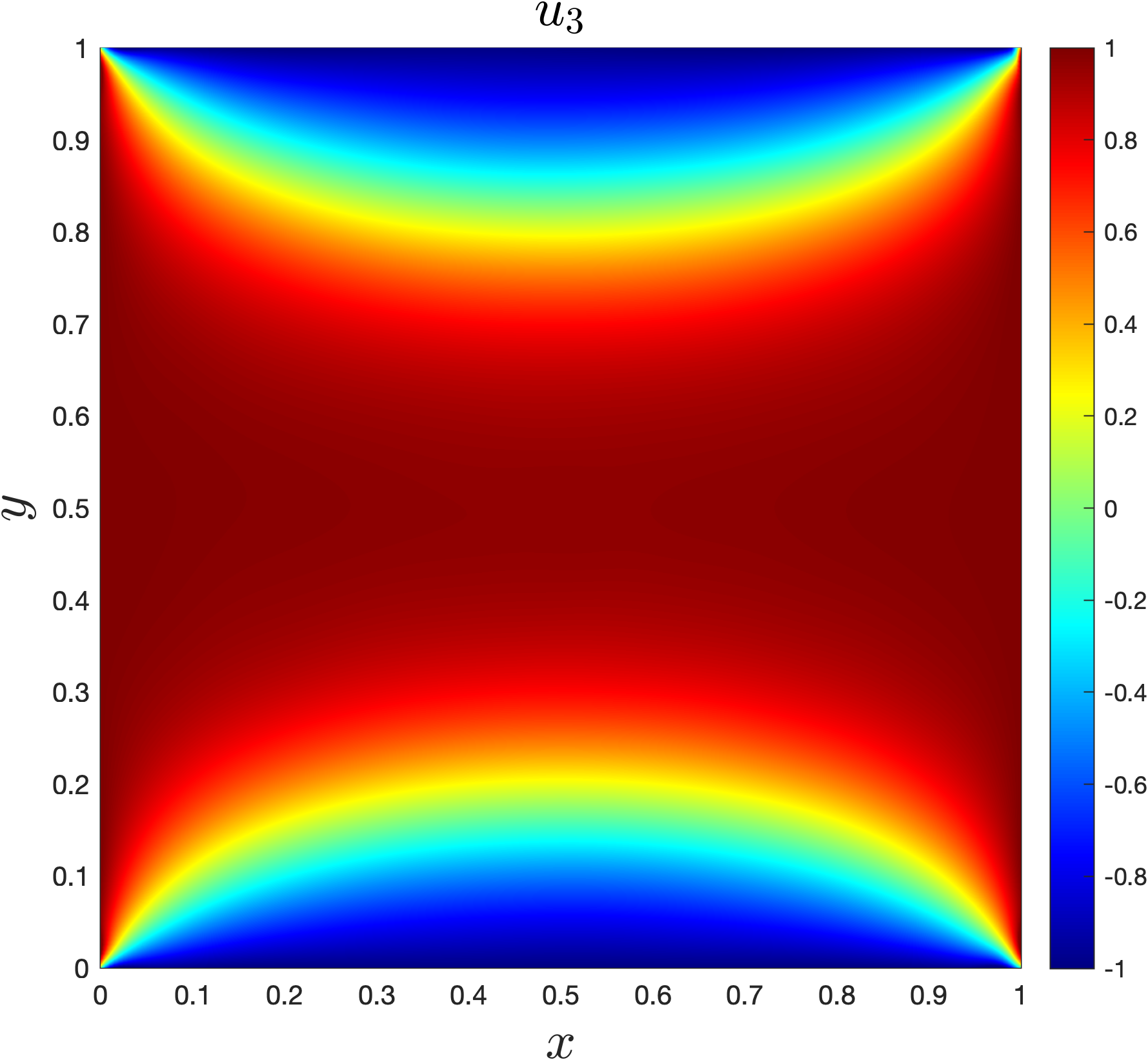}
    \caption{Three distinct PINN solutions to \eqref{eq:2d_allen} with $\epsilon=0.01$ chosen as the initial guesses for a FEM solver to solve \eqref{eq:2d_allen} with $\epsilon=0.01, 0.001$. The corresponding FEM solutions are presented in Figs. \ref{fig:example_5} and \ref{fig:example_5_2} for $\epsilon=0.01, 0.001$, respectively.}
    \label{fig:example_5_3}
\end{figure}

\begin{figure}[h!]
    \centering
    \subfigure[A PINN approximation of $u_1$ that transitions into $u_2$.]{
    \includegraphics[scale=0.2]{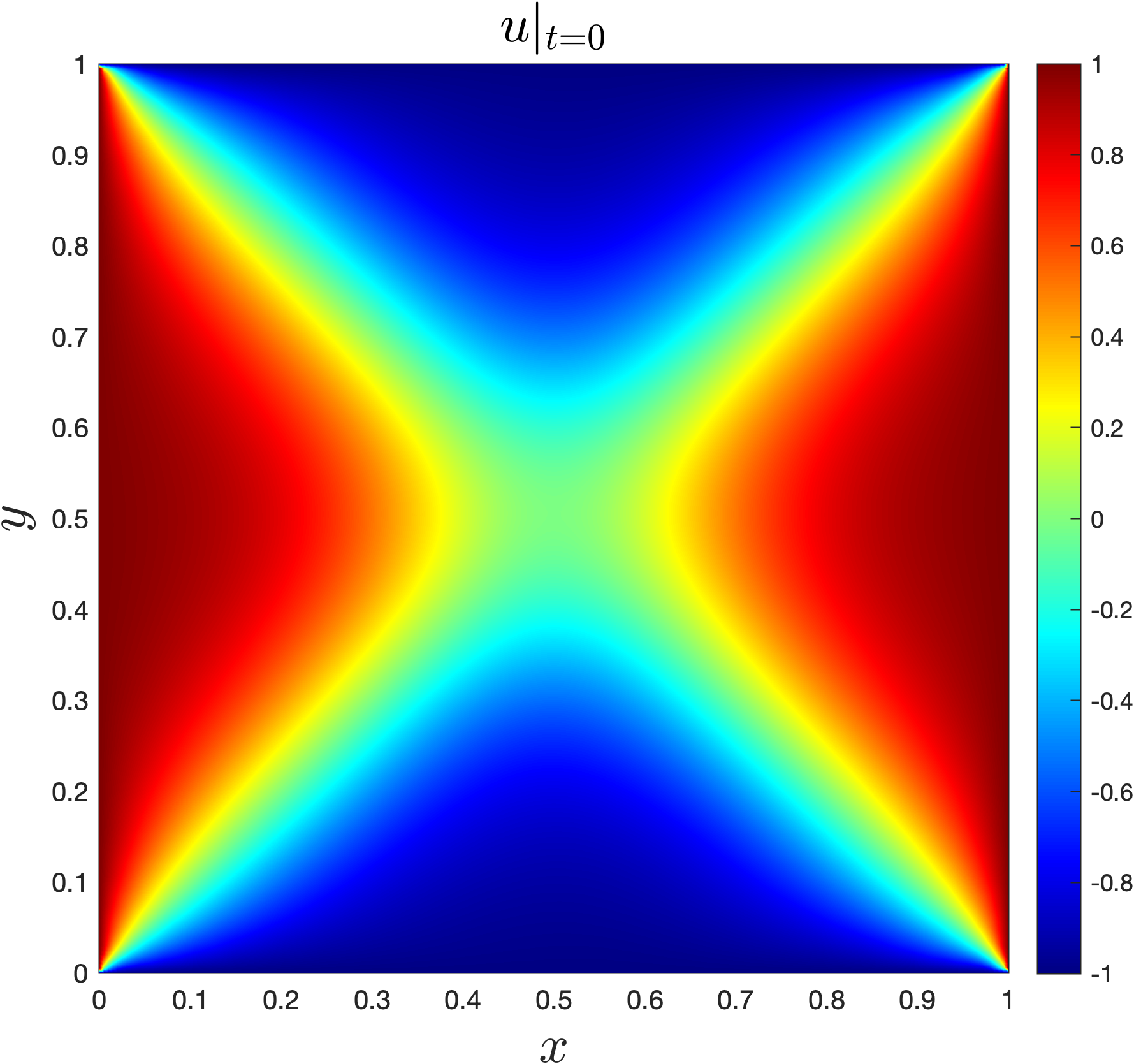}
    \includegraphics[scale=0.2]{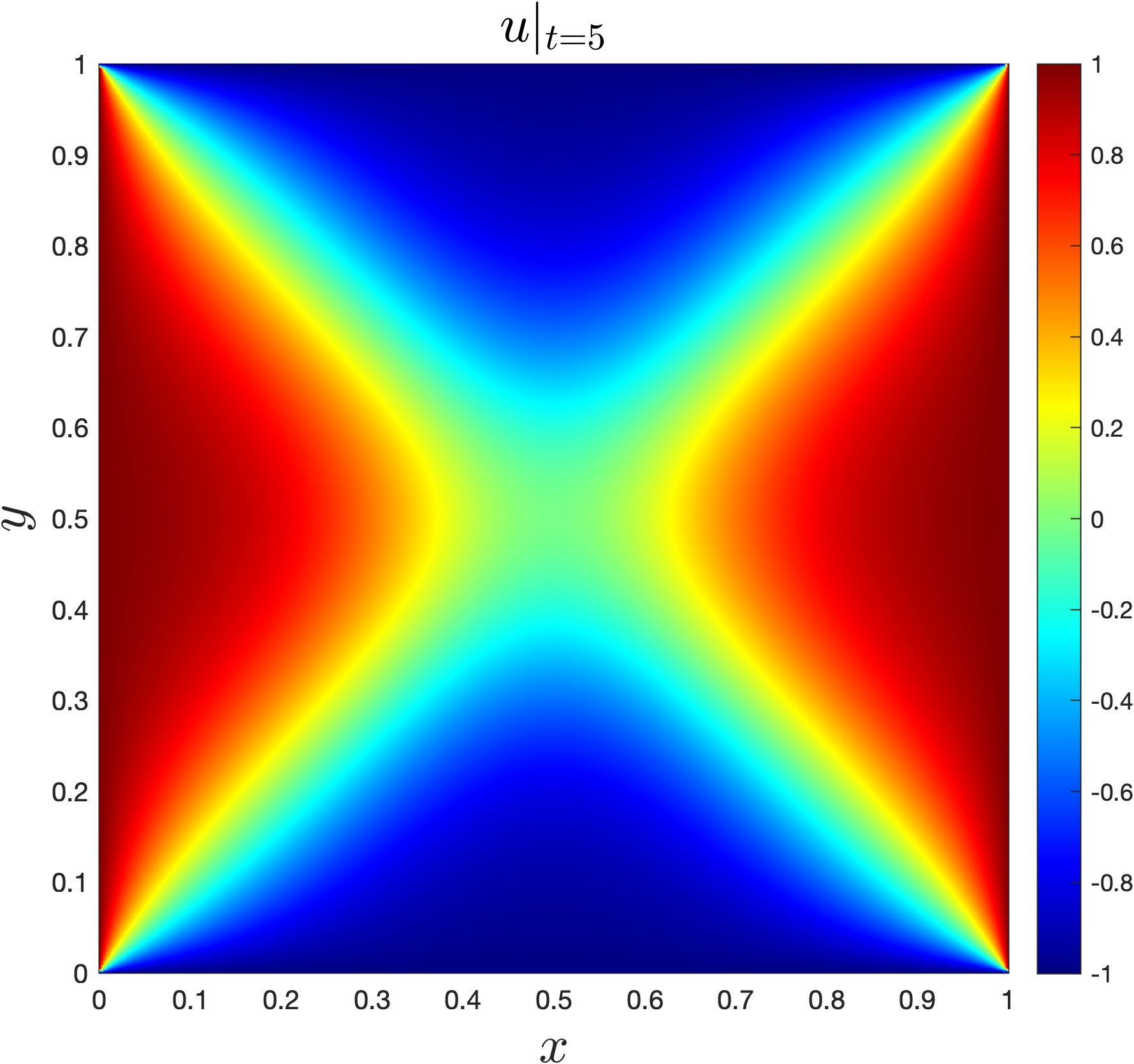}
    \includegraphics[scale=0.2]{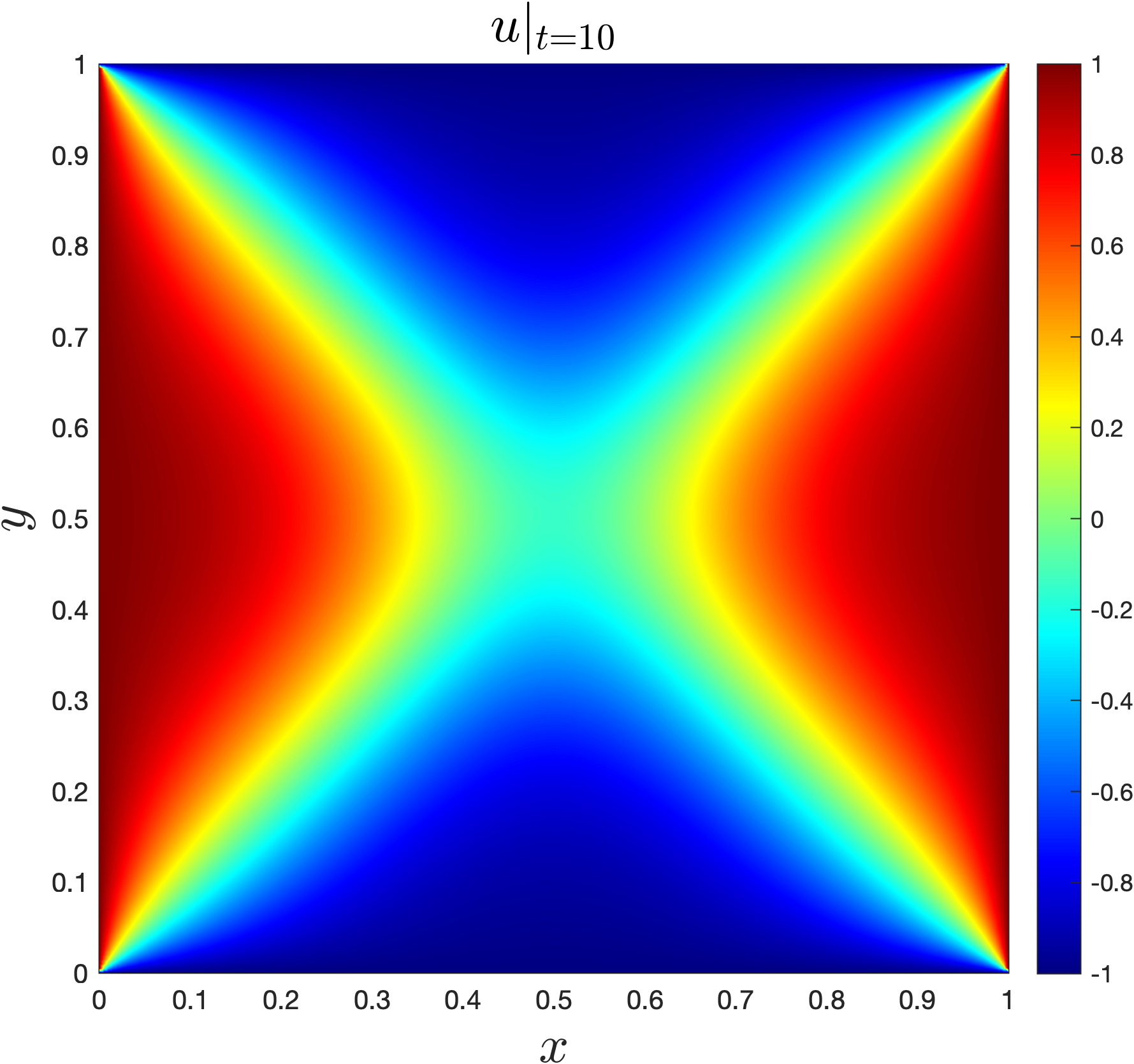}
    \includegraphics[scale=0.2]{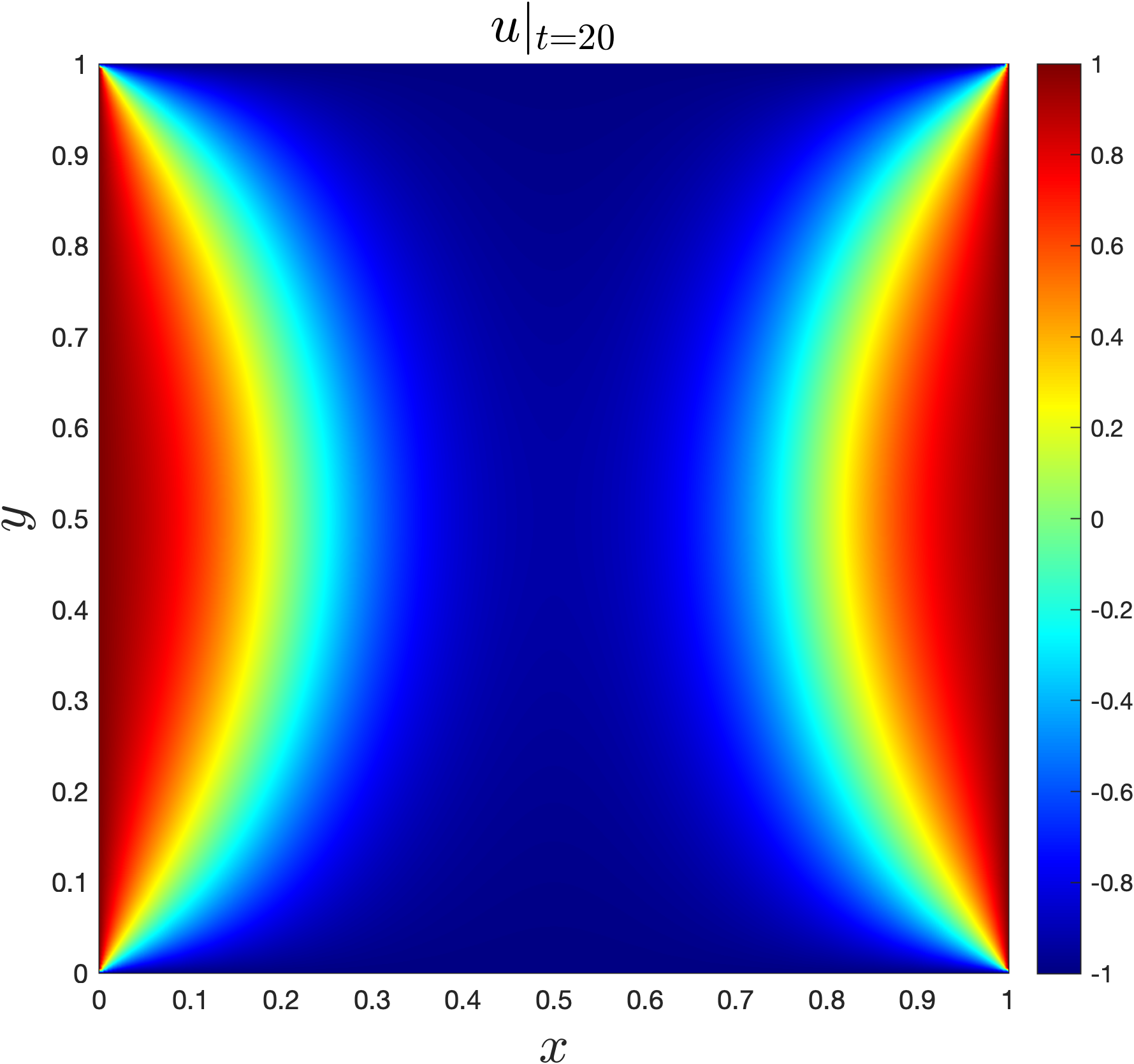}
    \includegraphics[scale=0.2]{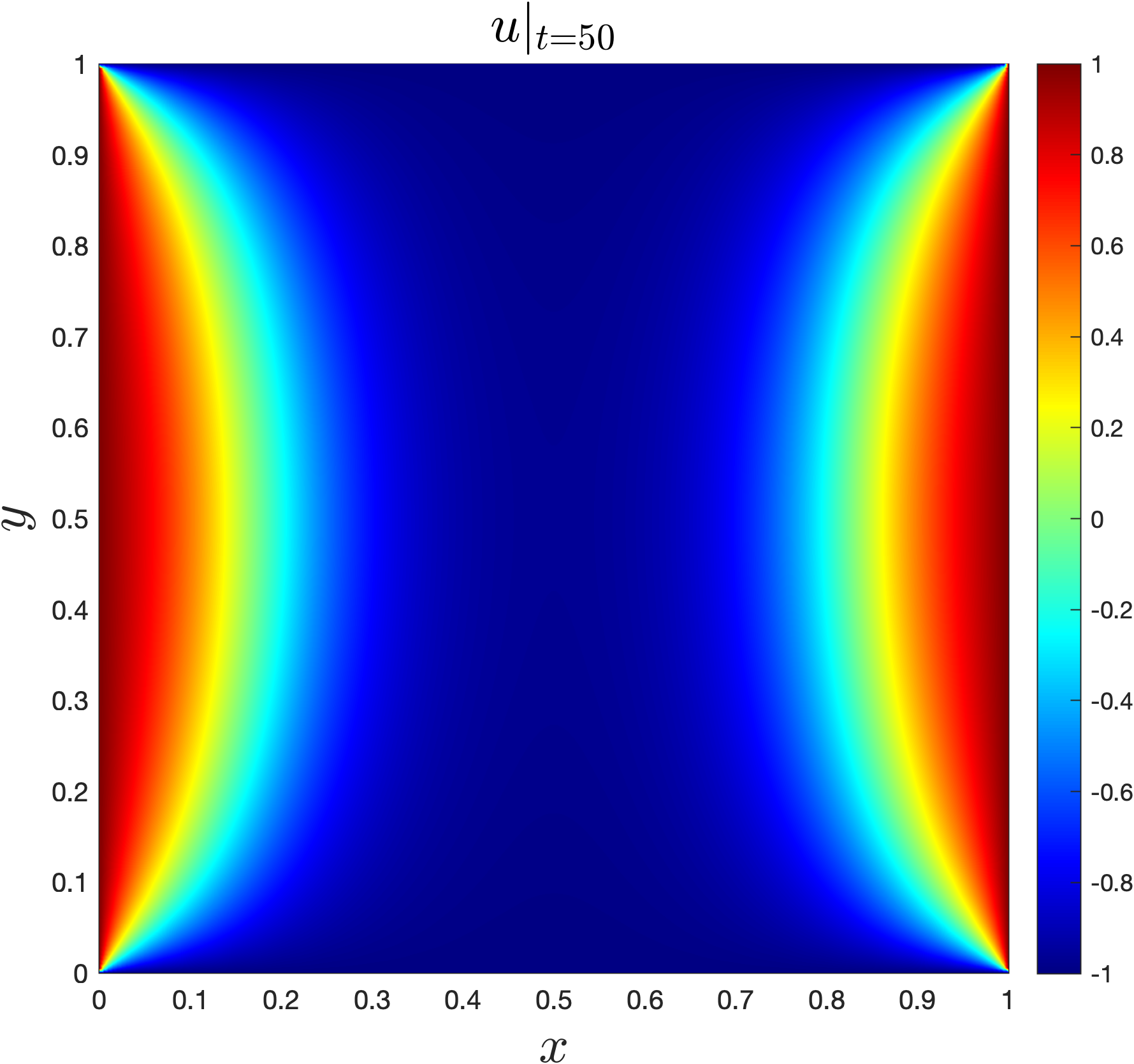}
    }
    \subfigure[A PINN approximation of $u_1$ that transitions into $u_3$.]{
    \includegraphics[scale=0.2]{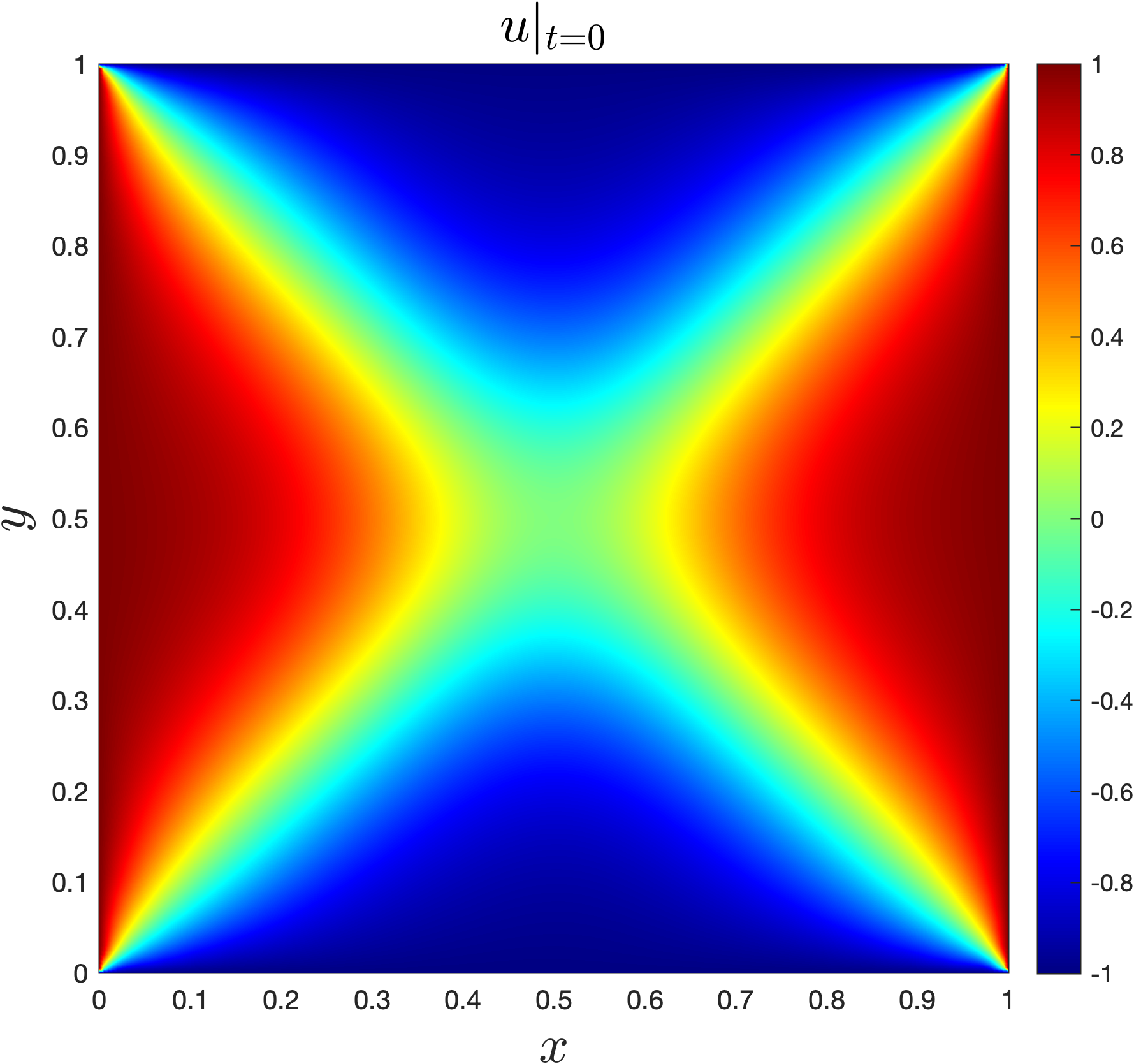}
    \includegraphics[scale=0.2]{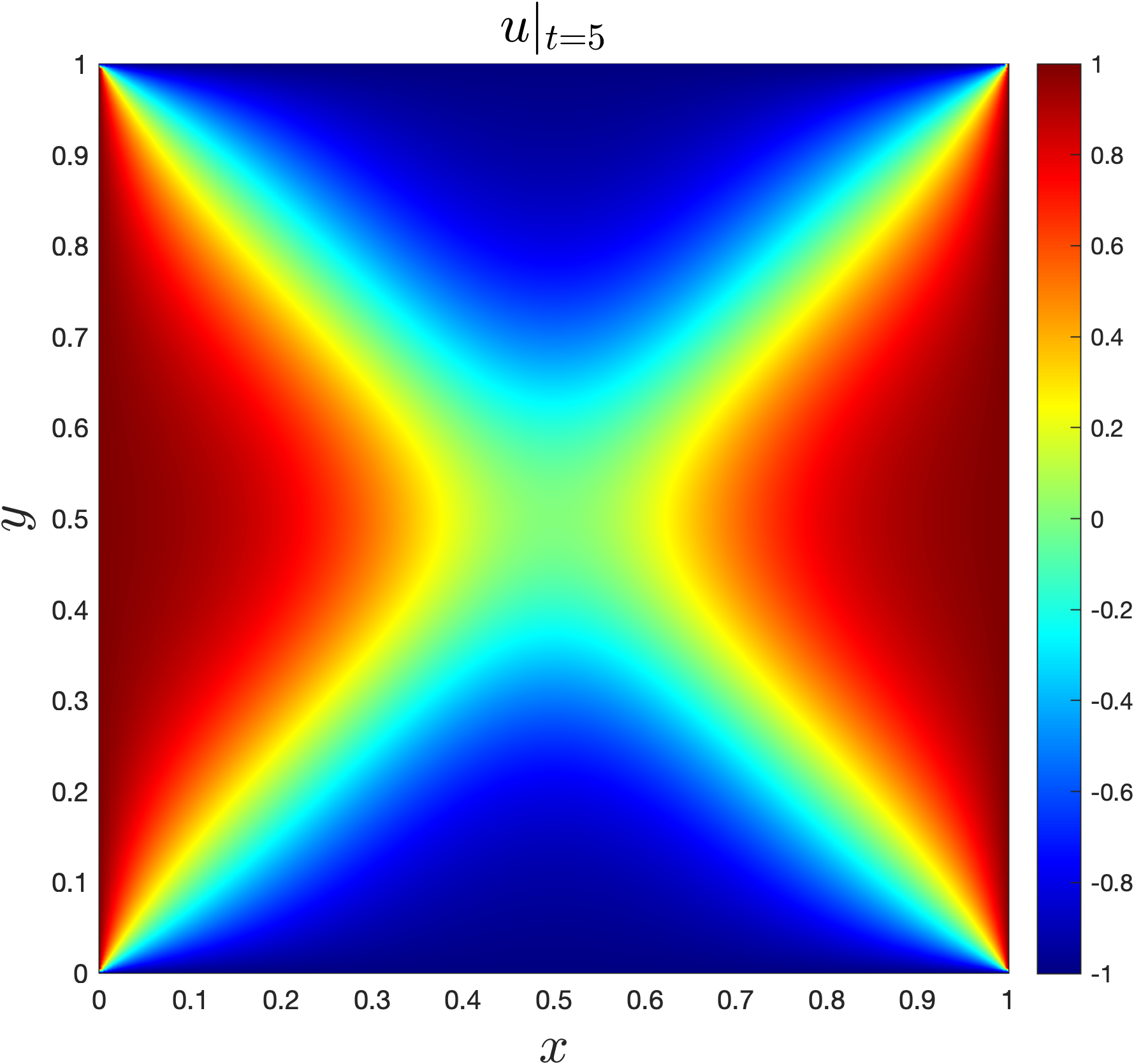}
    \includegraphics[scale=0.2]{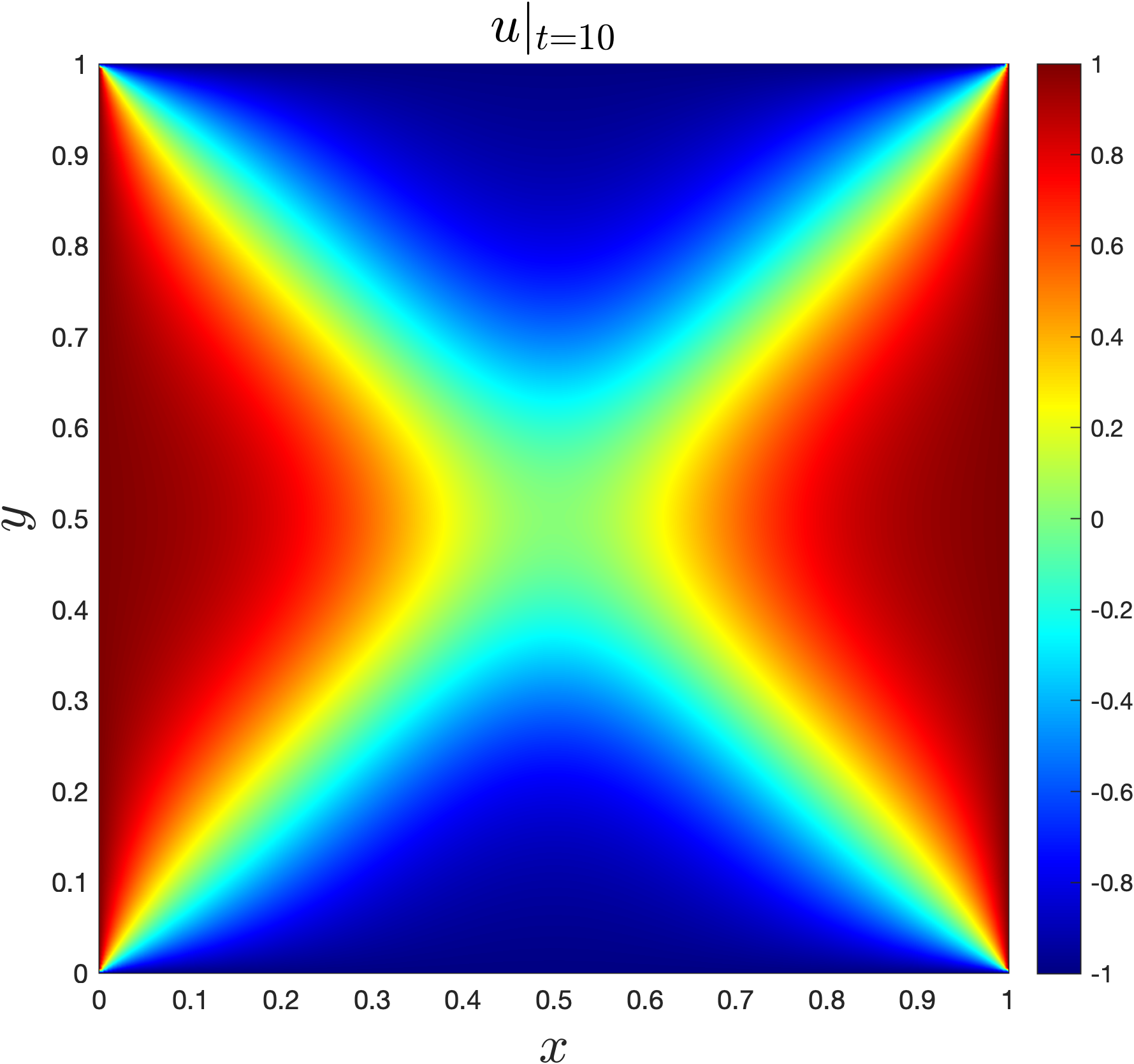}
    \includegraphics[scale=0.2]{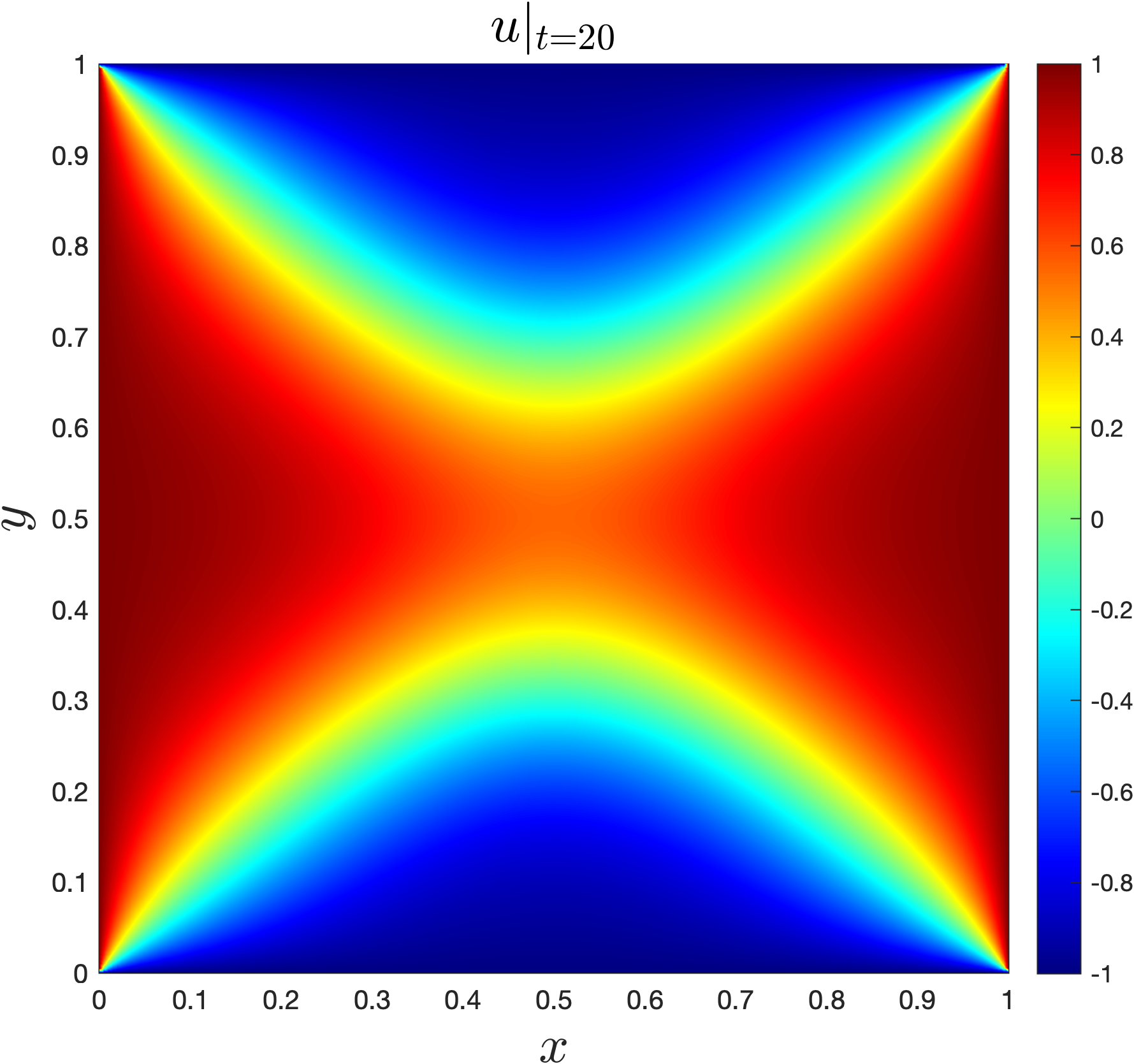}
    \includegraphics[scale=0.2]{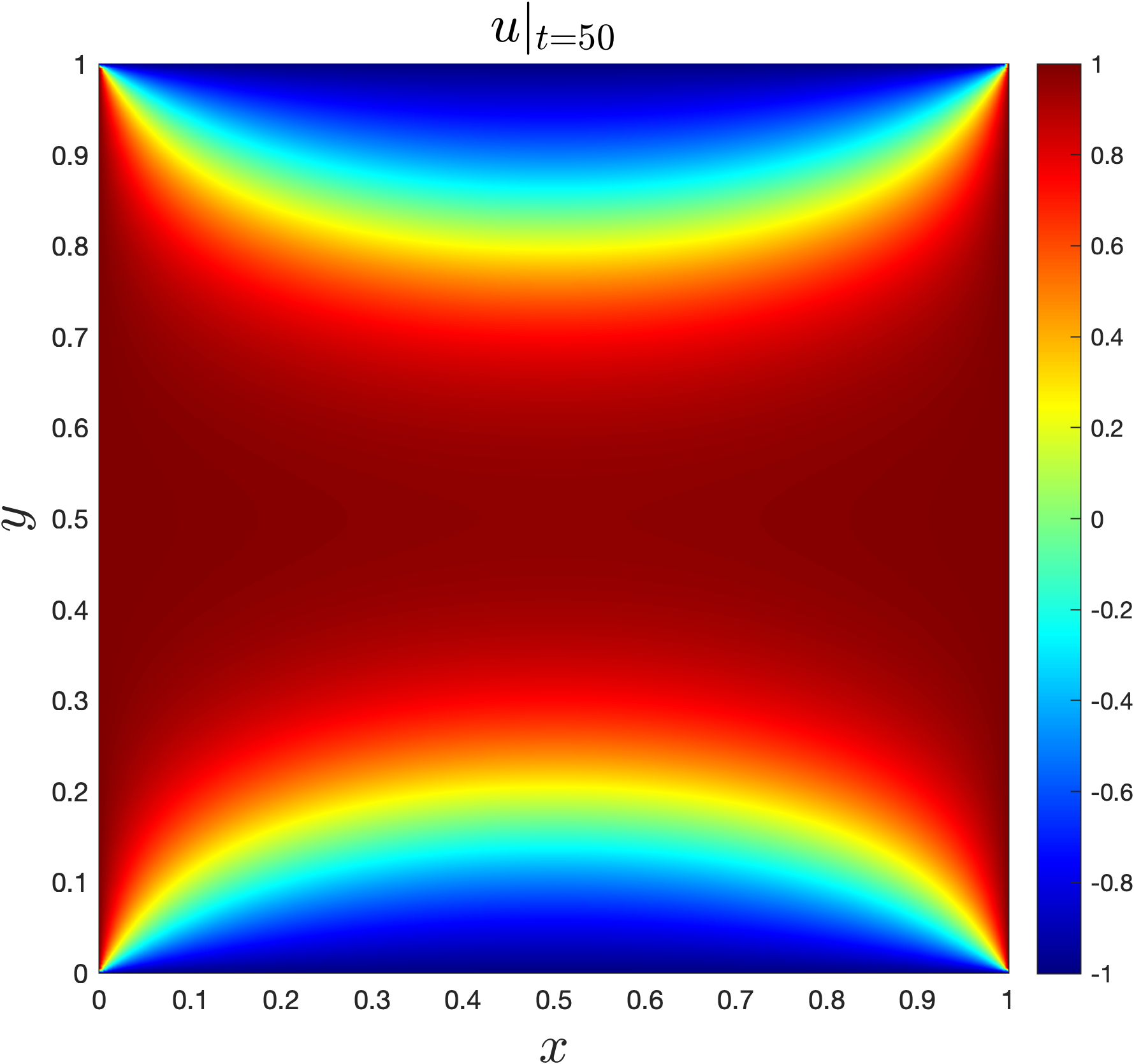}
    }
    \subfigure[A PINN approximation of $u_2$ that stays at $u_2$]{
    \includegraphics[scale=0.2]{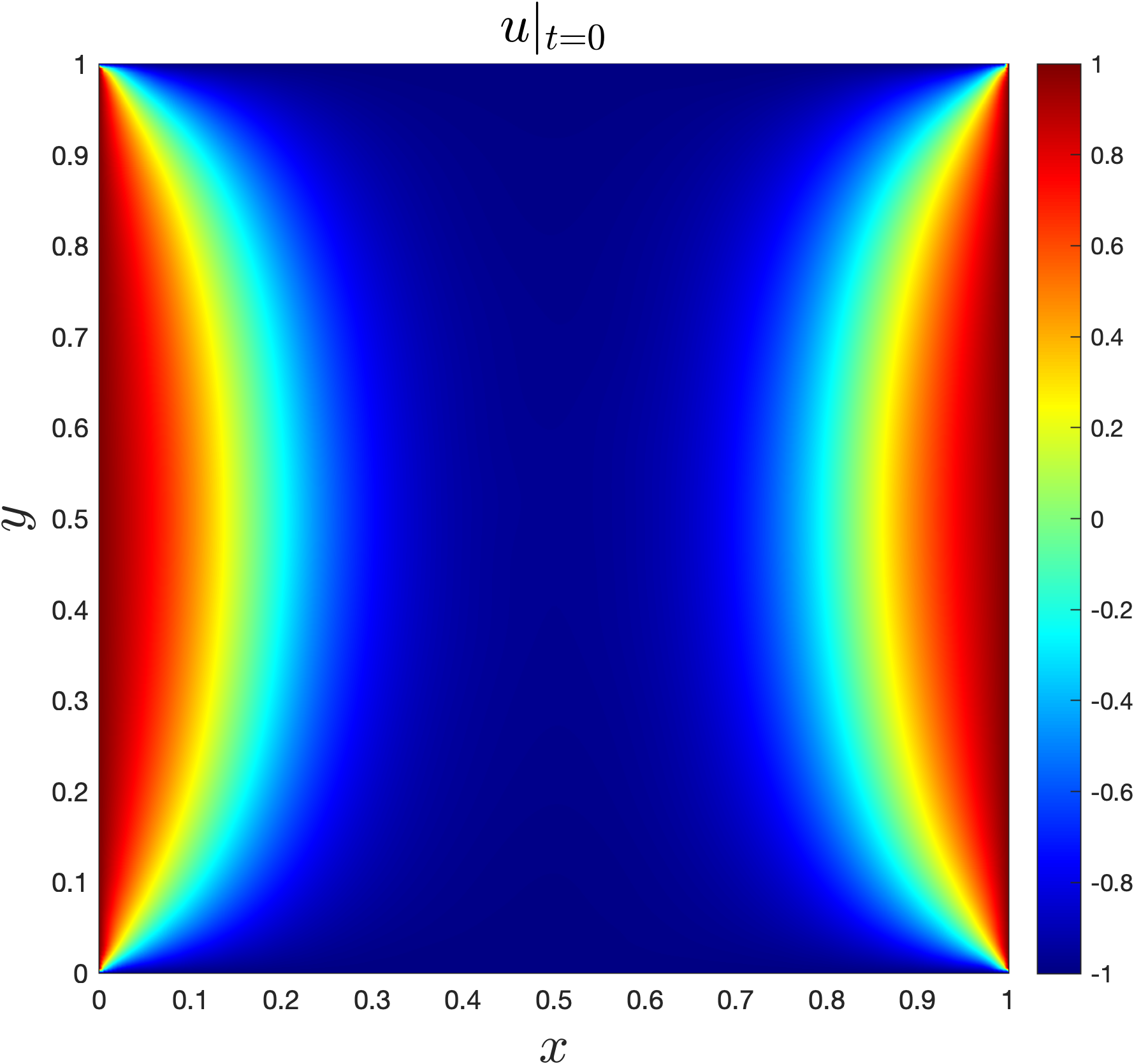}
    \includegraphics[scale=0.2]{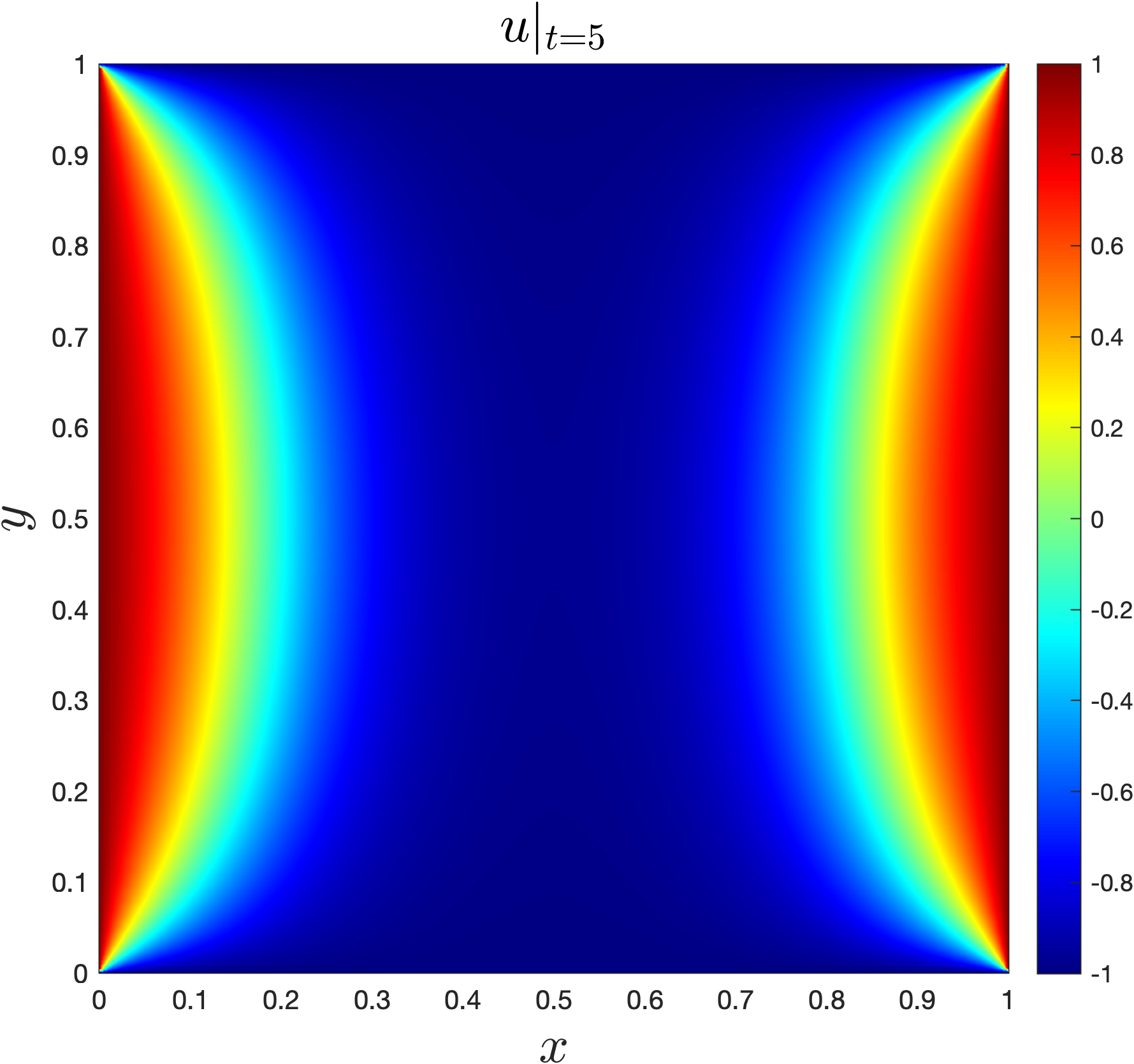}
    \includegraphics[scale=0.2]{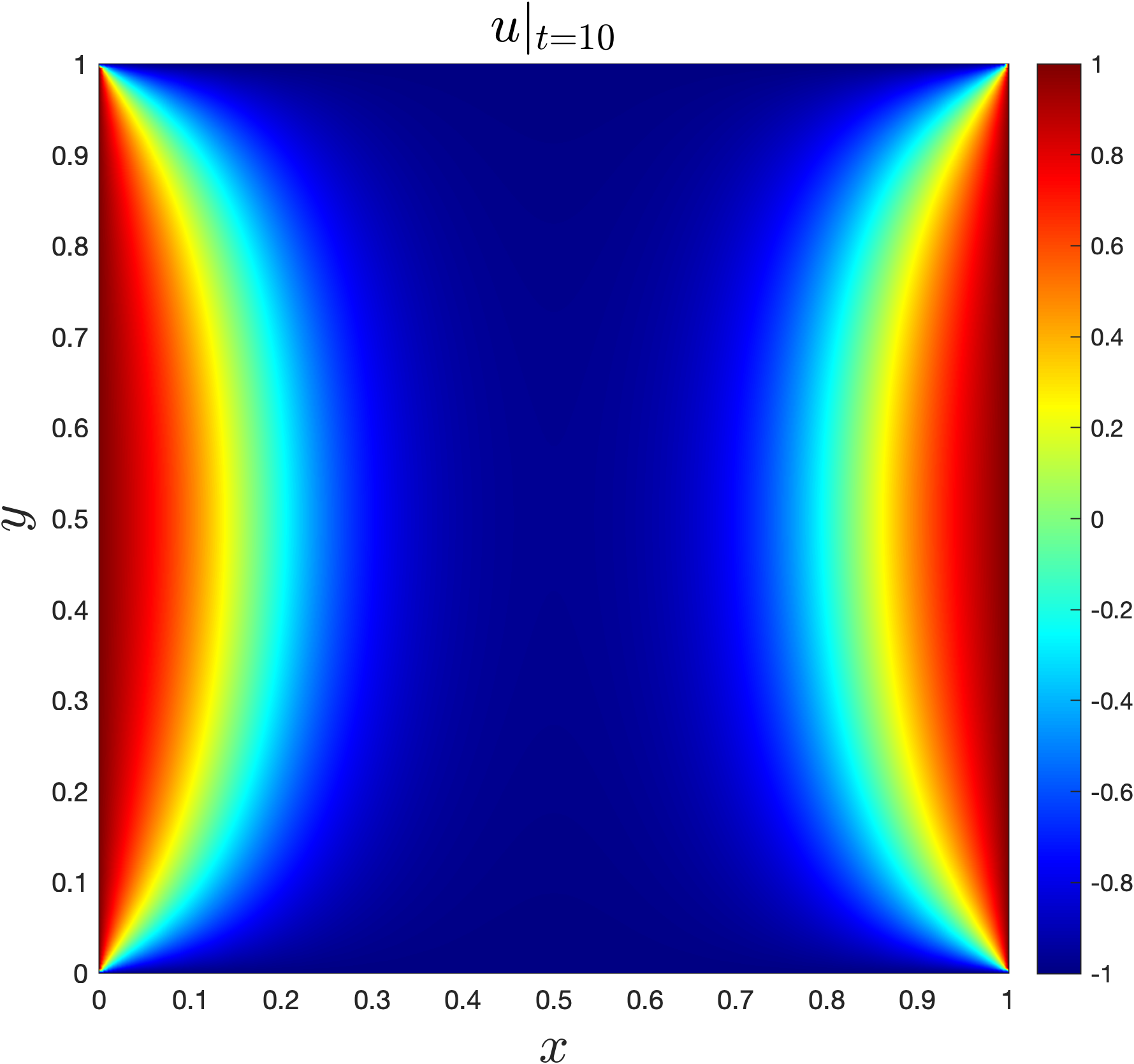}
    \includegraphics[scale=0.2]{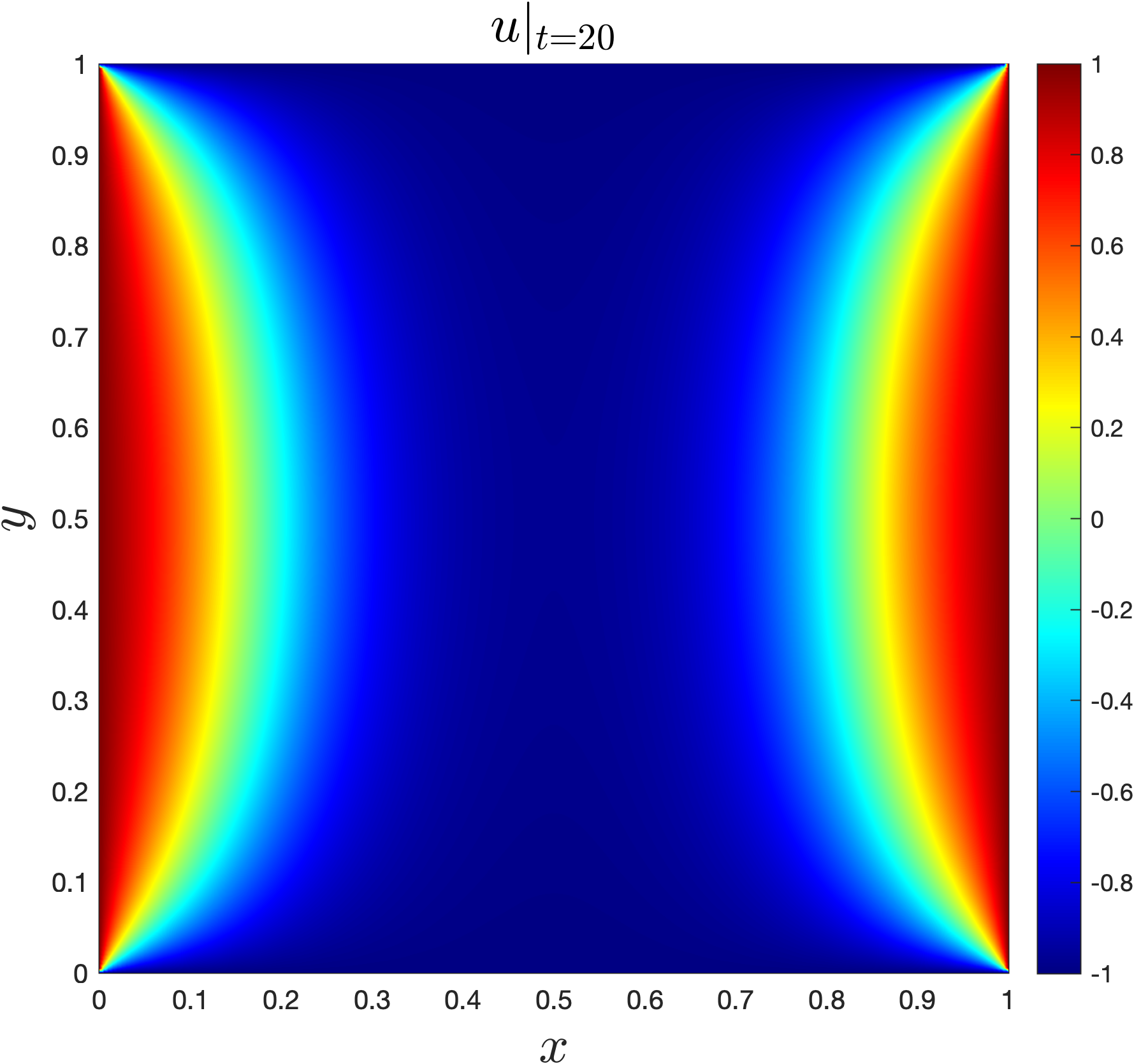}
    \includegraphics[scale=0.2]{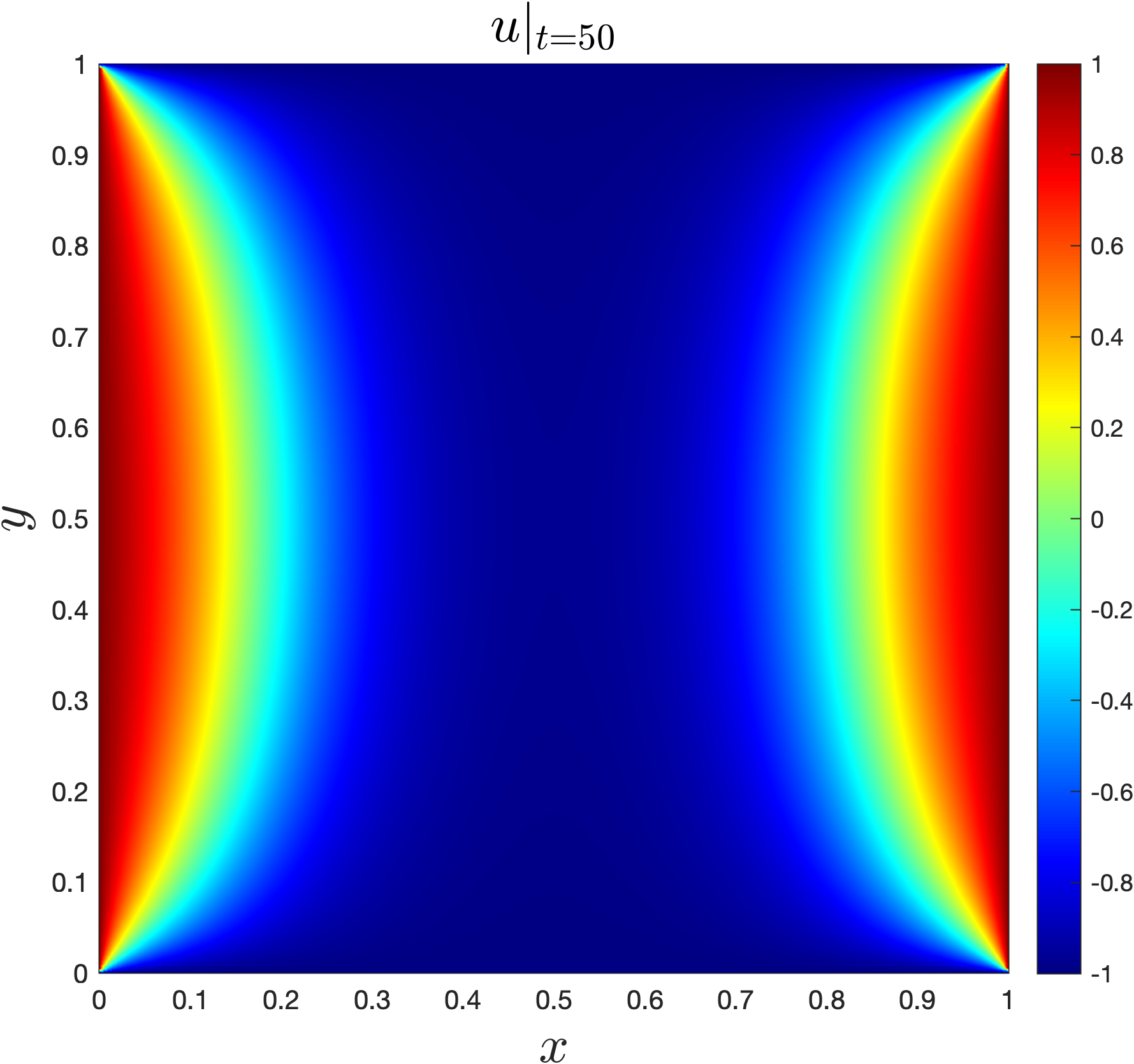}
    }
    \subfigure[A PINN approximation of $u_3$ that stays at $u_3$]{
    \includegraphics[scale=0.2]{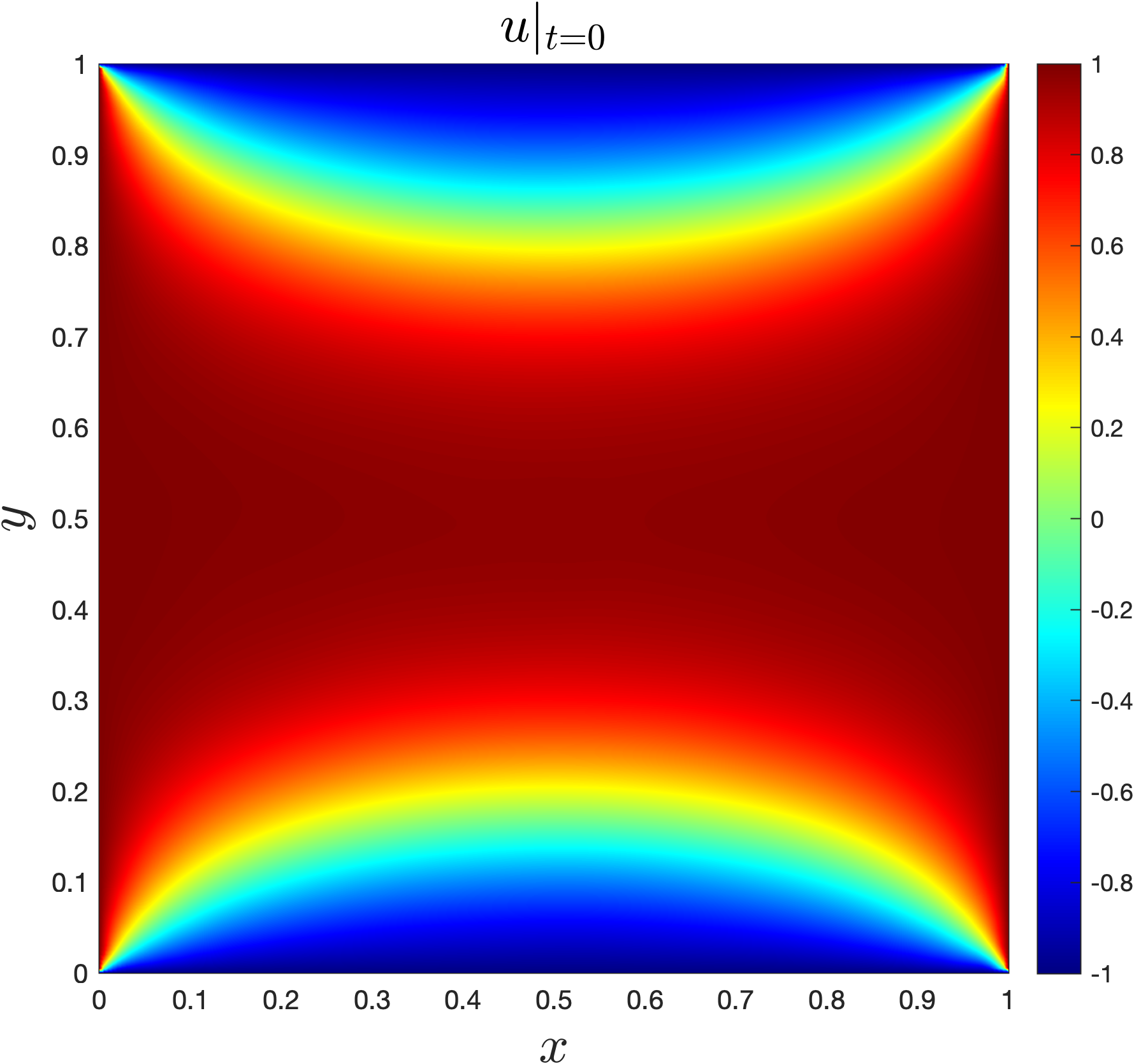}
    \includegraphics[scale=0.2]{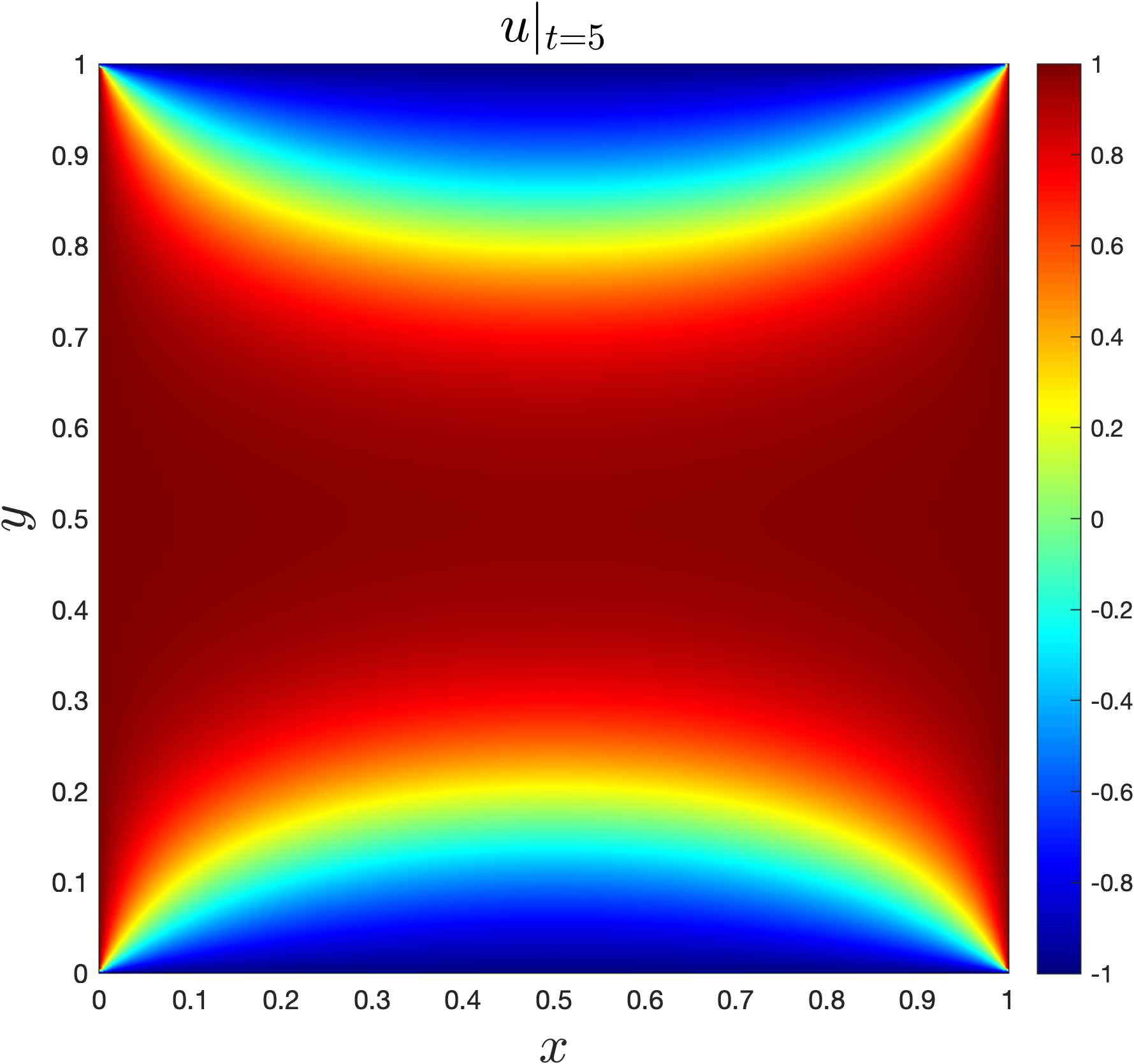}
    \includegraphics[scale=0.2]{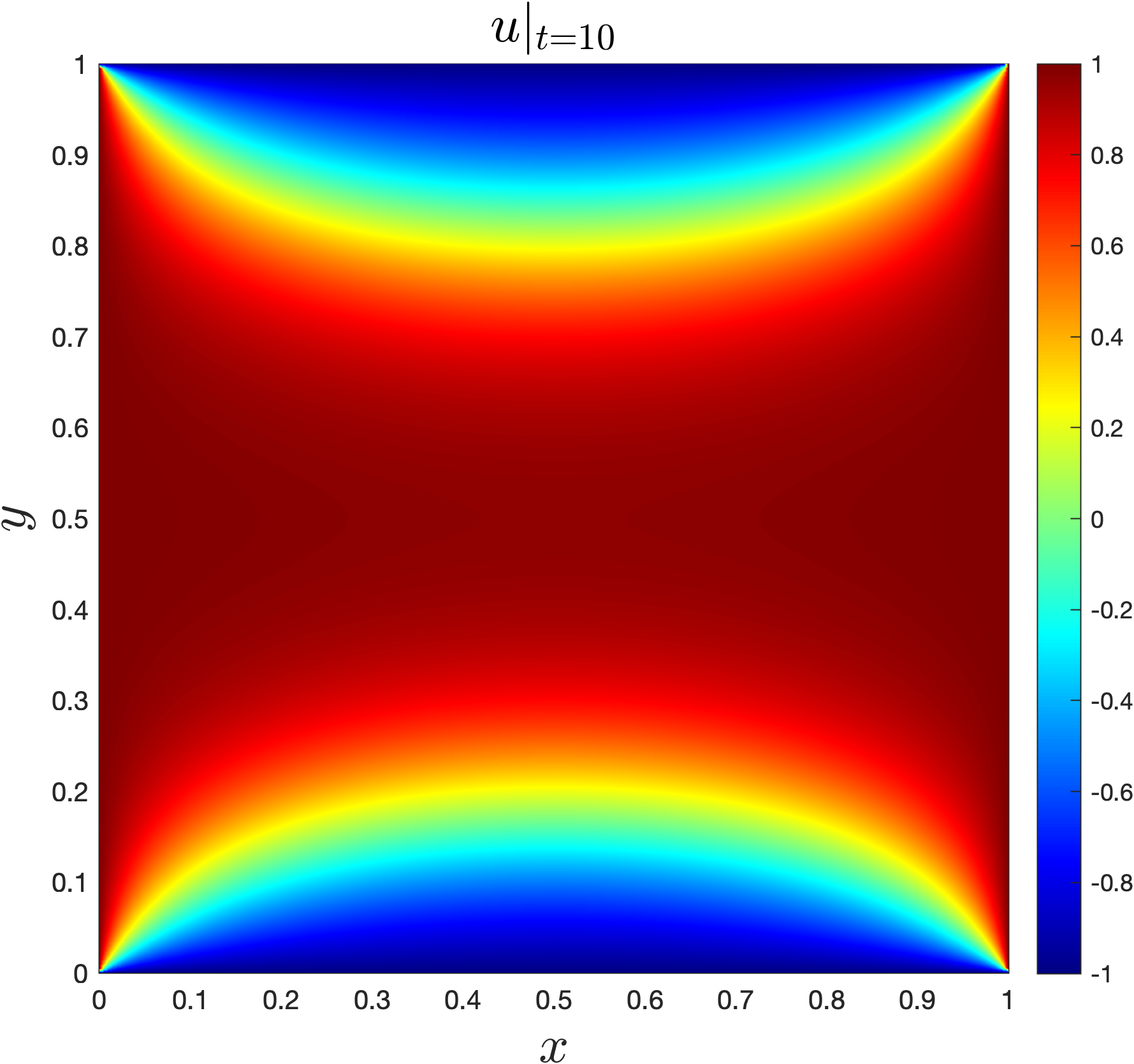}
    \includegraphics[scale=0.2]{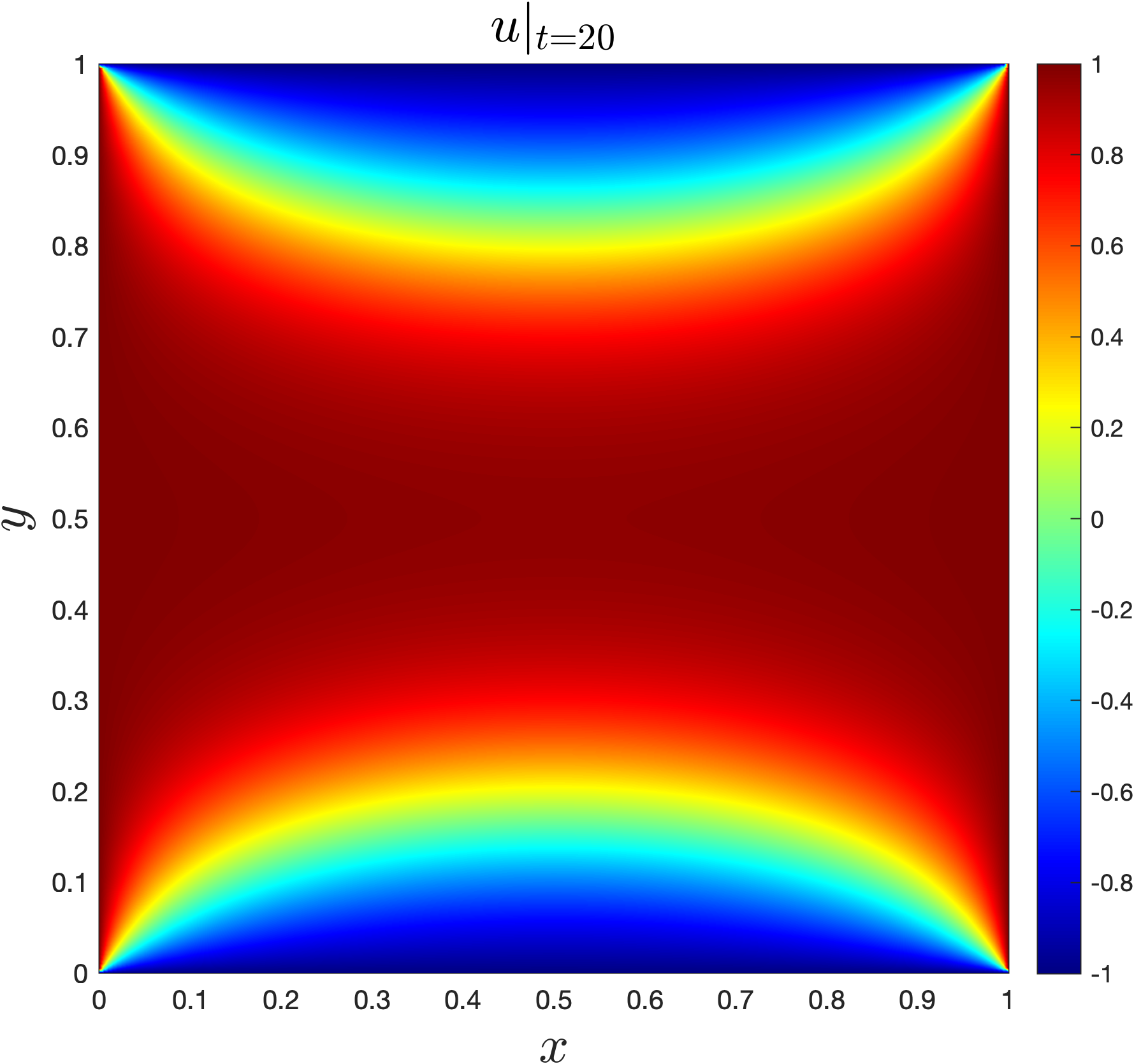}
    \includegraphics[scale=0.2]{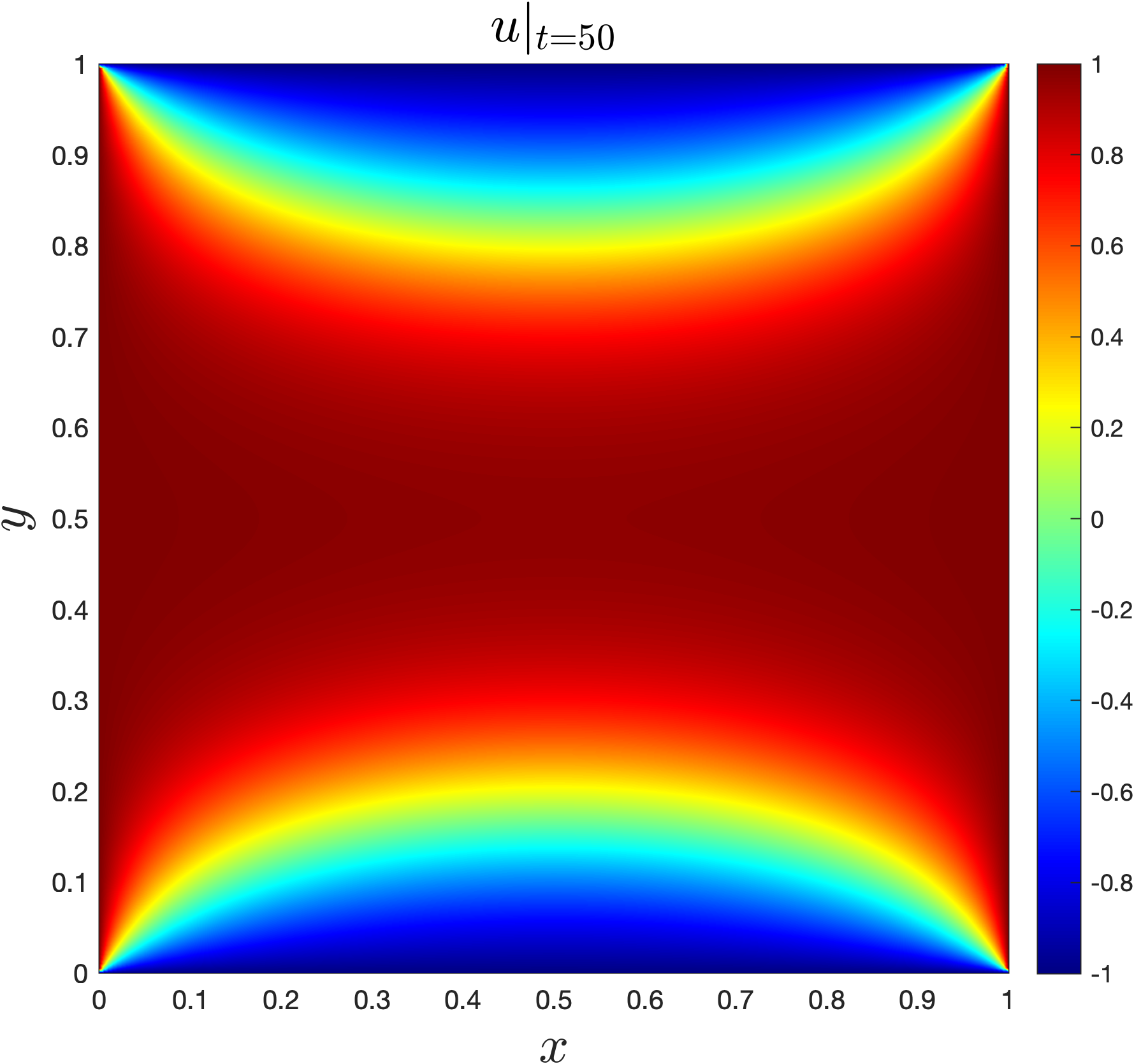}
    }
    \caption{Results from applying the time-marching approach to solve \eqref{eq:2d_allen} with $\epsilon=0.01$, using different PINN solutions as initial conditions. From left to right, the panels show $u$ at $t=0, 5, 10, 20, 50$ where $t$ represents the artificial time. Here, $u_1, u_2$ and $u_3$ correspond to the three distinct solutions shown in Figure \ref{fig:example_5}(a), with $u_1$ being an unstable solution, while $u_2$ and $u_3$ are stable.}
    \label{fig:example_5_4}
\end{figure}

We display three representative PINN solutions in Figure \ref{fig:example_5_3} from solving \eqref{eq:2d_allen} with $\epsilon=0.01$ using PINNs with random initialization and deep ensemble. These PINN solutions are used as initial guesses for solving \eqref{eq:2d_allen} with $\epsilon=0.01, 0.001$ using a FEM-based numerical solver to produce results shown in Figures \ref{fig:example_5} and \ref{fig:example_5_2}.

The time-marching approach is implemented as solving the following time-dependent PDE:
\begin{equation}
    \frac{\partial u}{\partial t} = \epsilon \Delta u - u^3 + u, t\in[0, 50].
\end{equation}
The same computational mesh and numerical solver are used.
The results are presented in Figure \ref{fig:example_5_4}, demonstrating that $u_1$ is an unstable solution and $u_2, u_3$ are stable, where $u_1, u_2, u_3$ refer to the three distinct solutions shown in Figure \ref{fig:example_5}(a).

\end{document}